\documentclass[review,3p]{elsarticle}
\usepackage{hyperref}
\usepackage{float}
\usepackage{verbatim}
\graphicspath{{./}}
\usepackage{xcolor}
\usepackage{xpatch}
\usepackage{graphicx}
\usepackage[section]{placeins}
\usepackage{float}
\definecolor{revisioncolor}{rgb}{1,0,0} 
\usepackage{amsthm,amsmath,amssymb}
\usepackage{mathrsfs}

\usepackage{amsmath}

\usepackage{graphicx}

\usepackage{soul, color, xcolor}
\graphicspath{{./}}

\usepackage{amssymb}
\usepackage{float}
\usepackage{natbib}

\usepackage{tabu}                     
\usepackage{multirow}                
\usepackage{multicol}                 
\usepackage{multirow}               
\usepackage{float}                    
\usepackage{makecell}                
\usepackage{booktabs} 
\usepackage{caption}
\graphicspath{ {./figuress/} }
\usepackage{hyperref}
\usepackage{float}
\usepackage{verbatim} %comments
\usepackage{apalike}
\restylefloat{figure}
\floatstyle{plaintop} %table caption at top
\restylefloat{table}
\usepackage{footmisc}

\usepackage{amsmath,amsfonts}
\usepackage{algorithmic}
\usepackage{algorithm}
\usepackage{array}
\usepackage[caption=false,font=normalsize,labelfont=sf,textfont=sf]{subfig}
\usepackage{textcomp}
\usepackage{stfloats}
\usepackage{url}
\usepackage{verbatim}
\usepackage{graphicx}
\usepackage{amsthm,amsmath,amssymb}
\usepackage{mathrsfs}
\usepackage{cuted}
\usepackage{float}
\usepackage[section]{placeins}

\begin{document}
\begin{frontmatter}

\begin{titlepage}
\begin{center}
\vspace*{1cm}

\textbf{ \large RefLSM: Linearized Structural-Prior Reflectance Model for Medical Image Segmentation and  Bias-Field Correction }

\vspace{1.5cm}

 %Author names and affiliations

 Wenqi Zhao$^{a,\footnotemark[1]}$ (zwq@stu.cqu.edu.cn), Jiacheng Sang$^{b,\footnotemark[1]}$(jiacheng.sang@duke.edu), Fenghua Cheng$^c$ (fenghua.cheng@uq.edu.au), Yonglu Shu$^a$ (shuyonglu@cqu.edu.cn), Dong Li$^a$(lid@cqu.edu.cn), Xiaofeng Yang$^d$(xyang43@emory.edu)\\
 \footnotetext[1]{The two authors contribute equally to this work.}

\hspace{10pt}

 \begin{flushleft}
 \small  
 $^a$ College of Mathematics and Statistics, Chongqing University, 401331, Chongqing, China\\
 $^b$ Deparment of Computer Science, Duke University, NC 27708, Durham, USA \\
 $^c$ The University of Queensland Brisbane, Queensland, Australia\\
$^d$ Department of Radiation Oncology and Winship Cancer Institute, Emory University, Atlanta, GA 30322, United States of America
 \begin{comment}
% Clearly indicate who will handle correspondence at all stages of refereeing and publication, also post-publication. Ensure that phone numbers (with country and area code) are provided in addition to the e-mail address and the complete postal address. Contact details must be kept up to date by the corresponding author.
\end{comment}

 \vspace{1cm}
 \textbf{Corresponding Author:} \\
 Yonglu Shu \\
 College of Mathematics and Statistics, Chongqing University, 401331, Chongqing, China \\
 Tel: 18852128835 \\
 Email: shuyonglu@cqu.edu.cn
 \end{flushleft}        
\end{center}
\end{titlepage}

\title{ RefLSM: Linearized Structural-Prior Reflectance Model for Medical Image Segmentation and  Bias-Field Correction }
\tnotetext[t1]{Under review at Pattern Recognition}
%\author[1]{Wenqi Zhao\fnref{fn1}}
%\ead{zwq@stu.cqu.edu.cn}

%\author[2]{Jiacheng Sang\fnref{fn1}}
%\ead{jiacheng.sang@duke.edu}

%\author[3]{Fenghua Cheng}
%\ead{fenghua.cheng@uq.edu.cn}

%\author[1]{Yonglu Shu\corref{cor1}}
%\ead{shuyonglu@cqu.edu.cn}

%\author[1]{Dong Li}
%\ead{lid@cqu.edu.cn}

%\author[4]{Xiaofeng Yang}
%\ead{xyang43@emory.edu}

%\address[1]{College of Mathematics and Statistics, Chongqing University, 401331,  Chongqing, China}
%\address[2]{Department of Computer Science, Duke University, NC 27708, Durham, USA}
%\address[3]{The University of Queensland, Brisbane, Queensland, Australia}
%\address[4]{Department of Radiation Oncology and Winship Cancer Institute, Emory University, Atlanta, GA 30322, USA}

%\fntext[fn1]{The two authors contribute equally to this work.}

\begin{abstract}
Medical image segmentation remains challenging due to intensity inhomogeneity, noise, blurred boundaries, and irregular structures. 
Traditional level set methods, while effective in certain cases, often depend on approximate bias field estimations and therefore struggle under severe non-uniform imaging conditions. 
To address these limitations, we propose a novel variational Reflectance-based Level Set Model (RefLSM), which explicitly integrates Retinex-inspired reflectance decomposition into the segmentation framework. 
By decomposing the observed image into reflectance and bias field components, RefLSM directly segments the reflectance, which is invariant to illumination and preserves fine structural details. 
Building on this foundation, we introduce two key innovations for enhanced precision and robustness. First, a linear structural prior steers the smoothed reflectance gradients toward a data‑driven reference, providing reliable geometric guidance in noisy or low‑contrast scenes. Second, a relaxed binary level‑set is embedded in RefLSM and enforced via convex relaxation and sign projection, yielding stable evolution and avoiding reinitialization‑induced diffusion.
The resulting variational problem is solved efficiently using an ADMM-based optimization scheme.
Extensive experiments on multiple medical imaging datasets demonstrate that RefLSM achieves superior segmentation accuracy, robustness, and computational efficiency compared to state-of-the-art level set methods.

\end{abstract}

\begin{keyword}
	Retinex \sep Reflectance prior constraint \sep Binary level set  \sep Image segmentation \sep Bias correction 
\end{keyword}

\end{frontmatter}

\section{Introduction}
\label{introduction}

Image segmentation has emerged as a crucial technology with the field of image processing. It plays a vital role in various applications, including medical image analysis, facial recognition, and autonomous vehicles \cite{7684170,Kaymak2019}. In particular, during medical diagnosis aided by image segmentation, medical images frequently exhibit varying degrees of intensity inhomogeneity and noise, which can affect the diagnostic judgment of physicians. In general, intensity inhomogeneity and noise can arise from several factors, including imperfections in imaging devices, subject-induced susceptibility effects, and uneven illumination \cite{10.1007/978-3-540-85990-1_130,6932467}. These issues create overlaps among the intensity ranges of the regions that need to be partitioned. Consequently, precisely segmenting images with severe intensity inhomogeneity and high levels of noise remains a significant challenge.
Recently, advances in machine learning, particularly deep learning techniques, have led to important improvements in image segmentation methods. These learning-based approaches perform well in extracting image features with extensive labeled datasets for model training \cite{10.1007/978-3-319-24574-4_28,10.1007/978-3-319-55524-9_12,10.1007/978-3-030-32248-9_28,VENKATESH2020101748,unknown, 10822098, Ma2024}.

Although considering high-level features provides deep learning methods certain advantages over traditional active contour models, neural networks can struggle with issues such as imprecise segmentation boundaries and overly smooth contours. Their reliance on large datasets also limits their effectiveness, particularly when there are insufficient labeled samples available for training. In such instances, traditional active contour models can often accurately delineate target boundaries and do not require large-scale labeled datasets for model training \cite{HoangNganLe2020}. Thus, traditional methods remain irreplaceable for image segmentation tasks where labeled data is limited. Among various traditional active contour models, the level set method proposed by \citet{OSHER198812} is used formally for image segmentation \cite{Caselles1993,368173}. The level set method represents the level set function implicitly rather than a parametric evolution curve, enabling numerical computation on a fixed Cartesian grid without requiring the parameterization of points along the curve \cite{OSHER198812}. This feature also enables the level set function to depict contours with intricate topologies and to effectively manage topological changes like splitting and merging. Moreover, to overcome the challenges that irregularity and re-initializing often arise in the level set function during the evolution process, leading to numerical inaccuracies and instability in image segmentation. \citet{5557813} introduced the distance regularized level set evolution (DRLSE) to intrinsically maintain the regularity, and \citet{1621239} proposed a piecewise constant level set method that avoids the reinitialization procedure of typical level set methods. Moreover, \citet{4623242} presented a novel variational level set framework aimed at concurrently segmenting and correcting the bias field of MR images affected by intensity inhomogeneities, expanding the application range of the level set methods.

Typically, the level set methods can be classified into two kinds of basic models: edge-based level set \cite{537567, 5557813} and region-based level set \cite{902291,4270039}. The edge-based level set is suited for segmenting images with clear distinctions between the target and the background, and it often computes faster in comparison to region-based level sets. \citet{537567} introduced the geodesic formulation of active contour models.
Based on images' intrinsic geometric measures, this model can detect object boundaries during segmentation. The DRLSE model, another classic edge-based model, employs a distance regularization term for contour evolution toward object edges. More recently, \citet{Akram2017} developed an edge-based method using the difference of Gaussians as a parameter to indicate the edges and segment images globally. \citet{7847297} also proposed an edge-based model with energy terms weighted by their relative importance for boundary detection, utilizing local edge features from surrounding regions. However, the edge-based level set tends to be highly sensitive to the initialization of the level set function and is not effective for objects with weak boundaries.
\begin{figure}[]
	\centering
	\subfloat[]{\includegraphics[width=1.49in]{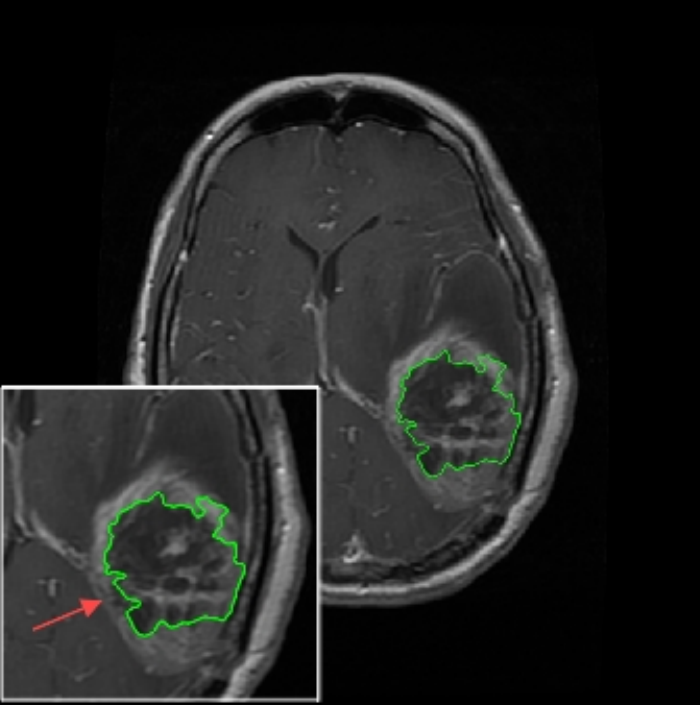}%
		\label{fig_1_case}}
	\subfloat[]{\includegraphics[width=1.5in]{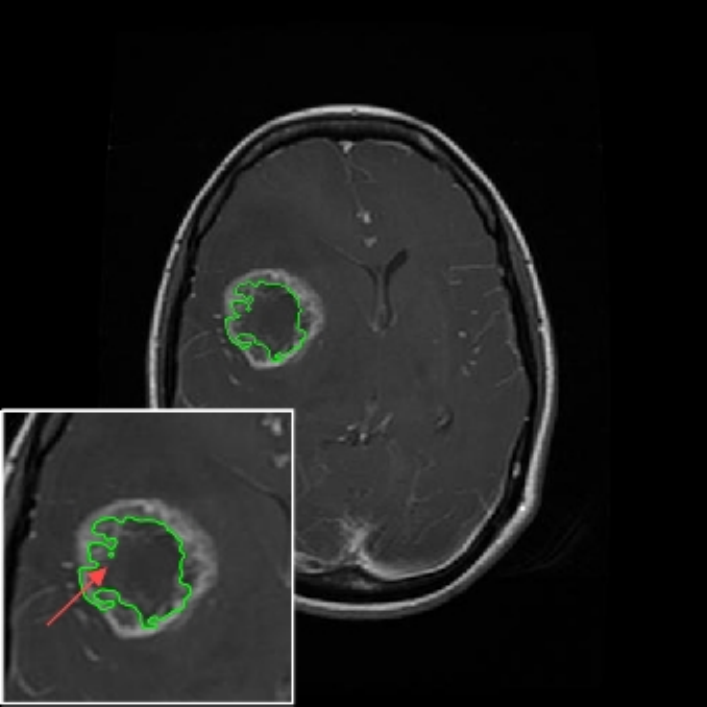}%
		\label{fig_2_case}}
	\caption{Segmentation results from the LIF model for 2 different brain tumor MR images.}
	\label{img1}
\end{figure}
 The CV model \cite{902291}, a classic region-based level set model, evolves the level set to detect objects with contours not defined by gradient. However, it struggles with intensity inhomogeneity because it presumes statistically homogeneous regions. To overcome this issue, \citet{4270039} introduced the local binary fitting (LBF) model with a Gaussian kernel function, allowing the model to precisely capture local image information and \citet{WANG2010603} introduced the Local CV (LCV) model, incorporating additional constraints and using intensity averages to enhance segmenting intensity inhomogeneous images. Moreover, \citet{ZHANG20101199} introduced the local image fitting (LIF) model that can also extract local image information while reducing computational cost by computing only two convolutions in each iteration. However, these model exhibit a sensitivity to initialization and images with noise, as they mainly rely on local information to direct the curve evolution. More recently, \citet{DING2017224} makes the region scalable-fitting model insensitive to the positions of initial contour by integrating the optimized Laplacian of Gaussian(LoG) energy. \citet{MA2019201} introduced an adaptive local fitting(ALF) model and \citet{SHU2023109257} proposed an adaptive local variances-based level set (ALVLS) model, both utilizing the adaptive techniques to fit the region of interest to address images exhibiting intensity inhomogeneity and noise. Additionally, \citet{WENG2021115633} developed an additive bias field correction (ABC) model, achieving faster calculation speeds than traditional multiplicative models by decomposing the observed image into three parts. Even though the aforementioned models outperform old models when dealing with intensity inhomogeneity and noise, there remains significant room for improvement in their segmentation accuracy and robustness when facing high levels of noise and severe intensity inhomogeneity. 
\begin{figure}[!t]
	\centering
	\subfloat[]{\includegraphics[width=1.14in]{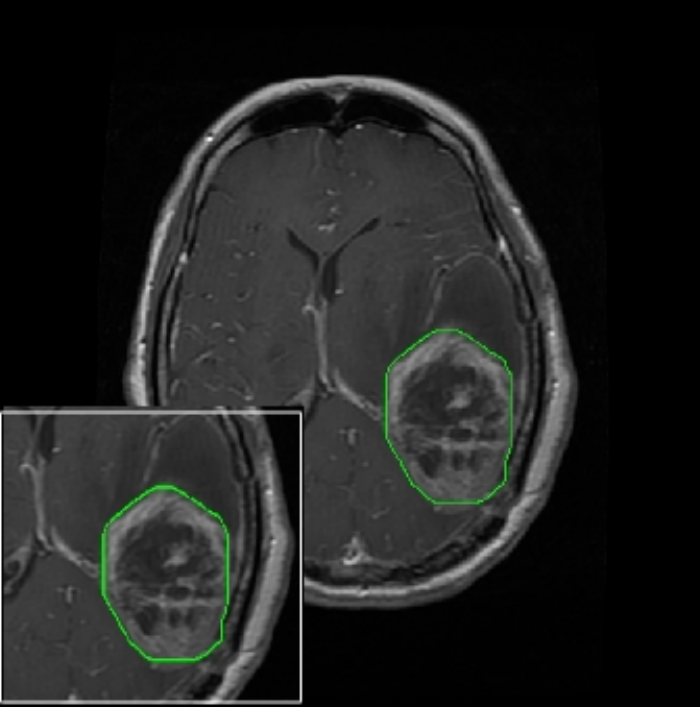}%
		\label{fig_first_case}}
	\subfloat[]{\includegraphics[width=1.15in]{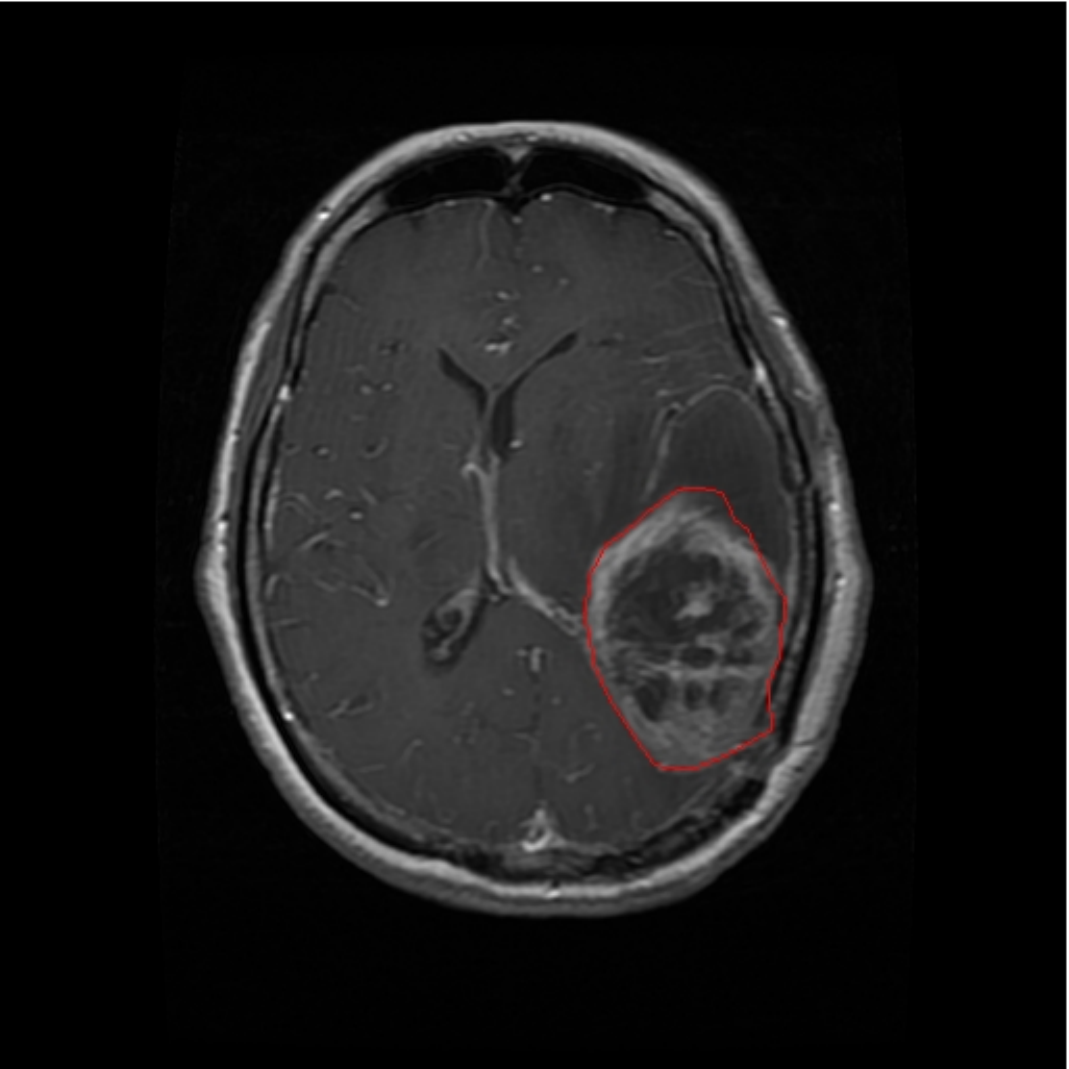}%
		\label{fig_second_case}}
	%\hfil
	\subfloat[]{\includegraphics[width=1.155in]{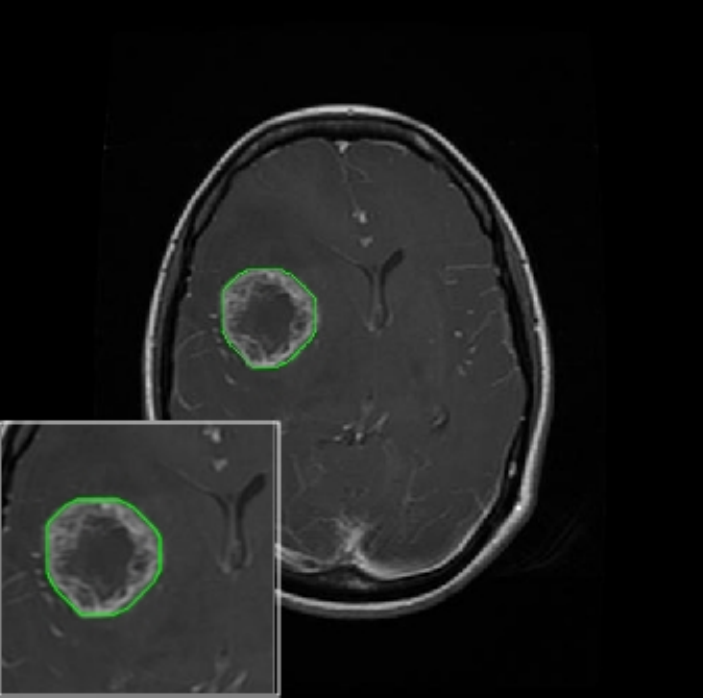}%
		\label{fig_third_case}}
	\subfloat[]{\includegraphics[width=1.155in]{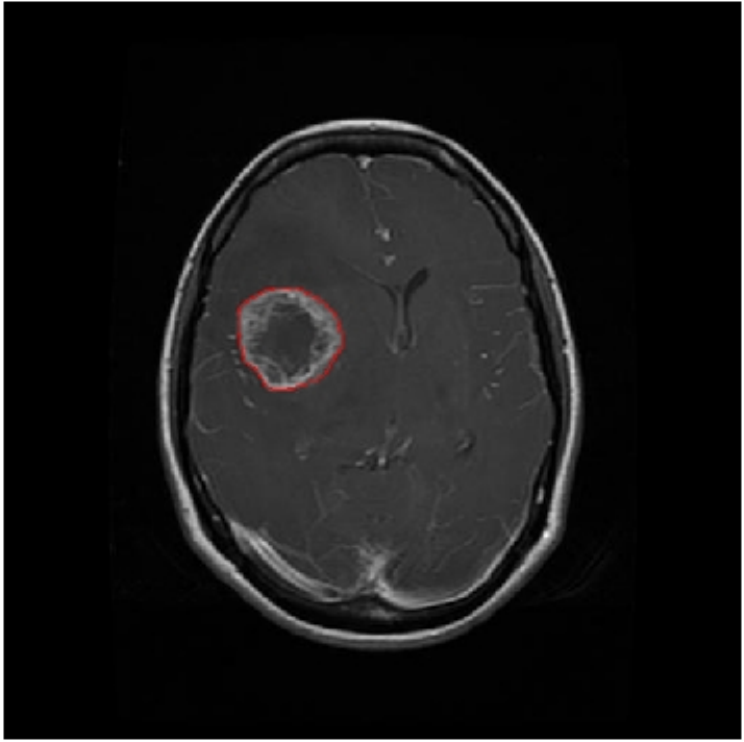}%
		\label{fig_fourth_case}}
	\caption{Segmentation results from the RefLSM for 2 different brain tumor MR images. Row 1 and Row 3 :Results from the RefLSM. Row 2 and Row 4 : Ground truth.}
	\label{img2}
\end{figure}

\indent To further improve the performance of level set models, some researchers \cite{8765635,9619954} combine edge information and region information and propose hybrid level set methods. Moreover, some researchers draw from different fields such as \citet{XUE2024110673} draws on learning-based models by utilizing high-level features and \citet{article} propose Retinex-based approaches. In particular, Retinex theory is widely applied in the image enhancement field \cite{9056796,9032356,CAI2023109195},  which is as fundamental as image segmentation in image processing. Since Retinex theory focuses on the underlying structure information of images, combining Retinex theory into image segmentation can deal with intensity inhomogeneity due to illuminance bias. Esedoglu and Otto proposed a CVB model that integrates Retinex theory into the CV model and developed an efficient algorithm using alternating minimization and threshold dynamics \cite{ESEDOGLU2006367}. While the model effectively corrects bias and segments images, it has unsatisfactory segmentation accuracy in details due to the incomplete removal of bias in the corrected images and its lack of consideration for image noise and multiphase inhomogeneities. Then, \citet{JIN202052} proposed the CVBS model, which improves upon the CVB model by incorporating a total variation term and focusing on the piecewise constant nature of structures. However,to our best knowledge, existing Retinex-based models are still limited and insufficient with regard to segmentation accuracy and noise resistance.

From the above discussions, we have discovered the significant potential of applying Retinex theory to image segmentation and explored its fundamental differences from traditional models. Traditional image segmentation models typically focus on the impact of intensity information on the segmentation results. Therefore, when faced with complex segmentation scenarios, the segmentation results are often affected by lighting, artifacts, and unclear boundaries in the image. As shown in Fig. \ref{img1}, we present the results of the classical local model LIF \cite{ZHANG20101199} for segmenting brain tumor images along with surrounding tissue edema. The irregular ring-like enhancement caused by the edematous tissue leads to irregular boundaries and low contrast in the images. Consequently, the LIF model can only identify the central necrotic and liquefied regions of the tumor, failing to detect the boundaries and becoming trapped in local minima. In this paper, we draw inspiration from the Retinex theory, which is widely applied in the field of image enhancement. According to Retinex theory, the reflectance component characterizes the intrinsic structural properties of the observed image and preserves texture information independent of illumination variations. By integrating this reflectance component into the level set framework, our model achieves robust segmentation of medical images even under severe intensity inhomogeneity. In addition, a linearized Structural-Prior is proposed to restore intensity consistency and capture local geometric features, thereby improving boundary localization in complex or blurred regions. Furthermore, a relaxed binary level set representation is employed to enhance robustness against noise and to enable accurate tracking of complex contours. Based on these innovations, we propose a novel variational reflectance-based level set model (RefLSM) that simultaneously corrects bias fields and performs segmentation. Experimental results demonstrate that RefLSM significantly outperforms conventional level set methods in both segmentation accuracy and robustness. We present the results of our model segmenting the two brain tumor images mentioned above in Fig. \ref{img2}. \\In summary, main contributions of this paper are as follows:
\begin{itemize}
	\item An innovative framework for the energy functional is constructed based on the reflectance component from Retinex theory, enabling the capture of intrinsic texture information in medical images even with bias field interference, thus enhancing the accuracy and stability of the RefLSM when segmenting images with significant intensity inhomogeneity.
	\item A linear structure‑guided prior is introduced to effectively restore image intensity and improve boundary localization, thereby enhancing the bias field correction and segmentation accuracy of the RefLSM for medical images exhibiting severe intensity inhomogeneity.               
	\item A relaxed binary level‑set representation is incorporated into RefLSM to improve robustness to noise and to capture sharp, complex interfaces; it is enforced via convex relaxation and sign projection, avoiding reinitialization and reducing numerical diffusion.
\end{itemize} 
\begin{figure}[]
	\centering
	\subfloat[]{\includegraphics[width=1.41in]{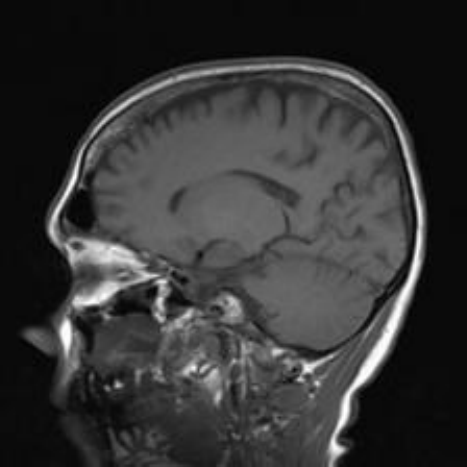}%
		\label{fig_5_case}}
	\subfloat[]{\includegraphics[width=1.395in]{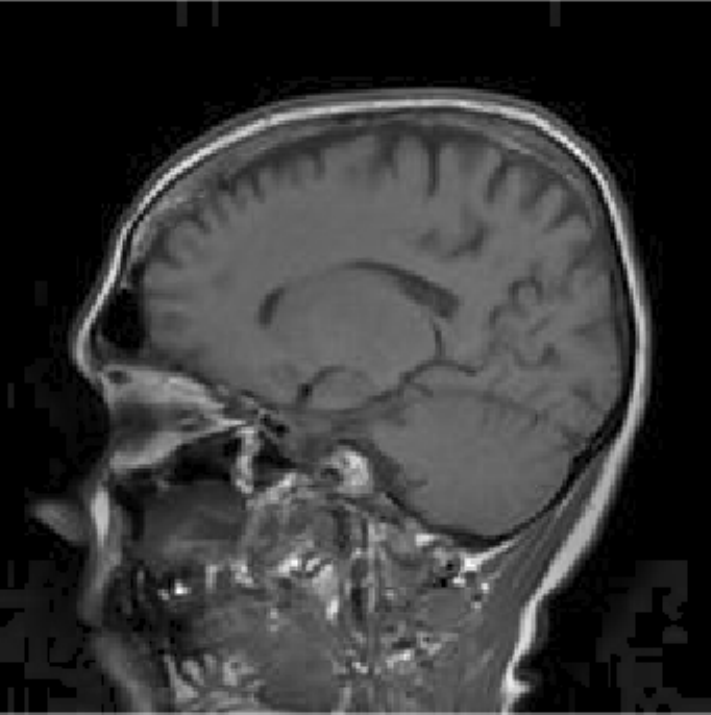}%
		\label{fig_6_case}}
	\subfloat[]{\includegraphics[width=1.402in]{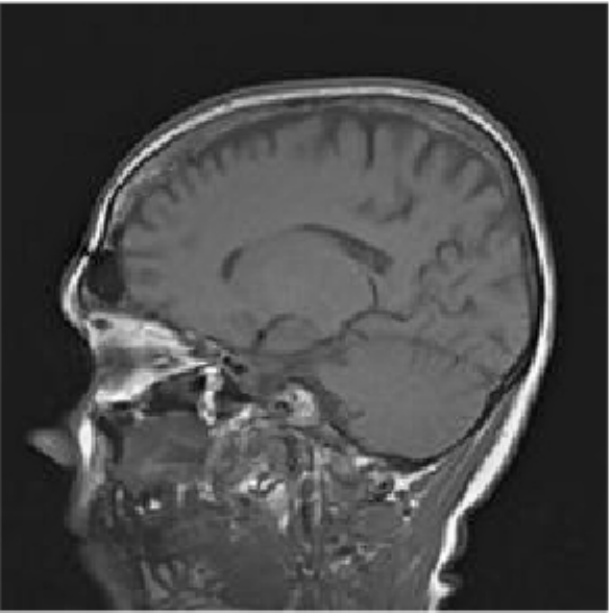}%
		\label{fig_7_case}}
	\hfil
	\subfloat[]{\includegraphics[width=1.5in]{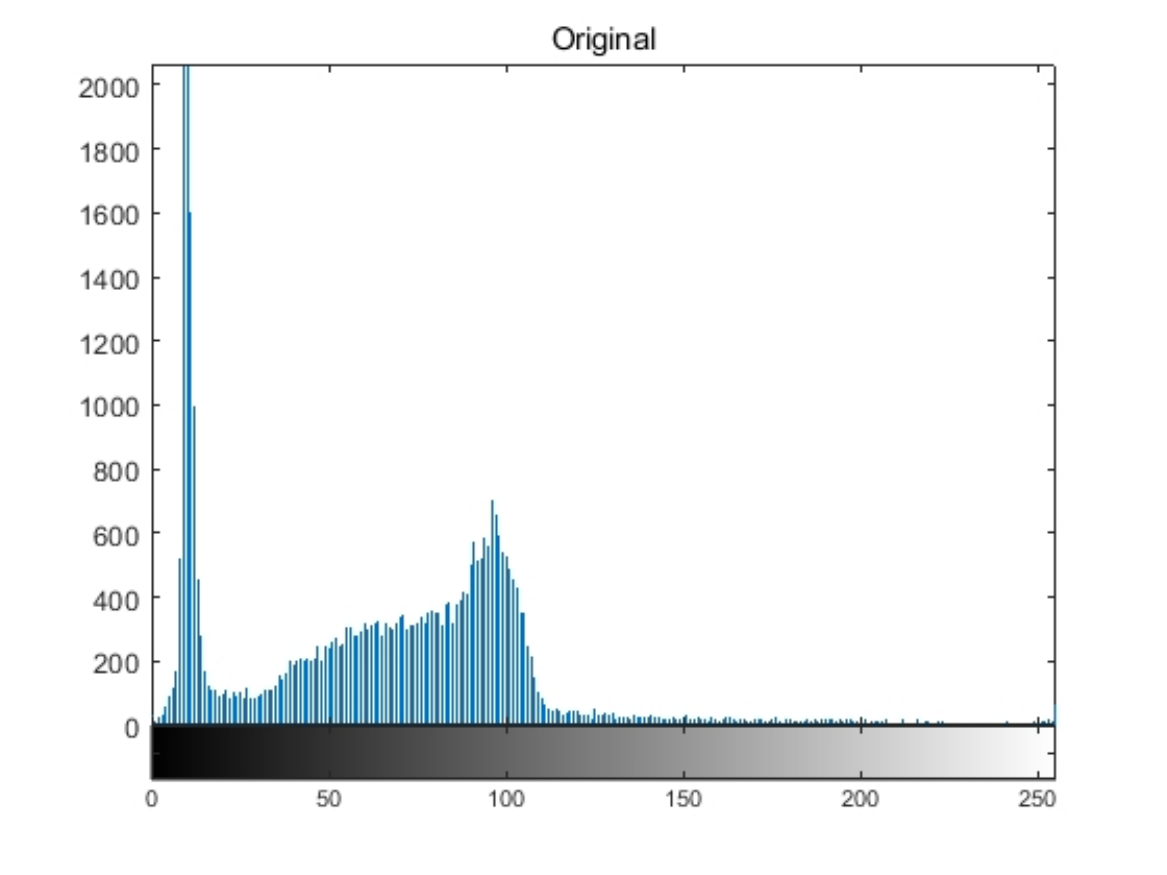}%
		\label{fig_8_case}}
	\subfloat[]{\includegraphics[width=1.5in]{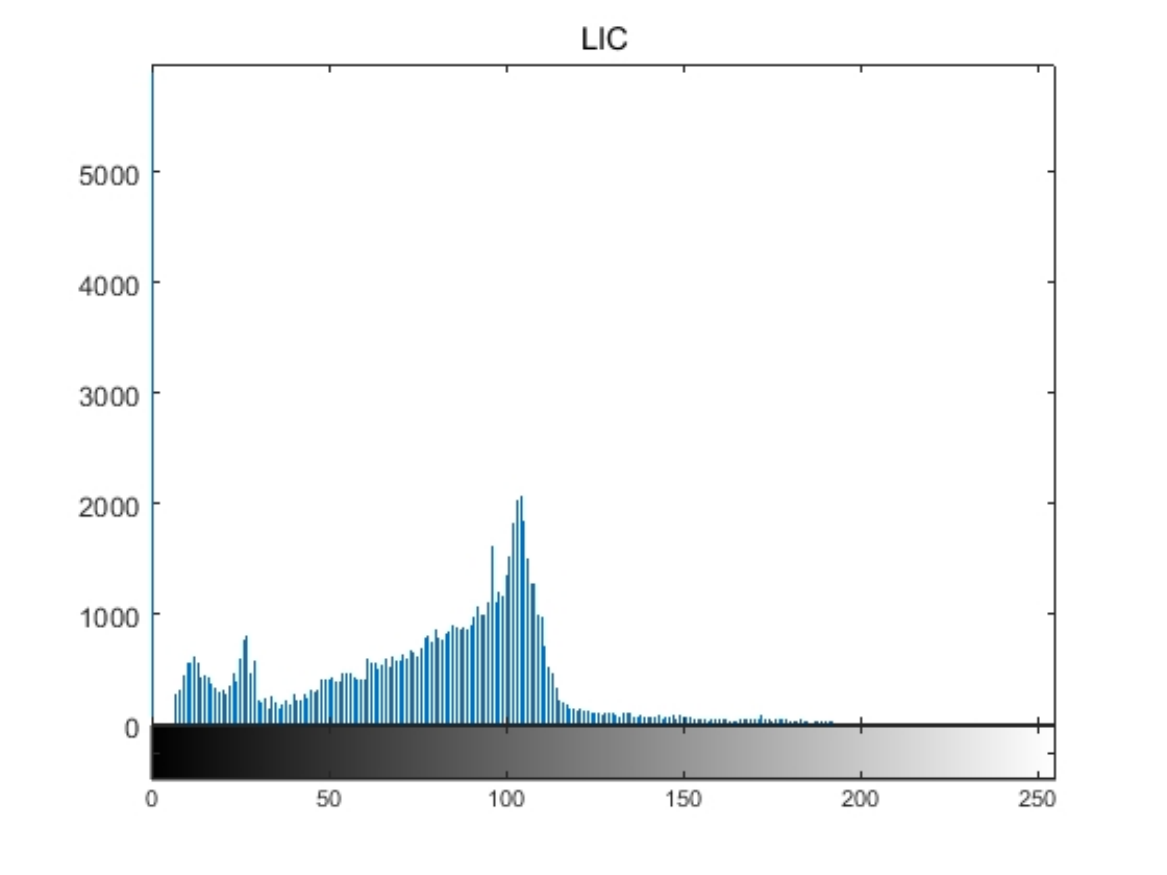}%
		\label{fig_9_case}}
	\subfloat[]{\includegraphics[width=1.5in]{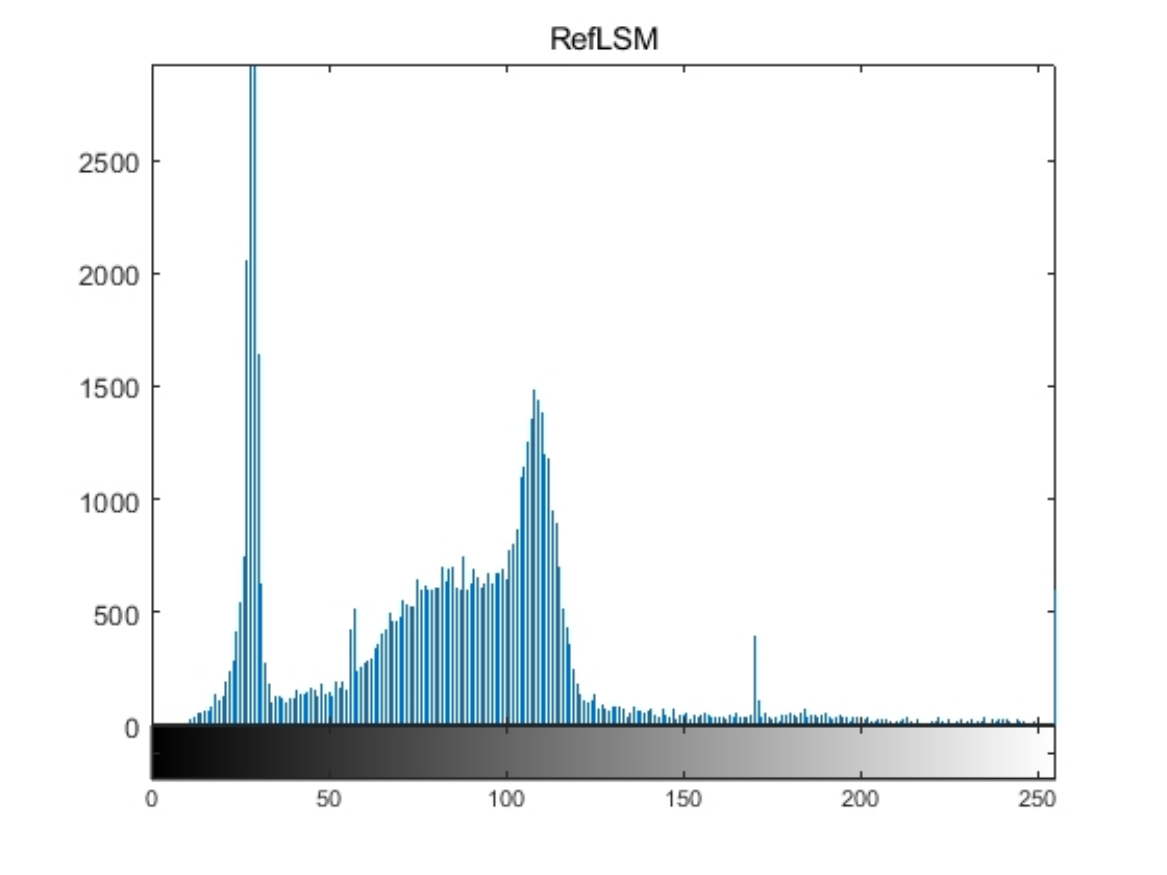}%
		\label{fig_10_case}}
	\caption{Image intensity inhomogeneity comparison with the bias field correction of the LIC model and the RefLSM. (a) is a brain MR image and (d) is its histogram. (b) is the result from the LIC model for bias field correction and (e) is its histogram. (c) is the result from the RefLSM for bias field correction and (f) is its histogram.}
	\label{img14}
\end{figure}
The structure of the paper is arranged as follows: Section \ref{section:A} outlines various models that relate to the RefLSM model. Section \ref{section:B} introduces the novel reflectance-based level set model and its advantages. Section \ref{section:C} shows the numerical implementation of the RefLSM model. Section \ref{section:D} assesses the RefLSM method through validation and analysis experiments. Finally, Section \ref{section:E} reviews the conclusions of this work.

.

\section{BACKGROUND}\label{section:A}
\subsection{LIF model}
\citet{ZHANG20101199} introduced the Local Image Fitting (LIF) model to handle images with intensity inhomogeneity. 
The core idea is to fit local image intensities within a moving window and minimize the discrepancy between the observed image and its locally fitted approximation. 
The energy functional is defined as
\begin{equation}
E^{\mathrm{LIF}}(\phi) = \frac{1}{2}\int_{\Omega} |I(x) - I_0(x)|^2\,dx,
\end{equation}
where $I:\Omega \subset \mathbb{R}^2 \to \mathbb{R}$ is the image, and $I_0$ is the locally approximated image:
\begin{equation}
I_0 = v_1 H(\phi) + v_2 (1-H(\phi)),
\end{equation}
with $H(\phi)$ denoting the Heaviside function. 
The local means $v_1$ and $v_2$ are computed within a window $P_k(x)$ (e.g., constant or Gaussian) centered at $x$: 
\begin{equation}
v_1 = \operatorname{mean}\!\big(I \,\big|\, \{x\in\Omega\mid \phi(x)<0\}\cap P_k(x)\big),\quad
v_2 = \operatorname{mean}\!\big(I \,\big|\, \{x\in\Omega\mid \phi(x)>0\}\cap P_k(x)\big).
\end{equation}

The LIF model leverages local information to better handle intensity inhomogeneity compared to global fitting methods. 
However, it suffers from several drawbacks: the segmentation result is highly sensitive to the initialization of the contour, the choice of window size, and it remains vulnerable to image noise. 
These limitations motivate the development of more robust models that incorporate structural priors and reflectance information.
\subsection{Binary Level Set}

The binary level set (BLS) framework of Lie et al.~\cite{1621239} replaces the signed distance function by a discontinuous binary label $\phi \in \{-1,+1\}$ and enforces the hard constraint $\phi^2=1$ through an augmented Lagrangian penalty.  This removes reinitialization and reduces numerical diffusion, but introduces non‑convex constraints that require careful parameter tuning. A representative BLS formulation is
\begin{equation}\label{BLS}
\min_{c_1,c_2,\phi} 
\;\; \tfrac{1}{2}\!\int_\Omega (I - (c_1 \upsilon_1+c_2 \upsilon_2))^2 dx
+ \xi \!\int_\Omega |\nabla \phi| dx ,
\quad \text{s.t. } \phi^2=1,
\end{equation}
where $\upsilon_1=\tfrac{1}{2}(1+\phi)$ and $\upsilon_2=\tfrac{1}{2}(1-\phi)$.

In contrast, our model adopts a relaxed binary variable $u \in [-1,1]$ with soft weights $w=(1+u)/2$ \cite{ChanEsedogluNikolova2006, Bresson2007FGM}and obtains the final binary segmentation by $\phi=sign(u)$. This convex relaxation removes the need for augmented‑Lagrangian penalties, improves numerical stability, and integrates naturally with our reflectance fitting and structure‑guided prior. 

\subsection{CVB model}
Before introducing the CVB model, it is important to recall the Retinex theory. 
The term ``Retinex,'' proposed by Land and McCann \cite{1977The}, is derived from a combination of ``retina'' and ``cortex'' 
and provides a perceptual model for explaining how the human visual system distinguishes reflectance from illumination under uncertain lighting conditions. 
According to the Retinex model, the observed image $i$ can be decomposed into the product of an illumination component $b$ and a reflectance component $s$:
\begin{equation}\label{A}
i = b \circ s ,
\end{equation}
where ``$\circ$'' denotes point-wise multiplication. 
Applying a logarithmic transform yields
\begin{equation}\label{B}
I = B + S, \qquad I=\log(i),\; B=\log(b),\; S=\log(s),
\end{equation}
where the reflectance $S$ captures the structural details (edges, textures), while the illumination $B$ varies smoothly across the image.

Motivated by this decomposition, \citet{article} proposed the Chan–Vese model with bias correction (CVB), which integrates Retinex theory into the segmentation process. 
The corresponding energy functional is given by
\begin{equation}\label{C}
\begin{aligned}
E_{\mathrm{CVB}}(c_1,c_2,\phi,B,S) &= 
\lambda_1 \int_\Omega (c_1 - S)^2 H(\phi)\,dx
+ \lambda_2 \int_\Omega (c_2 - S)^2 (1-H(\phi))\,dx \\
&\quad + \gamma \int_\Omega |\nabla H(\phi)|\,dx
+ \nu \int_\Omega |\nabla B|^2\,dx ,
\end{aligned}
\end{equation}
where $\lambda_1,\lambda_2,\gamma,\nu>0$ are weighting parameters, $c_1$ and $c_2$ represent the mean reflectance values in the respective regions, and the last term enforces smoothness of the bias field $B$. 

The CVB framework enables simultaneous bias field correction and object segmentation. 
However, its performance degrades when images exhibit severe boundary blurring or strong noise: the segmentation may fail to accurately delineate object boundaries, while the correction of intensity inhomogeneity remains limited. 
These challenges highlight the need for more robust formulations, motivating our proposed reflectance-guided model.

\section{The RefLSM}
In this section, we first discuss the motivation underlying our proposed approach. 
We then describe how the method is designed to address the challenges of image segmentation in the presence of severe intensity inhomogeneity and high levels of noise. 
Finally, we present the level set formulation together with the complete numerical implementation of the algorithm.
\subsection{Reflectance-based Level Set Framework}\label{section:B}

According to Retinex theory, an observed image $i$ can be decomposed as the element-wise product of reflectance $s$ and illumination $b$, i.e., $i=b \circ s$, where $\circ$ denotes point-wise multiplication. Transforming into the logarithmic domain yields
\[
I = B + S, \quad S=\log(s), \; B=\log(b), \; I=\log(i).
\]
Here, the reflectance $S$ encodes the intrinsic structural properties of objects and is invariant to illumination, thereby preserving fine details even under varying lighting conditions \cite{9056796}. In medical imaging, particularly MR images, intensity inhomogeneity induced by bias fields manifests in a manner similar to uneven illumination in natural images \cite{5995422}. Motivated by this analogy, we incorporate Retinex theory into a level set framework, enabling segmentation to be driven by intrinsic reflectance rather than corrupted raw intensities.
\subsubsection{Reflectance Component Energy Functional}

To address the challenges of segmenting medical images with low contrast, severe intensity inhomogeneity, or high levels of noise, we propose a novel variational model that explicitly incorporates the reflectance component. By fitting $S$ rather than the observed intensity $I$, the model effectively decouples the influence of bias fields, illumination variations, and overlapping tissue signals, thereby achieving robust segmentation even in challenging cases.
In practice, the reflectance-driven design yields two concrete advantages. First, it enhances robustness against spatially varying illumination and severe intensity inhomogeneity, since the bias field is explicitly modeled and separated. Second, it improves delineation of regions with weak contrast or blurred edges, as reflectance emphasizes structural consistency. 
Formally, we define a region-fitting function for the reflectance component $S$ in the continuous domain $\Omega$, 
\[
S^{\mathrm{FFR}}(u) = c_1 w + c_2 (1-w),
\]
Let $u:\Omega \to [-1, 1]$. Here, we adopt a relaxed binary soft level-set variable $u \in [-1,1]$; after convergence we obtain a truly binary level-set $\phi=sign(u^k)$. Thus, we define the soft foreground membership as $w = \frac{1+u}{2}$,  while the background membership is given by  $1-w = \frac{1-u}{2}.$
The regional means $c_1$ and $c_2$ are computed using the same soft weights.  
This formulation allows the segmentation process to exploit intrinsic structural cues, while remaining robust to intensity inhomogeneity and illumination artifacts.  
Consequently, the associated region-fitting energy is defined as:

\begin{equation}
E^{\mathrm{FFR}} = \frac{1}{2} \int_{\Omega} \big(S - S^{\mathrm{FFR}}(u)\big)^2 \, dx .
\end{equation}\\
This design improves robustness and avoids numerical diffusion during evolution.
As a result, the model becomes more suitable for images with complex shapes and strong intensity inhomogeneity.
In our binary level set implementation, the corresponding energy functional is simplified from the original form as follows:
\begin{equation}\label{K}
E(S,c_1,c_2,u) =
\frac{1}{2} \int_{\Omega} \big( S - S^{\text{FFR}}(u) \big)^2 dx
+ \theta \int_{\Omega} ||\nabla u||^2 dx,
\end{equation}
where the binary constraint $u\in[-1,1]$ is enforced directly 
via convex relaxation and sign projection during the numerical implementation, 
thus eliminating the need for the additional penalty term $\nu \int (\phi^2-1)^2 dx$ required in traditional formulations.

\subsubsection{A Novel Prior Constraint Term}

To address the challenge of segmenting images with severe intensity inhomogeneity, we propose a linearized structural prior that directly operates on the reflectance component $S$. Reflectance-based structural information is more robust to illumination variations and bias field distortions, helping preserve weak edges and subtle anatomical boundaries. As shown in Fig.~\ref{img2}, our method can accurately delineate tumor boundaries and surrounding edema even under severe inhomogeneity, where traditional intensity-based models often fail. The proposed prior aligns smoothed reflectance gradients with data-driven directions, enhancing inter-region contrast, preserving weak edges, and stabilizing the evolution of $u$. Formally, we define the linear structure operator $\mathcal{L}_\sigma[S]$ as the gradient field of the smoothed reflectance:
\begin{equation}
    \mathcal{L}_\sigma[S] = \nabla(G_\sigma * S), \quad  \quad \mathcal{L}_\sigma[S] = (\nabla G_\sigma) * S = G_\sigma * (\nabla S).
\end{equation}
This operator captures the essential first-order structure of $S$ in a computationally tractable manner. Where $*$  denotes convolution, and $G_\sigma$ is a symmetric Gaussian kernel. For brevity, we omit the spatial variable and write all quantities as functions over $\Omega$. The equivalence holds under standard boundary conditions.
The prior reference field, $V_{\text{pre}}$, is constructed to leverage the robust directional cues from the observed image $I$ while imposing a regularized and constant structural magnitude, $\alpha$. It is defined as:
\begin{equation}
     V_{\text{pre}} = \alpha \cdot \frac{\mathcal{L}_\sigma [I]}{\max(\|\mathcal{L}_\sigma [I]\|_2, \epsilon)},
\end{equation}
where $\alpha$ is the expected structural strength and $\epsilon >0$ to avoid division by zero. Therefore, our final prior energy becomes:
\begin{equation}
    \mathcal{M}_{ST}(S, V_{\text{pre}}) = \tau \int_{\Omega} w \| \mathcal{L}_{\sigma}[S] - V_{\text{pre}} \|_{2}^{2} \,dx,
\end{equation}
where  $w(x) = \frac{1+u}{2}$ is soft foreground mask.
With the application of our proposed structural prior, we have significantly enhanced the model's ability to guide the segmentation along correct anatomical boundaries, particularly in images with severe low-contrast or high levels of noise. A key feature of our prior is its linearized nature. Consequently, the energy functional can be solved efficiently, enabling the rapid and robust recovery of both the reflectance component $S$ and the bias field $B$. As a result, the model provides a powerful tool for medical diagnosis.
By integrating the above prior constraint into our Retinex-based fitting framework, we arrive at the final energy functional. The energy is formulated as:
\begin{equation}\label{final}
\begin{aligned}
    E(S, B, u) &= \frac{1}{2} \int_{\Omega} (S - S^{\mathrm{FFR}}(u))^2 \,dx 
    + \frac{\lambda_I}{2} \int_{\Omega} (I - S - B)^2 \,dx \\
    &\quad + \frac{\alpha_B}{2} \int_{\Omega} \|\nabla B\|^2 \,dx 
    + \beta \int_{\Omega} \|\nabla S\| \,dx 
    + \theta \int_{\Omega} \|\nabla u\|^2 \,dx \\
    &\quad + \tau \int_{\Omega}  w(x) \| \mathcal{L}_\sigma [S] - V_{\text{pre}} \|_2^2 \,dx ,
\end{aligned}
\end{equation}
 The piecewise-constant approximation $S^{\text{FFR}}(u)$ is given by:
\[
    S^{\text{FFR}}(u)
    = \frac{c_1+c_2}{2}+\frac{c_1-c_2}{2}\,u,
\]
Utilizing the energy in Eq.~\eqref{final}, segmentation with bias correction is formulated as the following minimization problem:
\begin{equation}
    \min_{S,B,u}\; E(S,B,u).
    \label{eq:main-min_final}
\end{equation}

\subsection{Implementation of RefLSM}\label{section:C}

To solve the minimization problem in Eq.~\eqref{final}, we adopt the alternating direction
method of multipliers (ADMM) with a three-step splitting scheme. By introducing an
auxiliary variable $d=(d_x,d_y)$ to decouple the total variation term of the
reflectance $S$, the constrained problem can be reformulated as Eq.~\eqref{B}:
\begin{align}\label{B}
\min_{S,B,u,d}\;\;&
\frac{1}{2}\!\int_{\Omega}\!\big(S-S^{\mathrm{FFR}}(u)\big)^{2} dx
+ \frac{\lambda_I}{2}\!\int_{\Omega}\!(I-S-B)^{2} dx \notag\\
&+ \frac{\alpha_B}{2}\!\int_{\Omega}\!\|\nabla B\|^{2} dx
+ \theta \!\int_{\Omega}\!\|\nabla u\|^{2} dx
+ \beta \!\int_{\Omega}\!|d|\,dx \notag\\
&+ \tau \!\int_{\Omega}\! w(x)\,\big\| \mathcal{L}_{\sigma}[S] - V_{\mathrm{pre}}\big\|_{2}^{2}\, dx \notag,
\quad \text{s.t. } d=\nabla S .
\tag{19}
\end{align}
The augmented Lagrangian corresponding to Eq.~\eqref{B} is given in Eq. \ref{I}
\begin{align}\label{I}
\mathcal{L}(S,B,u,d;p)=&\;
\frac{1}{2}\!\int_{\Omega}\!\big(S - S^{\mathrm{FFR}}(u)\big)^{2} dx
+ \frac{\lambda_I}{2}\!\int_{\Omega}\!(I - S - B)^{2} dx \notag\\
&+ \frac{\alpha_B}{2}\!\int_{\Omega}\!\|\nabla B\|^{2} dx
+ \theta \!\int_{\Omega}\!\|\nabla u\|^{2} dx \notag\\
&+ \beta \!\int_{\Omega}\!|d|\,dx
+ \tau \!\int_{\Omega}\! w(x)\,\big\|\mathcal{L}_{\sigma}[S](x) - V_{\mathrm{pre}}(x)\big\|_{2}^{2}\, dx \notag\\
&+ \frac{\rho_1}{2}\!\int_{\Omega}\!\|d - \nabla S - p\|^{2} dx .
\tag{20}
\end{align}
Here $p$ denotes the scaled dual variable and $\rho_1>0$, $\theta>0$ are positive regularization parameters. 
Given the solution at iteration step $k$, including $S^k$, $B^k$, $u^k$, $d^k=(d_x^k,d_y^k)$, and the dual variable $p^k$, 
the variables are updated by alternating minimization.

We first update $c_1^{k+1}, c_2^{k+1}$ given $u^k$, $S^k$, $B^k$, $d^k$, and $p^k$. 
In particular, the region statistics $c_1^{k+1}$ and $c_2^{k+1}$ are computed according to Eq.~\eqref{eq:c1c2}.

\begin{equation}
c_1^{k+1}=\frac{\displaystyle\int_{\Omega} S ^kw^k\,dx}{\displaystyle\int_{\Omega} w^k\,dx+\epsilon},
\qquad
c_2^{k+1}=\frac{\displaystyle\int_{\Omega} S^k(1-w^k)\,dx}{\displaystyle\int_{\Omega} (1-w^k)\,dx+\epsilon}.
\label{eq:c1c2}
\tag{21}
\end{equation}
where $\epsilon>0, w^k= \frac{(1+u^k)}{2}$, next, we update $u$ by minimizing the fitting term with a quadratic smoothness on $u$. 

 When updating $u^{k+1}$ we fix 
$c_1^{k+1}, c_2^{k+1}$, $S^k$, $B^k$, $d^k$, $p^k$ and the soft weight $w^k=(1+u^k)/2$.We employ a two–step splitting strategy, where each step corresponds to one term of the energy. The evolution process can be described as follows. First, we define:
\begin{equation}
m^{k+1} = \tfrac{1}{2}(c_1^{k+1}+c_2^{k+1}), 
\quad
d^{k+1} = \tfrac{1}{2}(c_1^{k+1}-c_2^{k+1}),
\tag{22}
\end{equation}
so that, 
\begin{equation}
S^{\mathrm{FFR}}(u) = m^{k+1} + d^{k+1} u.
\tag{23}
\end{equation}
The reduced energy depending on $u$ is
\begin{equation}
E_u(u) 
= \tfrac{1}{2}\int_{\Omega} \big(S^k - m^{k+1} - d^{k+1} u\big)^2 dx
+ \tfrac{\theta}{2}\int_{\Omega} \|\nabla u\|^2 dx,
\quad u \in [-1,1].
\tag{24}
\end{equation}
The Euler--Lagrange equation is
\begin{equation}
d^{k+1}(m^{k+1}+d^{k+1}u-S^k) - \theta \Delta u = 0,
\tag{25}
\end{equation}
which can be rewritten as
\begin{equation}
(d^{k+1})^2 u - \theta \Delta u = d^{k+1}(S^k - m^{k+1}). \tag{$26$}
\end{equation}
Thus, two equivalent solution strategies are possible:
Under Neumann boundary conditions, Eq.~($23$) is linear. 
In the cosine domain, its solution is
\begin{equation}\label{Fe}
\widehat{u}^{k+1}(\xi) 
= \frac{d^{k+1}\,\big(\widehat{S^k}(\xi) - m^{k+1}\,\widehat{\delta}(\xi)\big)}
{(d^{k+1})^2 + \theta \lambda_N(\xi)}, \tag{$27$}
\end{equation}
where $\lambda_N(\xi)\geq 0$ are the eigenvalues of $-\Delta$ under Neumann boundary conditions. After inverse transform, we project onto the feasible set:
\begin{equation}\label{Fi}
u^{k+1} \leftarrow \mathrm{clip}(u^{k+1},-1,1). \tag{$28$}
\end{equation}
Thus, by combining the above three steps, the final update $u^{k+1}$ is obtained by combining Eqs.~\eqref{Fe}–\eqref{Fi}, ensuring both data fidelity and regularization while maintaining a binary level set representation.

Next, given $c_1^{k+1}$, $c_2^{k+1}$, $u^{k+1}$, $S^k$, $d^k$, and the dual variable $p^k$, we solve the quadratic subproblem for $B$ in Eq. \eqref{I}. Its Euler--Lagrange equation is shown in Eq. \eqref{D}, and the closed-form FFT solution is provided in Eq. \eqref{F}
Given \(S^k\) and parameters \(\varepsilon>0\), \(\alpha_B\ge 0\),
we solve the quadratic subproblem
\begin{equation}
B^{k+1} = \arg\min_{B}\;
\frac{\lambda_I}{2} \int_\Omega (I - S^k - B)^2 \, dx
+ \frac{\alpha_B}{2} \int_\Omega \|\nabla B\|^2 \, dx,
\tag{29}
\end{equation}
subject to homogeneous Neumann boundary conditions.
The Euler--Lagrange equation reads
\begin{equation}\label{D}
\varepsilon (B - (I-S^k)) - \alpha_B \Delta B = 0,
\quad
\partial_{\mathbf n} B = 0
\tag{30}
\end{equation}
Let \(\widehat{\Delta}(\xi)\le 0\) be the discrete Laplacian symbol in the Fourier domain.
Then the closed-form FFT solution is
\begin{equation}\label{F}
\widehat{B}^{\,k+1}(\xi) = \frac{\varepsilon \, \widehat{(I-S^k)}(\xi)}{\varepsilon - \alpha_B \, \widehat{\Delta}(\xi)}, 
\qquad
B^{k+1} = \mathcal{F}^{-1} \big( \widehat{B}^{\,k+1} \big)
\tag{31}
\end{equation}
Given $c_1^{k+1}$, $c_2^{k+1}$, $u^{k+1}$, $B^{k+1}$, $d^k$, and the dual variable $p^k$, the $S$–subproblem is obtained by collecting all terms in Eq.~\eqref{I} that depend on $S$: 
\begin{equation}\label{S-sub}
\begin{aligned}
S^{k+1} = \arg\min_{S}\;\;
 \frac{1}{2}\int_{\Omega} (S - S^{\mathrm{FFR}}(u^{k+1}))^2 dx 
 + \frac{\lambda_I}{2}\int_{\Omega} (I - S - B^{k+1})^2 dx 
& + \frac{\rho_1}{2}\int_{\Omega}\|d^k - \nabla S - p^k\|^2 dx \\
& + \tau \int_{\Omega} w^k \| \mathcal{L}_\sigma[S] - V_{\mathrm{pre}}\|_2^2 dx ,
\end{aligned}
\tag{32}
\end{equation}
where $w^{k+1}=(1+u^{k+1})/2$.
The corresponding Euler--Lagrange equation reads:
\begin{equation}\label{S-euler}
(1+\lambda_I)\,S +\rho_1\,\Delta S +2\tau\,\mathcal{L}_\sigma^*\!\big(w^{k+1}\,\mathcal{L}_\sigma[S]\big)
= S^{\mathrm{FFR}}(u^{k+1}) + \lambda_I(I-B^{k+1})+ \rho_1\big(\operatorname{div}d^k - \operatorname{div}p^k\big)
+ 2\tau\,\mathcal{L}_\sigma^*\!\big(w^{k+1} V_{\mathrm{pre}}\big),
\tag{33}
\end{equation}
where $\mathcal{L}_\sigma^*$ denotes the adjoint of $\mathcal{L}_\sigma$. 
Thus $S^{k+1}$ is obtained by solving the linear system
\begin{equation}\label{S-update}
\Big[(1+\lambda_I)I + \rho_1\Delta + 2\tau\,\mathcal{L}_\sigma^*(w^{k+1}\mathcal{L}_\sigma[\cdot])\Big] S^{k+1}
= S^{\mathrm{FFR}}(u^{k+1}) + \lambda_I(I-B^{k+1}) + \rho_1(\operatorname{div}d^k - \operatorname{div}p^k) + 2\tau\,\mathcal{L}_\sigma^*(w^{k+1} V_{\mathrm{pre}}).
\tag{34}
\end{equation}
In practice, Eq.~\eqref{S-update} can be solved either by conjugate gradient on the discretized operator, when $\mathcal{L}_\sigma$ and $\Delta$ are simultaneously diagonalizable.
By introducing the splitting variable $d = \nabla S$, we obtain the update problem:
\begin{equation}
d^{k+1} = \arg\min_d \;
\beta \int_\Omega \|d\| \, dx
+ \frac{\rho_1}{2}\int_\Omega \|d- \nabla S^{k+1} - p^k\|^2 dx.
\tag{35}
\end{equation}
This admits the standard shrinkage update:
\begin{equation}
d^{k+1} 
= \operatorname{shrink}\!\left(\nabla S^{k+1} + p^k, \; \tfrac{\beta}{\rho_1}\right),
\tag{36}
\end{equation}
where,
\begin{equation}
\operatorname{shrink}(z,\lambda) = \max\!\left(1-\frac{\lambda}{\|z\|_2},0\right) z .
\tag{36}
\end{equation}
The dual variable $p$ is updated in the standard ADMM form:
\begin{equation}
p^{k+1} = p^k + \nabla S^{k+1} - d^{k+1}.
\tag{37}
\end{equation}
In summary, we present the implementation of the RefLSM framework in Algorithm 1:
\begin{algorithm}[H]
\caption{Compact Implementation of RefLSM with Retinex and Structural Prior}\label{alg:RefLSM_compact}
\begin{algorithmic}[1]
\STATE \textbf{Input:} Image $I$, parameters $\varepsilon$, $\alpha_B$, $\beta$, $\lambda_I$, $\theta$, $\tau$, $\rho_1$, tolerance $\delta$, max iter $K$.
\STATE \textbf{Init:} $S^0 = I$, $B^0 = 0$, $u^0 = \mathrm{sign}(I)$, $d^0 = 0$, $p^0 = 0$; precompute $\mathcal{L}_\sigma$, $V_{\mathrm{pre}}$.
\FOR{$k=0$ \textbf{to} $K$}
    \STATE \textbf{(a) Region stats:} 
    $w^k = (1+u^k)/2$,  
    $c_1^{k+1} = \frac{\int S^k w^k}{\int w^k + \epsilon}$,  
    $c_2^{k+1} = \frac{\int S^k (1-w^k)}{\int (1-w^k)+\epsilon}$,  
    $S^{\mathrm{FFR}} = \frac{c_1+c_2}{2} + \frac{c_1-c_2}{2} u^k$.
    \STATE \textbf{(b) Update $u$:} Solve $E_u(u) = \frac12\|S^k - S^{\mathrm{FFR}}\|^2 + \frac{\theta}{2}\|\nabla u\|^2$, then clip $u^{k+1} \in [-1,1]$.
    \STATE \textbf{(c) Update $B$:} $B^{k+1} = \arg\min_B \frac{\lambda_I}{2}\|I-S^k-B\|^2 + \frac{\alpha_B}{2}\|\nabla B\|^2$ (FFT solution).
    \STATE \textbf{(d) Update $S$:} Solve $[(1+\lambda_I)I + \rho_1\Delta + 2\tau \mathcal{L}_\sigma^* w^k \mathcal{L}_\sigma] S^{k+1} = S^{\mathrm{FFR}} + \lambda_I(I-B^{k+1}) + \rho_1(\operatorname{div} d^k - \operatorname{div} p^k) + 2\tau \mathcal{L}_\sigma^*(w^k V_{\mathrm{pre}})$.
    \STATE \textbf{(e) Update $d$:} $d^{k+1} = \operatorname{shrink}(\nabla S^{k+1} + p^k, \beta/\rho_1)$.
    \STATE \textbf{(f) Update $p$:} $p^{k+1} = p^k + \nabla S^{k+1} - d^{k+1}$.
   \IF{$\|u^{k+1}-u^k\| < \delta$}
   \STATE \textbf{break}
   \ENDIF
\ENDFOR
\STATE \textbf{Output:} $u$, $S$, $B$.
\end{algorithmic}
\end{algorithm}

\section{Experimental results}\label{section:D}
In this section, we present comprehensive experimental studies. All experiments were conducted on a device equipped with an AMD Ryzen 7 4800U 1.8 GHz CPU, Radeon Graphics, and 16 GB of RAM. The models were implemented using Matlab 2024a on a Windows 11 operating system. 
\begin{figure*}[!ht]
	\centering
	\subfloat{\rotatebox{0}{(a)}
		\includegraphics[width=0.7in]{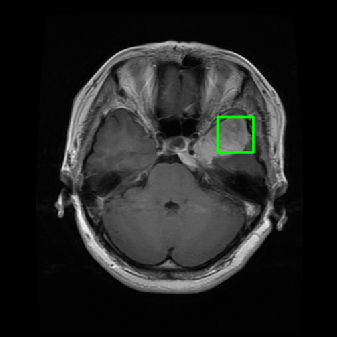}%
		\label{fig_11_case}}\vspace{-3.5mm}\hspace{-1.5mm}
	\subfloat{\includegraphics[width=0.7in]{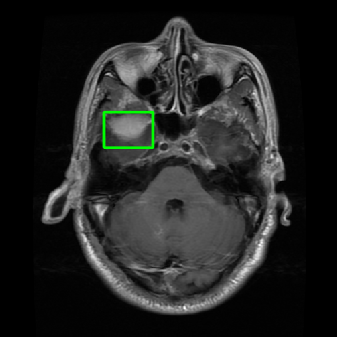}%
		\label{fig_12_case}}\hspace{-1.5mm}
	\subfloat{\includegraphics[width=0.7in]{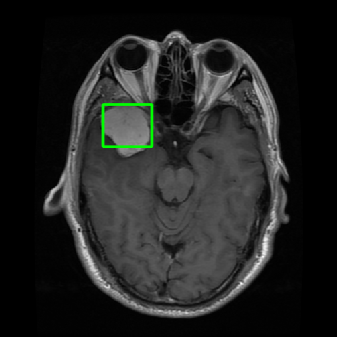}%
		\label{fig_13_case}}\hspace{-1.5mm}
	\subfloat{\includegraphics[width=0.7in]{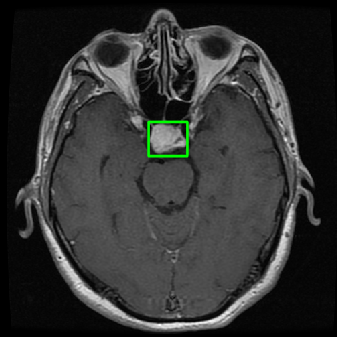}%
		\label{fig_14_case}}\hspace{-1.5mm}
	\subfloat{\includegraphics[width=0.7in]{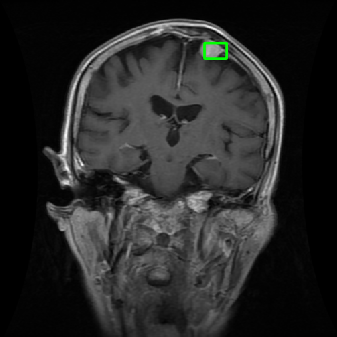}%
		\label{fig_15_case}}\hspace{-1.5mm}
	\subfloat{\includegraphics[width=0.7in]{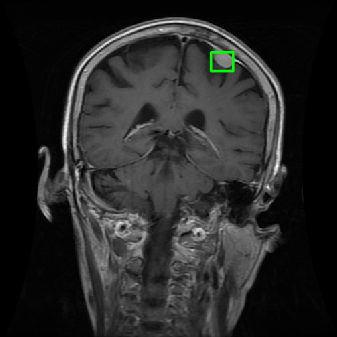}%
		\label{fig_16_case}}\hspace{-1.5mm}
	\subfloat{\includegraphics[width=0.7in]{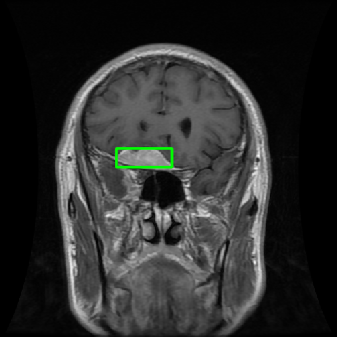}%
		\label{fig_17_case}}\hspace{-1.5mm}
	\subfloat{\includegraphics[width=0.7in]{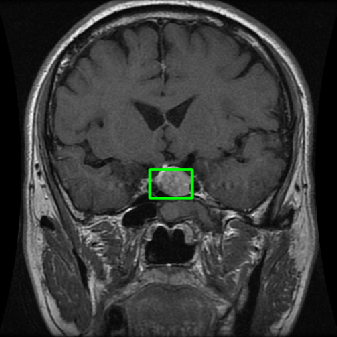}%
		\label{fig_18_case}}\hspace{-1.5mm}
	\hfil	
	\subfloat{\rotatebox{0}{(b)}
		\includegraphics[width=0.7in]{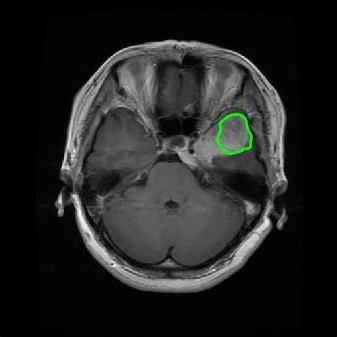}%
		\label{fig_19_case}}\vspace{-3.5mm}\hspace{-1.5mm}
	\subfloat{\includegraphics[width=0.7in]{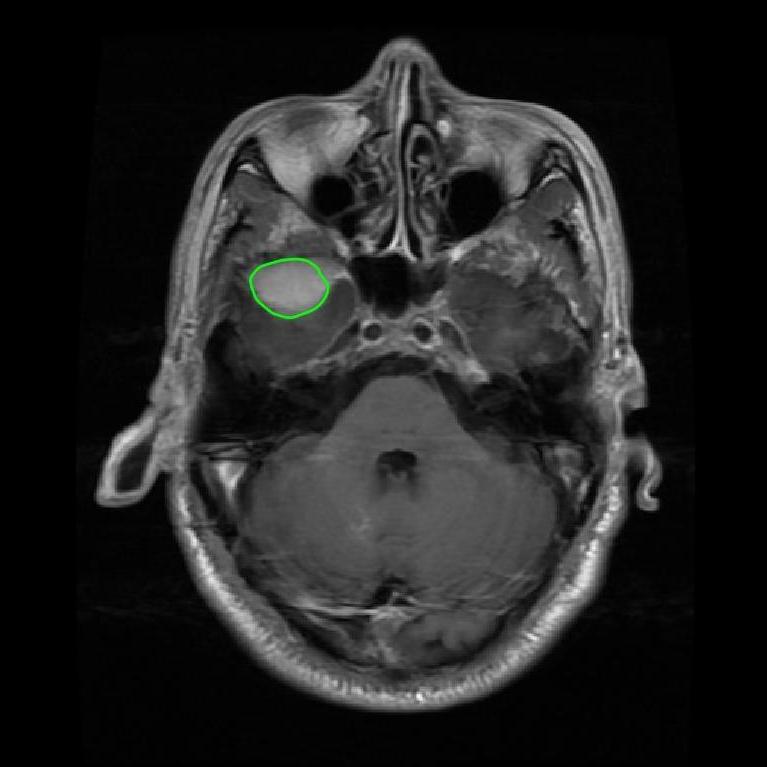}%
		\label{fig_20_case}}\hspace{-1.5mm}
	\subfloat{\includegraphics[width=0.7in]{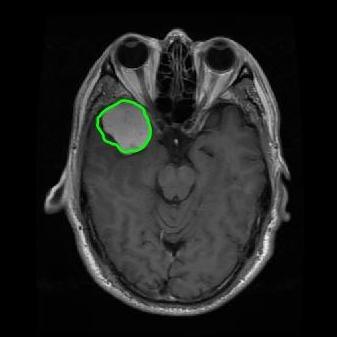}%
		\label{fig_21_case}}\hspace{-1.5mm}
	\subfloat{\includegraphics[width=0.7in]{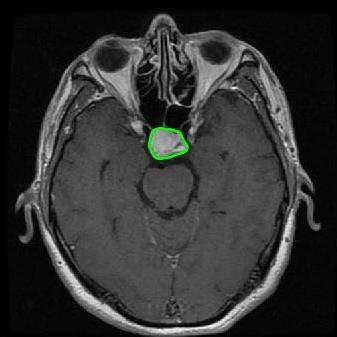}%
		\label{fig_22_case}}\hspace{-1.5mm}
	\subfloat{\includegraphics[width=0.7in]{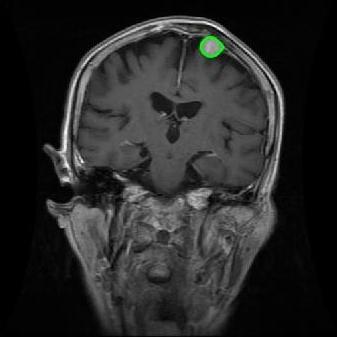}%
		\label{fig_23_case}}\hspace{-1.5mm}
	\subfloat{\includegraphics[width=0.7in]{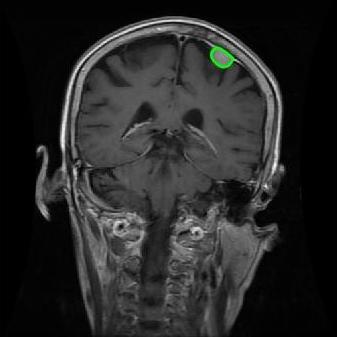}%
		\label{fig_24_case}}\hspace{-1.5mm}
	\subfloat{\includegraphics[width=0.7in]{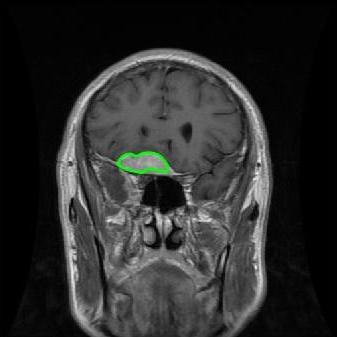}%
		\label{fig_25_case}}\hspace{-1.5mm}
	\subfloat{\includegraphics[width=0.7in]{ 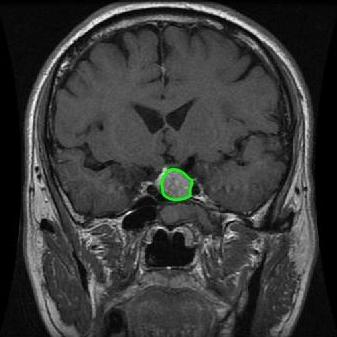}%
		\label{fig_26_case}}\hspace{-1.5mm}
	\hfil
	\subfloat{\rotatebox{0}{(c)}
		\includegraphics[width=0.7in]{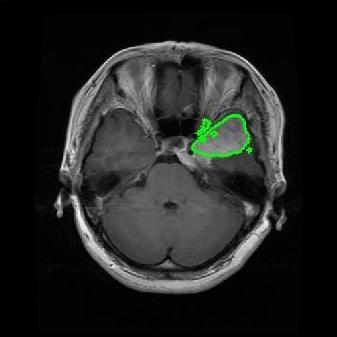}%
		\label{fig_27_case}}\vspace{-3.5mm}\hspace{-1.5mm}
	\subfloat{\includegraphics[width=0.7in]{ 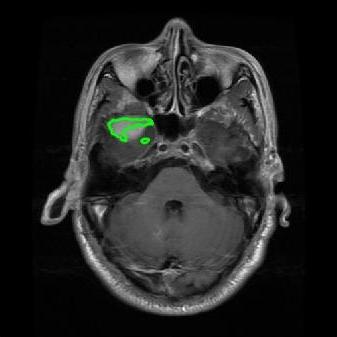}%
		\label{fig_28_case}}\hspace{-1.5mm}
	\subfloat{\includegraphics[width=0.7in]{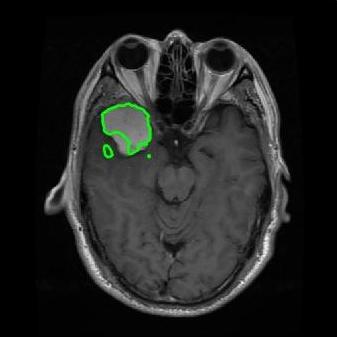}%
		\label{fig_29_case}}\hspace{-1.5mm}
	\subfloat{\includegraphics[width=0.7in]{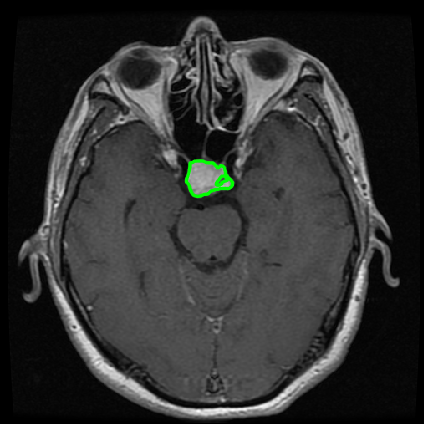}%
		\label{fig_30_case}}\hspace{-1.5mm}
	\subfloat{\includegraphics[width=0.7in]{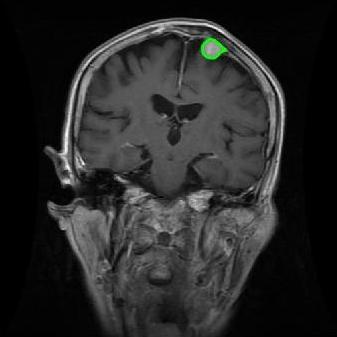}%
		\label{fig_31_case}}\hspace{-1.5mm}
	\subfloat{\includegraphics[width=0.7in]{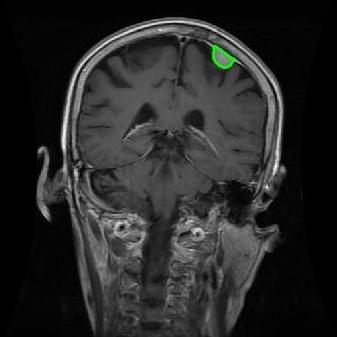}%
		\label{fig_32_case}}\hspace{-1.5mm}
	\subfloat{\includegraphics[width=0.7in]{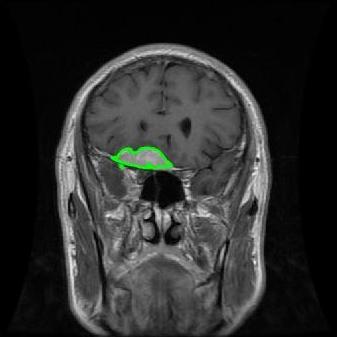}%
		\label{fig_33_case}}\hspace{-1.5mm}
	\subfloat{\includegraphics[width=0.7in]{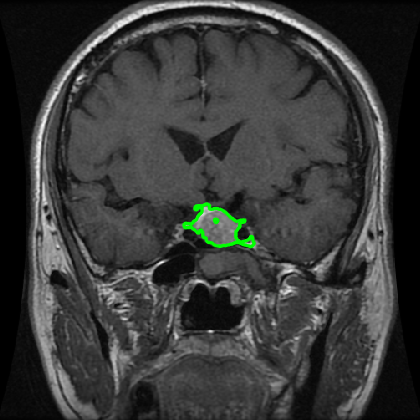}%
		\label{fig_34_case}}\hspace{-1.5mm}
	\hfil
	\subfloat{\rotatebox{0}{(d)}
		\includegraphics[width=0.7in]{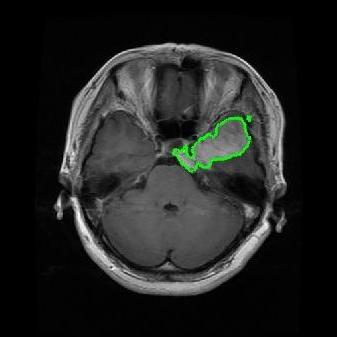}%
		\label{fig_35_case}}\vspace{-3.5mm}\hspace{-1.5mm}
	\subfloat{\includegraphics[width=0.7in]{ 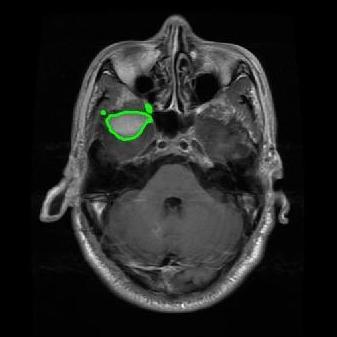}%
		\label{fig_36_case}}\hspace{-1.5mm}
	\subfloat{\includegraphics[width=0.7in]{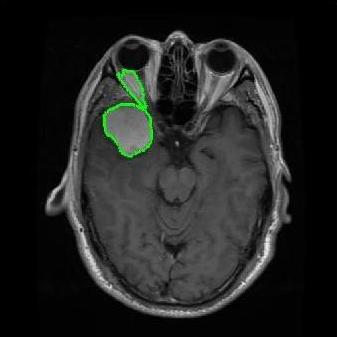}%
		\label{fig_37_case}}\hspace{-1.5mm}
	\subfloat{\includegraphics[width=0.7in]{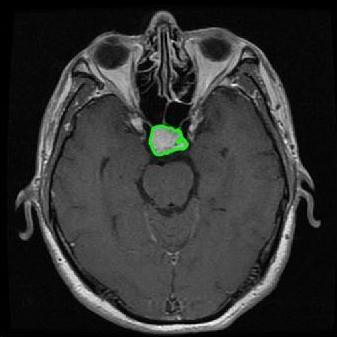}%
		\label{fig_38_case}}\hspace{-1.5mm}
	\subfloat{\includegraphics[width=0.7in]{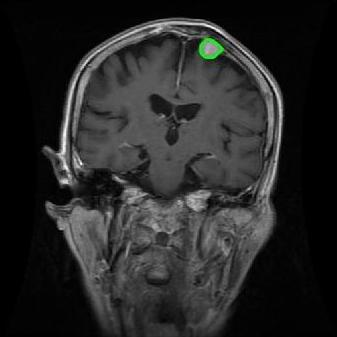}%
		\label{fig_39_case}}\hspace{-1.5mm}
	\subfloat{\includegraphics[width=0.7in]{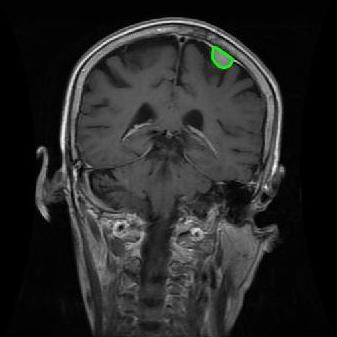}%
		\label{fig_40_case}}\hspace{-1.5mm}
	\subfloat{\includegraphics[width=0.7in]{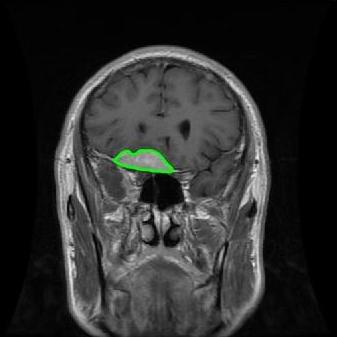}%
		\label{fig_41_case}}\hspace{-1.5mm}
	\subfloat{\includegraphics[width=0.7in]{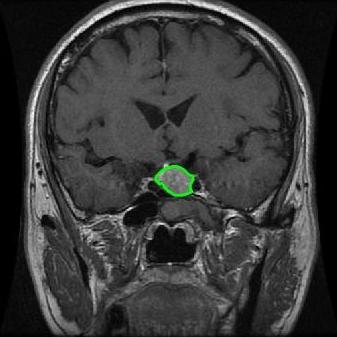}%
		\label{fig_42_case}}\hspace{-1.5mm}
	\hfil
	
	\subfloat{\rotatebox{0}{(e)}
		\includegraphics[width=0.7in]{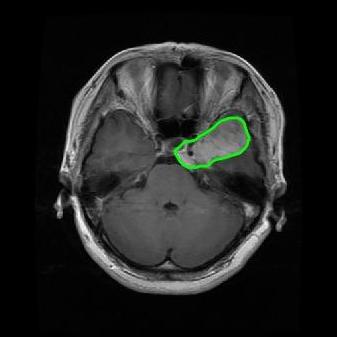}%
		\label{fig_43_case}}\vspace{-3.5mm}\hspace{-1.5mm}
	\subfloat{\includegraphics[width=0.7in]{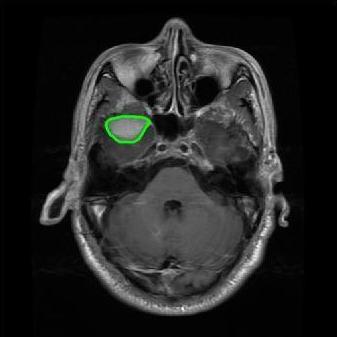}%
		\label{fig_44_case}}\hspace{-1.5mm}
	\subfloat{\includegraphics[width=0.7in]{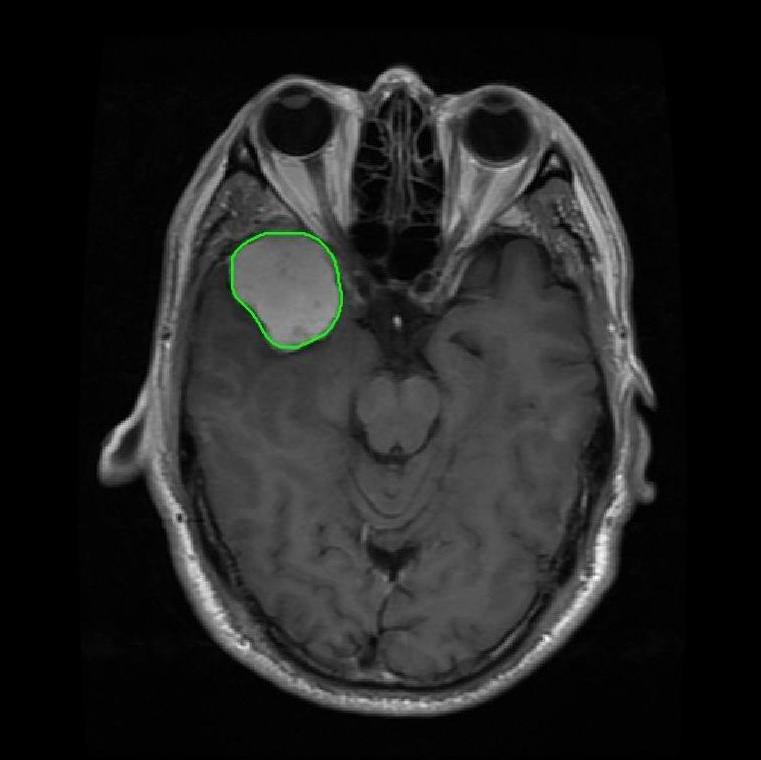}%
		\label{fig_45_case}}\hspace{-1.5mm}
	\subfloat{\includegraphics[width=0.7in]{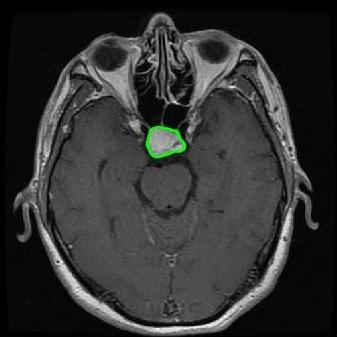}%
		\label{fig_46_case}}\hspace{-1.5mm}
	\subfloat{\includegraphics[width=0.7in]{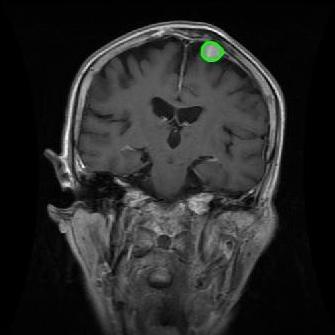}%
		\label{fig_47_case}}\hspace{-1.5mm}
	\subfloat{\includegraphics[width=0.7in]{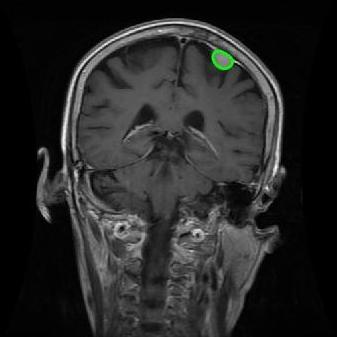}%
		\label{fig_48_case}}\hspace{-1.5mm}
	\subfloat{\includegraphics[width=0.7in]{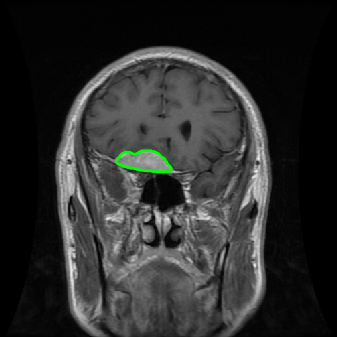}%
		\label{fig_49_case}}\hspace{-1.5mm}
	\subfloat{\includegraphics[width=0.7in]{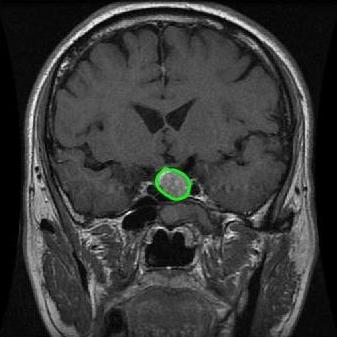}%
		\label{fig_50_case}}\hspace{-1.5mm}
	\hfil
	\subfloat{\rotatebox{0}{(f)}
		\includegraphics[width=0.7in]{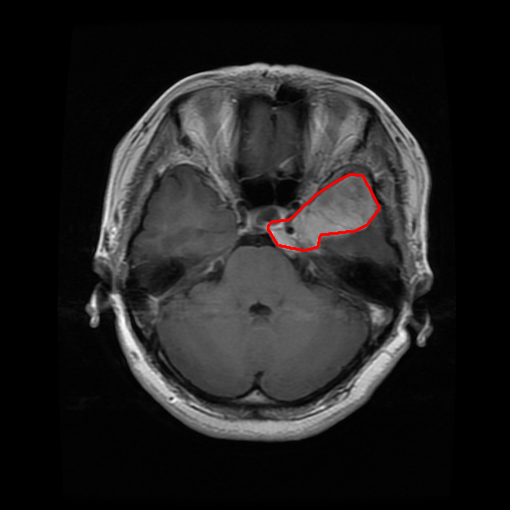}%
		\label{fig_51_case}}\hspace{-1.5mm}
	\subfloat{\includegraphics[width=0.7in]{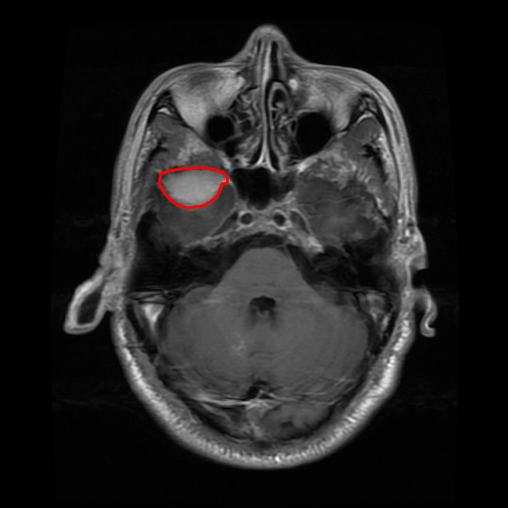}%
		\label{fig_52_case}}\hspace{-1.5mm}
	\subfloat{\includegraphics[width=0.7in]{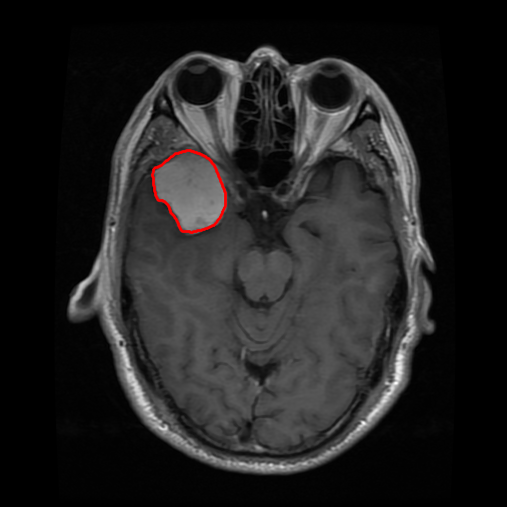}%
		\label{fig_53_case}}\hspace{-1.5mm}
	\subfloat{\includegraphics[width=0.7in]{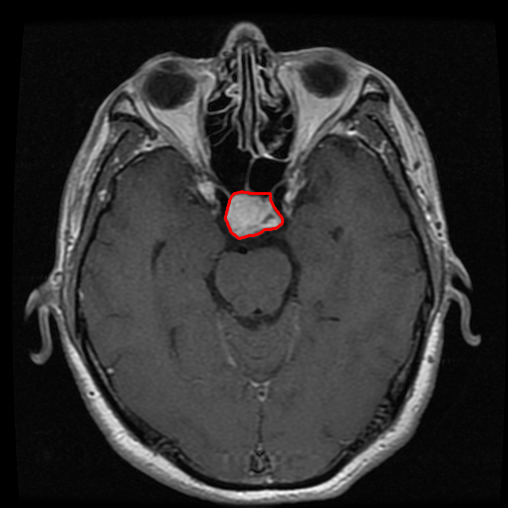}%
		\label{fig_54_case}}\hspace{-1.5mm}
	\subfloat{\includegraphics[width=0.7in]{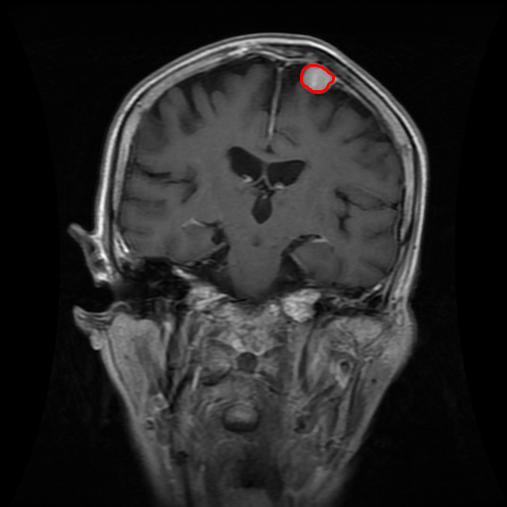}%
		\label{fig_55_case}}\hspace{-1.5mm}
	\subfloat{\includegraphics[width=0.7in]{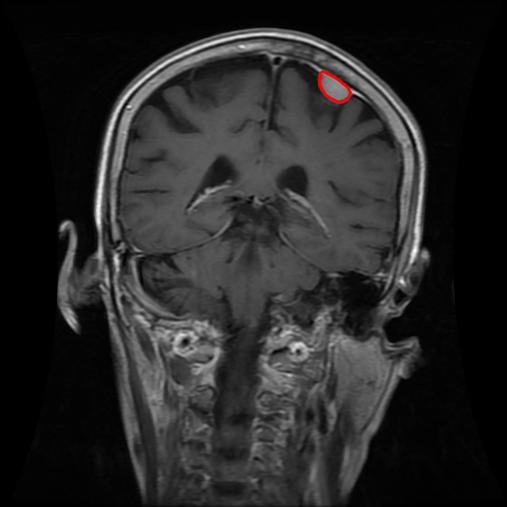}%
		\label{fig_56case}}\hspace{-1.5mm}
	\subfloat{\includegraphics[width=0.7in]{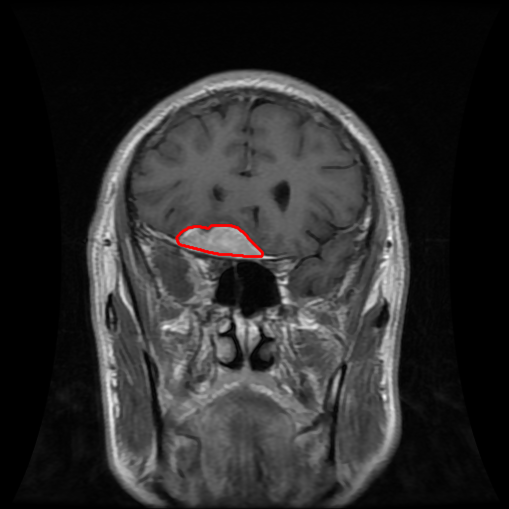}%
		\label{fig_57_case}}\hspace{-1.5mm}
	\subfloat{\includegraphics[width=0.7in]{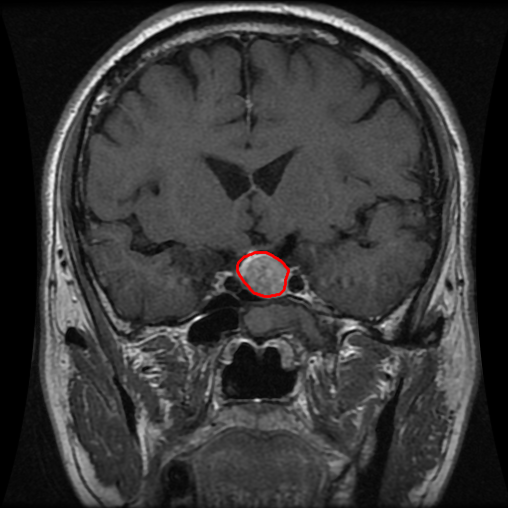}%
		\label{fig_58_case}}\hspace{-1.5mm}
	\hfil
	\caption{Segmentation results from evaluated models for brain tumor MR images. Row 1:Original images and initial contours. Row 2-4: Results from the RESLS model, ALF model, L1 model, and the RefLSM. Row 6: Ground truth.}
	\label{img3}
	
\end{figure*}
\subsection{Evaluation Criteria and Parameters}
To comprehensively evaluate the performance of our proposed model, RefLSM, we conducted extensive experiments on multiple medical image datasets.
We set a maximum of $K_{\max}=30$ iterations, with an early stopping criterion when the relative change of the main variables between iterations fell below a threshold of $\delta=10^{-4}$. Unless otherwise stated, the model parameters were consistently set as follows: the Retinex data-fidelity weight $\lambda_I=1.0$, the bias-field smoothness weight $\alpha_B=15$, the reflectance TV weight $\beta=0.02$, and the level-set regularization weight $\theta=0.1$. For our linearized structural prior, the main weight was set to $\tau=0.5$, with an internal Gaussian scale of $\sigma=3$ and a desired structural magnitude of $\alpha=0.1$. The ADMM penalty parameter was set to $\rho_1=1.0$. In practice, we found the numerical performance of RefLSM to be relatively insensitive to the choice of $\rho_1$ and $\theta$ within a broad range, while the parameters $\tau$ and $\sigma$ primarily controlled the strength and scale of the structural guidance. For certain images, we will employ tailored parameters that align with specific characteristics of the images used. To quantify the segmentation results, we employed two key evaluation metrics: the Dice coefficient and Precision:
\begin{equation}
	\begin{split}
		Dice=\frac{2*TP}{TP+FP+TP+FN},
	\end{split}
\end{equation}
\begin{equation}
	\begin{split}
		Precision=\frac{TP}{TP+FP}.
	\end{split}
\end{equation}
Here, TP (true positive) represents the areas correctly identified as belonging to a specific class, TN (true negative) signifies the regions correctly recognized as not belonging to that class, FP (false positive) indicates areas erroneously identified as part of the class, and FN (false negative) captures the actual areas that belong to the class but were not detected. The Dice coefficient is a well-established metric for gauging the similarity between two sets and is widely utilized in image segmentation evaluation. Additionally, the precision can be used to effectively measure the purity of positive detections in comparison to the actual ground truth.
We also used the relative Tenengrad (RTG) \cite{CAI2023109195} to assess the performance of our model with different $\tau$ for bias field segmentation. RTG is defined as:
\begin{equation}
	\begin{split}
		R T G=\frac{1}{P Q} \sum_{u=1}^{P} \sum_{v=1}^{Q} \frac{\sqrt{\left(\frac{\partial I(u, v)}{\partial u}\right)^{2}+\left(\frac{\partial I(u, v)}{\partial v}\right)^{2}}}{I(u, v)},
	\end{split}
\end{equation}
here I represents an image, $P\times Q$ indicates the image size, $(u,v)$ denotes the pixel locations. The RTG is a commomly used non-reference metric for assessing image quality, specially focusing on the sharpness and clarity of images. By calculating the gradients in both the $u$ and $v$ directions, we can then compute the sum of the squares of these gradients to quantify the sharpness.

\begin{figure*}[h]
	\centering
	\subfloat{\includegraphics[width=0.8in]{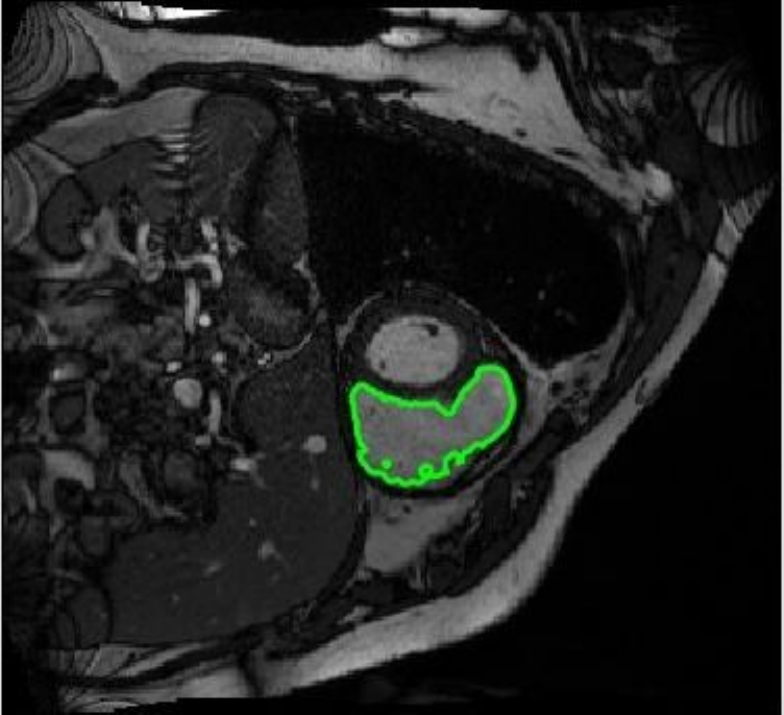}%
		\label{fig_61_case}}\vspace{-3mm}\hspace{-1.5mm}
	\subfloat{\includegraphics[width=0.8in]{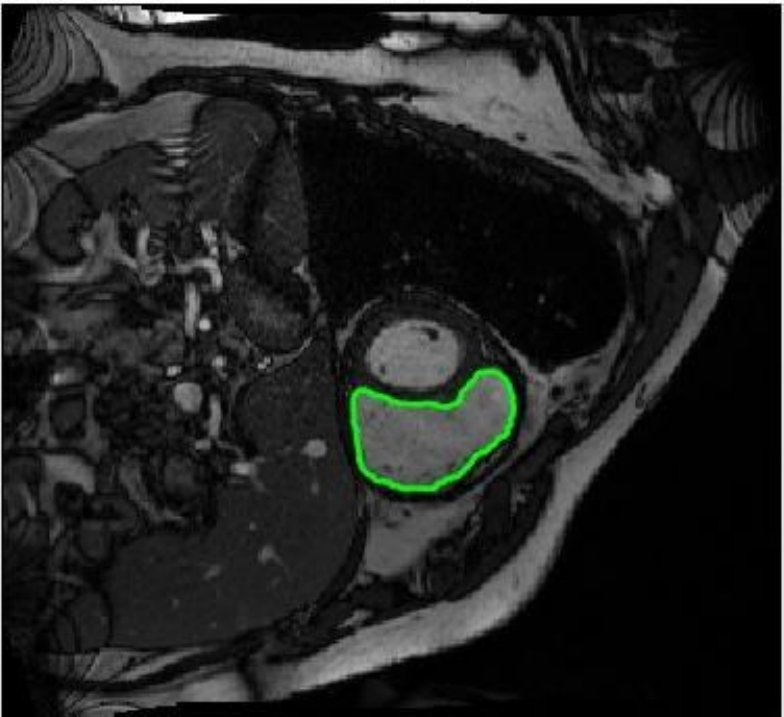}%
		\label{fig_62_case}}
	\subfloat{\includegraphics[width=0.8in]{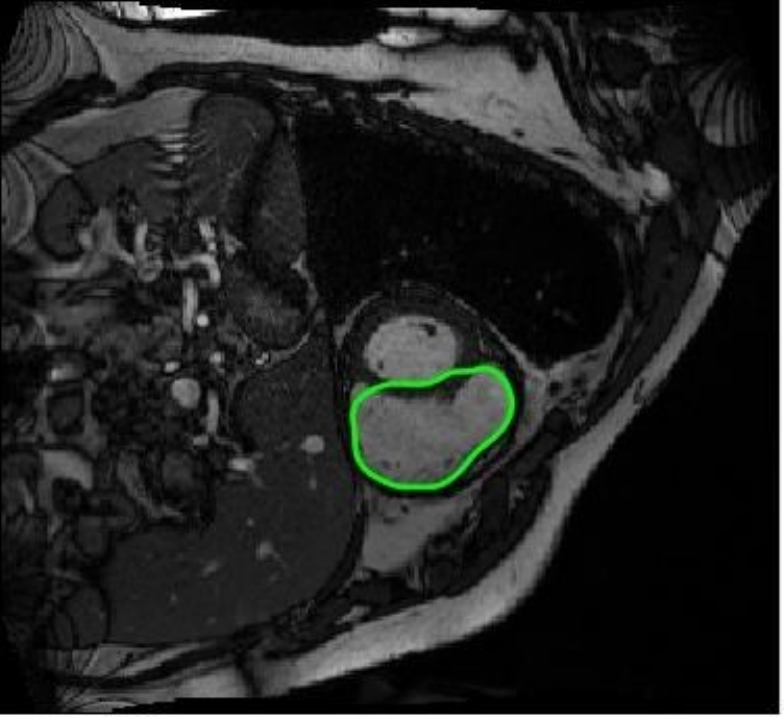}%
		\label{fig_63_case}}
	\subfloat{\includegraphics[width=0.8in]{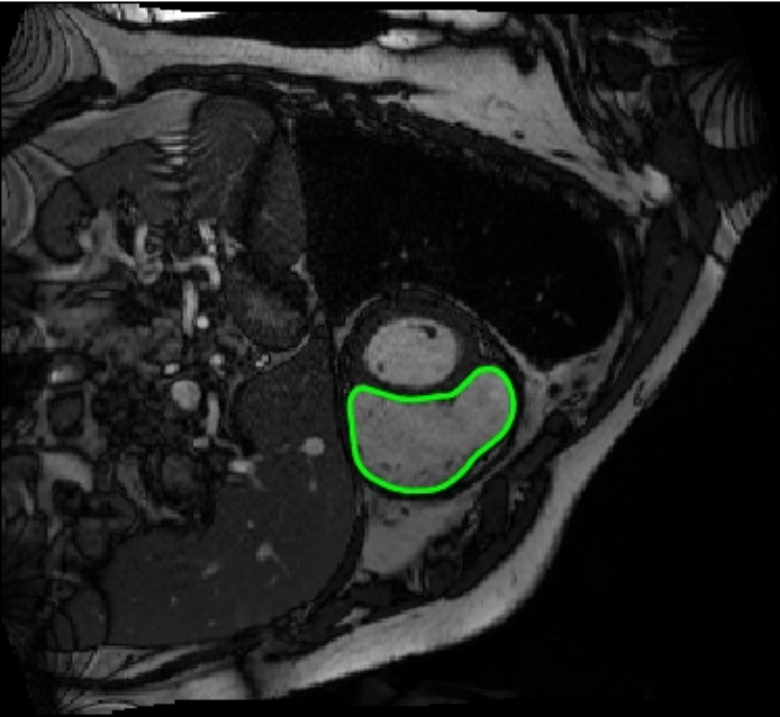}%
		\label{fig_64_case}}
	\subfloat{\includegraphics[width=0.8in]{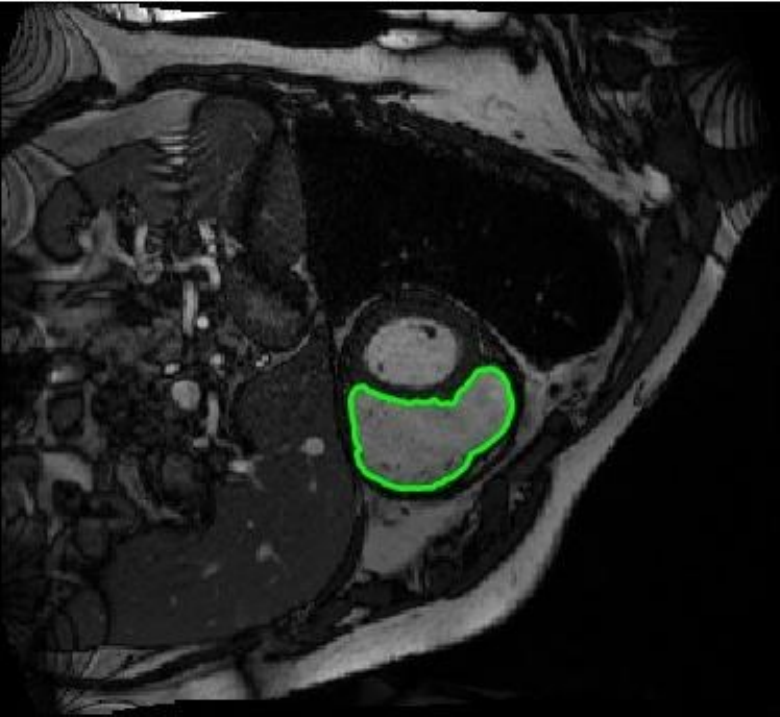}%
		\label{fig_65_case}}
	\subfloat{\includegraphics[width=0.8in]{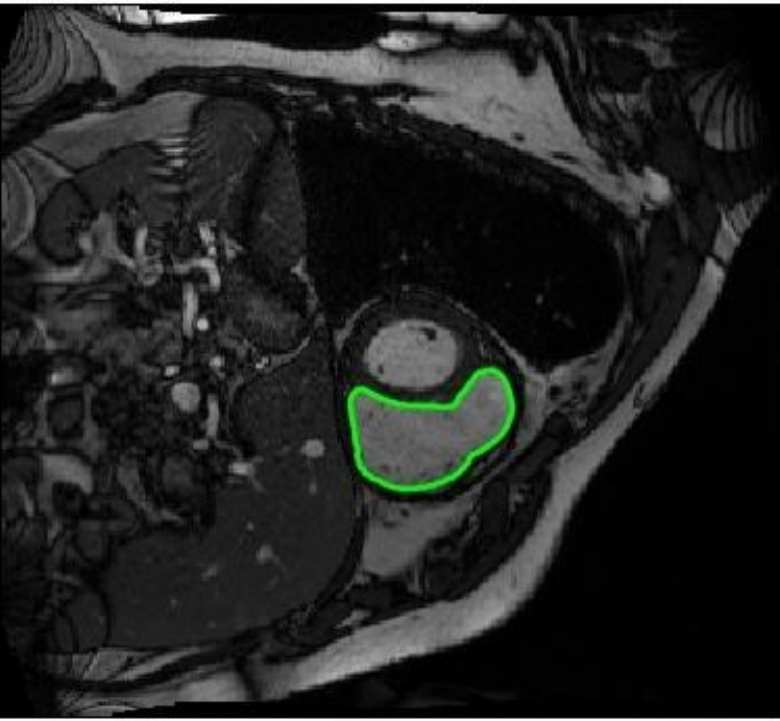}%
		\label{fig_66_case}}
	\subfloat{\includegraphics[width=0.8in]{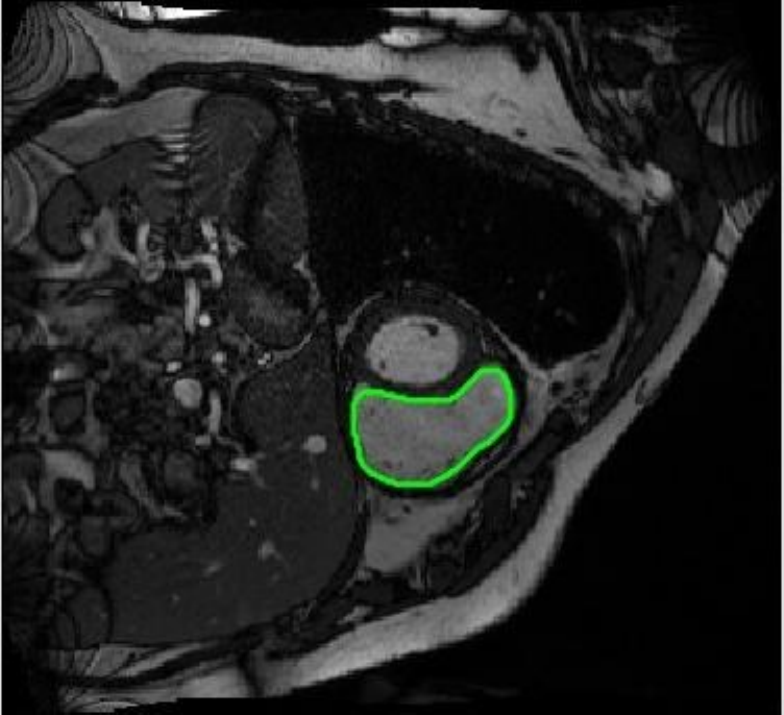}%
		\label{fig_67_case}}
	\subfloat{\includegraphics[width=0.8in]{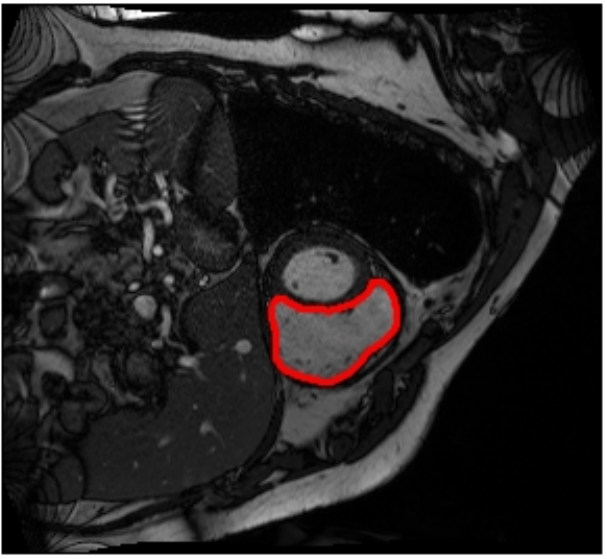}%
		\label{fig_68_case}}
	\hfil
	\subfloat{\includegraphics[width=0.8in]{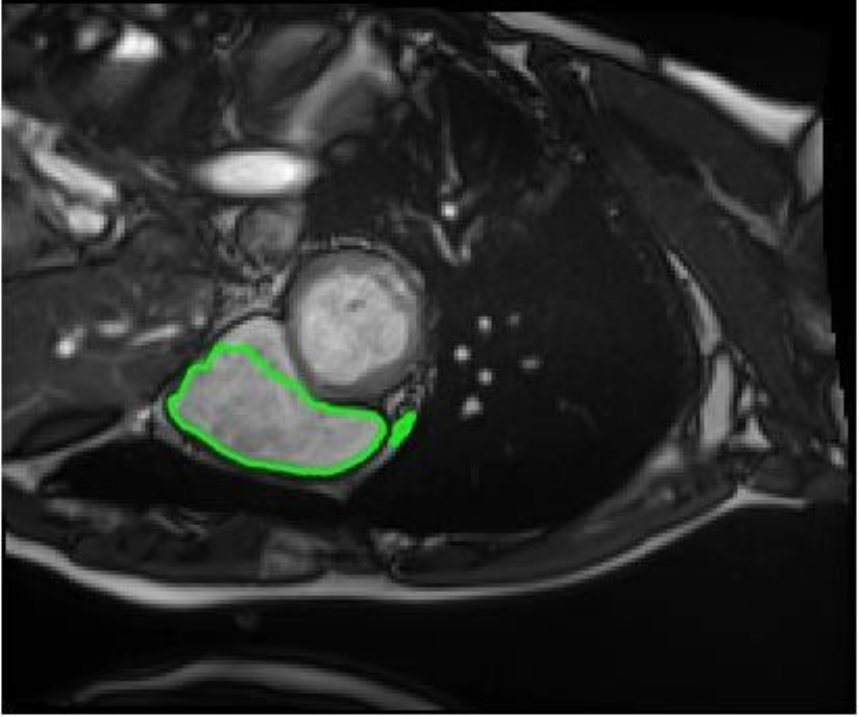}%
		\label{fig_69_case}}\vspace{-3mm}\hspace{-1.5mm}
	\subfloat{\includegraphics[width=0.8in]{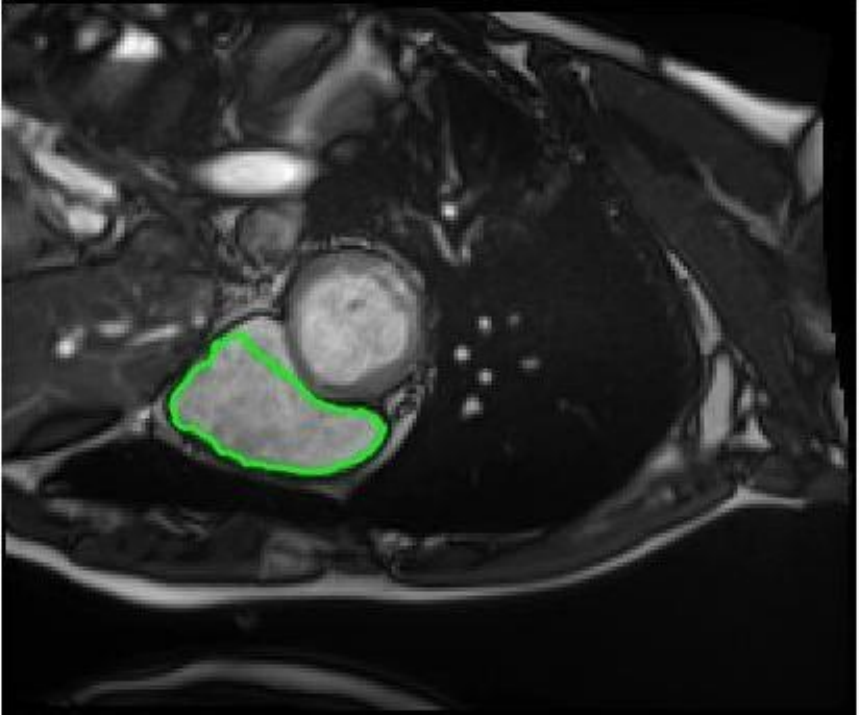}%
		\label{fig_70_case}}
	\subfloat{\includegraphics[width=0.8in]{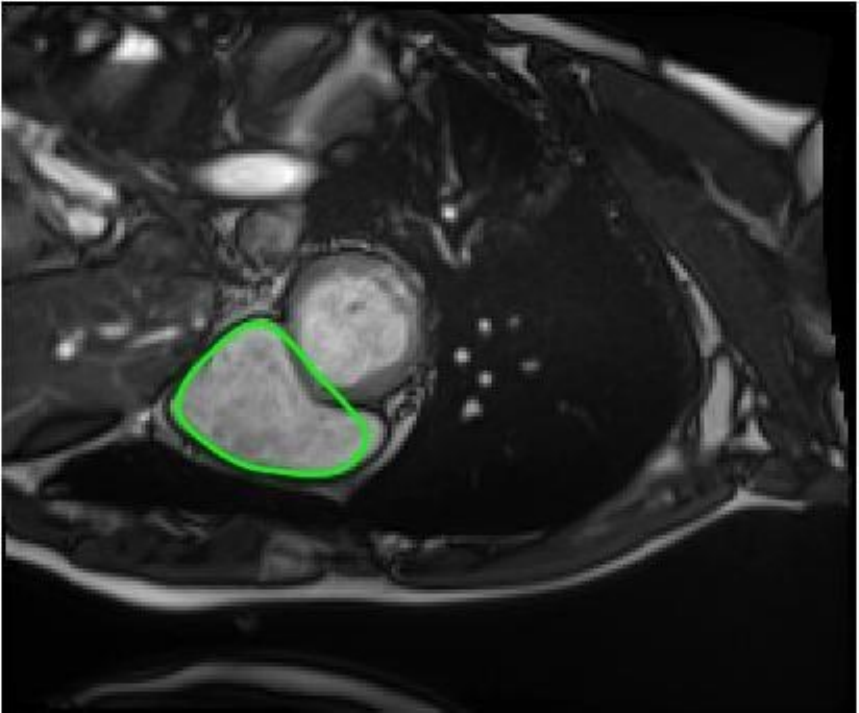}%
		\label{fig_71_case}}
	\subfloat{\includegraphics[width=0.795in]{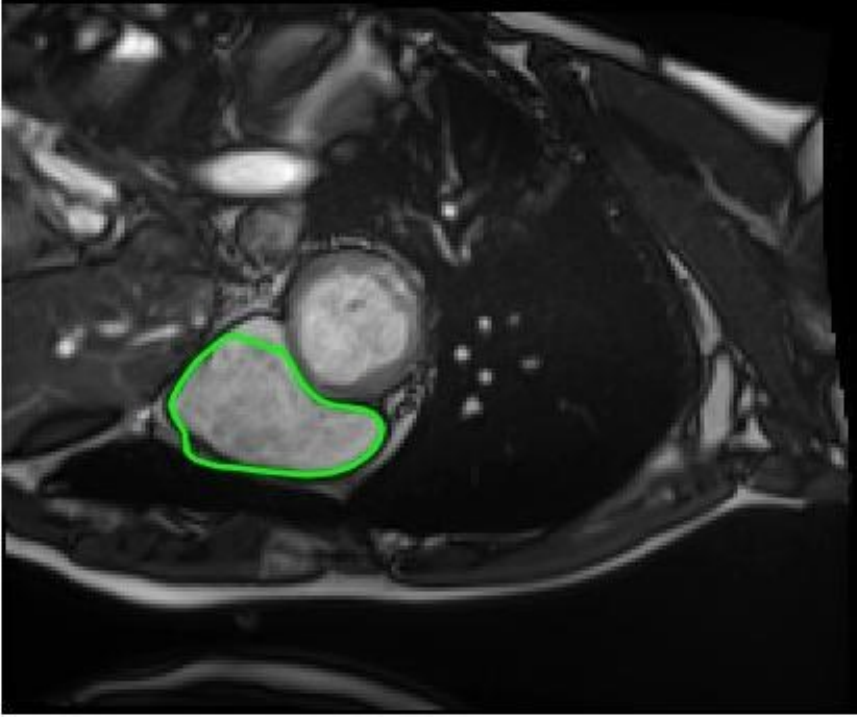}%
		\label{fig_72_case}}
	\subfloat{\includegraphics[width=0.8in]{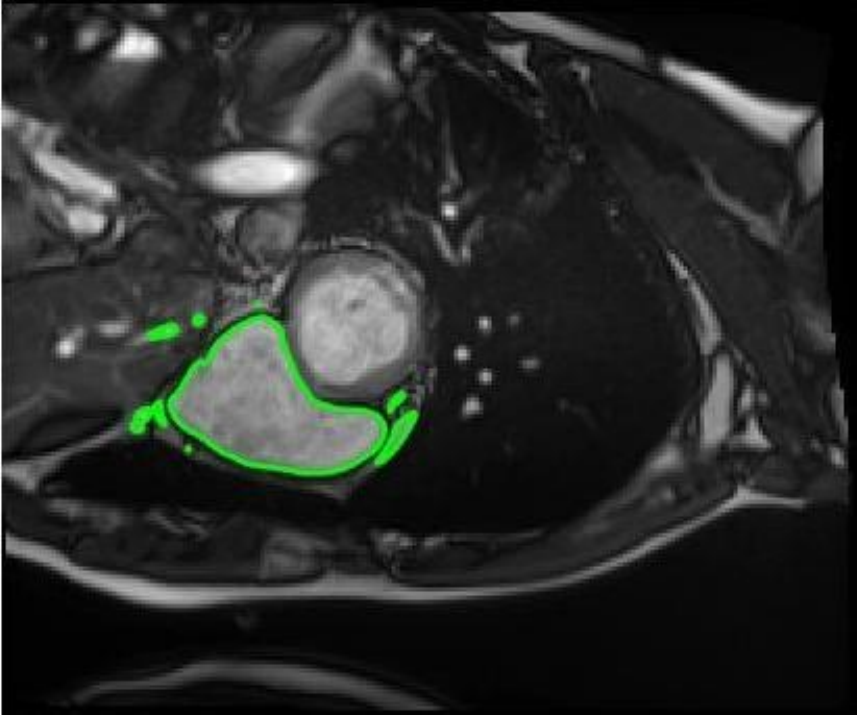}%
		\label{fig_73_case}}
	\subfloat{\includegraphics[width=0.8in]{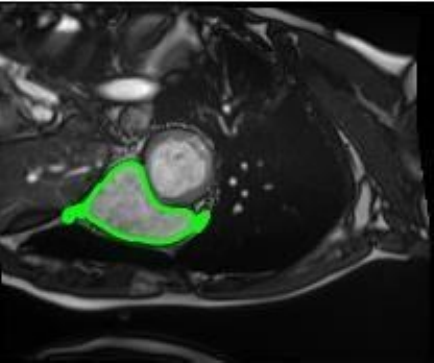}%
		\label{fig_74_case}}
	\subfloat{\includegraphics[width=0.8in]{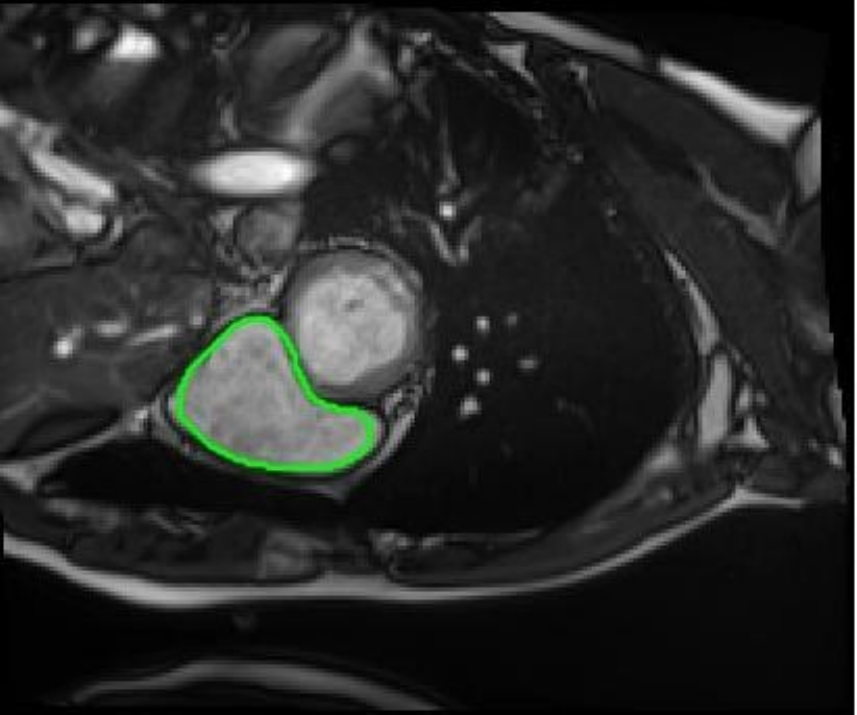}%
		\label{fig_75_case}}
	\subfloat{\includegraphics[width=0.8in]{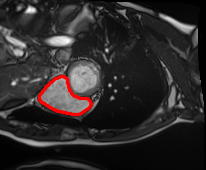}%
		\label{fig_76_case}}
	\hfil
	\subfloat{\includegraphics[width=0.8in]{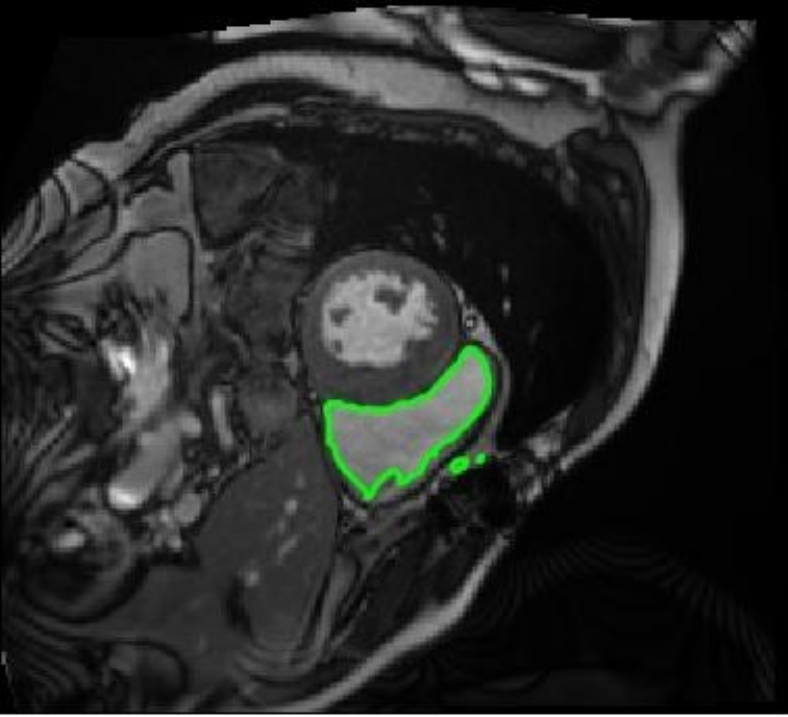}%
		\label{fig_77_case}}\vspace{-3mm}\hspace{-1.5mm}
	\subfloat{\includegraphics[width=0.8in]{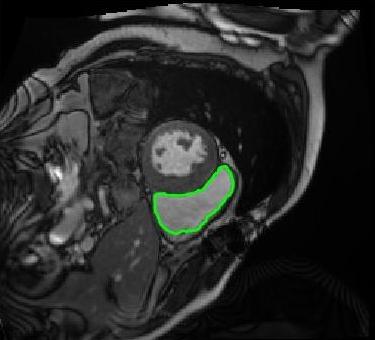}%
		\label{fig_82_case}}
	\subfloat{\includegraphics[width=0.795in]{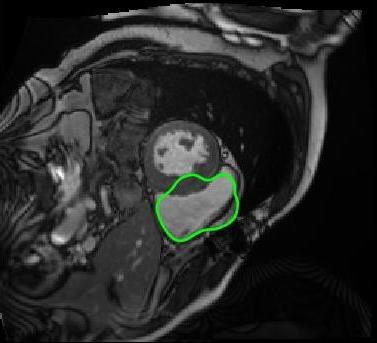}%
		\label{fig_78_case}}
	\subfloat{\includegraphics[width=0.8in]{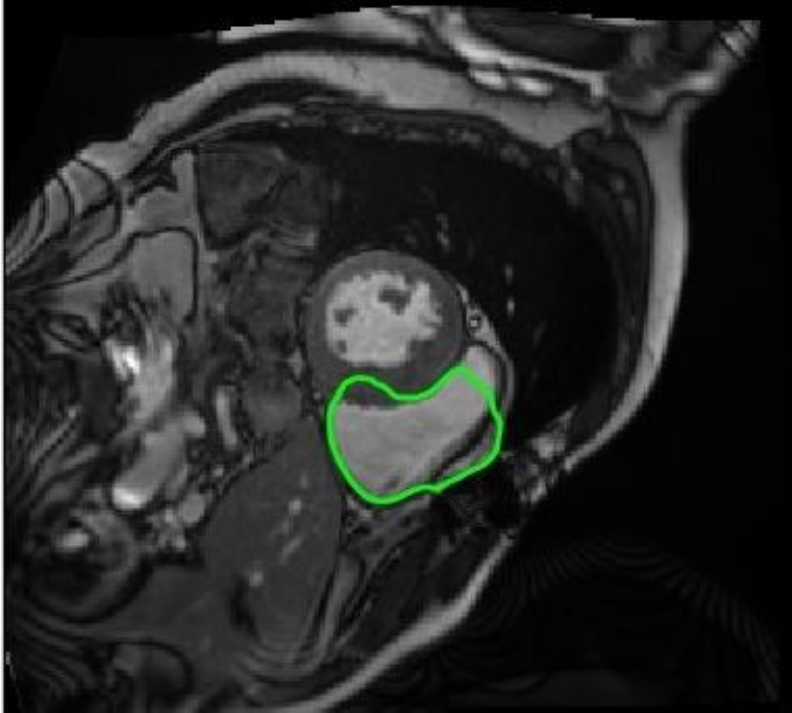}%
		\label{fig_79_case}}
	\subfloat{\includegraphics[width=0.795in]{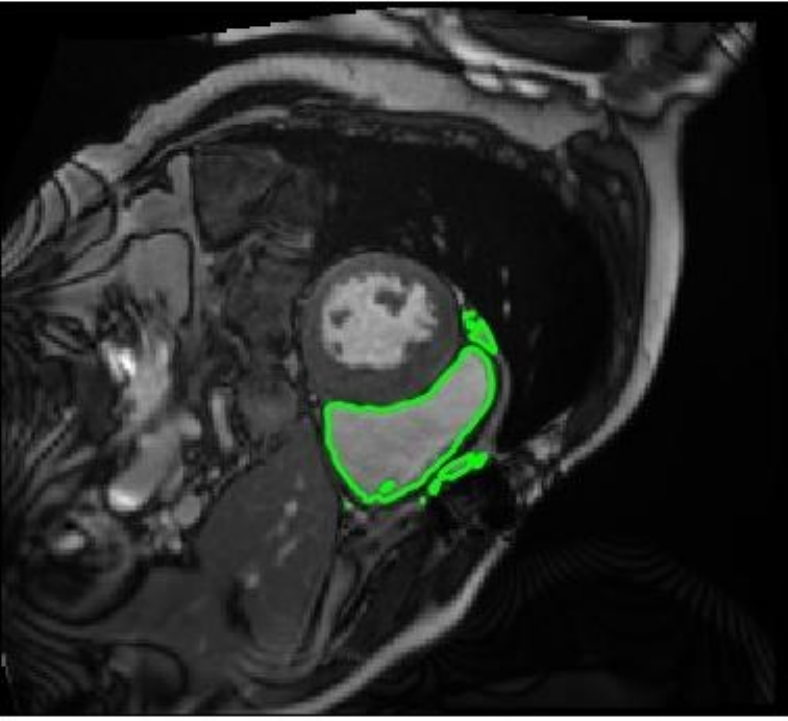}%
		\label{fig_80_case}}
	\subfloat{\includegraphics[width=0.795in]{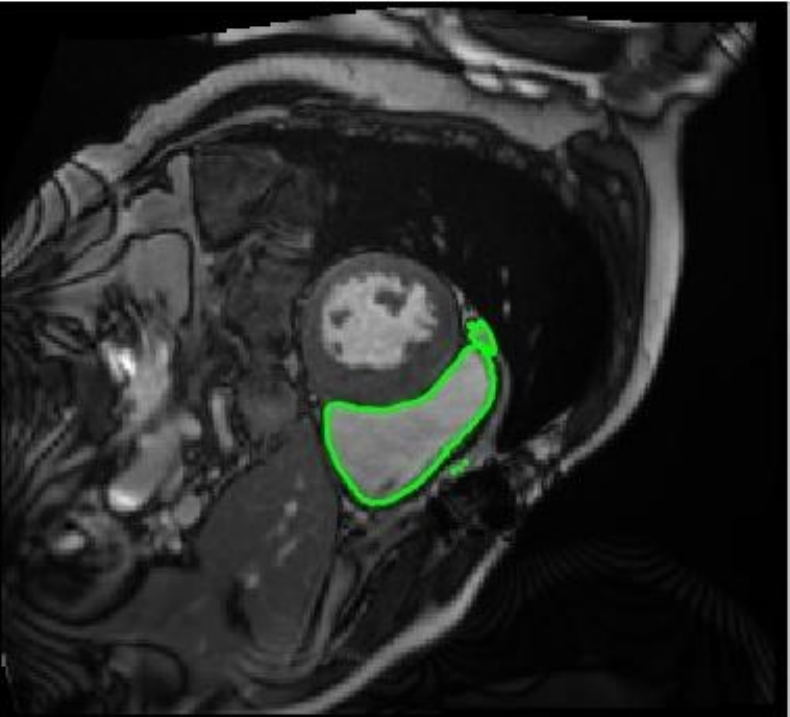}%
		\label{fig_81_case}}
	\subfloat{\includegraphics[width=0.8in]{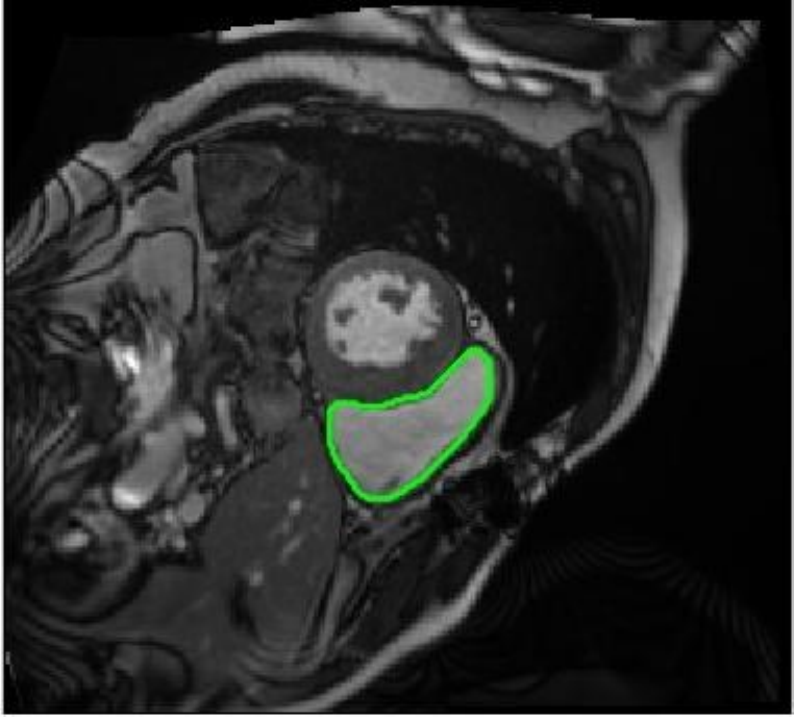}%
		\label{fig_83_case}}
	\subfloat{\includegraphics[width=0.8in]{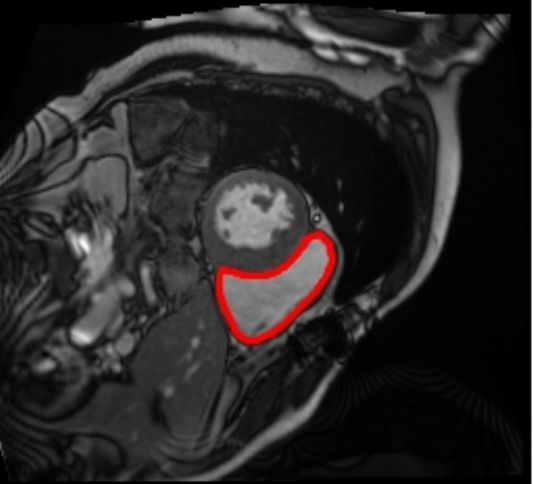}%
		\label{fig_84_case}}
	\hfil
	\subfloat{\includegraphics[width=0.8in]{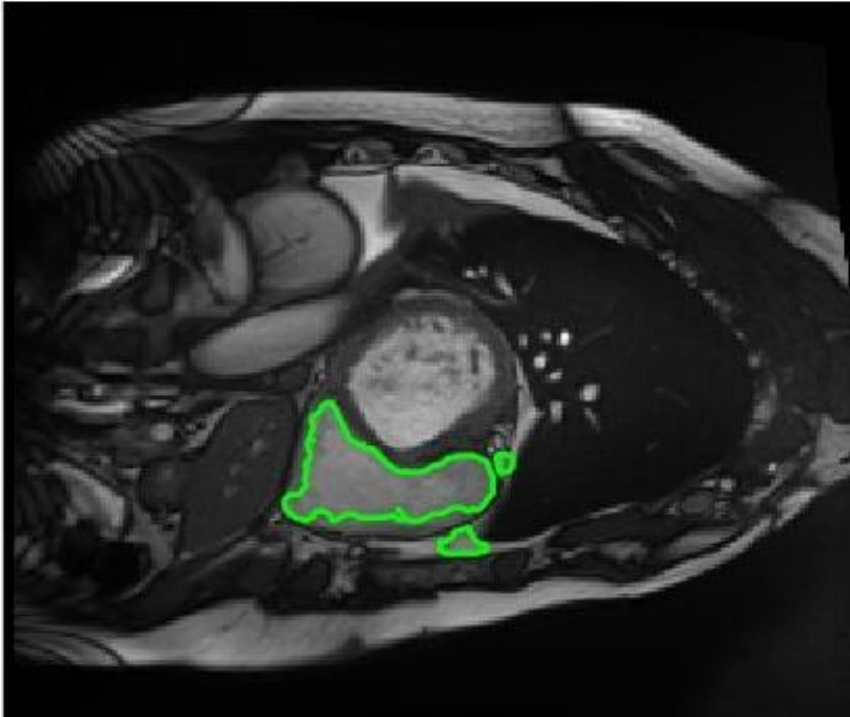}%
		\label{fig_85_case}}\vspace{-3mm}\hspace{-1.5mm}
	\subfloat{\includegraphics[width=0.8in]{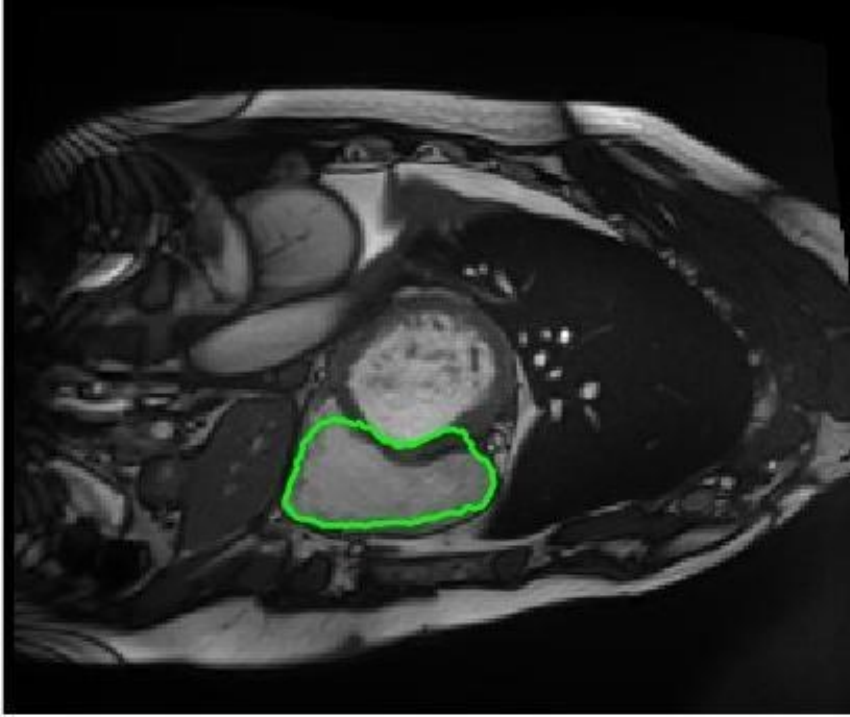}%
		\label{fig_86_case}}
	\subfloat{\includegraphics[width=0.8in]{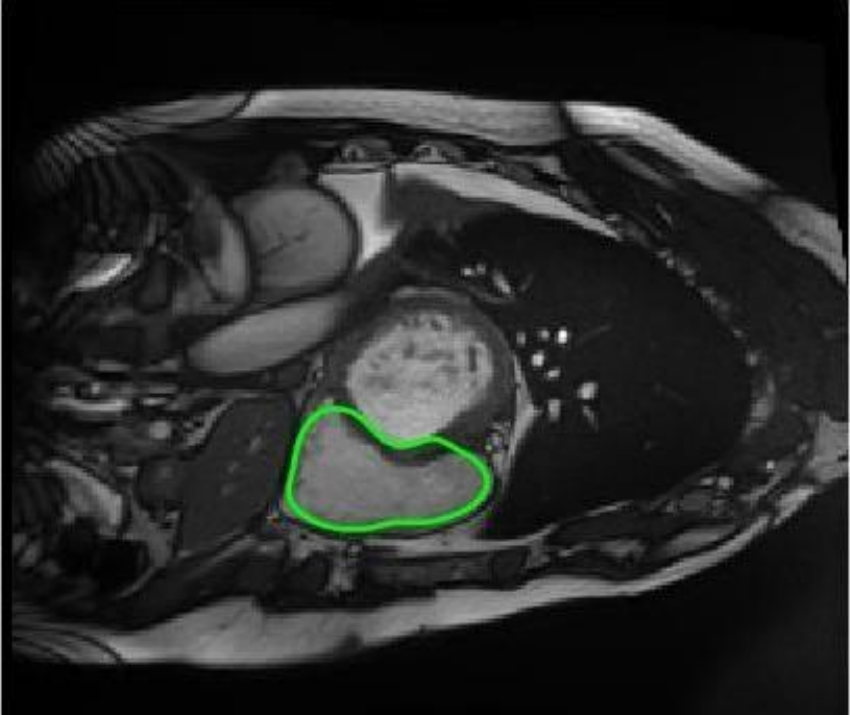}%
		\label{fig_87_case}}
	\subfloat{\includegraphics[width=0.8in]{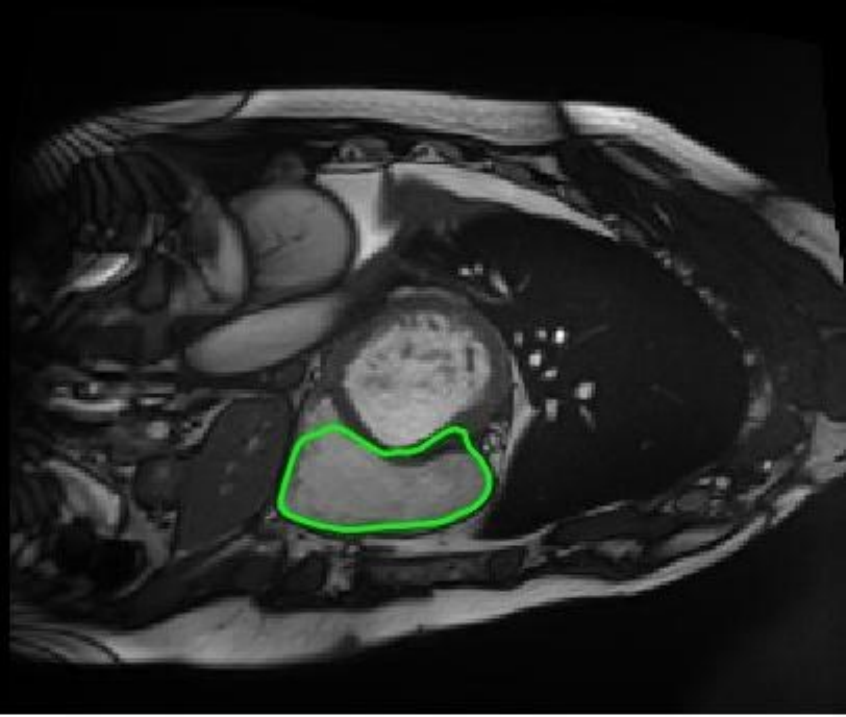}%
		\label{fig_88_case}}
	\subfloat{\includegraphics[width=0.8in]{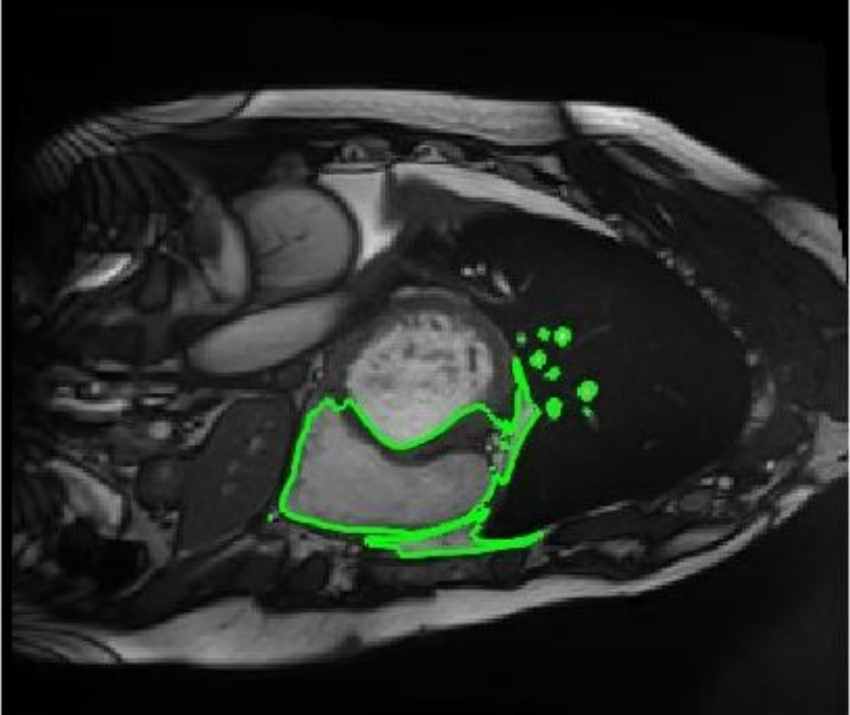}%
		\label{fig_89_case}}
	\subfloat{\includegraphics[width=0.8in]{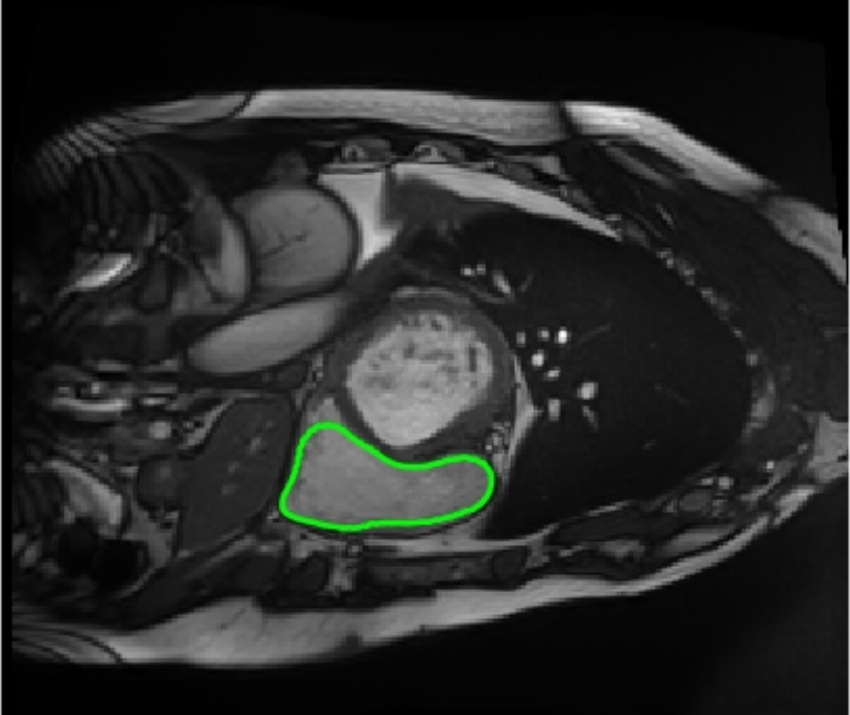}%
		\label{fig_90_case}}
	\subfloat{\includegraphics[width=0.8in]{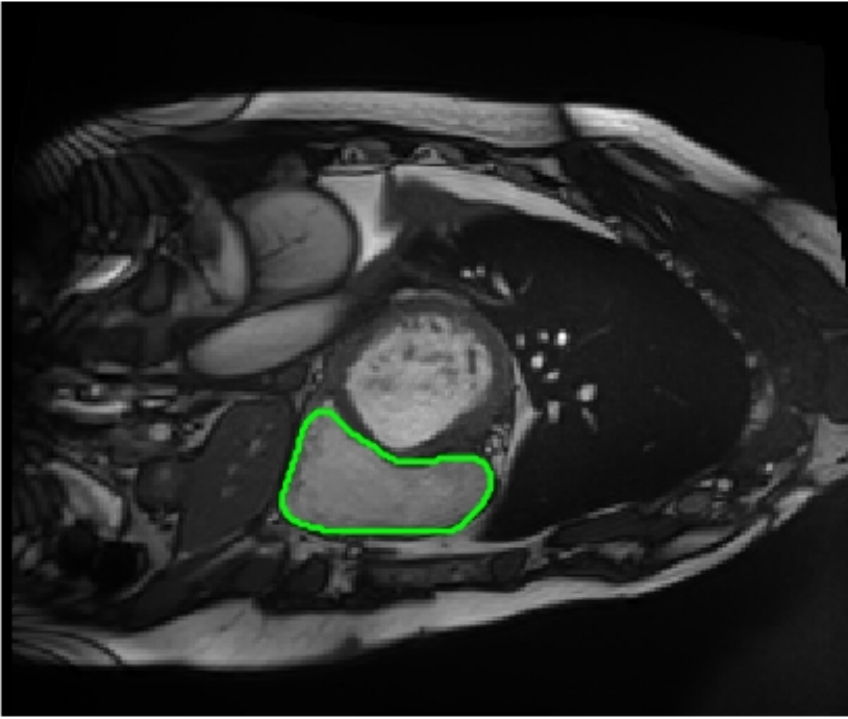}%
		\label{fig_91_case}}
	\subfloat{\includegraphics[width=0.8in]{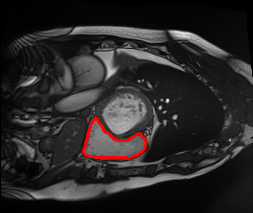}%
		\label{fig_92_case}}
	\hfil
	\subfloat{\includegraphics[width=0.8in]{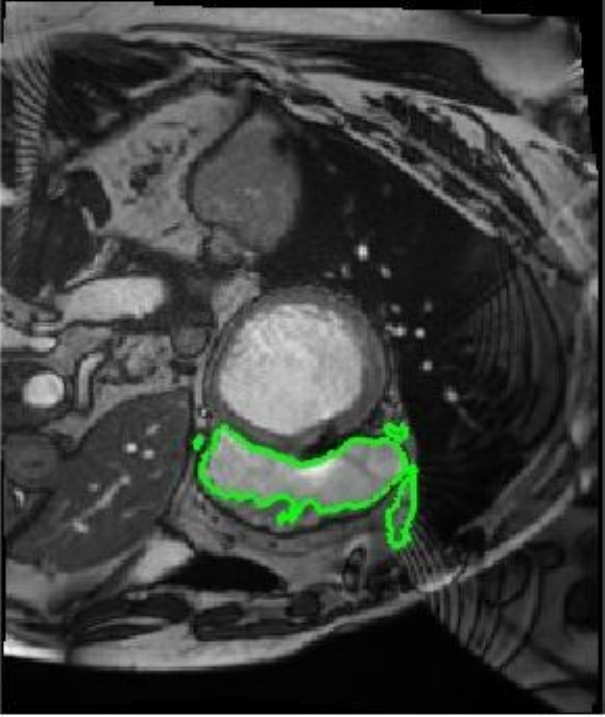}%
		\label{fig_93_case}}
	\subfloat{\includegraphics[width=0.795in]{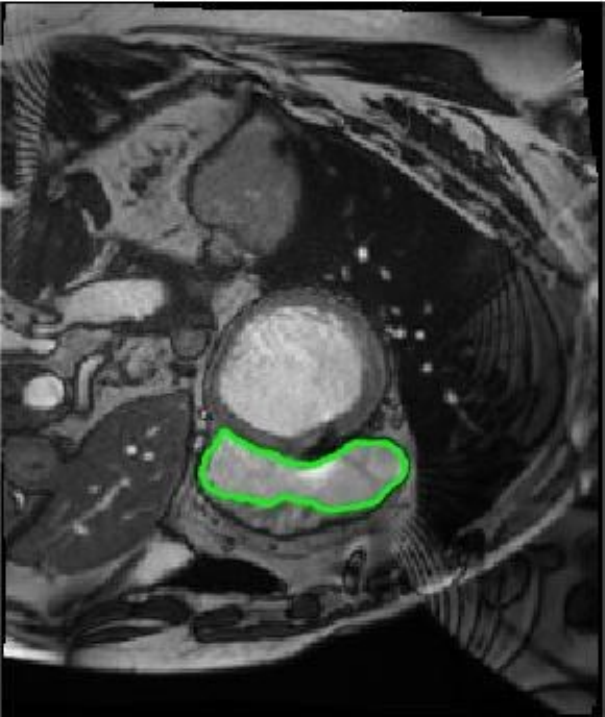}%
		\label{fig_94_case}}
	\subfloat{\includegraphics[width=0.8in]{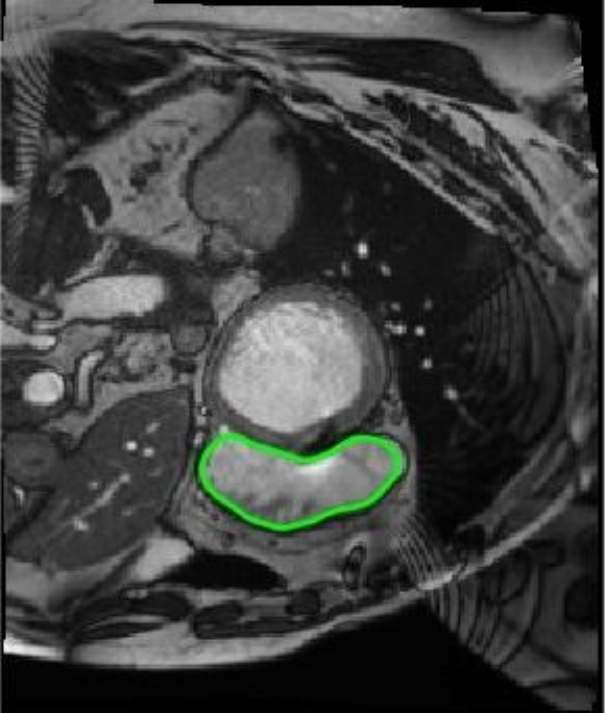}%
		\label{fig_95_case}}
	\subfloat{\includegraphics[width=0.8in]{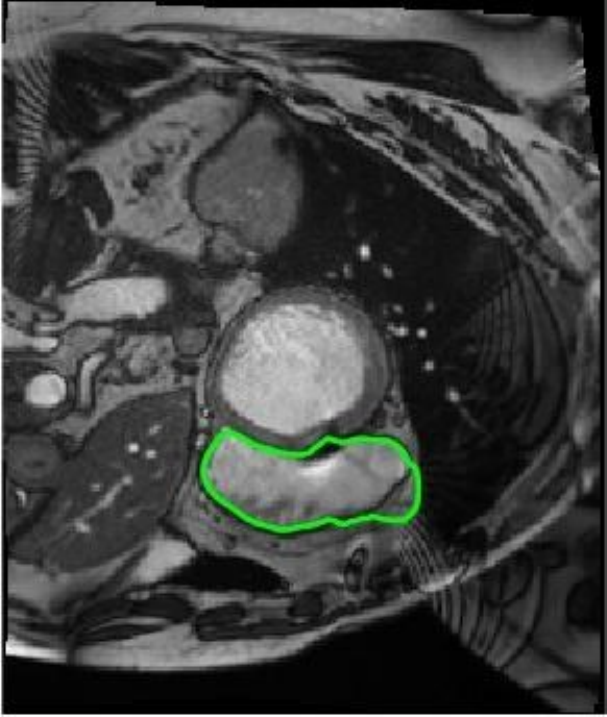}%
		\label{fig_96_case}}
	\subfloat{\includegraphics[width=0.8in]{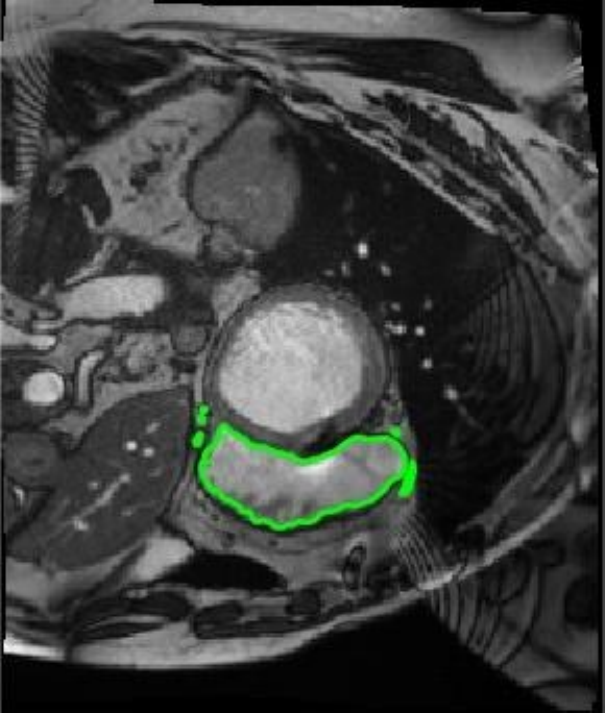}%
		\label{fig_97_case}}
	\subfloat{\includegraphics[width=0.795in]{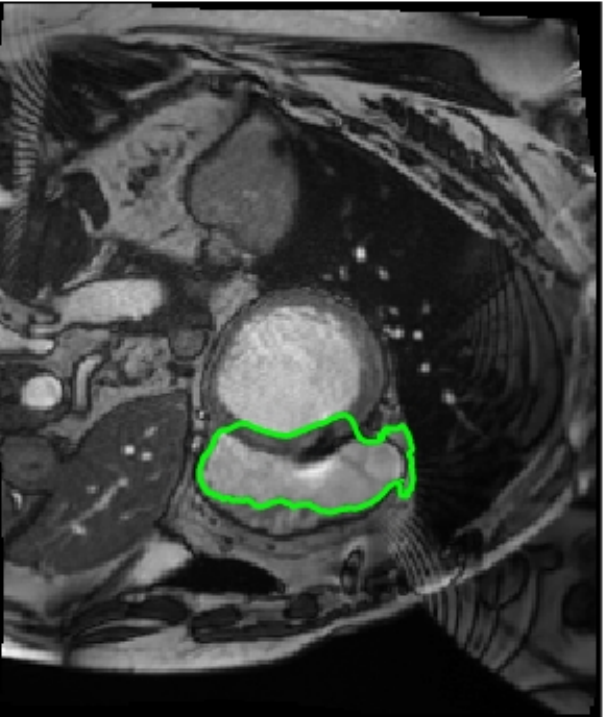}%
		\label{fig_98_case}}
	\subfloat{\includegraphics[width=0.8in]{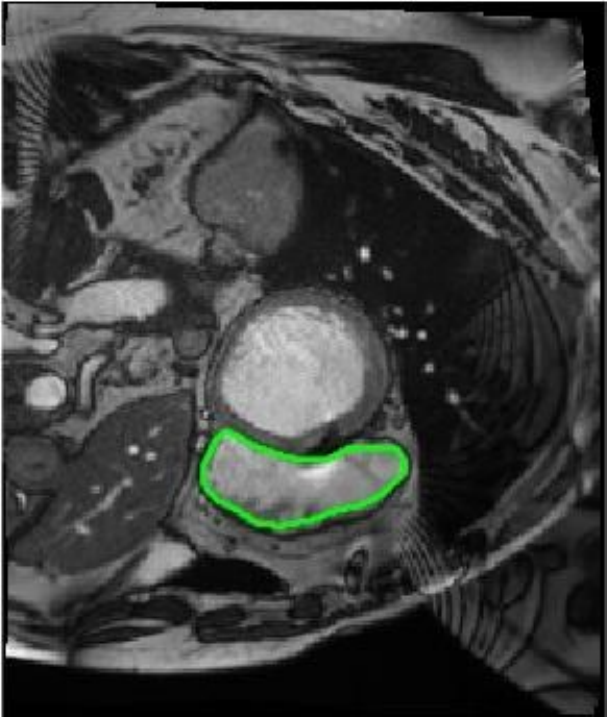}%
		\label{fig_99_case}}
	\subfloat{\includegraphics[width=0.8in]{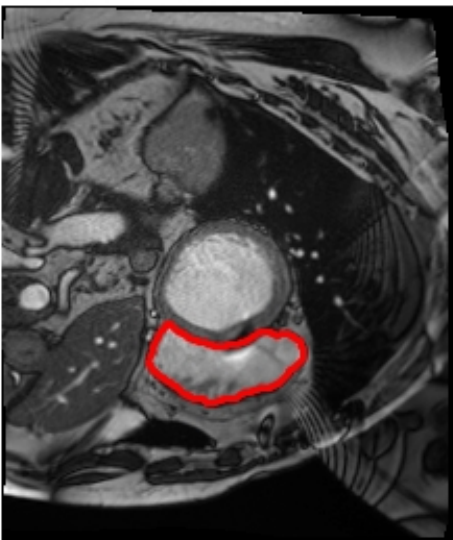}%
		\label{fig_100_case}}
	\hfil
	\caption{Segmentation results from evaluated models for right ventricle. Column 1-7: Results from the ALF model, LoGRSF model, ABC model, RESLS model, ICTM model, FeaACM model and the RefLSM. Column 8: Ground truth.}
	\label{img4}
\end{figure*}

\begin{figure}[!ht]
	\centering
	\subfloat[]{\includegraphics[width=2.3in]{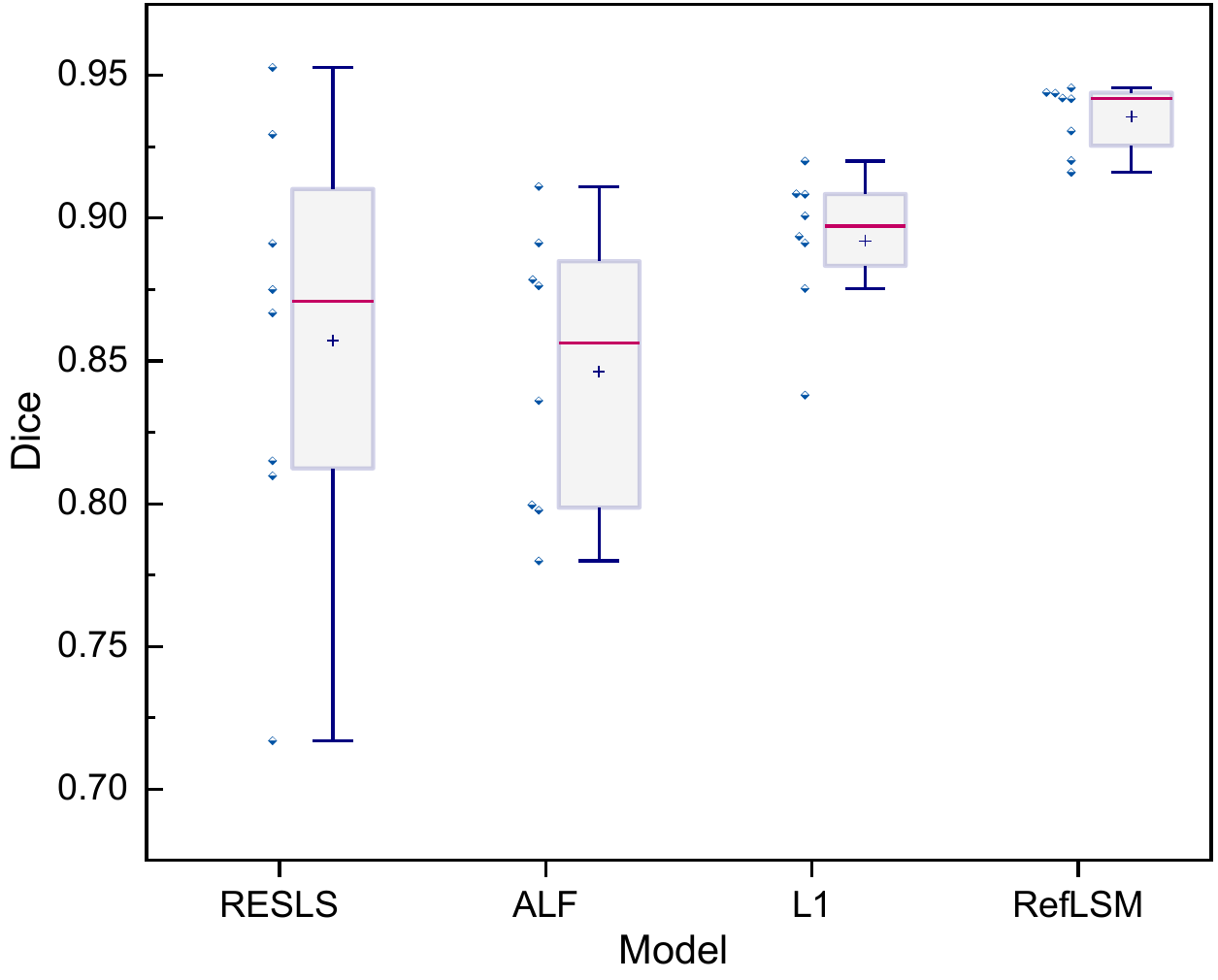}%
		\label{fig_59_case}}
	\subfloat[]{\includegraphics[width=2.3in]{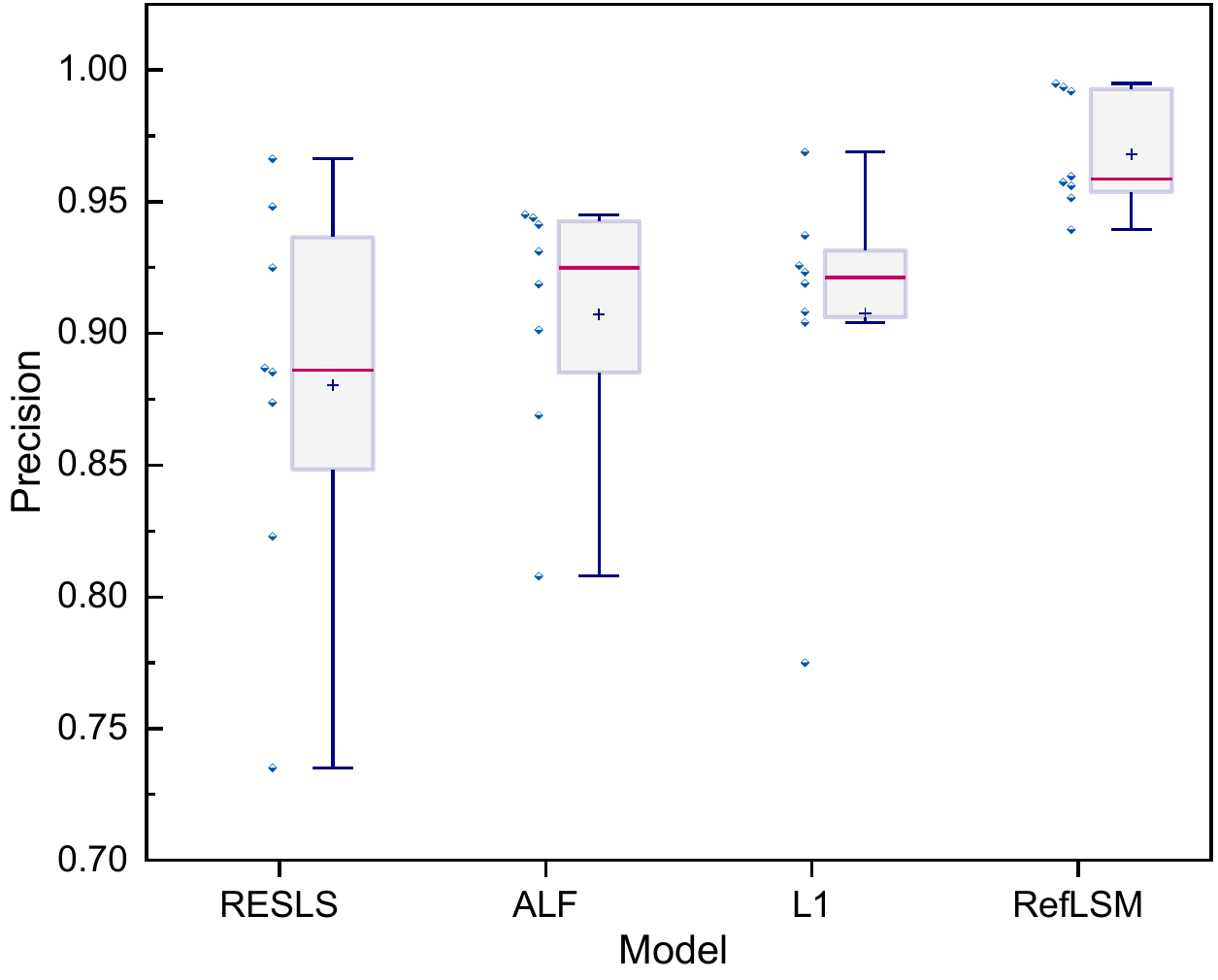}%
		\label{fig_60_case}}
	\caption{ Dice values and Precision values of different models for brain tumor MR images.}
	\label{img16}
\end{figure}

\begin{table*}[!ht]
\centering
\caption{Comparison of Precision and Dice values for different models on brain tumor MR images shown in Fig.~\ref{img3}. 
The best results are highlighted in bold.}
\renewcommand{\arraystretch}{1.15}
\small
\setlength{\tabcolsep}{6pt}
\begin{tabular}{l|lcccccccc|c}
\toprule
Metric & Model & Img.1 & Img.2 & Img.3 & Img.4 & Img.5 & Img.6 & Img.7 & Img.8 & Mean \\
\midrule
\multirow{4}{*}{Precision $\uparrow$} 
& RESLS   & 0.7351 & 0.8230 & 0.9249 & 0.9662 & 0.8737 & 0.8853 & 0.9481 & 0.8869 & 0.8804 \\
& ALF     & 0.9311 & 0.8079 & 0.9413 & 0.9187 & 0.9013 & 0.9439 & 0.9451 & 0.8689 & 0.9072 \\
& L1      & 0.9042 & 0.9190 & 0.7750 & 0.9688 & 0.9233 & 0.9372 & 0.9258 & 0.9082 & 0.9076 \\
& RefLSM  & \textbf{0.9574} & \textbf{0.9919} & \textbf{0.9596} & \textbf{0.9948} 
           & \textbf{0.9394} & \textbf{0.9514} & \textbf{0.9560} & \textbf{0.9935} & \textbf{0.9680} \\
\midrule
\multirow{4}{*}{Dice $\uparrow$} 
& RESLS   & 0.7169 & 0.8150 & 0.9293 & 0.9528 & 0.8750 & 0.8911 & 0.8668 & 0.8097 & 0.8571 \\
& ALF     & 0.7977 & 0.7995 & 0.8360 & 0.8763 & 0.8785 & 0.9111 & 0.8913 & 0.7799 & 0.8462 \\
& L1      & 0.9084 & 0.8753 & 0.8380 & 0.9085 & 0.8913 & 0.9200 & 0.9008 & 0.8936 & 0.8919 \\
& RefLSM  & \textbf{0.9421} & \textbf{0.9160} & \textbf{0.9440} & \textbf{0.9438} 
           & \textbf{0.9202} & \textbf{0.9418} & \textbf{0.9305} & \textbf{0.9456} & \textbf{0.9355} \\
\bottomrule
\end{tabular}
\label{t1}
\end{table*}

\subsection {Comparison with brain tumor MR images segmentation}

Brain tumor MRI images represent a critical challenge within the field of medical image analysis, making them a focal point of our research. Various level set models for image segmentation have been applied to brain tumor MRI images, and it is essential to consider the unique characteristics of these images. To validate the excellent performance of the RefLSM, we conduct segmentation comparison experiments on brain MR images and compute the Dice coefficient and Precision values for all evaluated models.

Fig. \ref{img3} displays the visual comparison results for segmenting eight brain tumor MR images between the RefLSM and three representative models: the RESLS \cite{8765635}, ALF\cite{MA2019201}, and L1 model\cite{LIU2019193}. These images exhibit severe intensity inhomogeneity and noise attributable to inconsistent bias field and imaging equipment. The initial level set function (LSF) initialization was consistent across all models, as shown in Fig. \ref{img3} (a), which illustrates the initial placement of the zero level contour. For visual clarity, we included the ground truth in Fig. \ref{img3} (f). The segmentation outcomes from the RESLS, ALF, and L1 methods are presented in Fig. \ref{img3} (b), (c), and (d), respectively.

It is evident from the first and last columns of Fig. \ref{img3} (b) that the ALF model tends to become trapped in local minima under severe intensity inhomogeneity. Furthermore, segmentation results presented in the columns 1-4 of Fig. \ref{img3} (b) and (c) reveal that both the ALF and L1 models mistakenly segment small isolated or irrelevant regions due to the effects of irregular intensity patterns. Additionally, when weak boundaries are present, as seen in the 1st and 3rd columns of Fig. \ref{img3}, the RESLS, ALF, and L1 methods fail to accurately identify object boundaries. Consequently, the zero level contours of these models deviate significantly from the objects throughout the level set evolution, causing drastic misalignment after further iterations.\\
\indent In stark contrast, our proposed method exhibits greater robustness against images exhibiting severe intensity inhomogeneity and weak boundaries. The proposed prior constraint term effectively corrects for intensity inhomogeneity, while the proposed binary level set adeptly reduces noise during the segmentation process.

\begin{figure}[!t]
	\centering
	\subfloat{\includegraphics[width=3in]{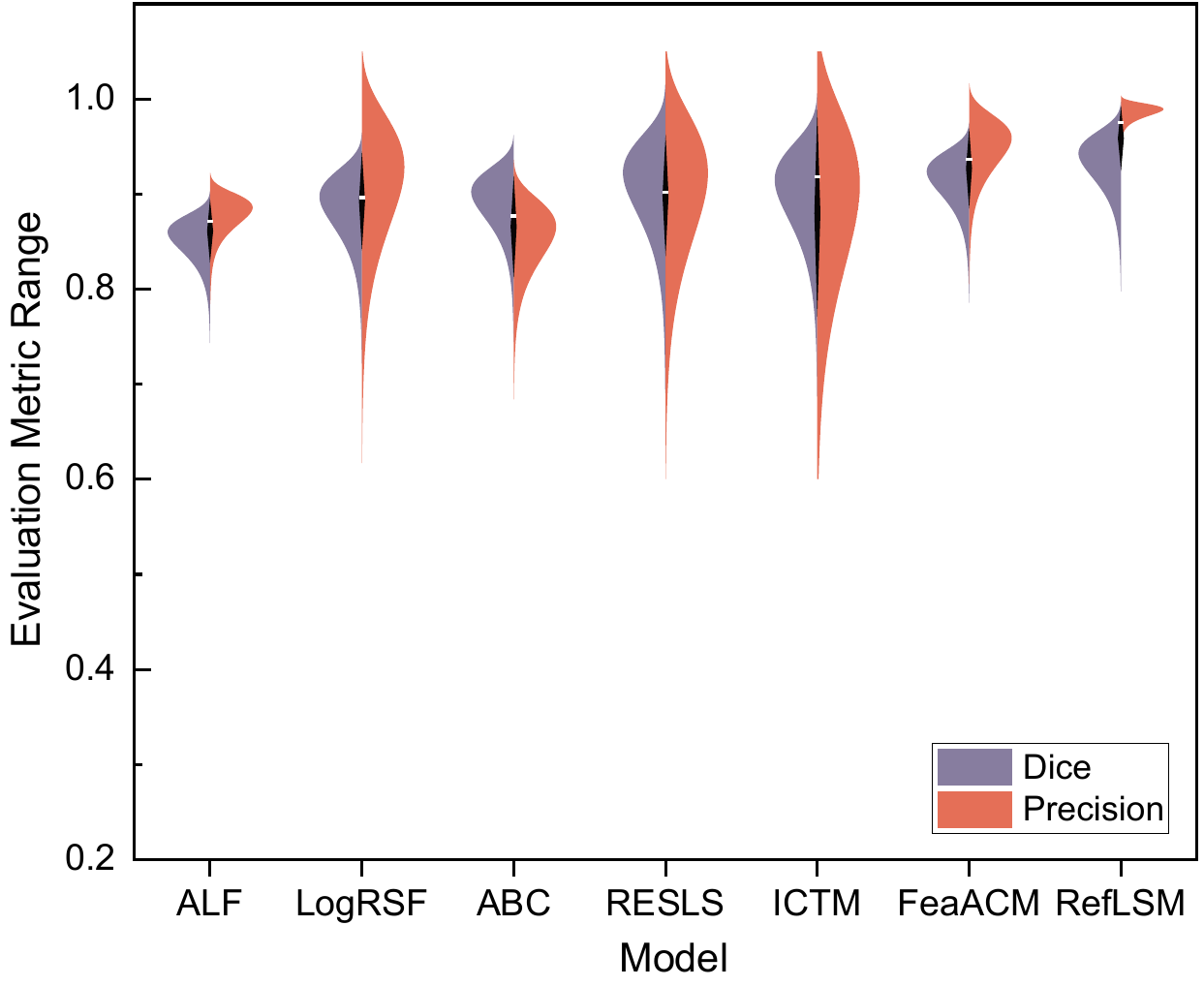}%
		\label{fig_101_case}}
	\caption{The violin plot of Dice values and Precision values of different models for right ventricle.}
	\label{img17}
\end{figure}
Overall, these findings demonstrate that the segmentation outcomes produced by our model are more accurate. Besides visual assessment, we calculate the Dice and Precision values of evaluated models for brain tumor segmentation in table \ref{t1}. Based on these results, we draw a boxplot with scatterplot overlay, as shown in Fig. \ref{img16}, which displays the data distribution and facilitates comparison of different models' performance across various brain tumor MR images. We analyze the maximum, median, and minimum, which correspond to the best, median segmentation results, and worst segmentation results, respectively. It is clear that the RefLSM achieves more accurate segmentation results compared to the other models. Moreover, the smaller range of the RefLSM in the boxplot indicates robustness and stability to different images of the brain tumor. 

\begin{figure*}[!t]
	\centering
	\subfloat{\includegraphics[width=0.8in,height=0.8in]{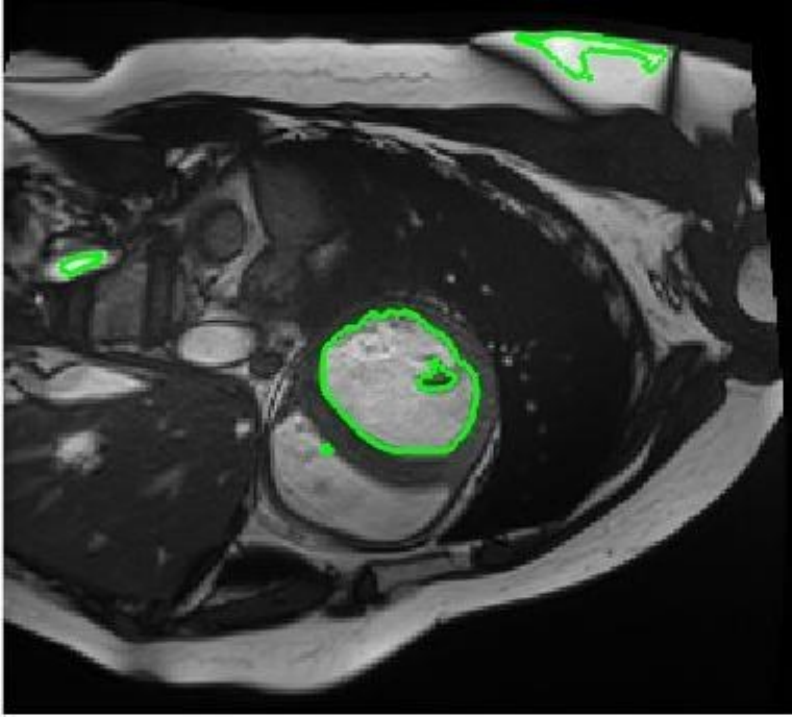}% 
		\label{fig1_case}}\vspace{-3mm}\hspace{-1.5mm}
	\subfloat{\includegraphics[width=0.8in,height=0.8in]{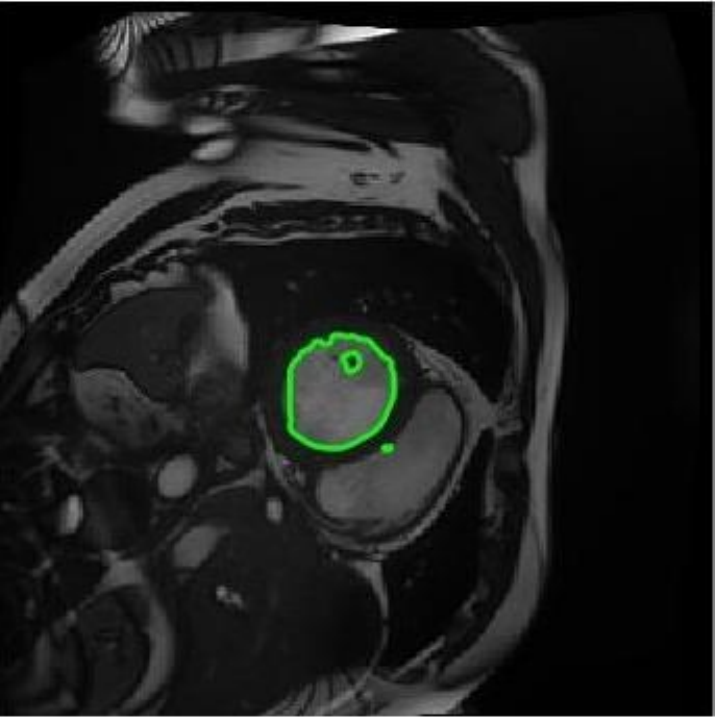}%
		\label{fig2_case}}
	\subfloat{\includegraphics[width=0.8in,height=0.8in]{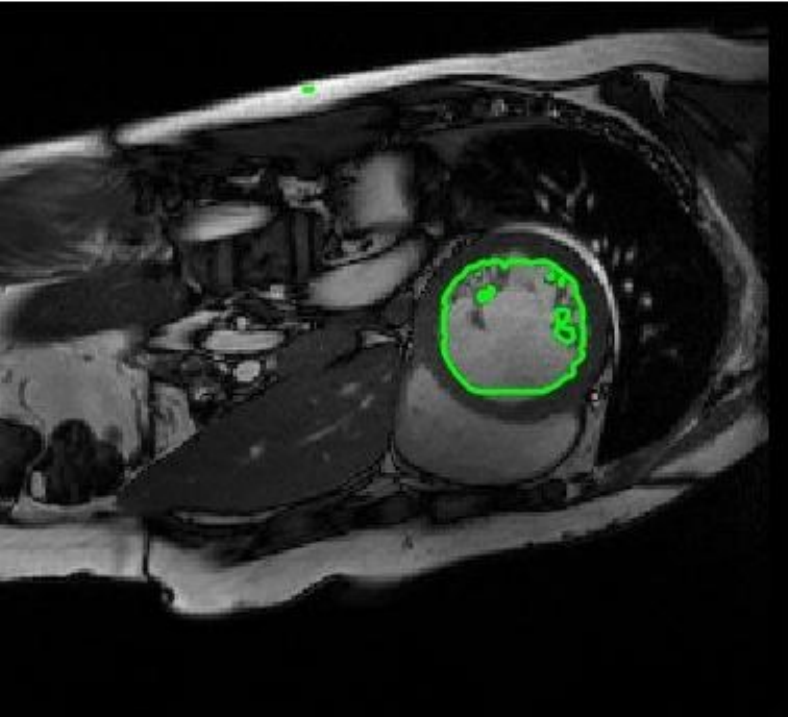}%
		\label{fig3_case}}
	\subfloat{\includegraphics[width=0.8in,height=0.8in]{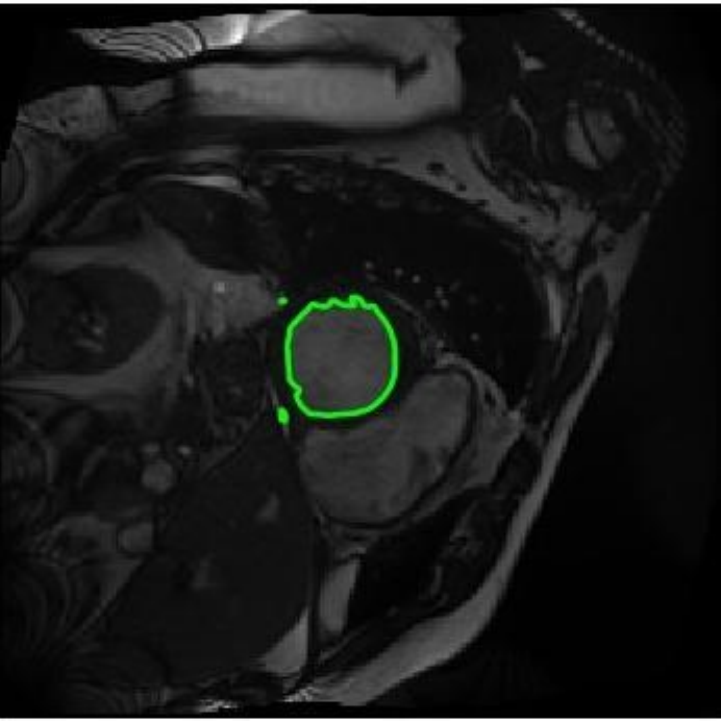}%
		\label{fig4_case}}
	\subfloat{\includegraphics[width=0.8in,height=0.8in]{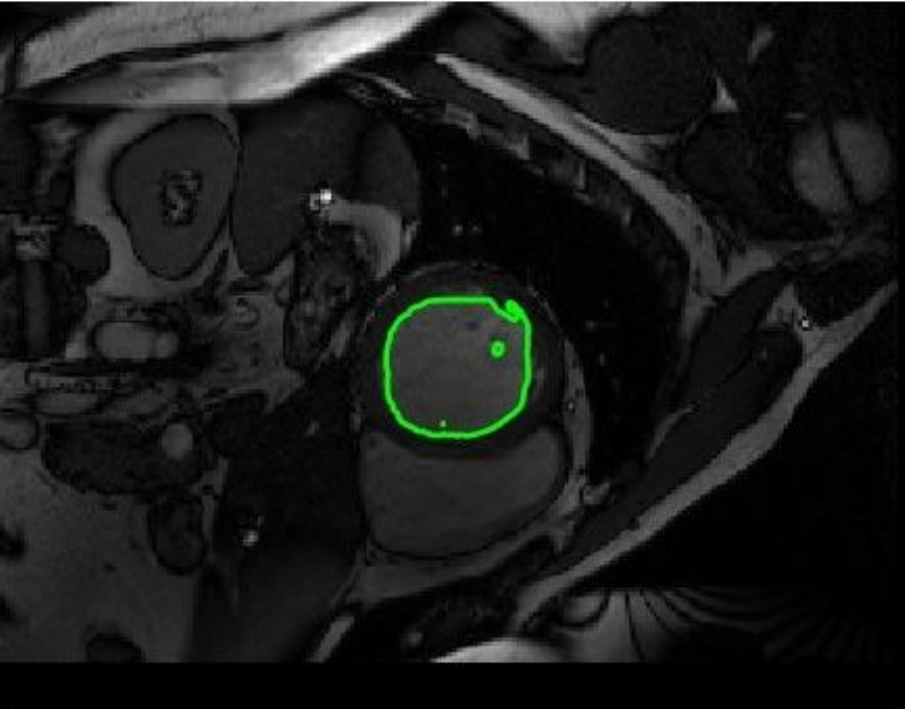}%
		\label{fig5_case}}
	\subfloat{\includegraphics[width=0.8in,height=0.8in]{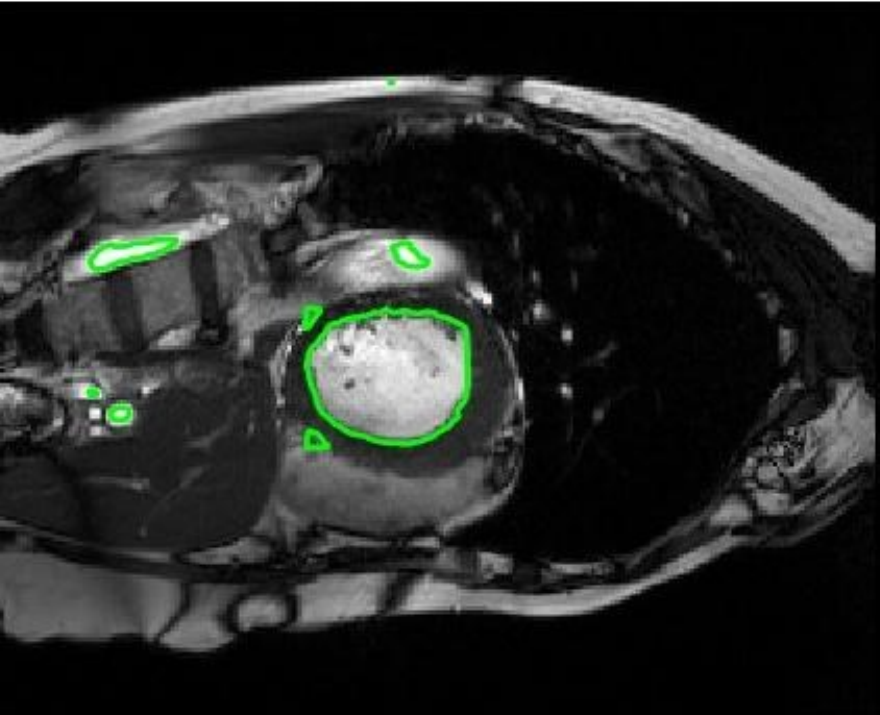}%
		\label{fig6_case}}
	\subfloat{\includegraphics[width=0.8in,height=0.8in]{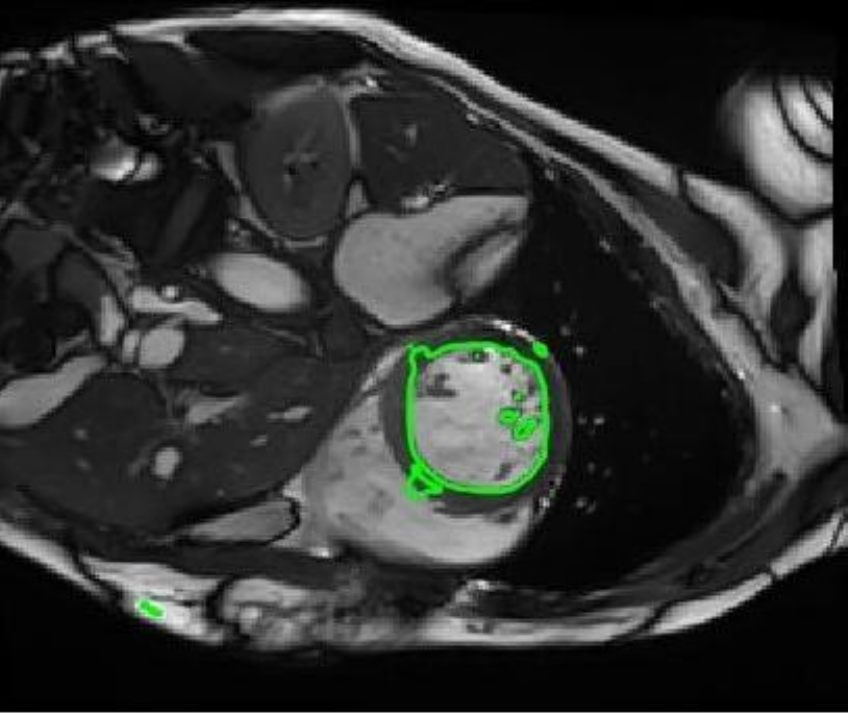}%
		\label{fig7_case}}
	\hfil
	
	\subfloat{\includegraphics[width=0.8in,height=0.8in]{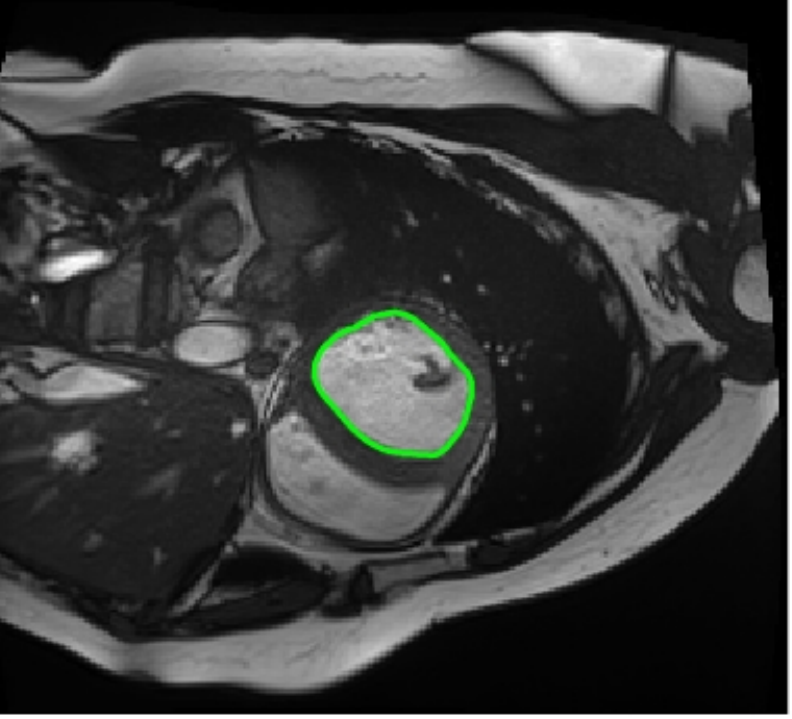}%
		\label{fig9_case}}\vspace{-3mm}\hspace{-1.5mm}
	\subfloat{\includegraphics[width=0.8in,height=0.8in]{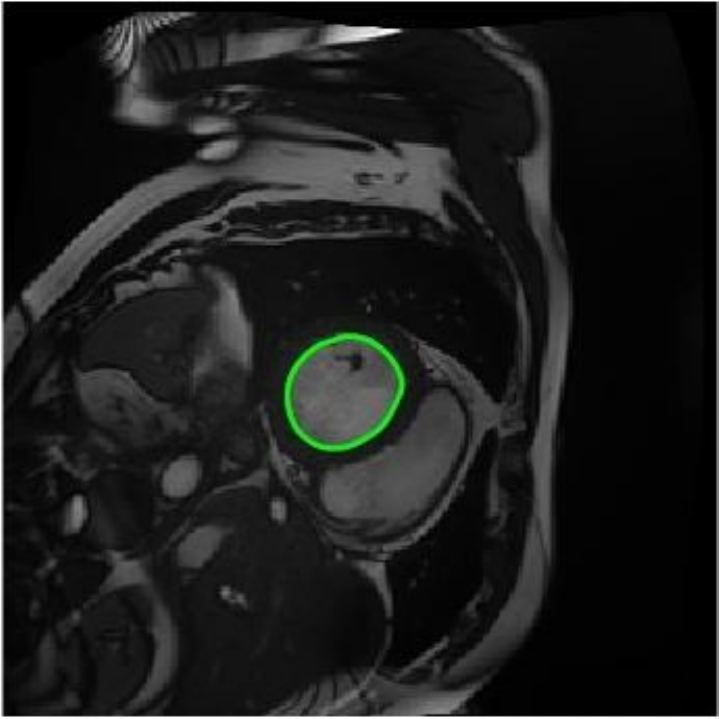}%
		\label{fig10_case}}
	\subfloat{\includegraphics[width=0.8in,height=0.8in]{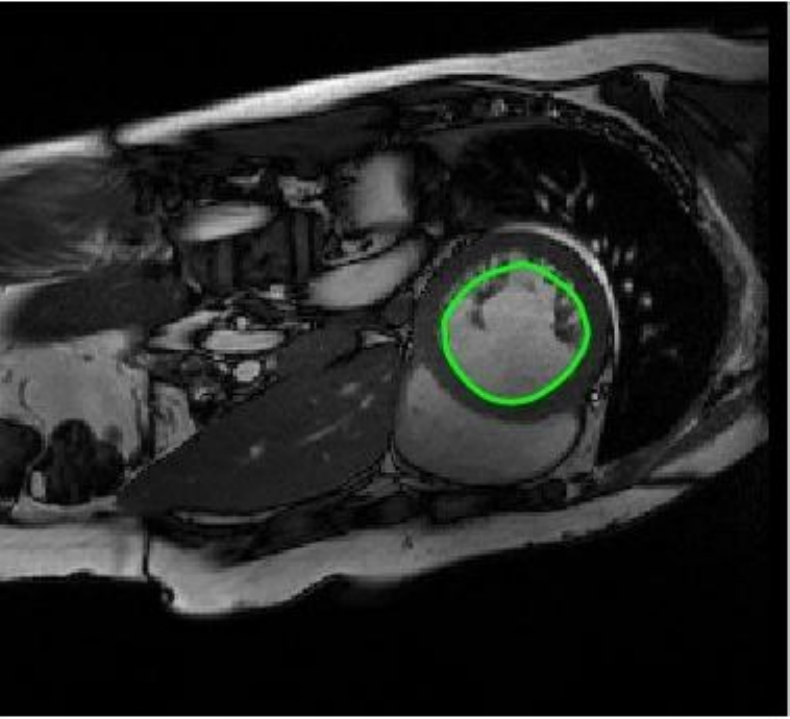}%
		\label{fig11_case}}
	\subfloat{\includegraphics[width=0.8in,height=0.8in]{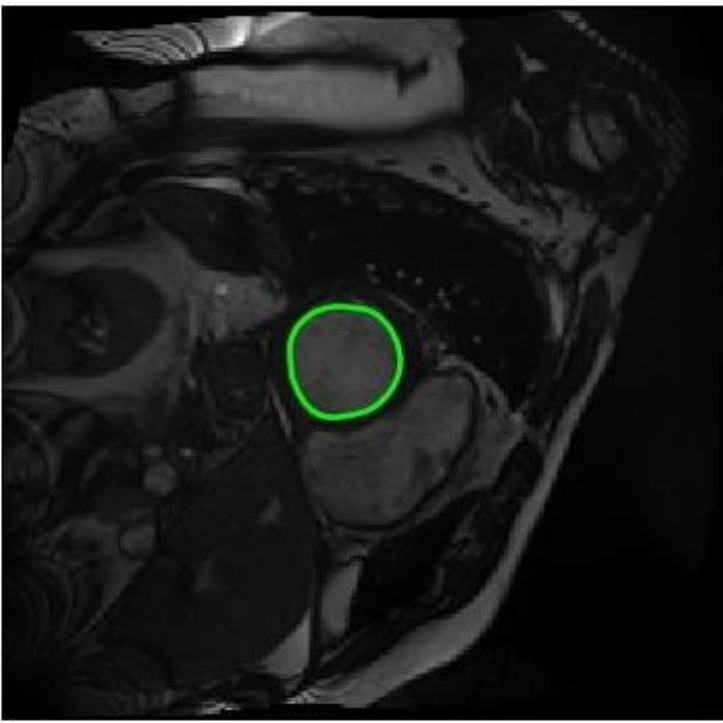}%
		\label{fig12_case}}
	\subfloat{\includegraphics[width=0.8in,height=0.8in]{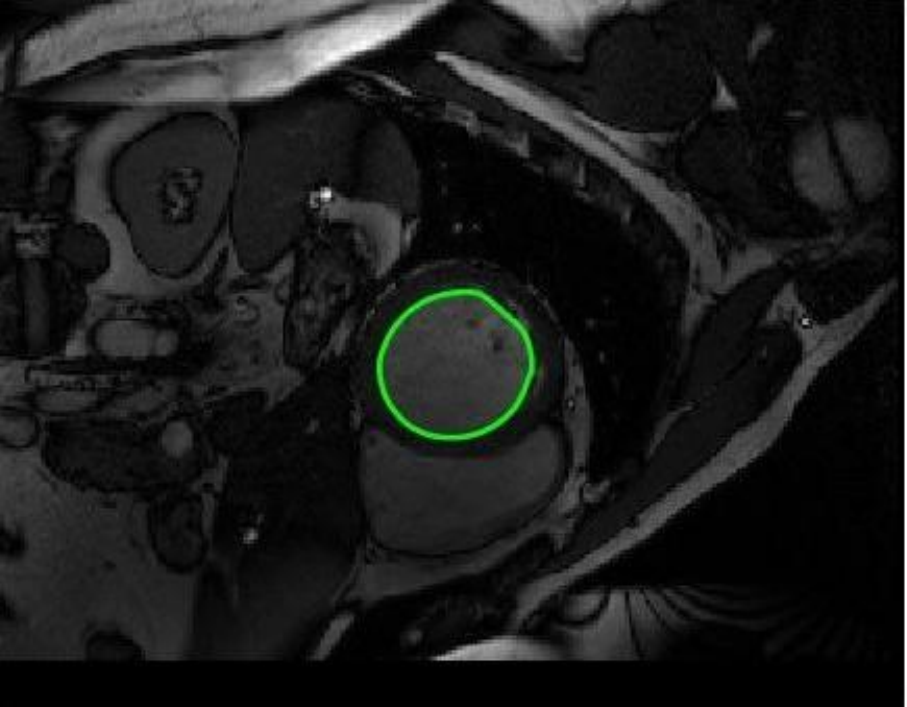}%
		\label{fig13_case}}
	\subfloat{\includegraphics[width=0.8in,height=0.8in]{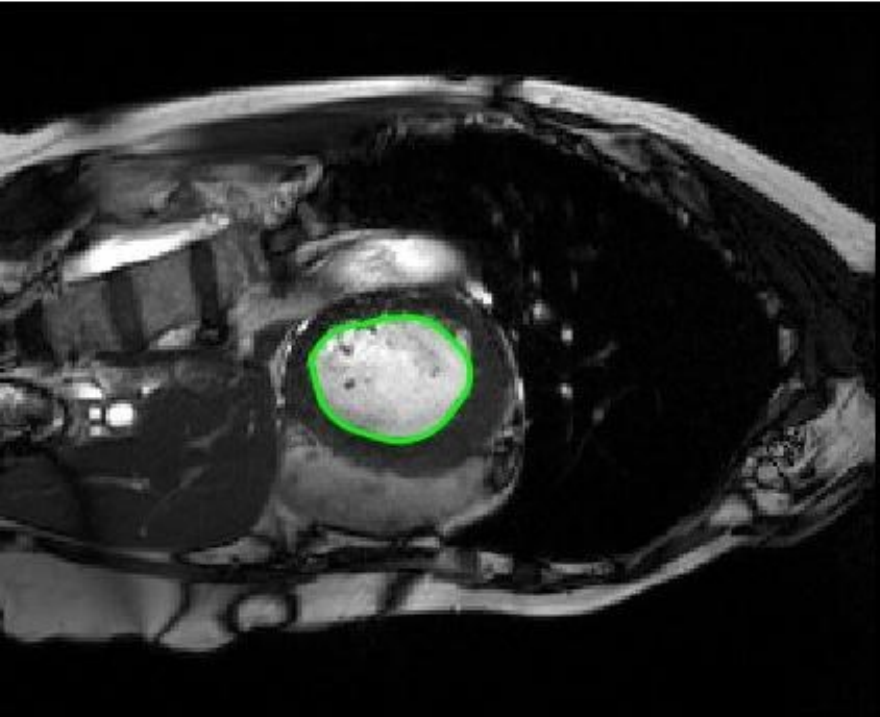}%
		\label{fig14_case}}
	\subfloat{\includegraphics[width=0.8in,height=0.8in]{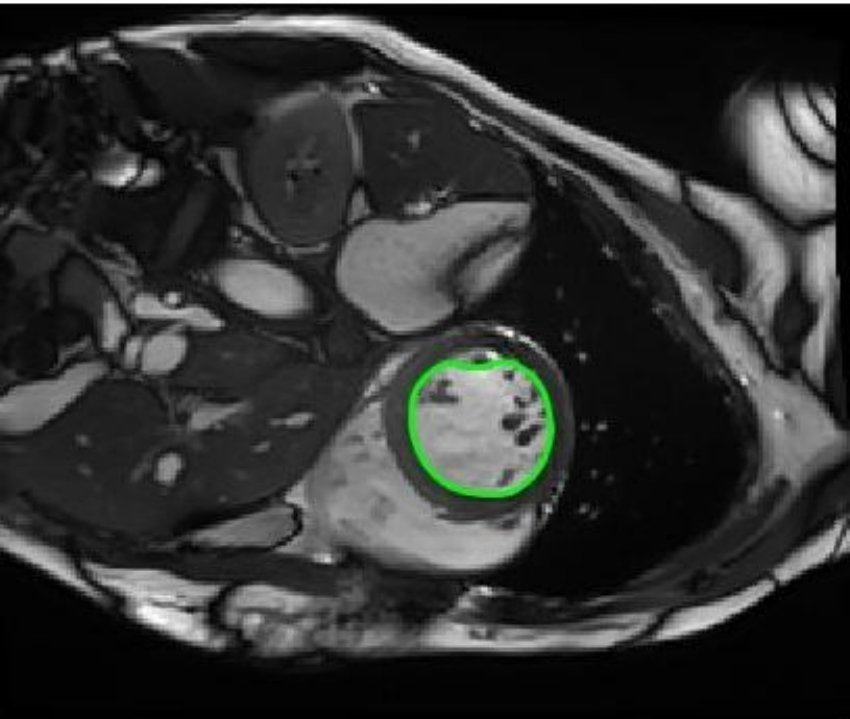}%
		\label{fig15_case}}
	\hfil
	
	\subfloat{\includegraphics[width=0.8in,height=0.8in]{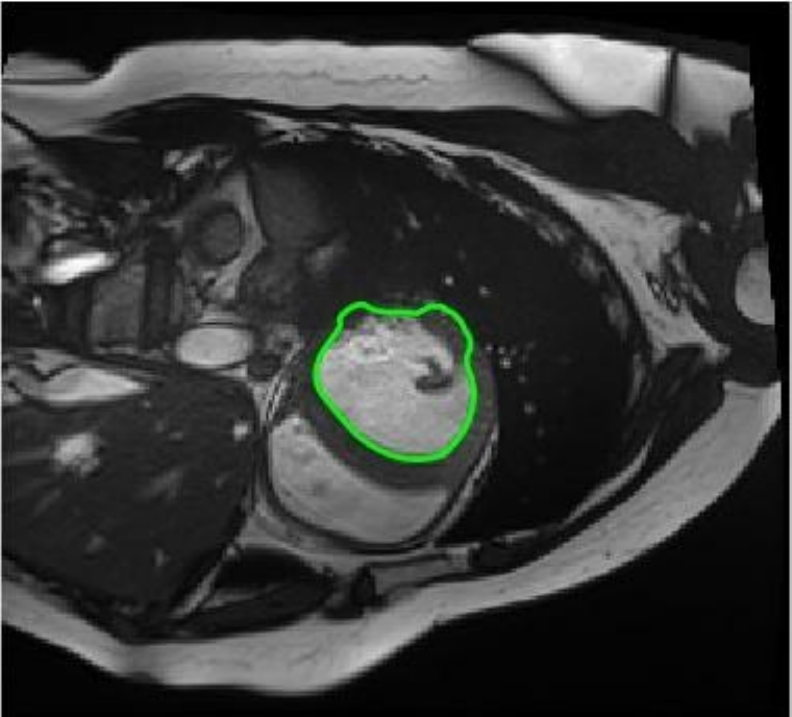}%
		\label{fig17_case}}\vspace{-3mm}\hspace{-1.5mm}
	\subfloat{\includegraphics[width=0.8in,height=0.8in]{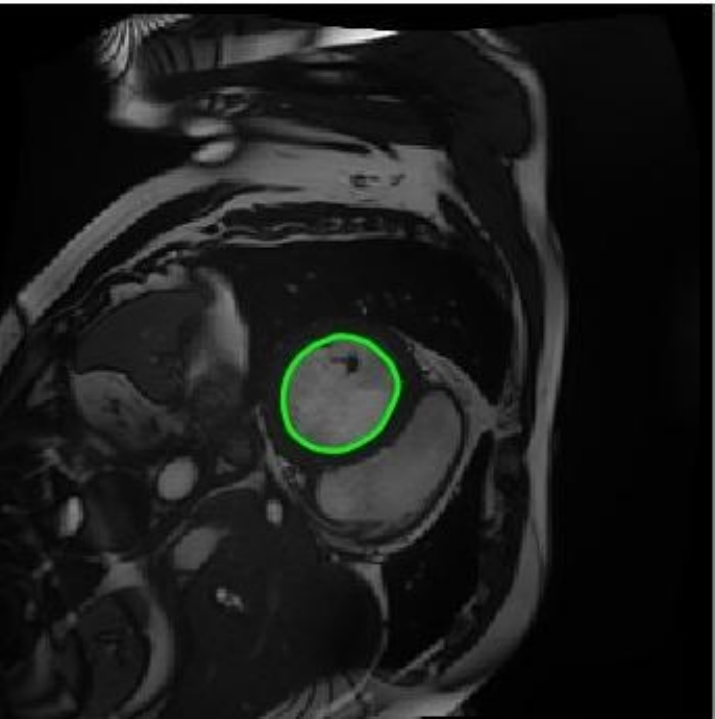}%
		\label{fig18_case}}
	\subfloat{\includegraphics[width=0.8in,height=0.8in]{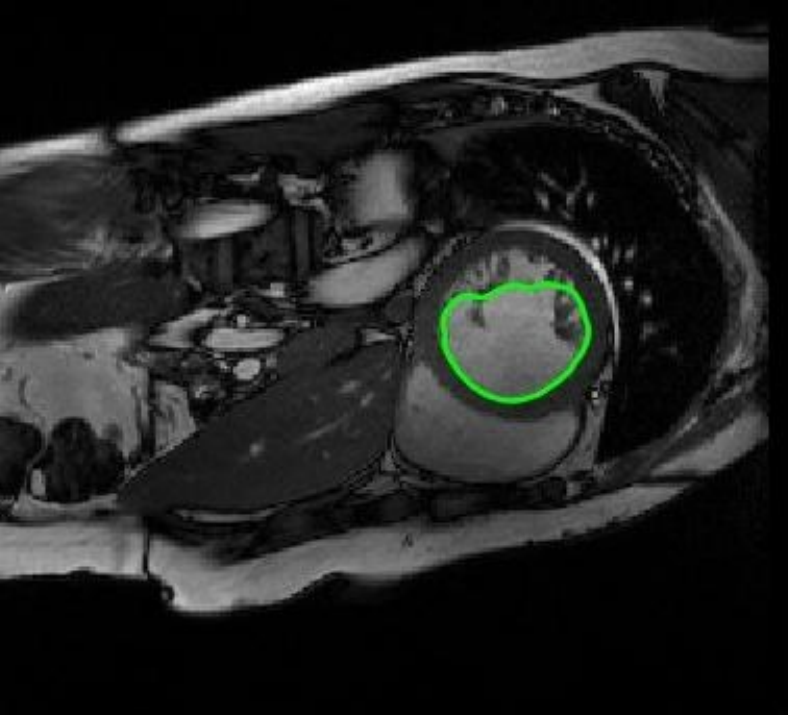}%
		\label{fig19_case}}
	\subfloat{\includegraphics[width=0.8in,height=0.8in]{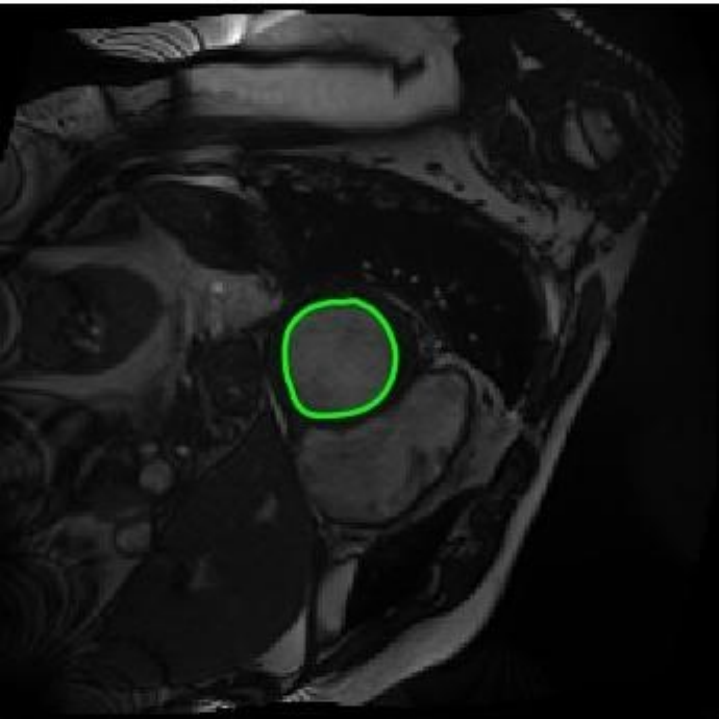}%
		\label{fig20_case}}
	\subfloat{\includegraphics[width=0.8in,height=0.8in]{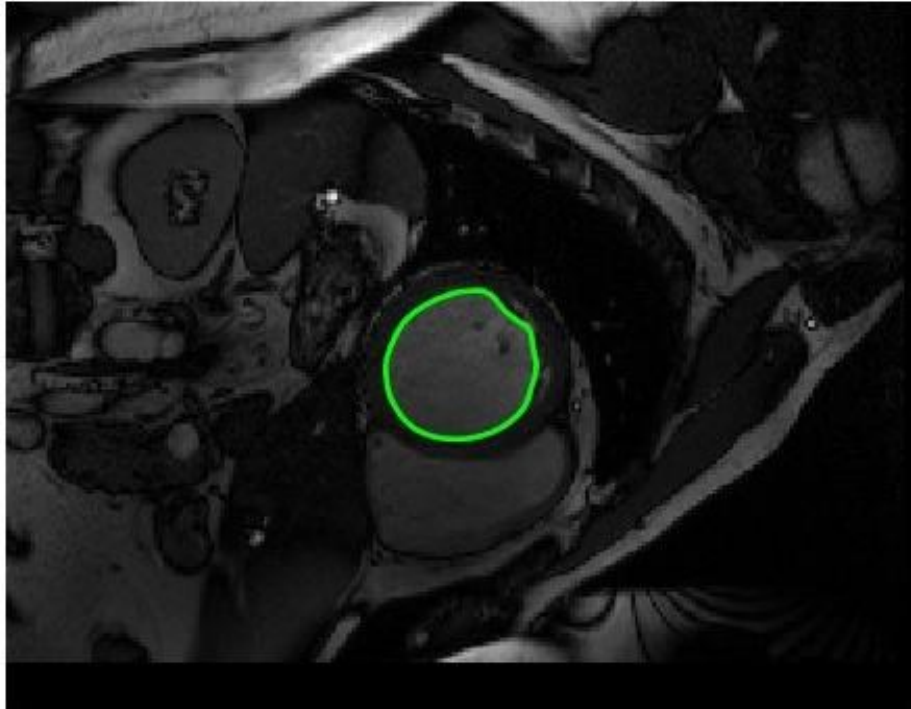}%
		\label{fig21_case}}
	\subfloat{\includegraphics[width=0.8in,height=0.8in]{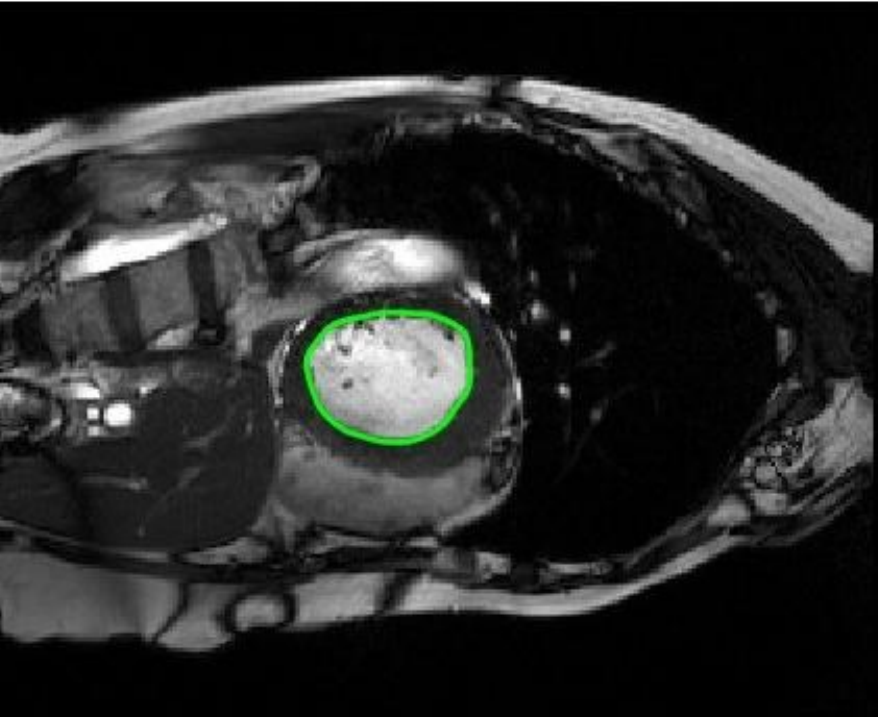}%
		\label{fig22_case}}
	\subfloat{\includegraphics[width=0.8in,height=0.8in]{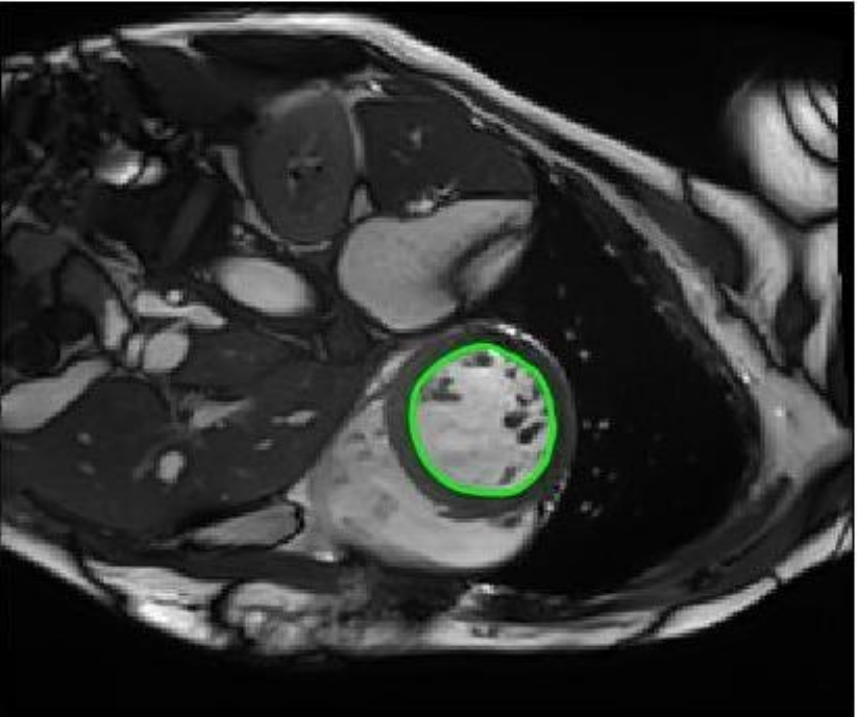}%
		\label{fig23_case}}
	\hfil
	
	\subfloat{\includegraphics[width=0.8in,height=0.8in]{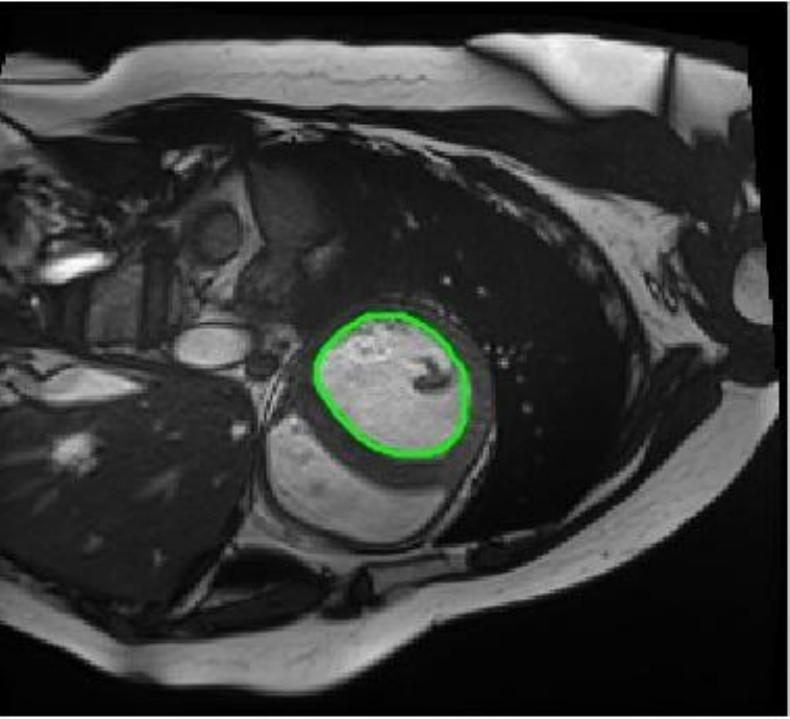}%
		\label{fig25_case}}\vspace{-3mm}\hspace{-1.5mm}
	\subfloat{\includegraphics[width=0.8in,height=0.8in]{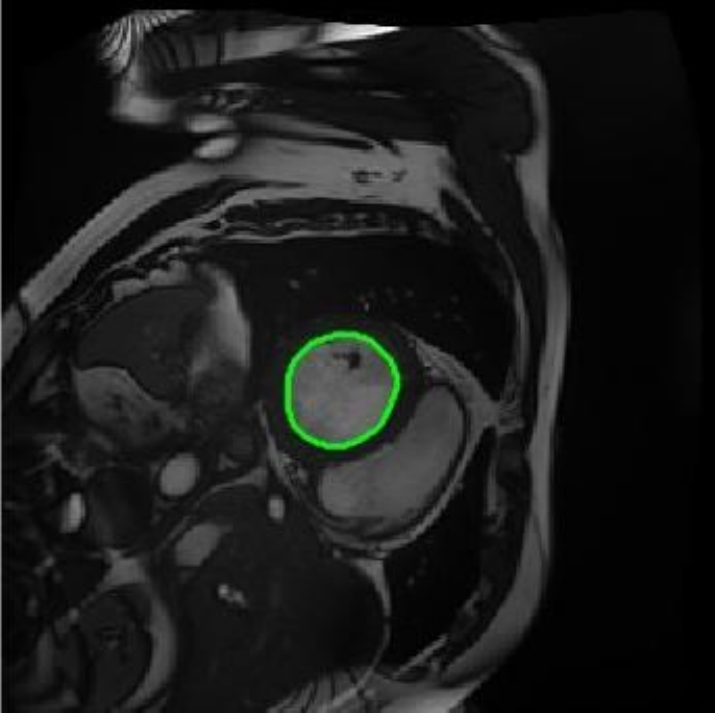}%
		\label{fig26_case}}
	\subfloat{\includegraphics[width=0.8in,height=0.8in]{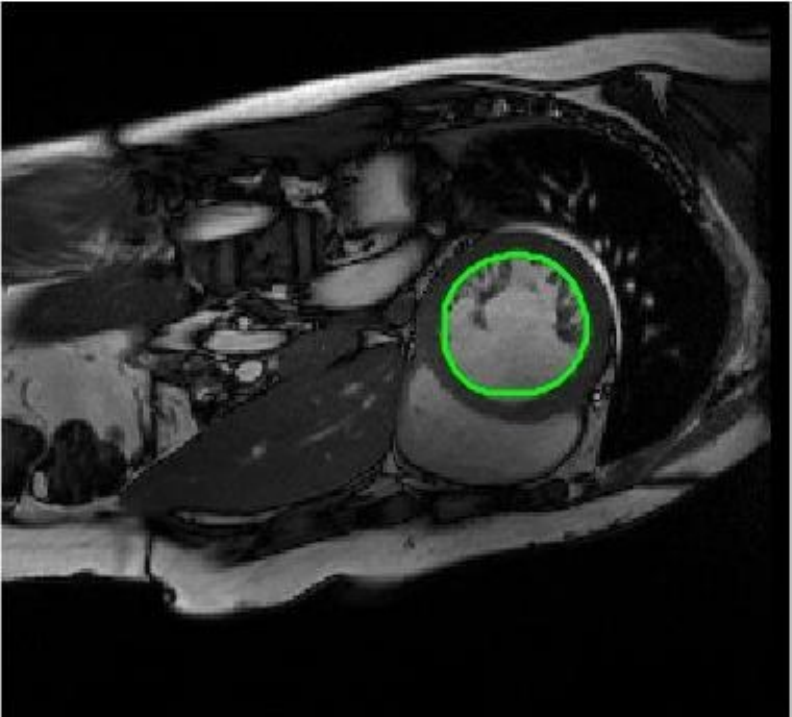}%
		\label{fig27_case}}
	\subfloat{\includegraphics[width=0.8in,height=0.8in]{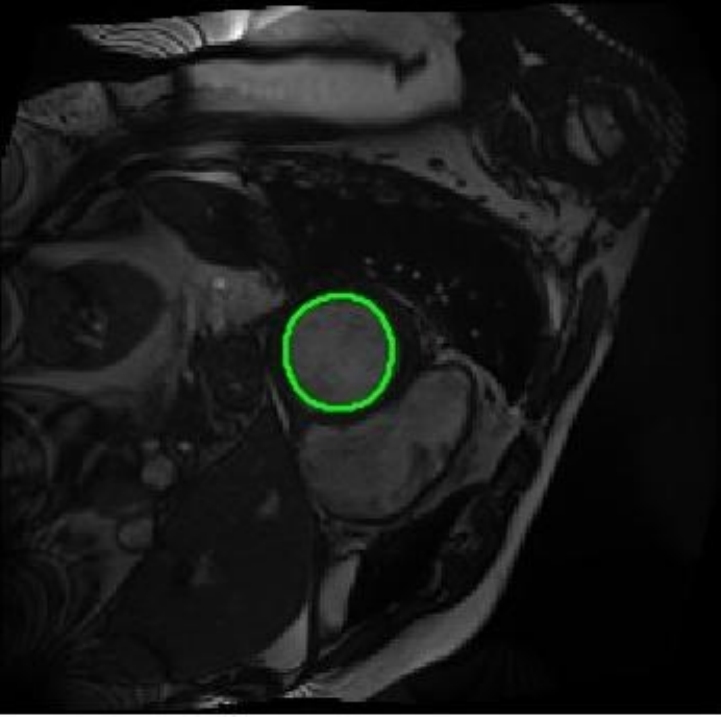}%
		\label{fig28_case}}
	\subfloat{\includegraphics[width=0.8in,height=0.8in]{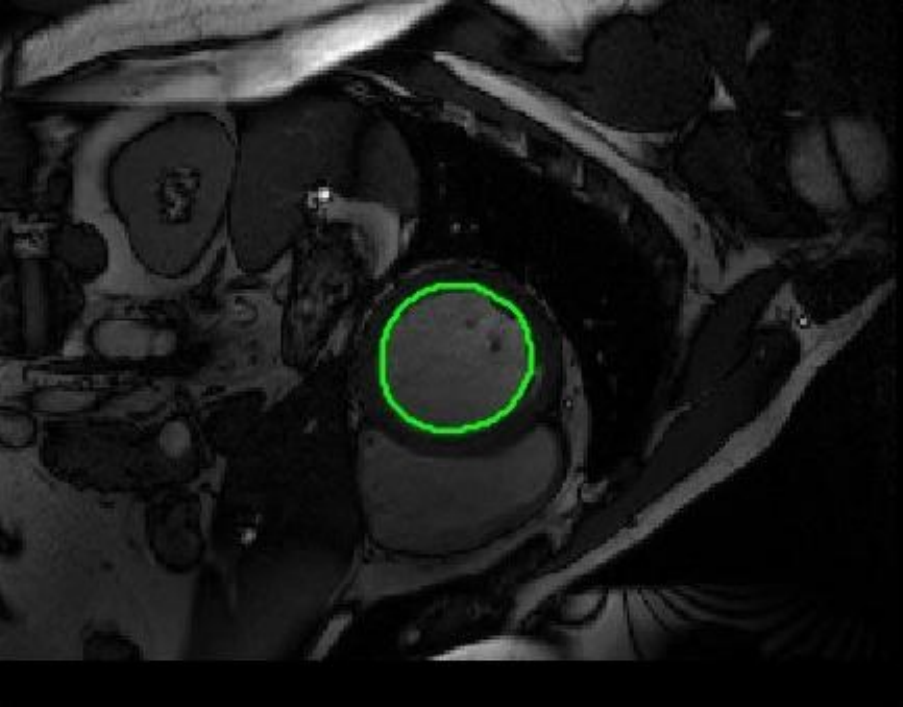}%
		\label{fig29_case}}
	\subfloat{\includegraphics[width=0.8in,height=0.8in]{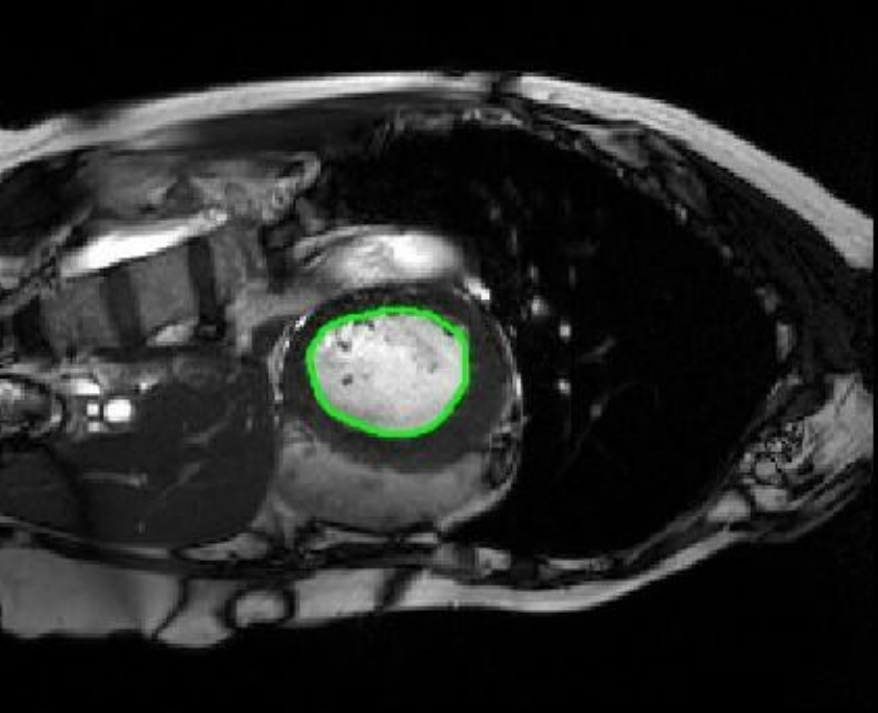}%
		\label{fig30_case}}
	\subfloat{\includegraphics[width=0.8in,height=0.8in]{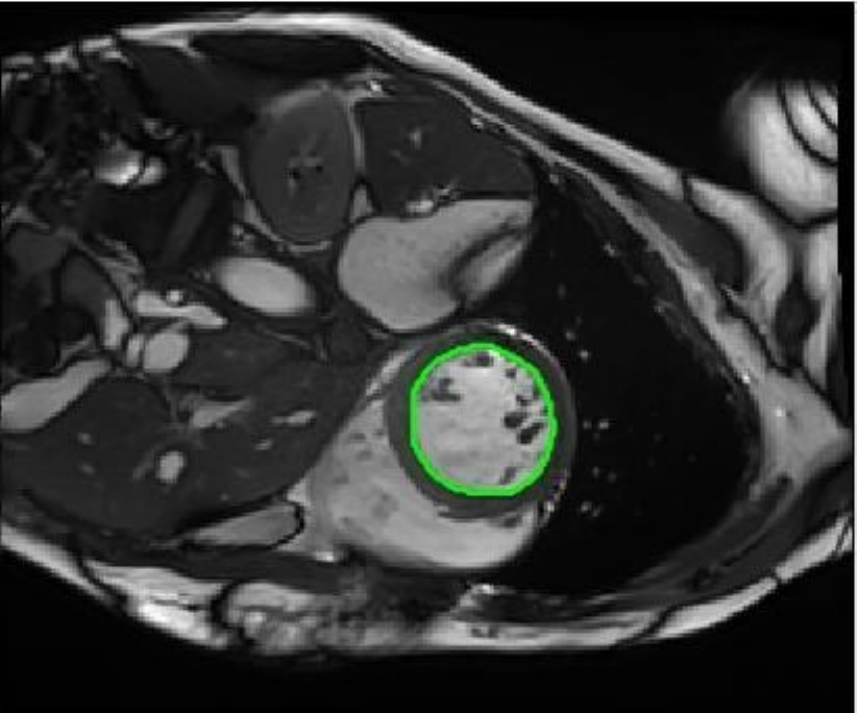}%
		\label{fig31_case}}
	\hfil
	\subfloat{\includegraphics[width=0.8in,height=0.8in]{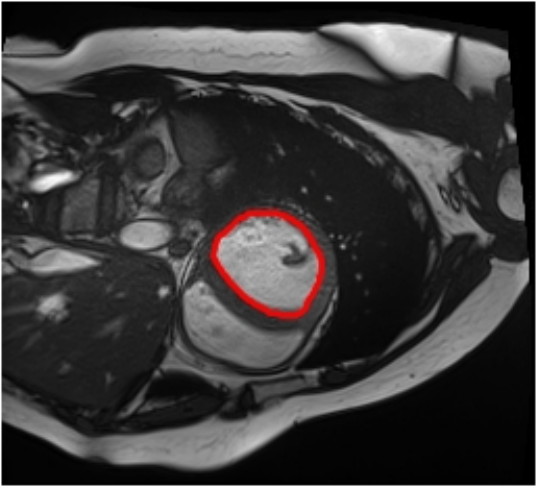}%
		\label{fig32_case}}
	\subfloat{\includegraphics[width=0.8in,height=0.8in]{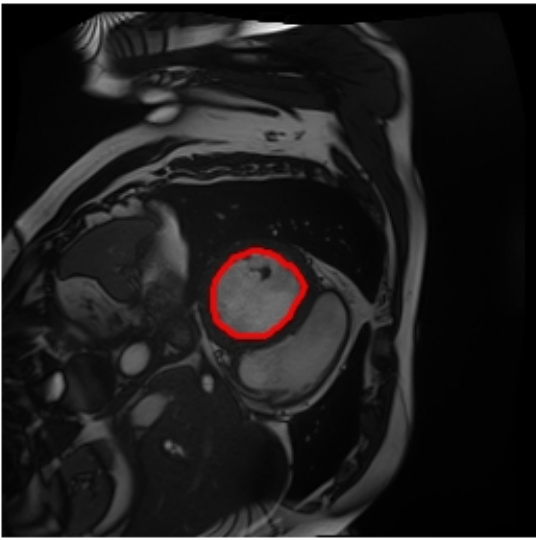}%
		\label{fig33_case}}
	\subfloat{\includegraphics[width=0.8in,height=0.8in]{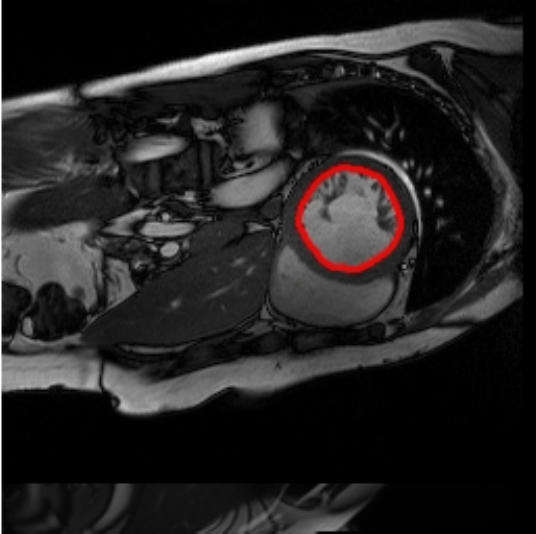}%
		\label{fig34_case}}
	\subfloat{\includegraphics[width=0.8in,height=0.8in]{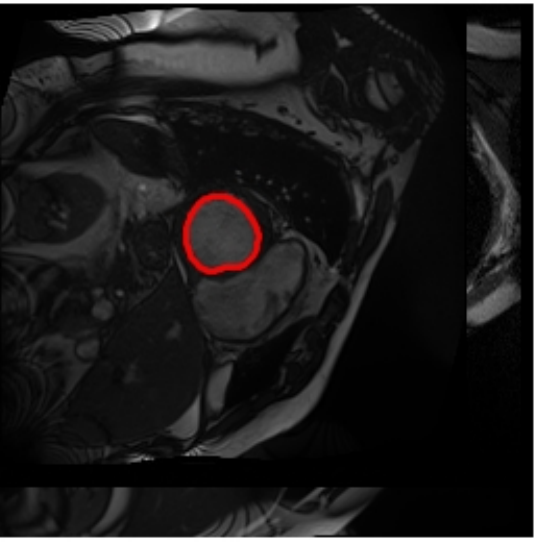}%
		\label{fig35_case}}
	\subfloat{\includegraphics[width=0.8in,height=0.8in]{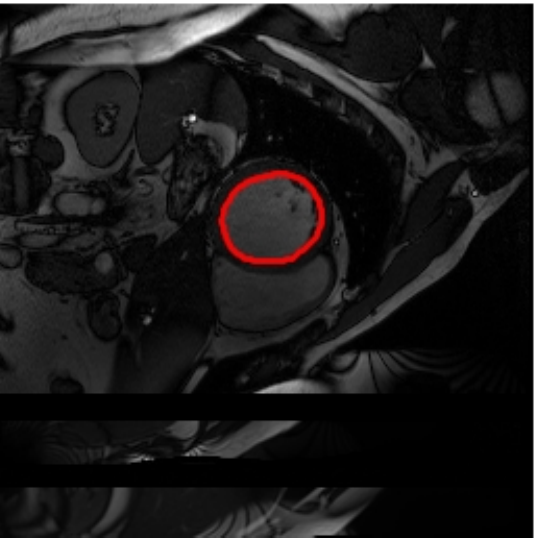}%
		\label{fig36_case}}
	\subfloat{\includegraphics[width=0.8in,height=0.8in]{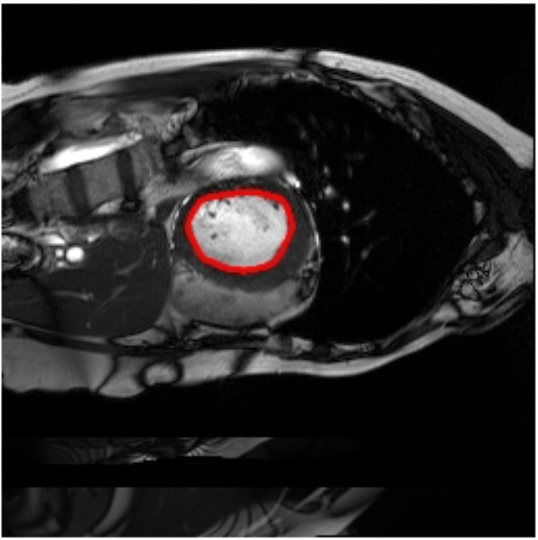}%
		\label{fig37_case}}
	\subfloat{\includegraphics[width=0.8in,height=0.8in]{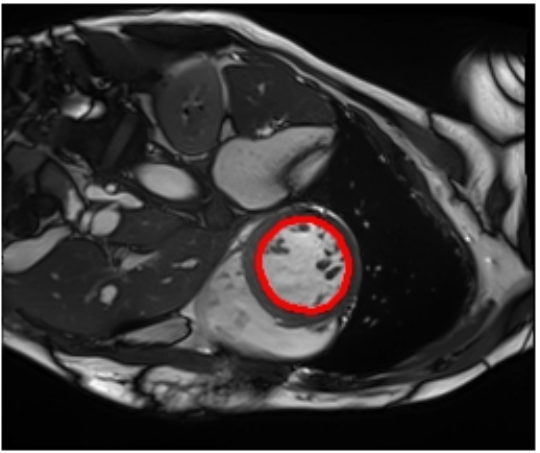}%
		\label{fig38_case}}
	\hfil
	\caption{Segmentation results from evaluated models for left ventricle. Row 1-4: Results from the DRLSE-ADMM model, ABC model, RESLS model and RefLSM. Row 5: Ground truth. }
	\label{img6}
\end{figure*}
\begin{figure}[!ht]
	\centering
	\subfloat[]{\includegraphics[width=2.5in]{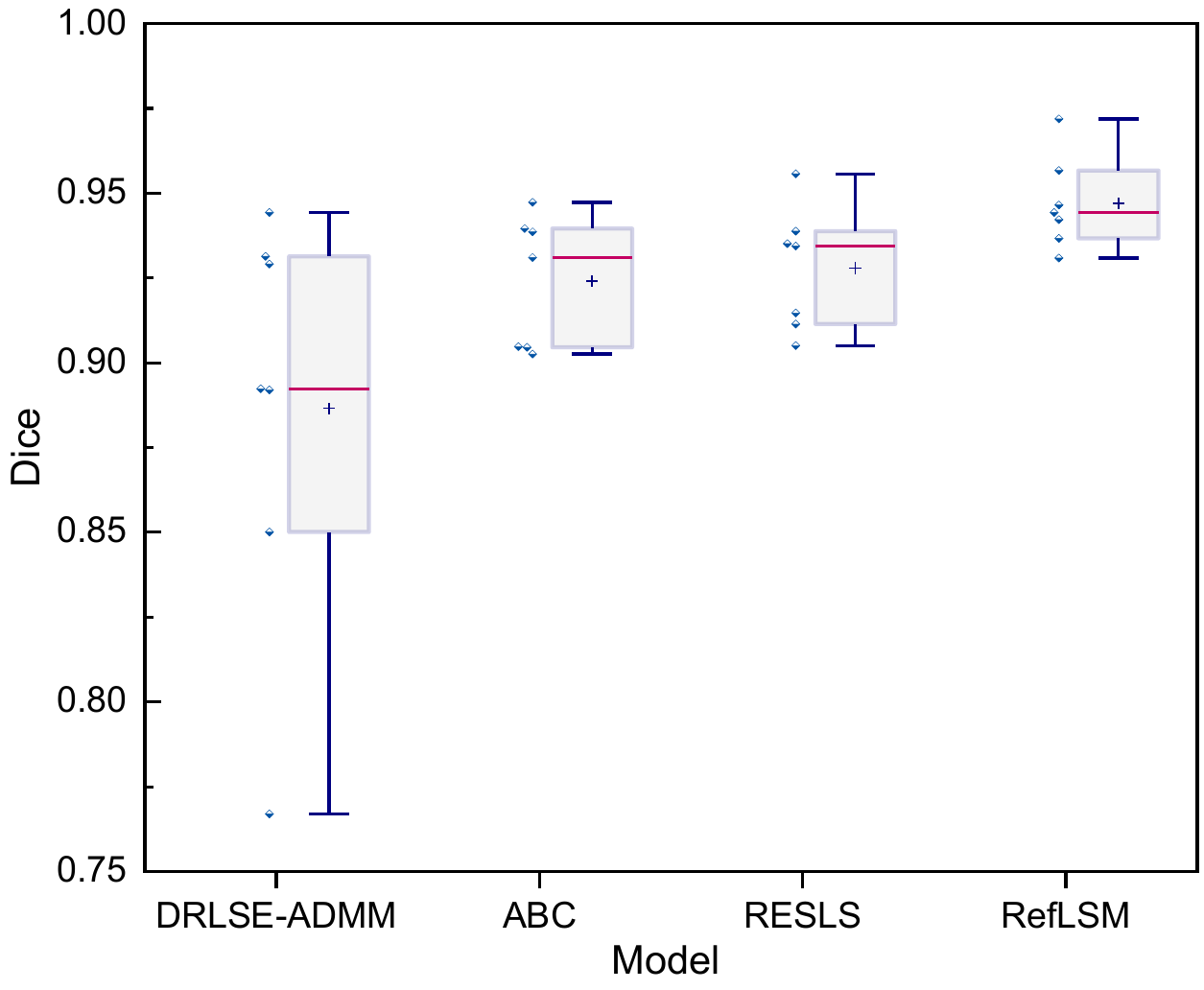}%
		\label{fig_112_case}}
	\subfloat[]{\includegraphics[width=2.5in]{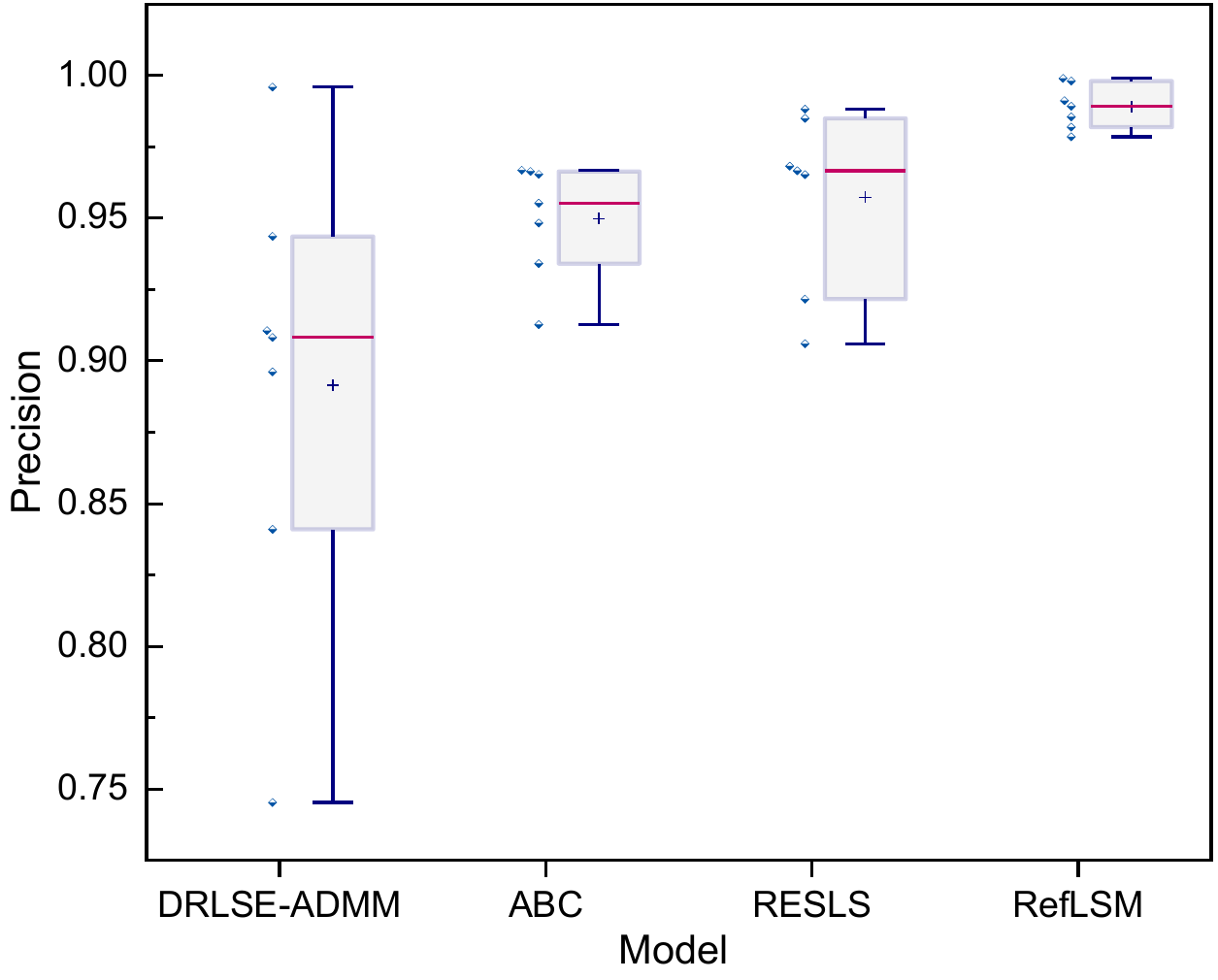}%
		\label{fig_114_case}}
	\caption{Dice values and Precision values of different models for left ventricle.}
	\label{img18}
\end{figure}

\begin{figure*}[!ht]
	\centering
	\subfloat{\includegraphics[width=0.8in,height=0.8in]{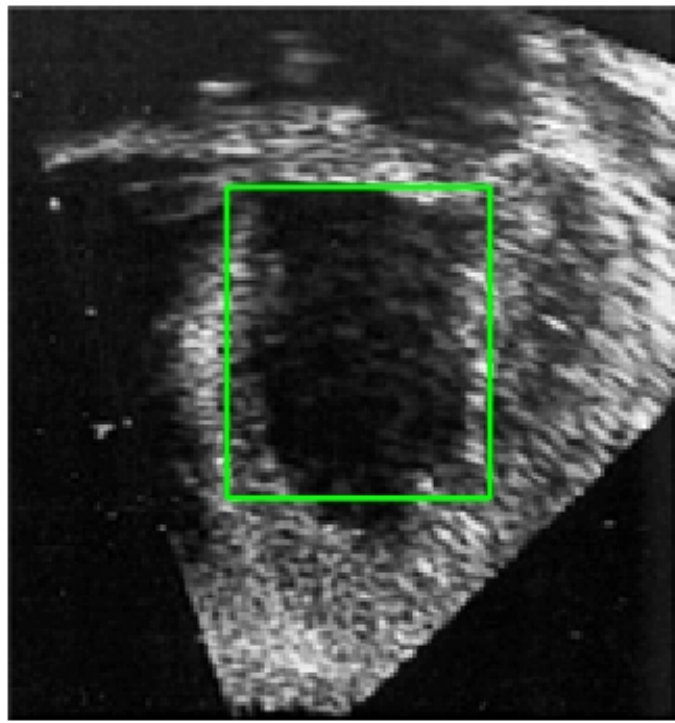}%
		\label{fig61_case}}\vspace{-3mm}\hspace{-1.5mm}
	\subfloat{\includegraphics[width=0.8in,height=0.8in]{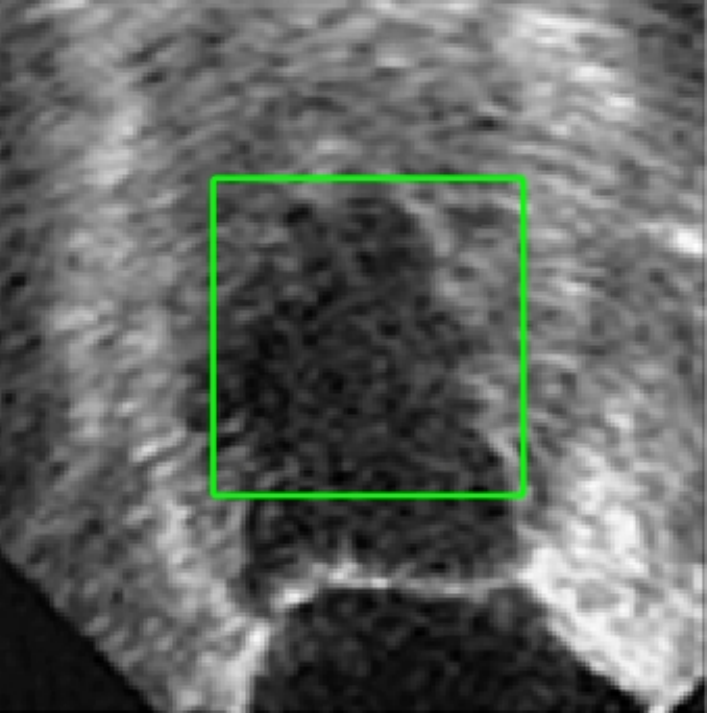}%
		\label{fig62_case}}
	\subfloat{\includegraphics[width=0.8in,height=0.8in]{a2_in.pdf}%
		\label{fig63_case}}
	\subfloat{\includegraphics[width=0.8in,height=0.8in]{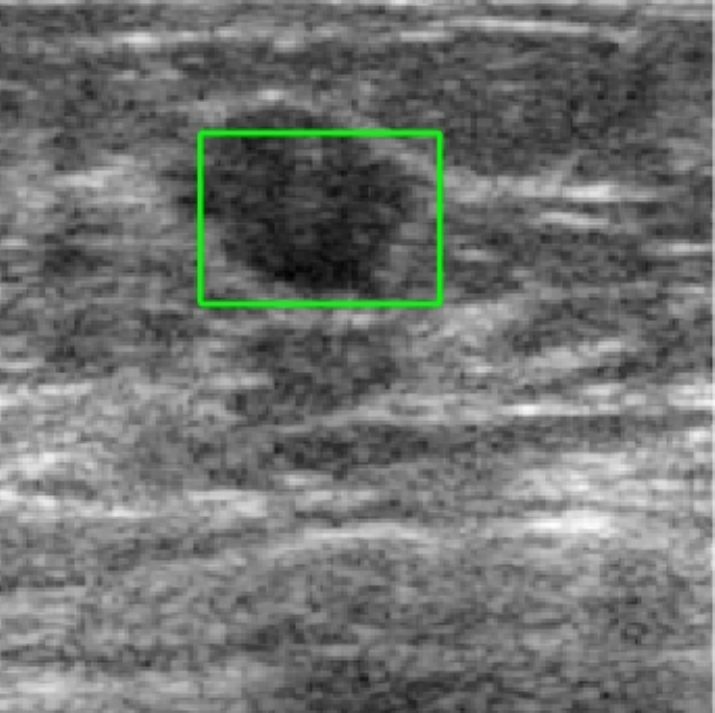}%
		\label{fig64_case}}
	\subfloat{\includegraphics[width=0.8in,height=0.8in]{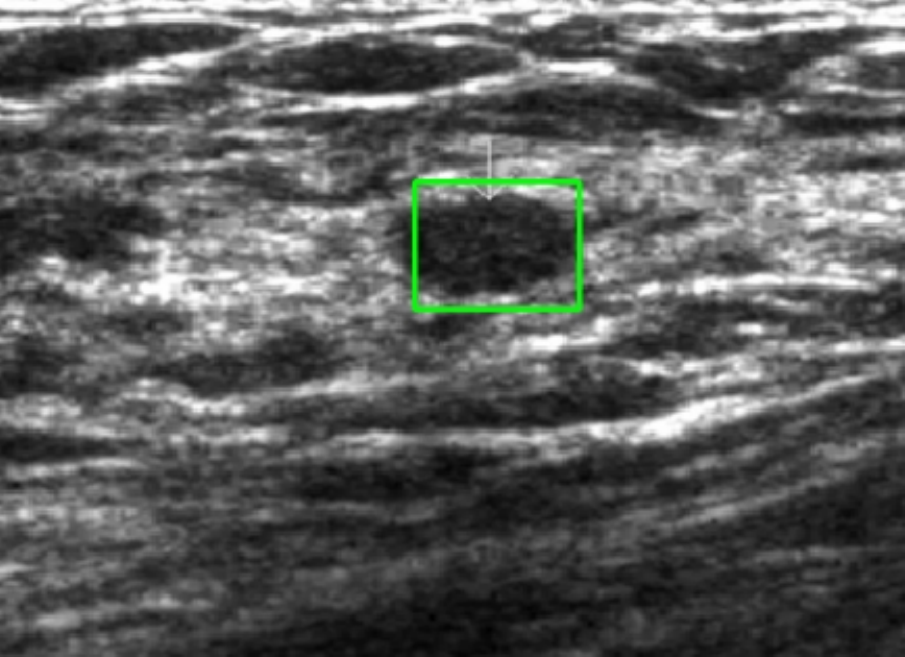}%
		\label{fig65_case}}
	\subfloat{\includegraphics[width=0.8in,height=0.8in]{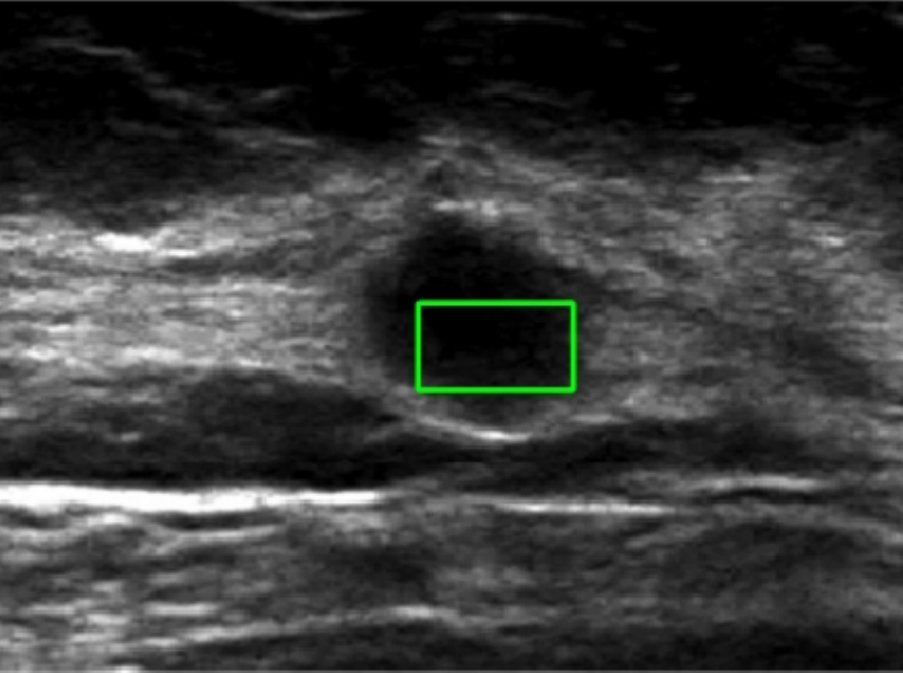}%
		\label{fig66_case}}
	\subfloat{\includegraphics[width=0.8in,height=0.8in]{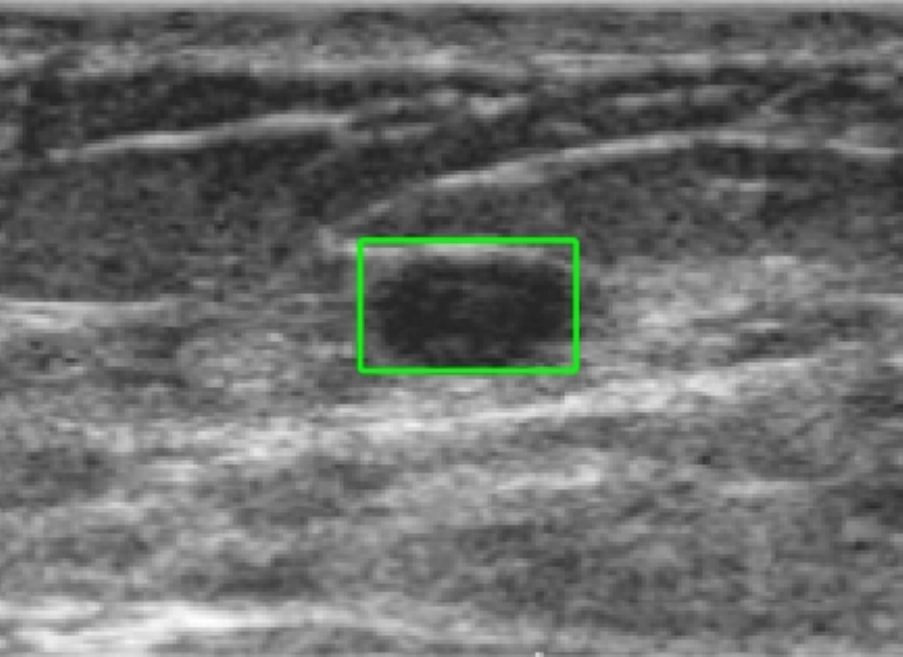}%
		\label{fig67_case}}
	\subfloat{\includegraphics[width=0.8in,height=0.8in]{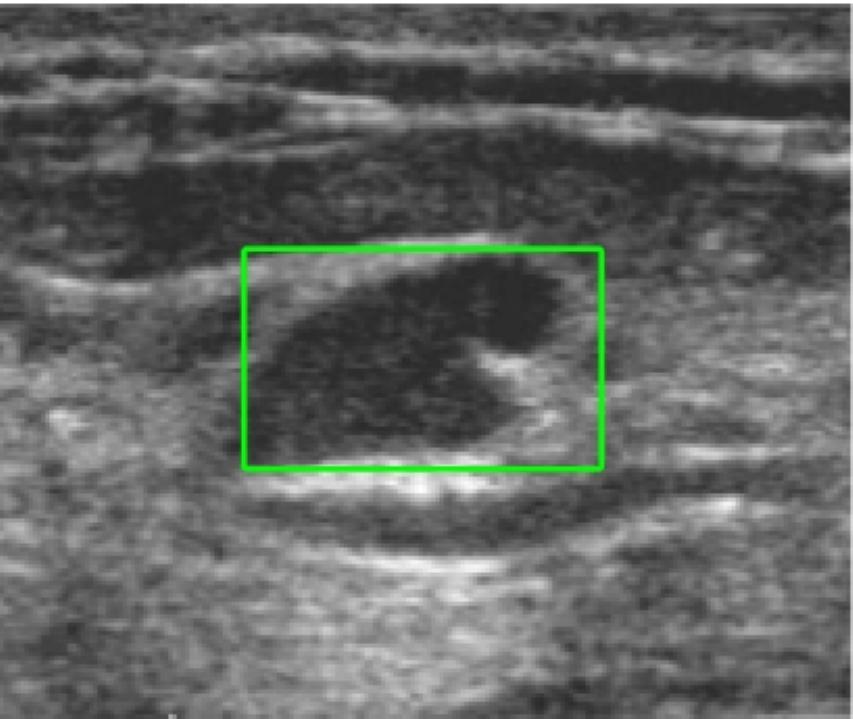}%
		\label{fig68_case}}
	\hfil
	
	\subfloat{\includegraphics[width=0.8in,height=0.8in]{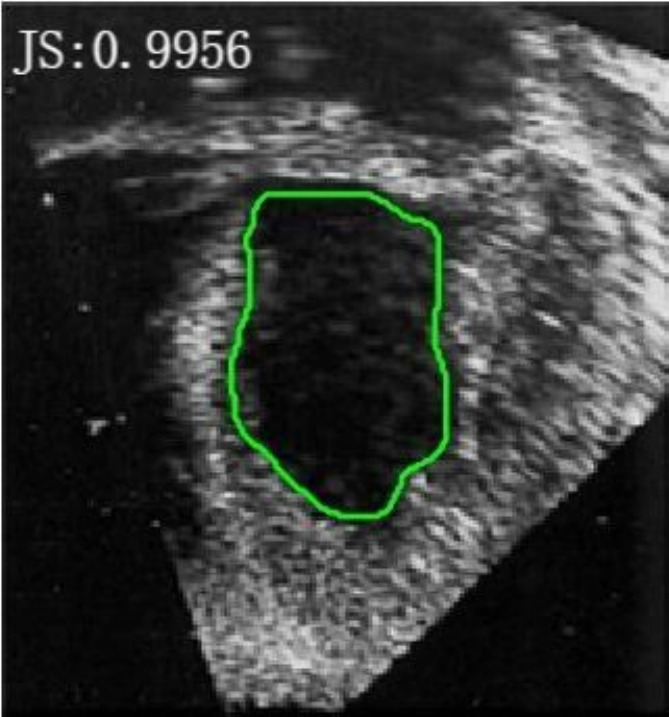}%
		\label{fig69_case}}
	\subfloat{\includegraphics[width=0.8in,height=0.8in]{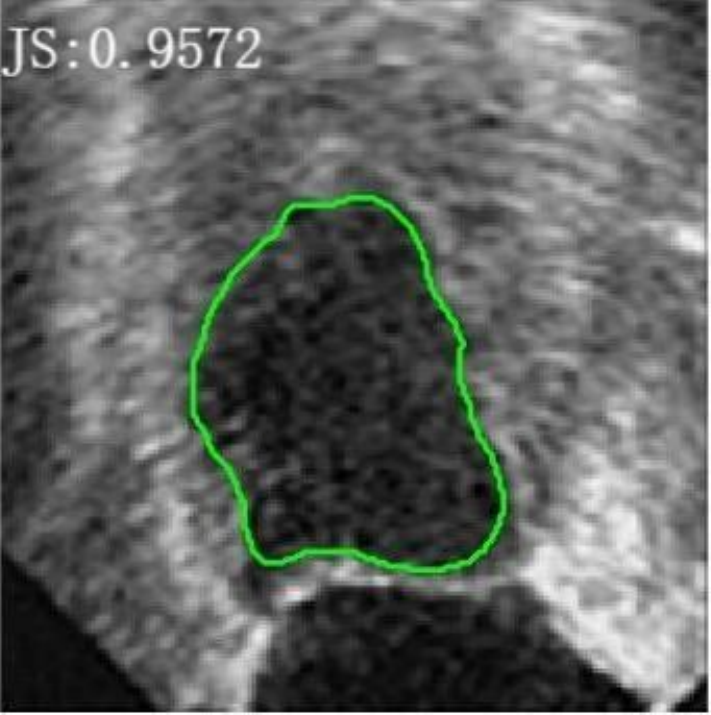}%
		\label{fig70_case}}
	\subfloat{\includegraphics[width=0.8in,height=0.8in]{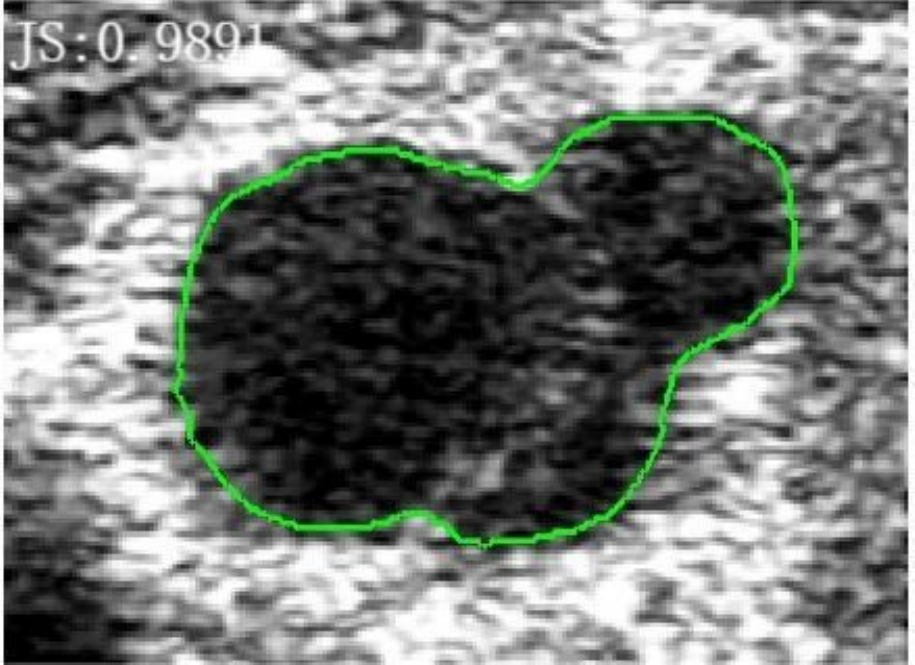}%
		\label{fig71_case}}
	\subfloat{\includegraphics[width=0.8in,height=0.8in]{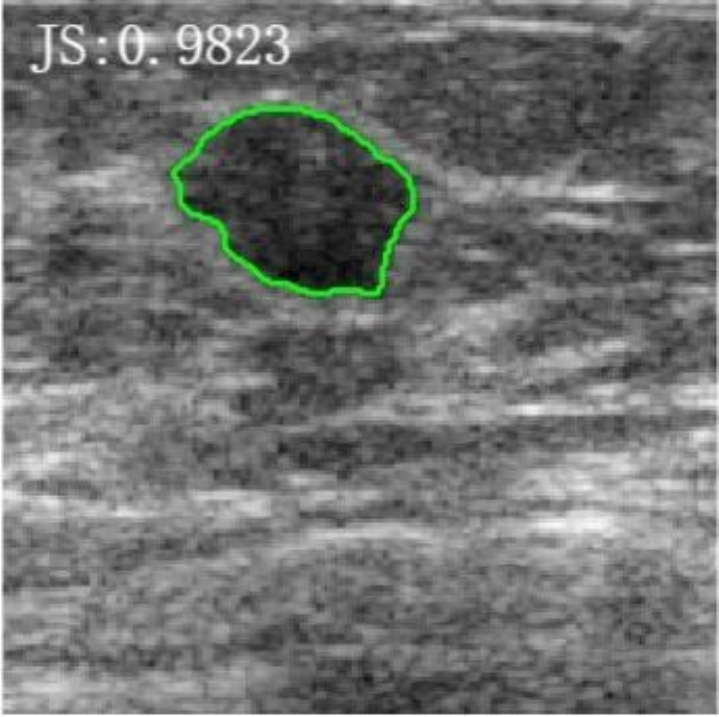}%
		\label{fig72_case}}
	\subfloat{\includegraphics[width=0.8in,height=0.8in]{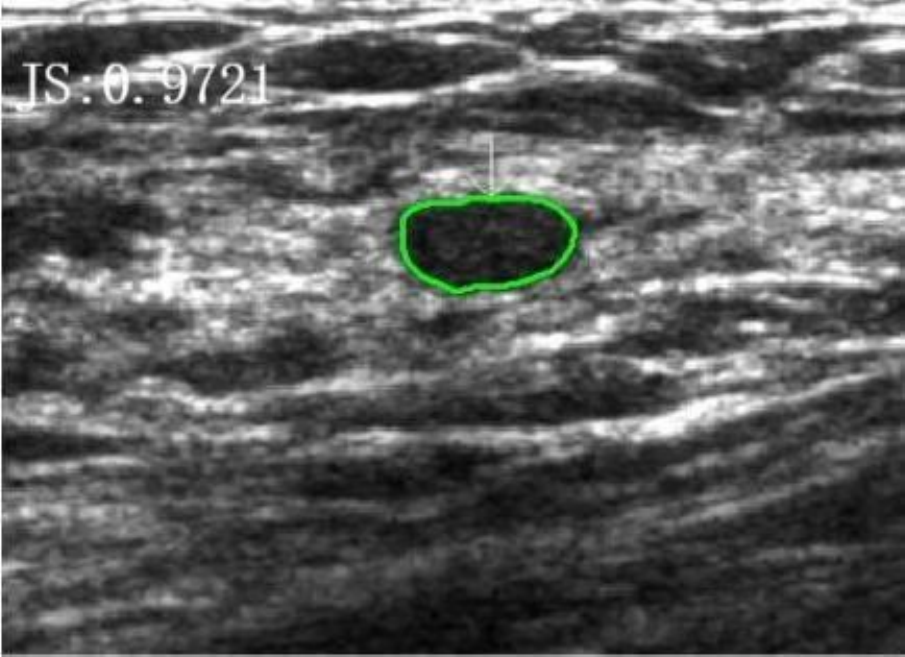}%
		\label{fig73_case}}
	\subfloat{\includegraphics[width=0.8in,height=0.8in]{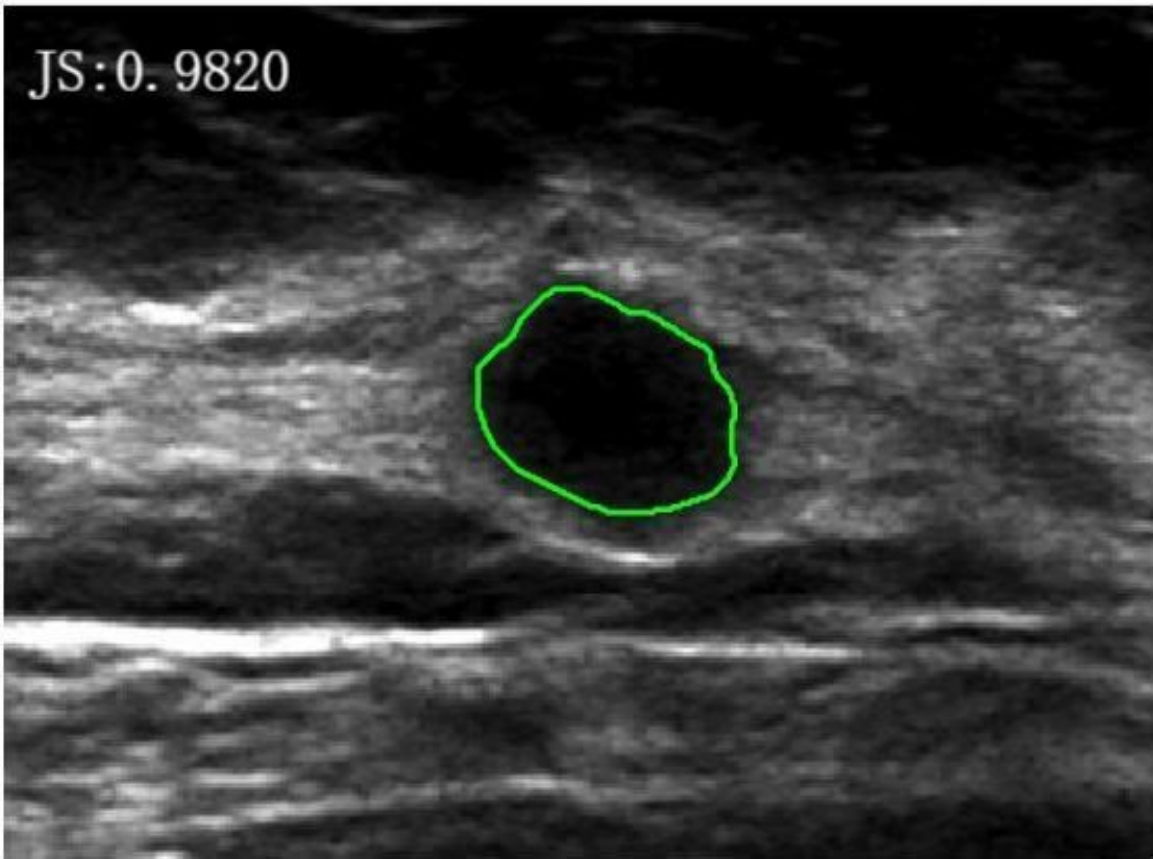}%
		\label{fig74_case}}
	\subfloat{\includegraphics[width=0.8in,height=0.8in]{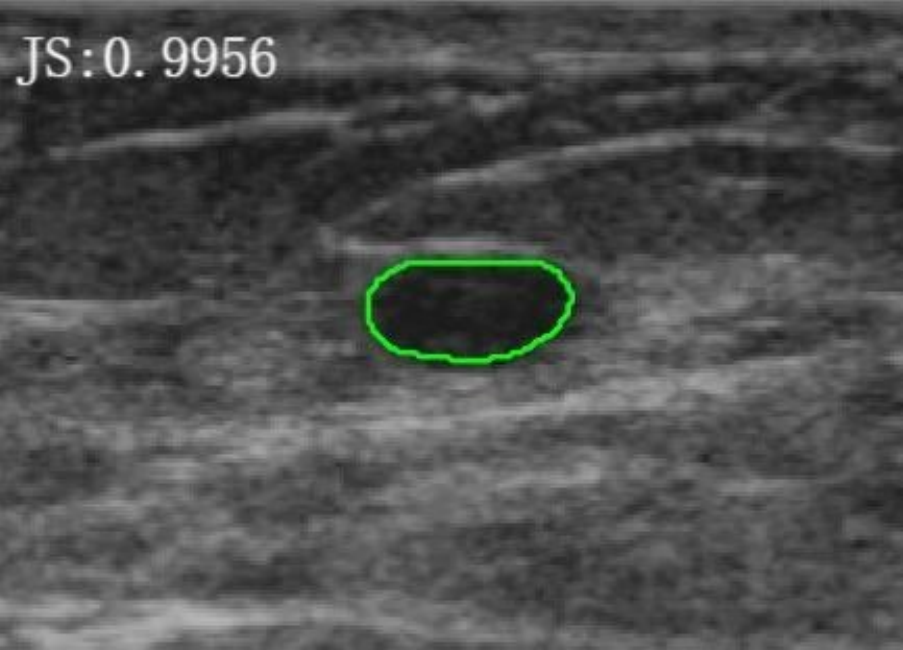}%
		\label{fig75_case}}
	\subfloat{\includegraphics[width=0.8in,height=0.8in]{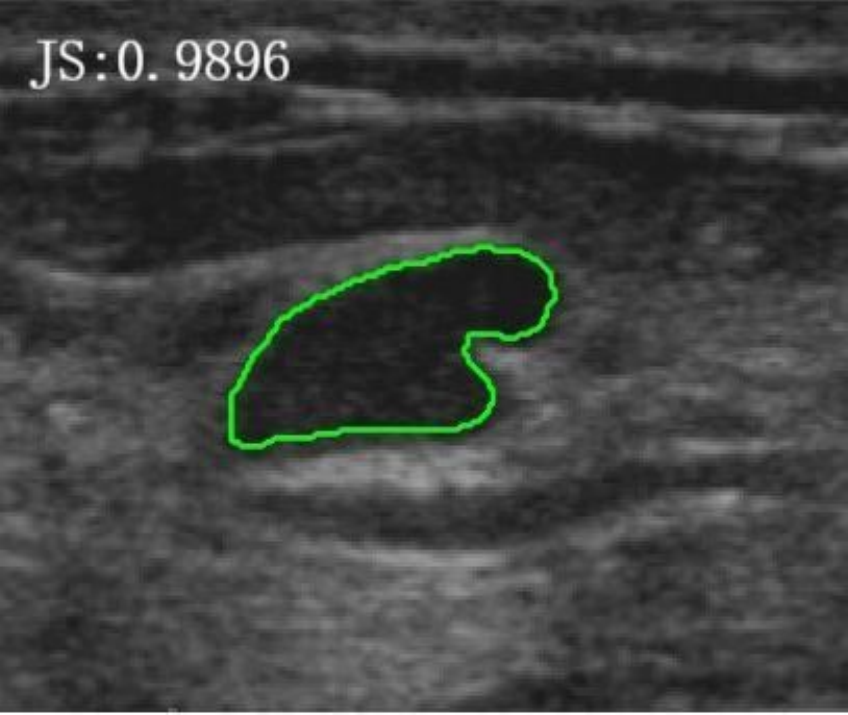}%
		\label{fig76_case}}
	\hfil
	\caption{Segmentation results from the RefLSM for ultrasound images. Row 1: Original images and initial contours. Row 2: Results from the RefLSM.}
	\label{img7}
\end{figure*}
\begin{figure*}[!ht]
	\centering
	\subfloat{\includegraphics[width=0.8in,height=0.7in]{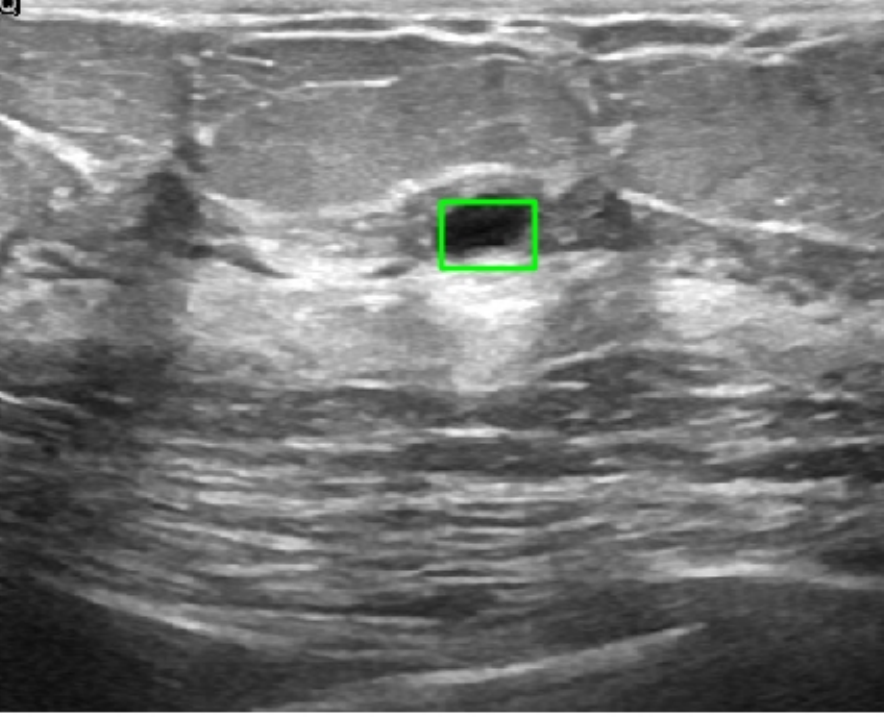}%
		\label{fig_11}}\vspace{-3mm}\hspace{-1.5mm}
	\subfloat{\includegraphics[width=0.8in,height=0.7in]{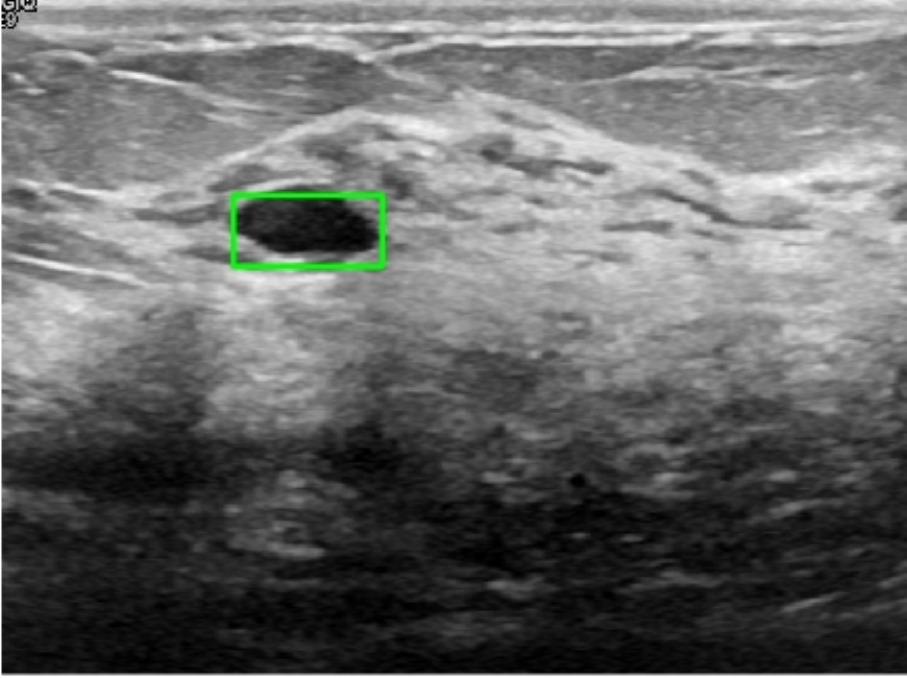}%
		\label{fig_12}}
	\subfloat{\includegraphics[width=0.8in,height=0.7in]{ 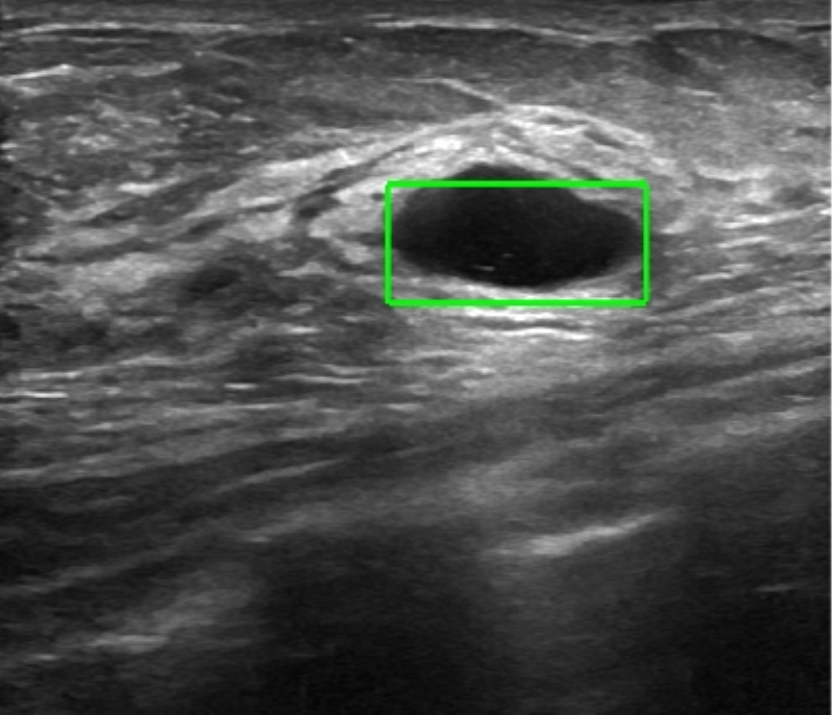}%
		\label{fig_13}}
	\subfloat{\includegraphics[width=0.8in,height=0.7in]{ 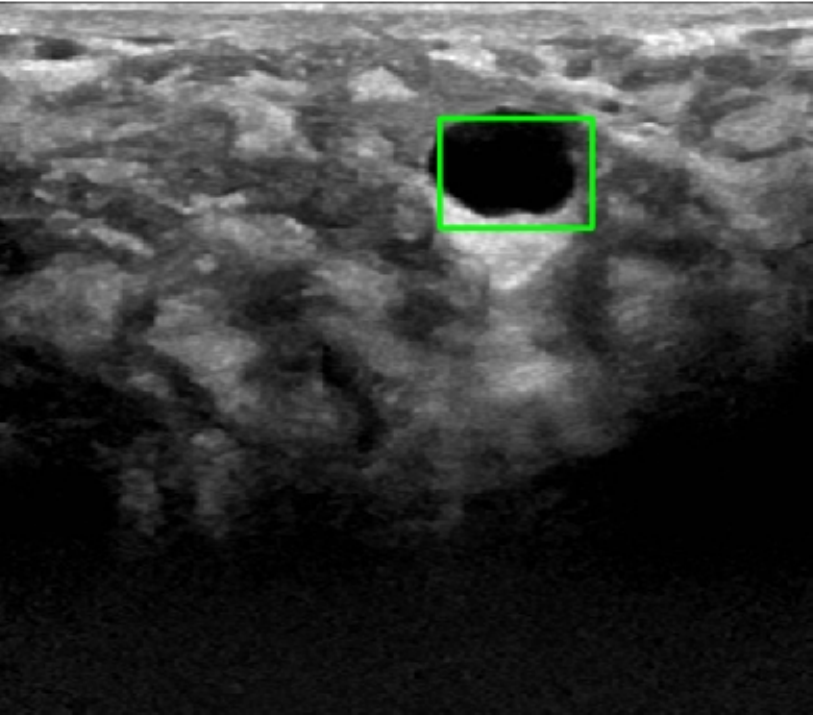}%
		\label{fig_14}}
	\subfloat{\includegraphics[width=0.8in,height=0.7in]{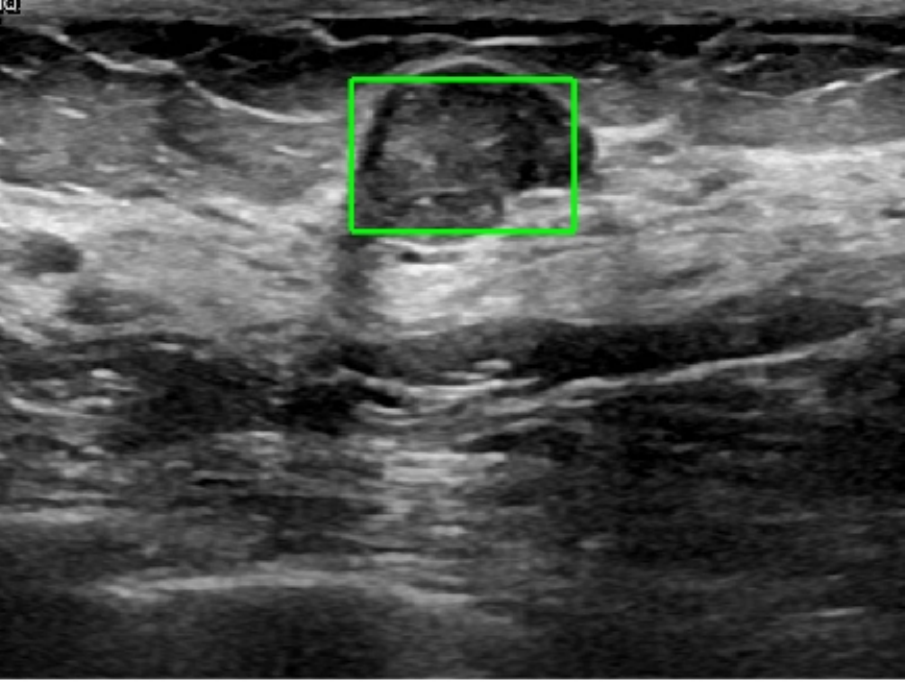}%
		\label{fig_15}}
	\subfloat{\includegraphics[width=0.8in,height=0.7in]{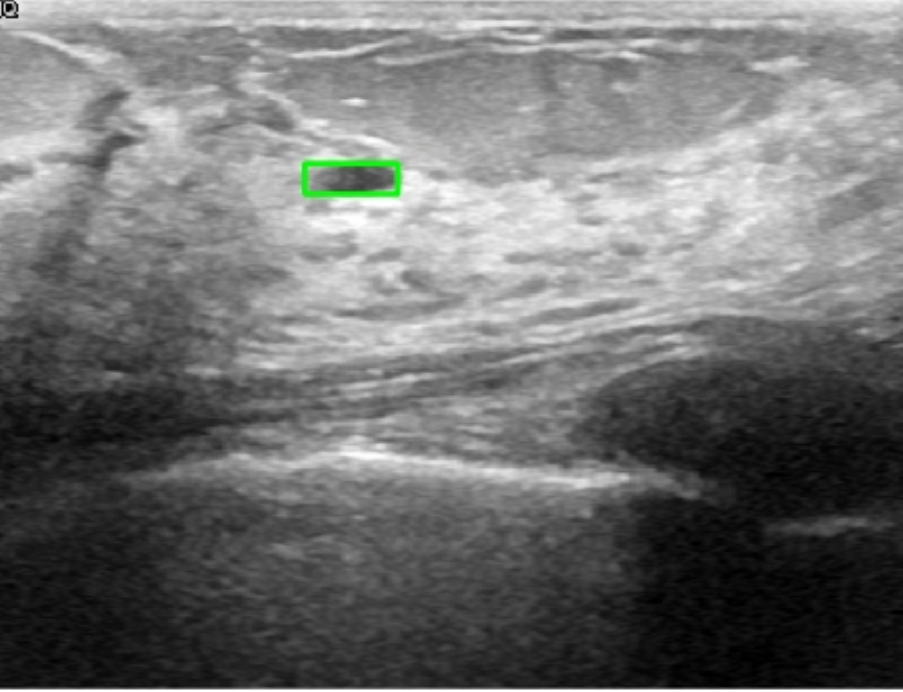}%
		\label{fig_16}}
	\subfloat{\includegraphics[width=0.8in,height=0.7in]{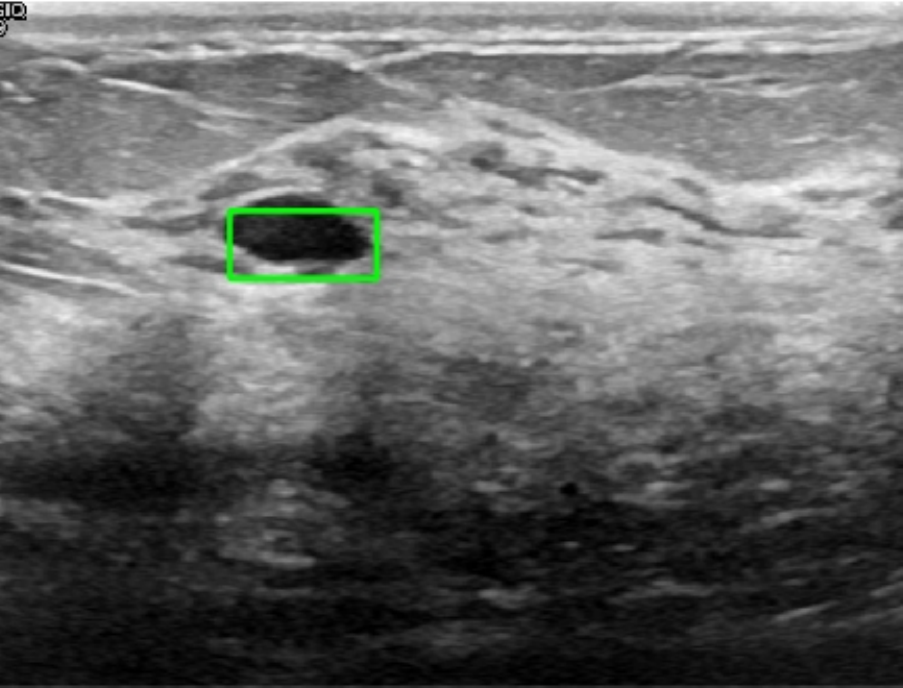}%
		\label{fig_17}}
	\subfloat{\includegraphics[width=0.8in,height=0.7in]{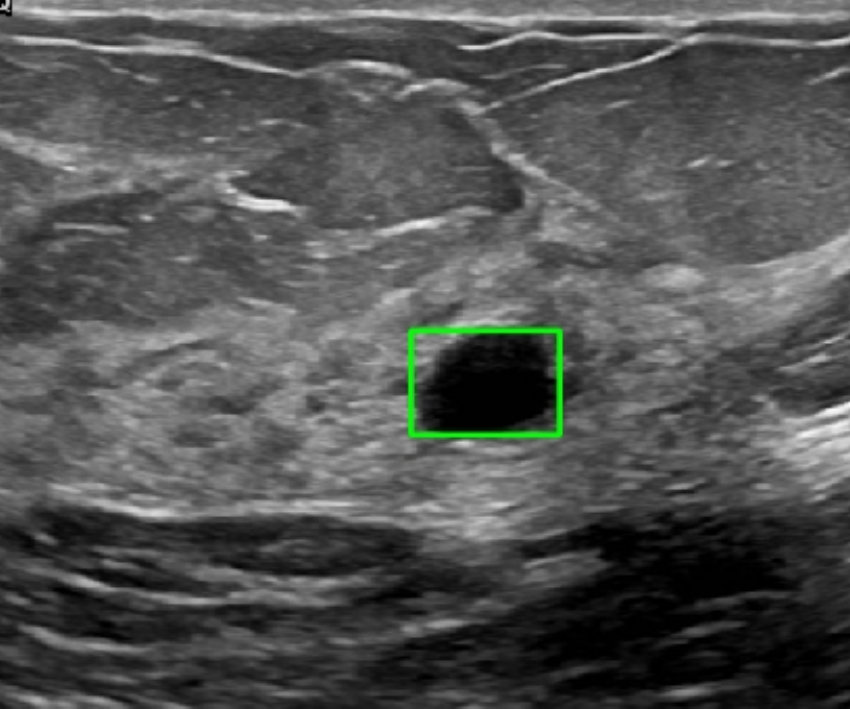}%
		\label{fig_18}}
	\hfil
	\subfloat{\includegraphics[width=0.8in,height=0.7in]{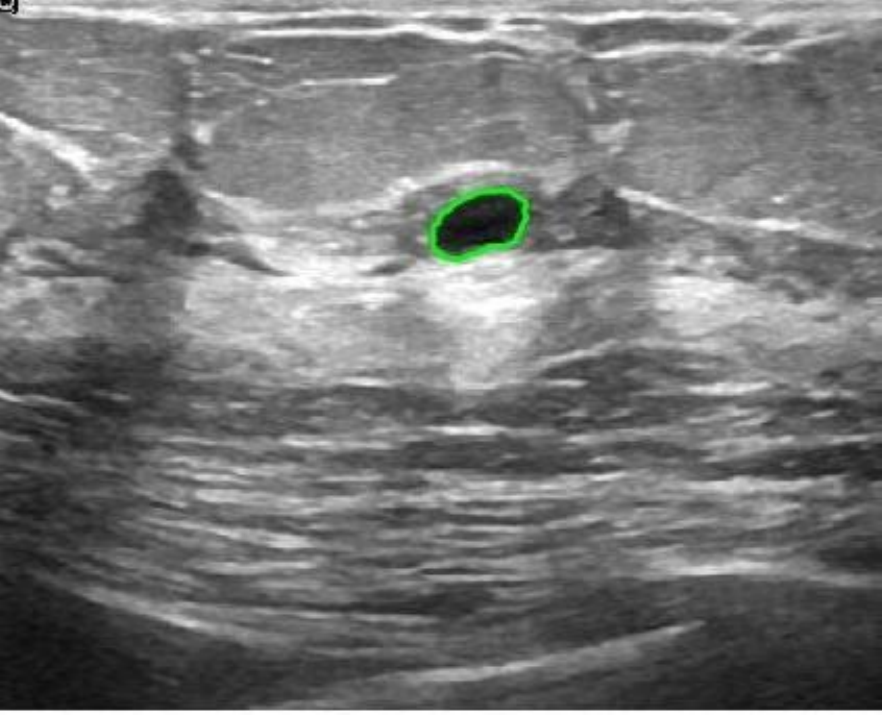}%
		\label{fig_19}}\vspace{-3mm}\hspace{-1.5mm}
	\subfloat{\includegraphics[width=0.8in,height=0.7in]{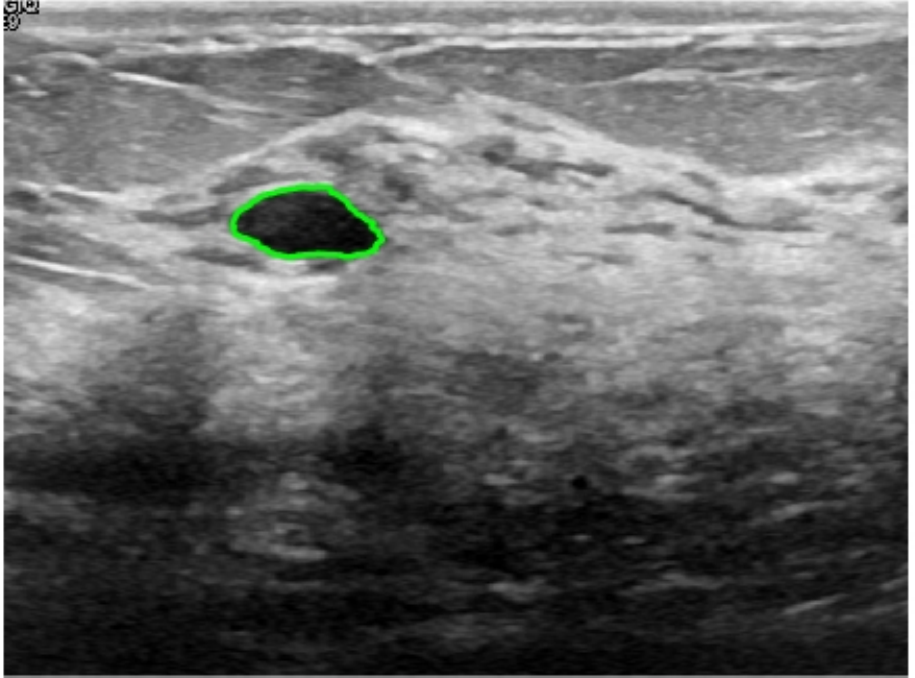}%
		\label{fig_20}}
	\subfloat{\includegraphics[width=0.8in,height=0.7in]{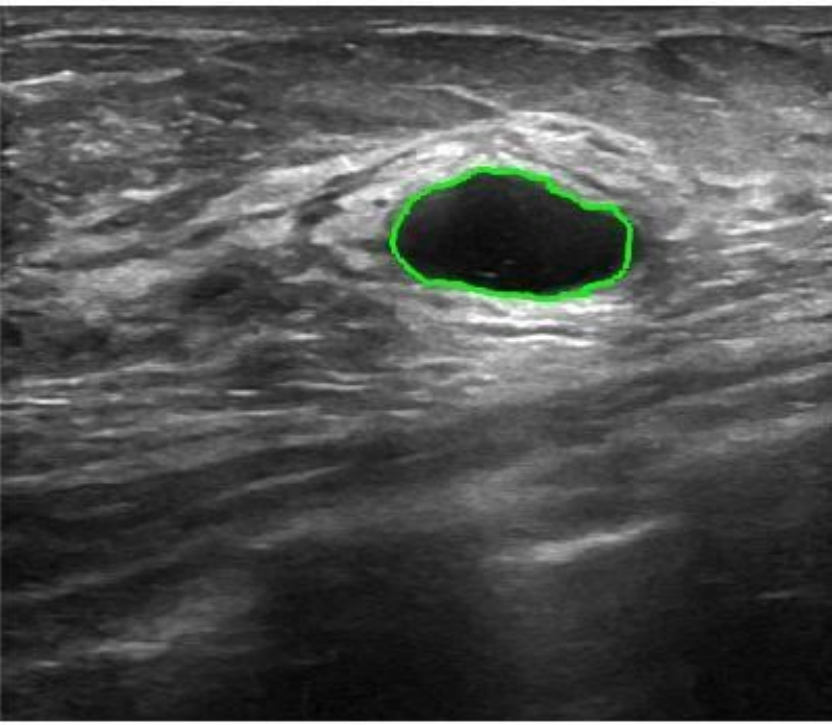}%
		\label{fig_21}}
	\subfloat{\includegraphics[width=0.8in,height=0.7in]{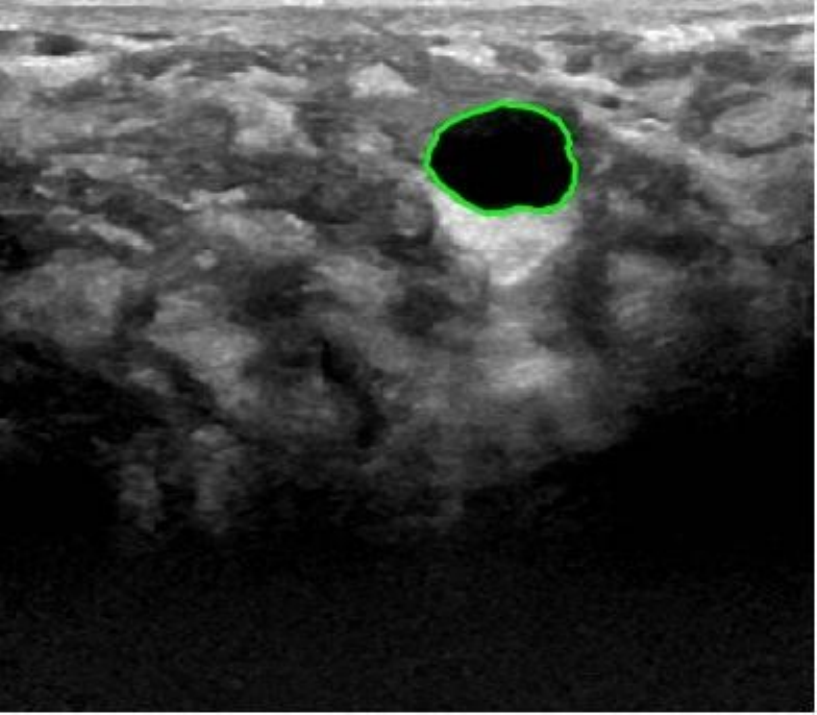}%
		\label{fig_22}}
	\subfloat{\includegraphics[width=0.8in,height=0.7in]{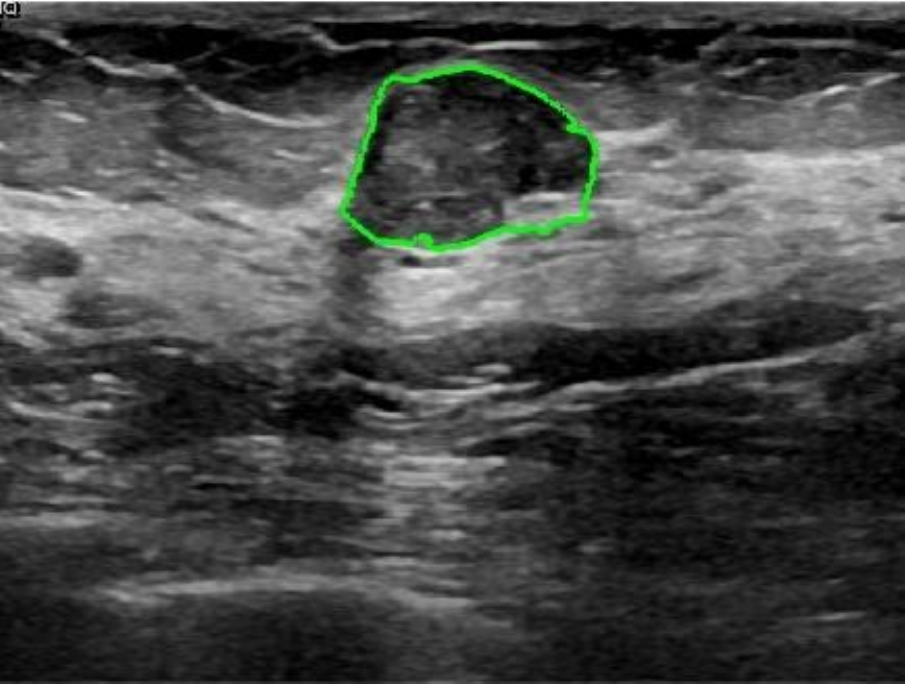}%
		\label{fig_23}}
	\subfloat{\includegraphics[width=0.8in,height=0.7in]{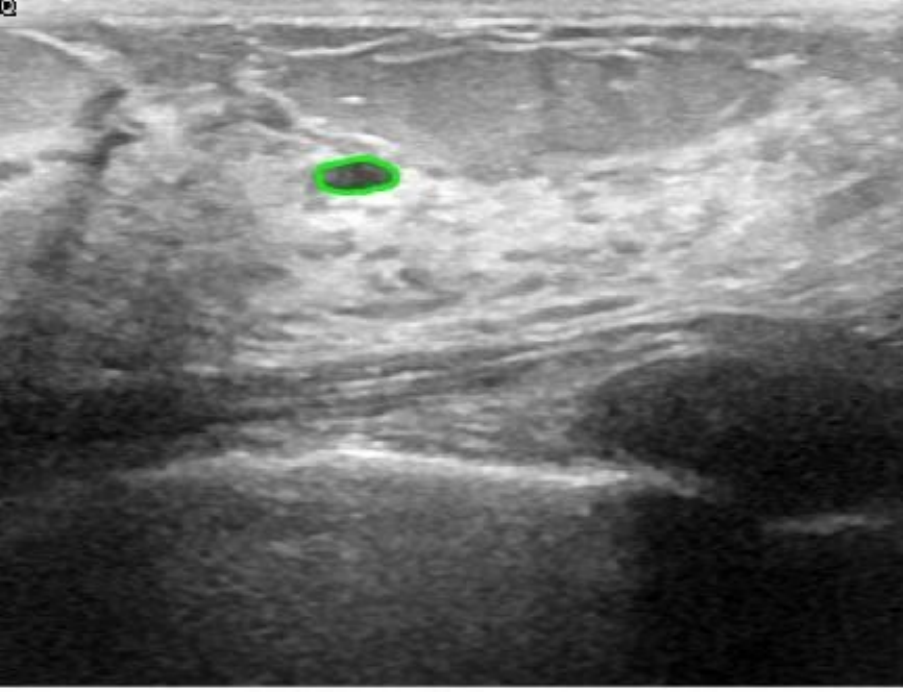}%
		\label{fig_24}}
	\subfloat{\includegraphics[width=0.8in,height=0.7in]{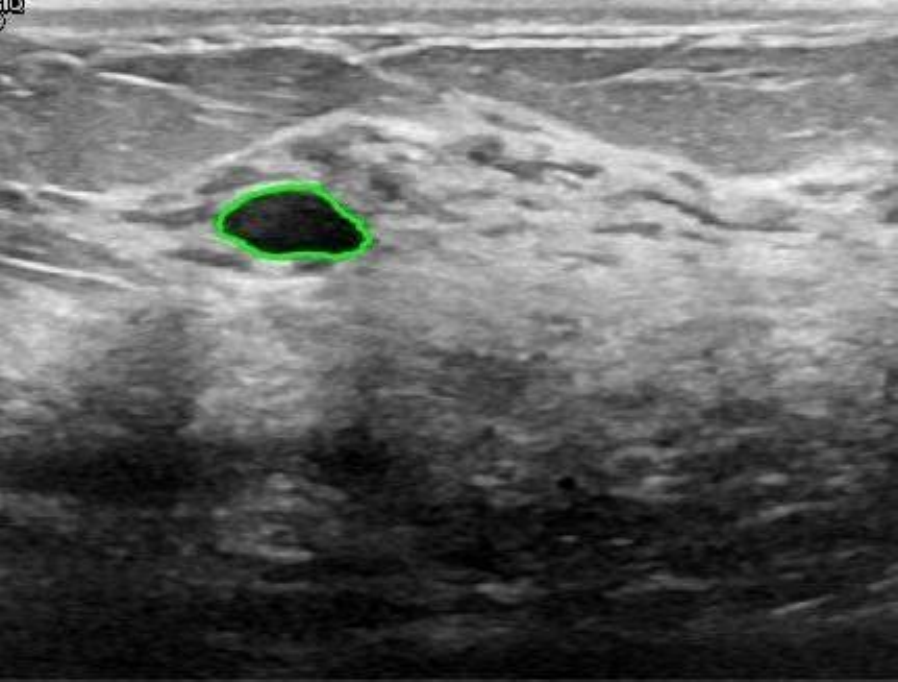}%
		\label{fig_25}}
	\subfloat{\includegraphics[width=0.8in,height=0.7in]{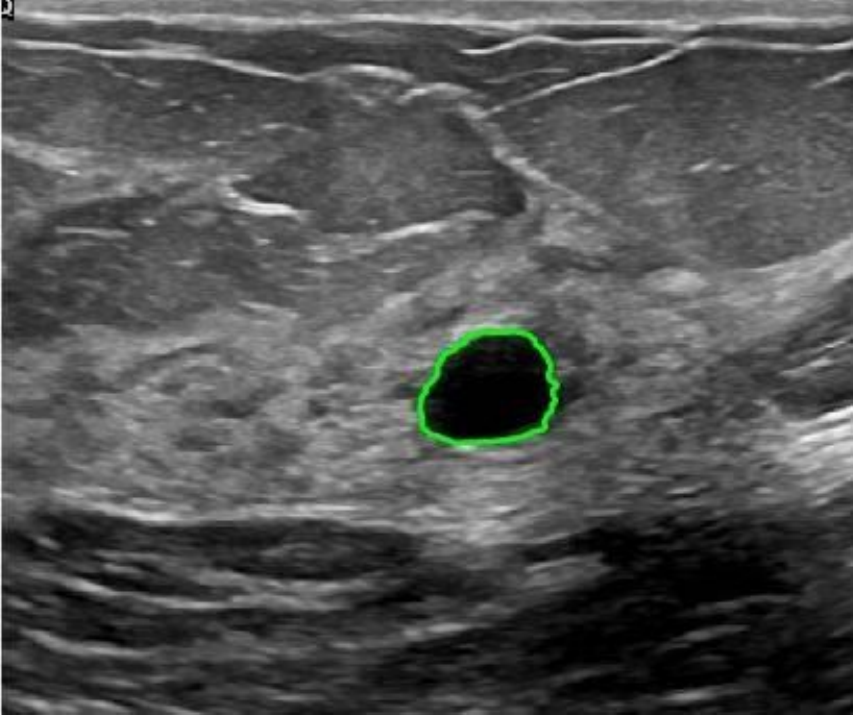}%
		\label{fig_26}}
	\hfil
	\subfloat{\includegraphics[width=0.8in,height=0.7in]{ 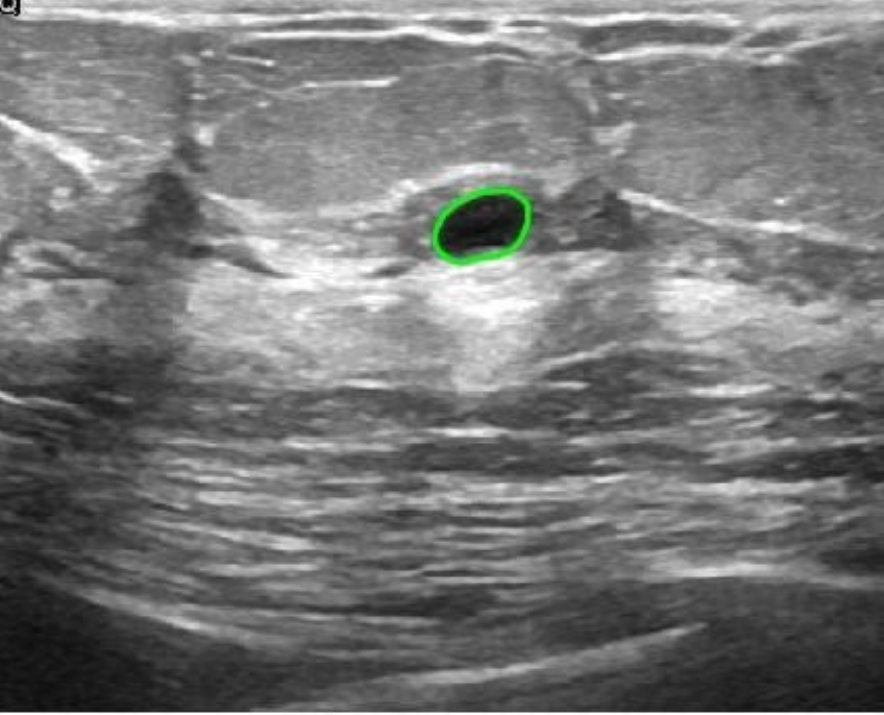}%
		\label{fig_27}}\vspace{-3mm}\hspace{-1.5mm}
	\subfloat{\includegraphics[width=0.8in,height=0.7in]{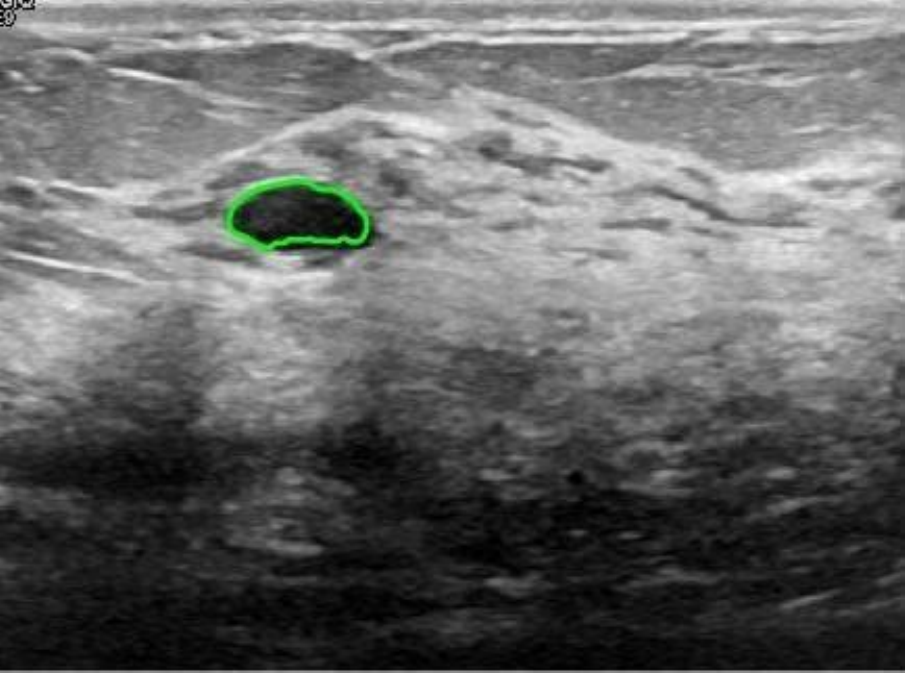}%
		\label{fig_28}}
	\subfloat{\includegraphics[width=0.8in,height=0.7in]{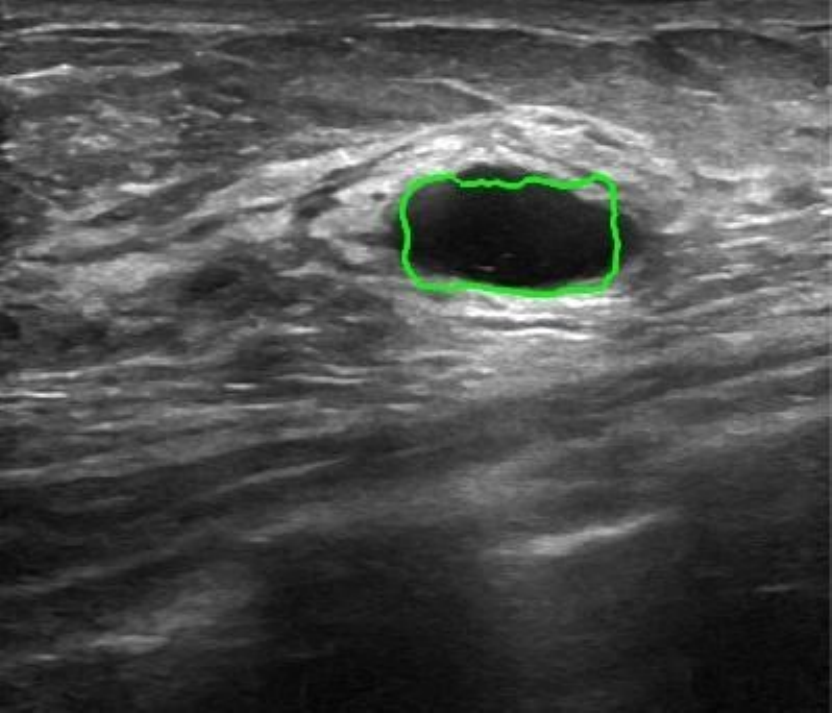}%
		\label{fig_29}}
	\subfloat{\includegraphics[width=0.8in,height=0.7in]{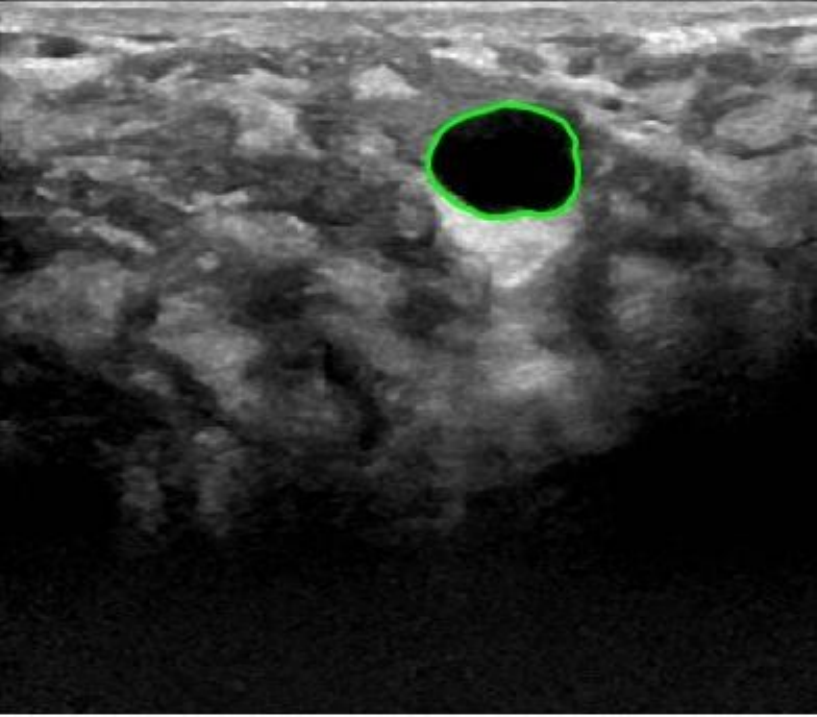}%
		\label{fig_30}}
	\subfloat{\includegraphics[width=0.795in,height=0.7in]{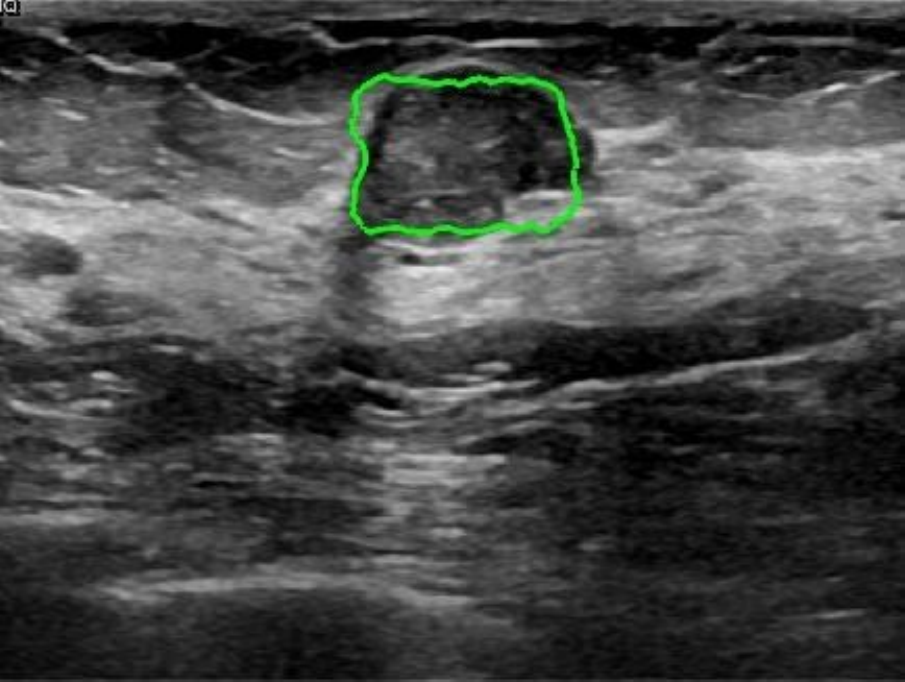}%
		\label{fig_31}}
	\subfloat{\includegraphics[width=0.8in,height=0.7in]{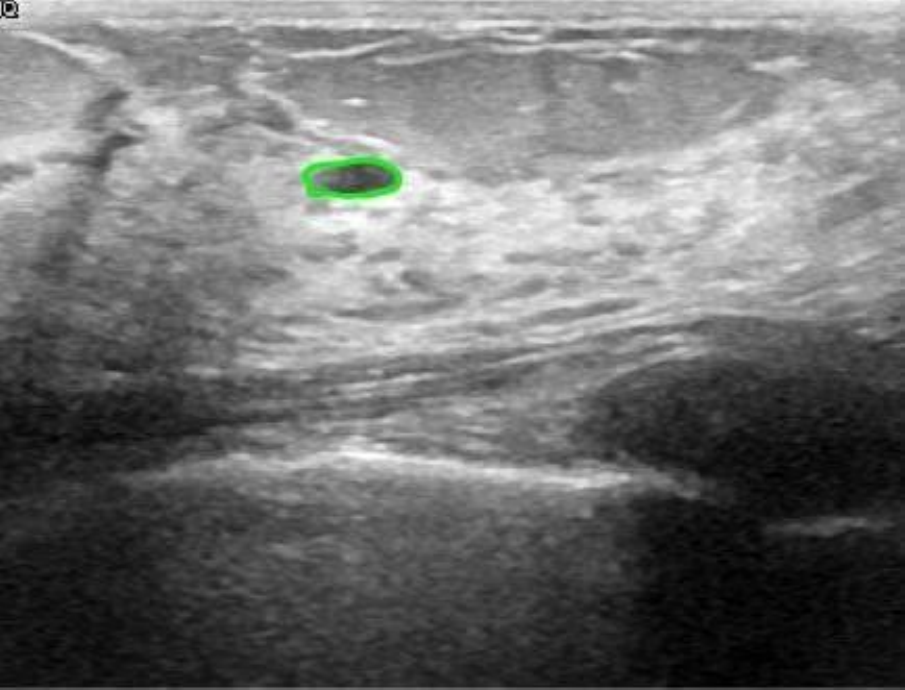}%
		\label{fig_32}}
	\subfloat{\includegraphics[width=0.8in,height=0.7in]{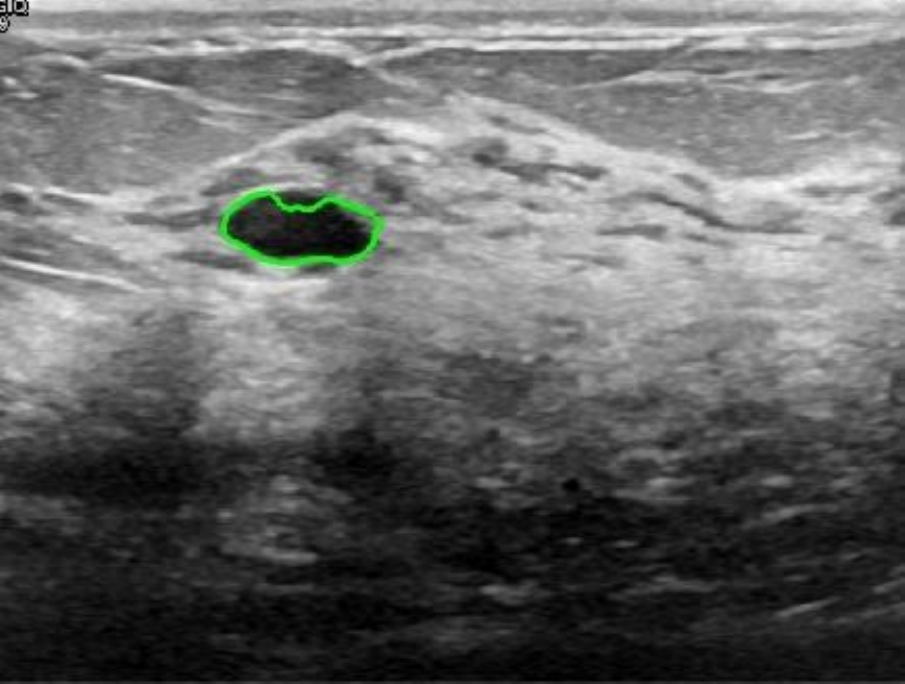}%
		\label{fig_33}}
	\subfloat{\includegraphics[width=0.8in,height=0.7in]{ 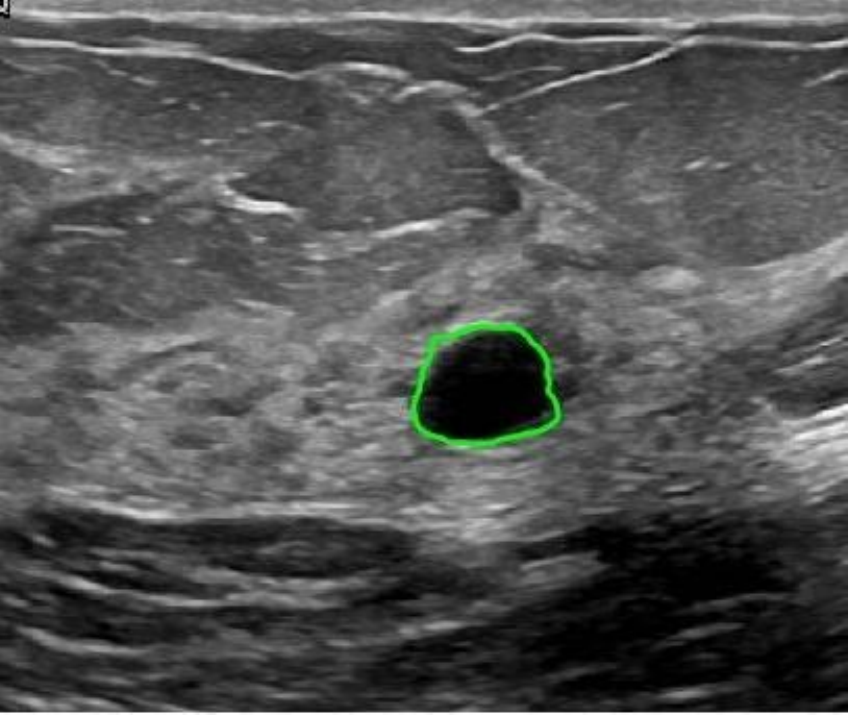}%
		\label{fig_34}}
	\hfil
	\subfloat{\includegraphics[width=0.8in,height=0.7in]{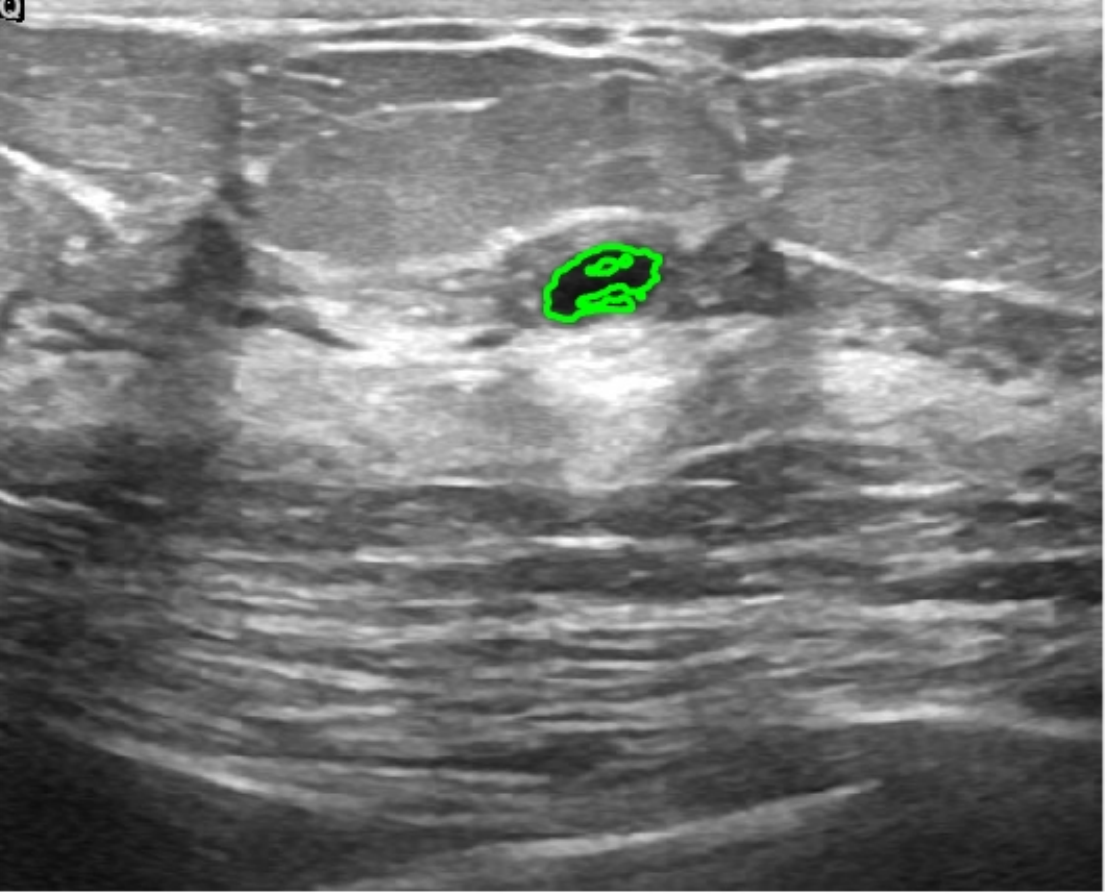}%
		\label{fig_35}}\vspace{-3mm}\hspace{-1.5mm}
	\subfloat{\includegraphics[width=0.8in,height=0.7in]{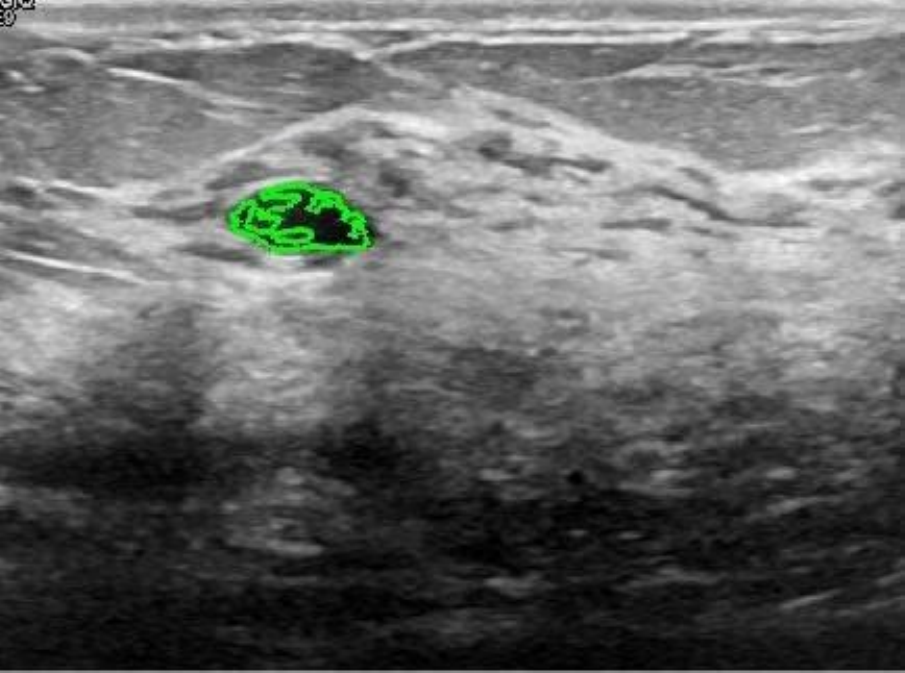}%
		\label{fig_36}}
	\subfloat{\includegraphics[width=0.8in,height=0.7in]{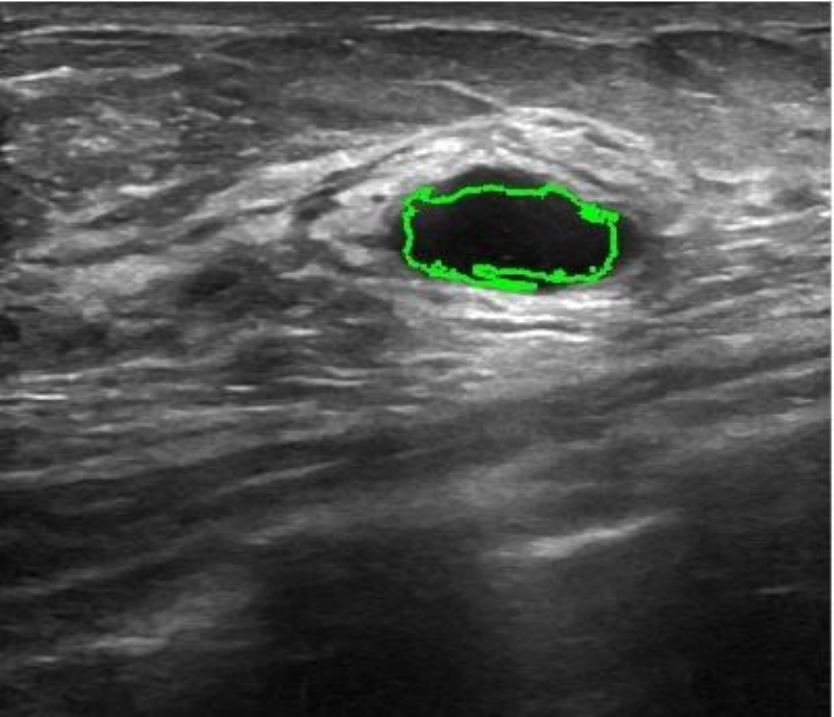}%
		\label{fig_37}}
	\subfloat{\includegraphics[width=0.8in,height=0.7in]{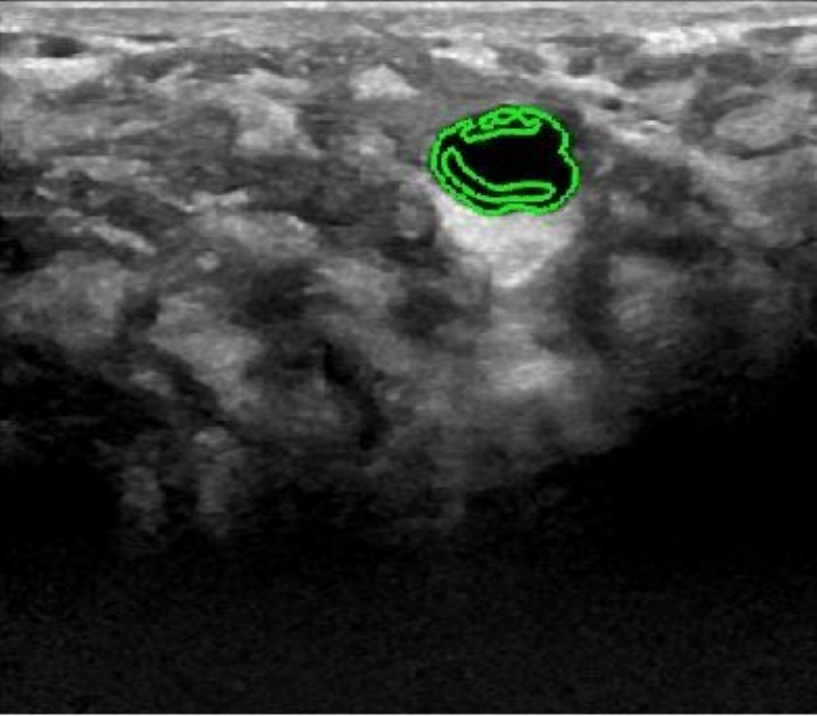}%
		\label{fig_38}}
	\subfloat{\includegraphics[width=0.8in,height=0.7in]{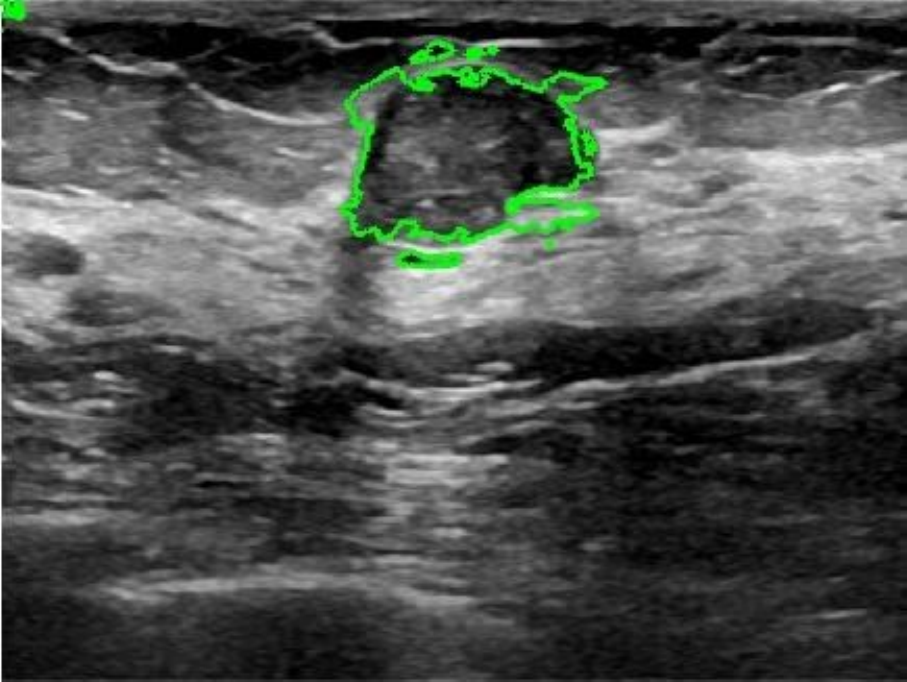}%
		\label{fig_39}}
	\subfloat{\includegraphics[width=0.8in,height=0.7in]{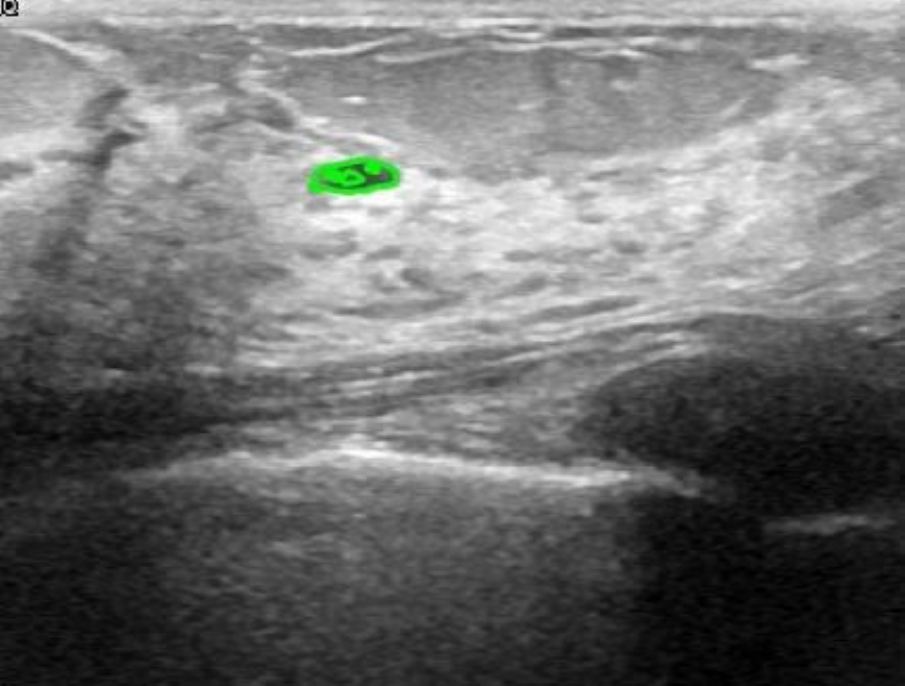}%
		\label{fig_40}}
	\subfloat{\includegraphics[width=0.8in,height=0.7in]{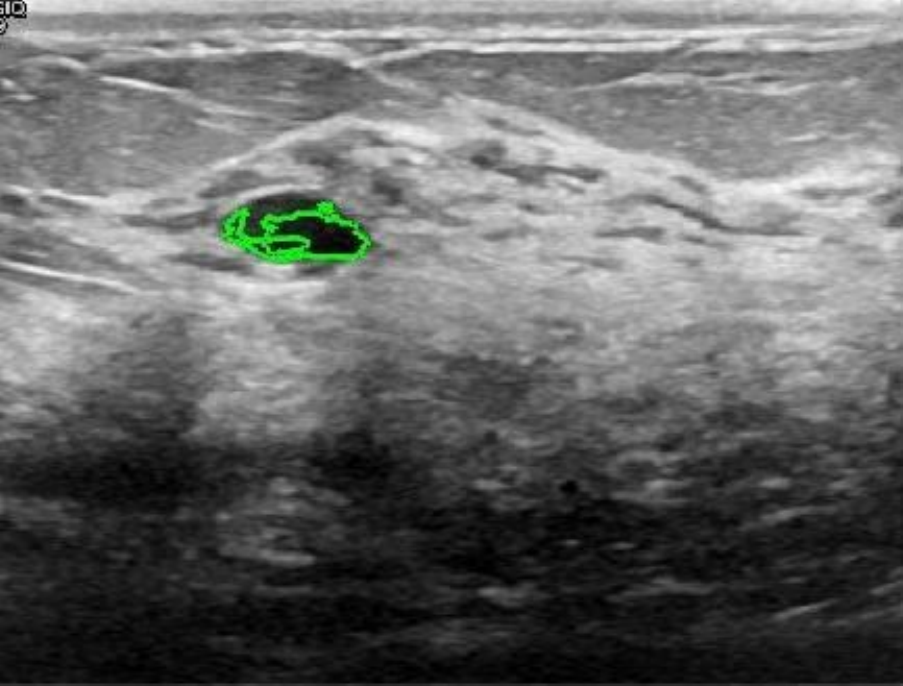}%
		\label{fig_41}}
	\subfloat{\includegraphics[width=0.8in,height=0.7in]{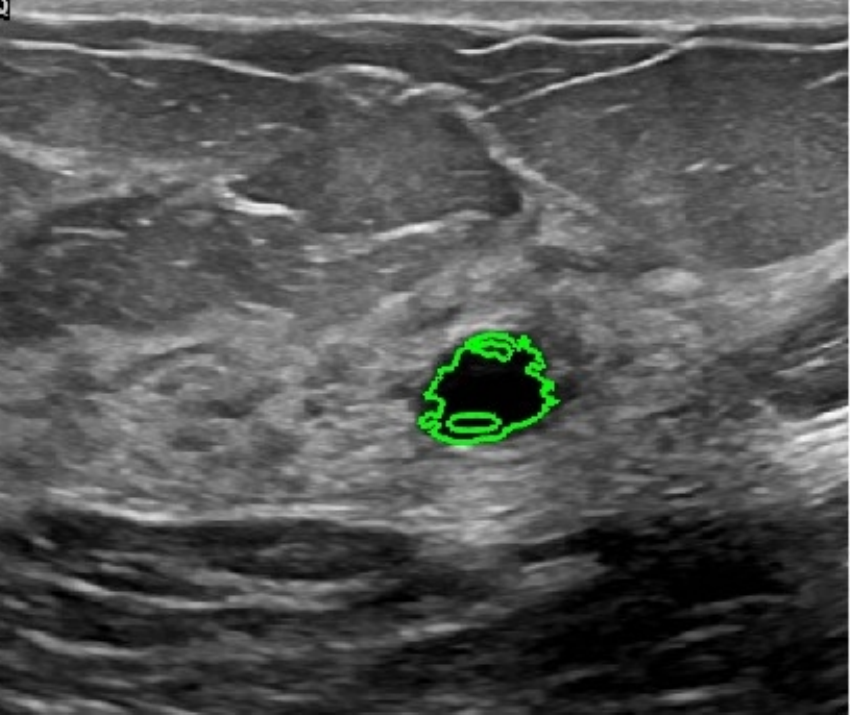}%
		\label{fig_42}}
	\hfil
	\subfloat{\includegraphics[width=0.80in,height=0.7in]{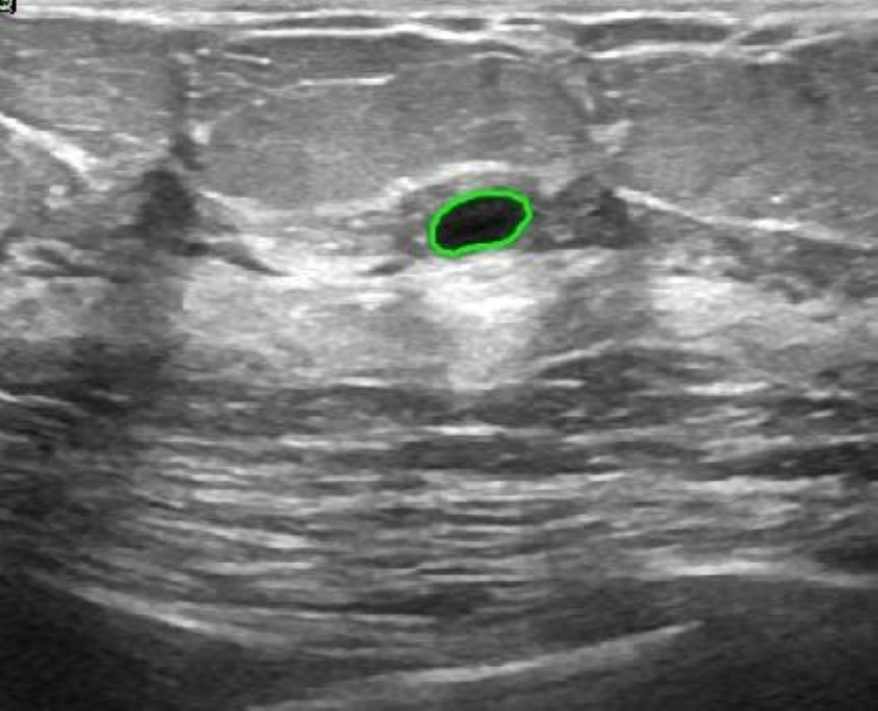}%
		\label{fig_43}}\vspace{-3mm}\hspace{-1.5mm}
	\subfloat{\includegraphics[width=0.8in,height=0.7in]{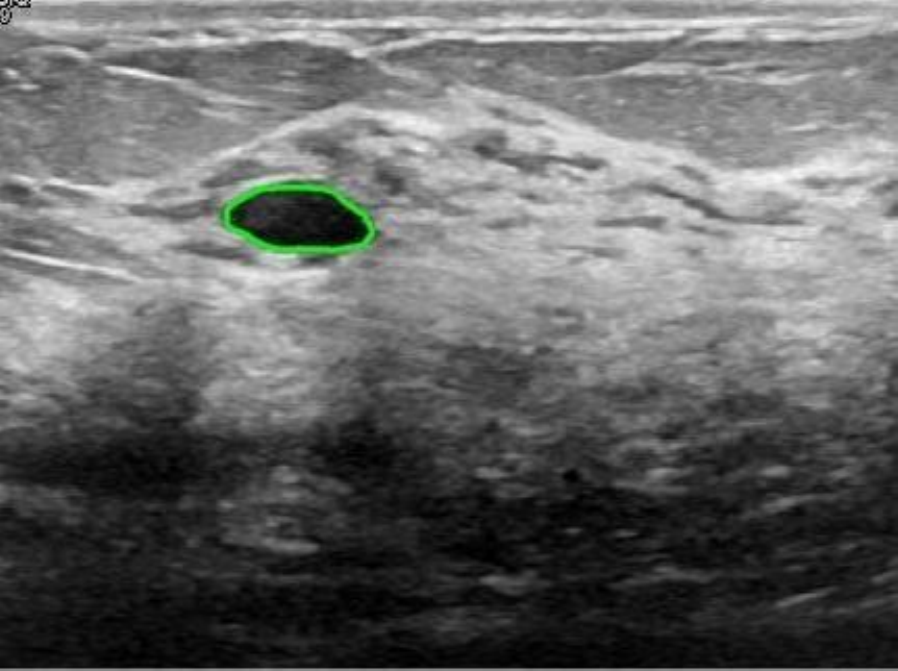}%
		\label{fig_44_}}
	\subfloat{\includegraphics[width=0.8in,height=0.7in]{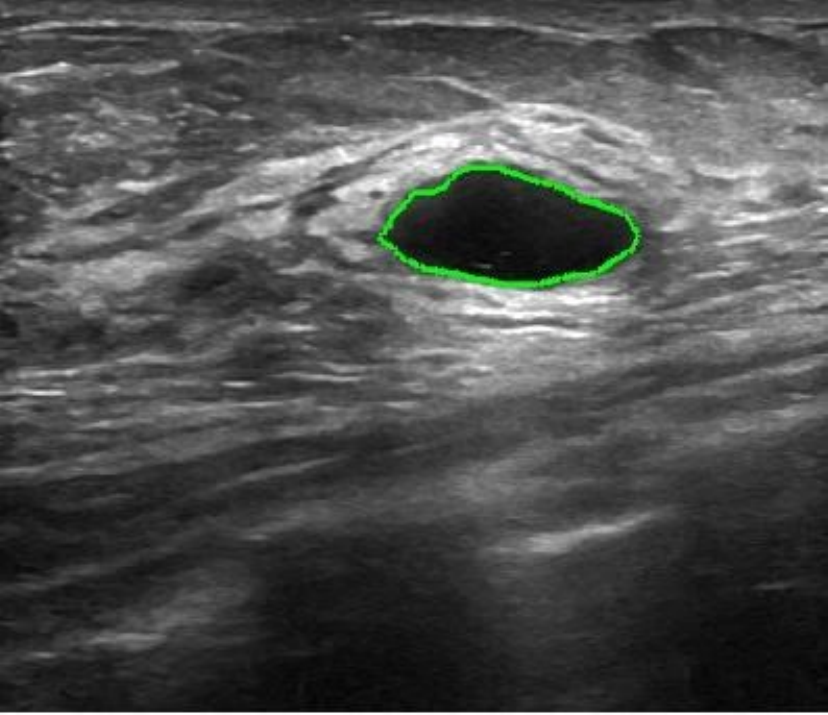}%
		\label{fig_45}}
	\subfloat{\includegraphics[width=0.8in,height=0.7in]{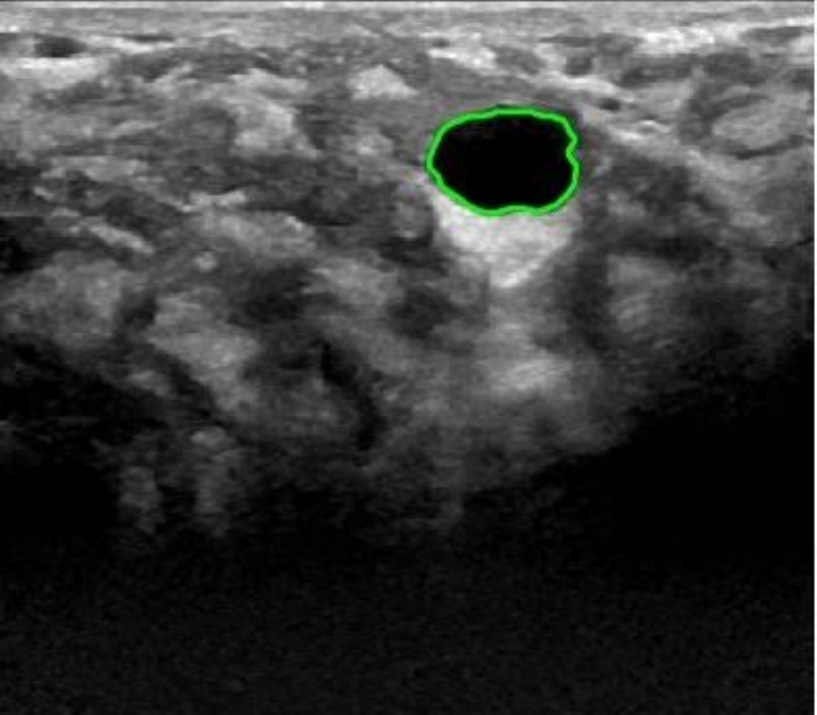}%
		\label{fig_46}}
	\subfloat{\includegraphics[width=0.8in,height=0.7in]{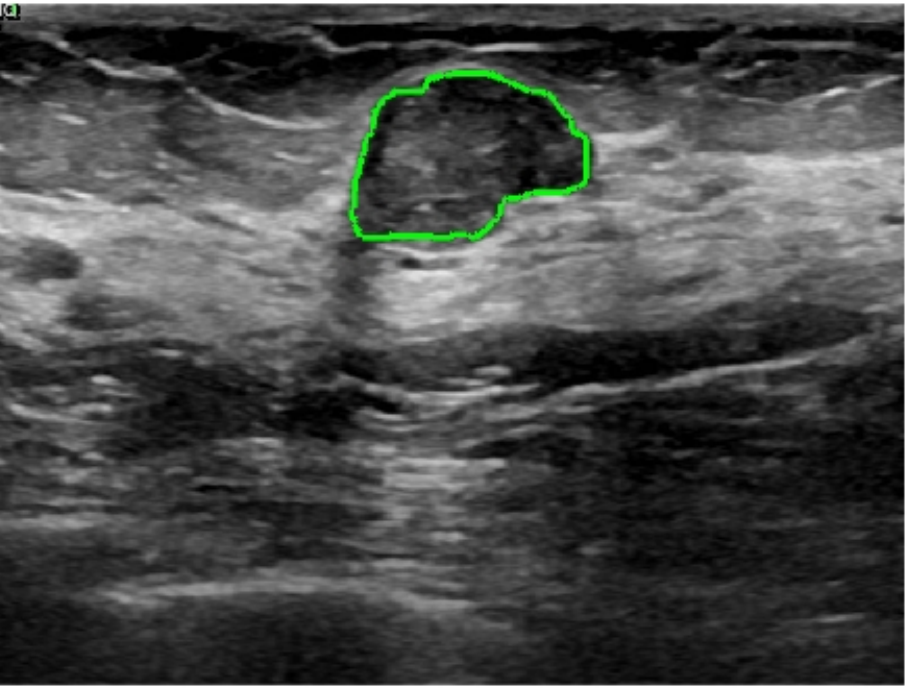}%
		\label{fig_47}}
	\subfloat{\includegraphics[width=0.8in,height=0.7in]{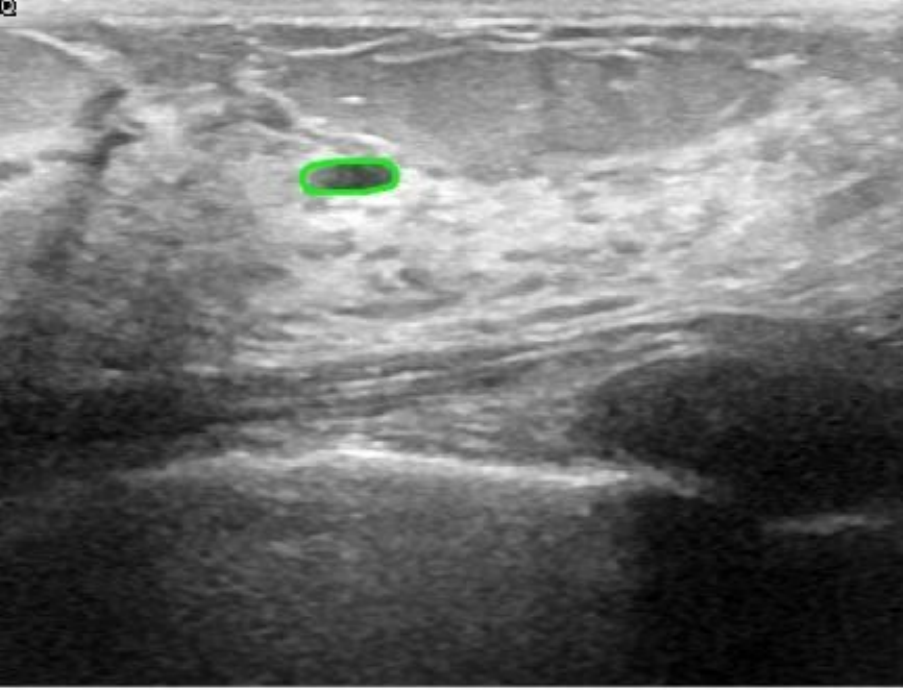}%
		\label{fig_48}}
	\subfloat{\includegraphics[width=0.8in,height=0.7in]{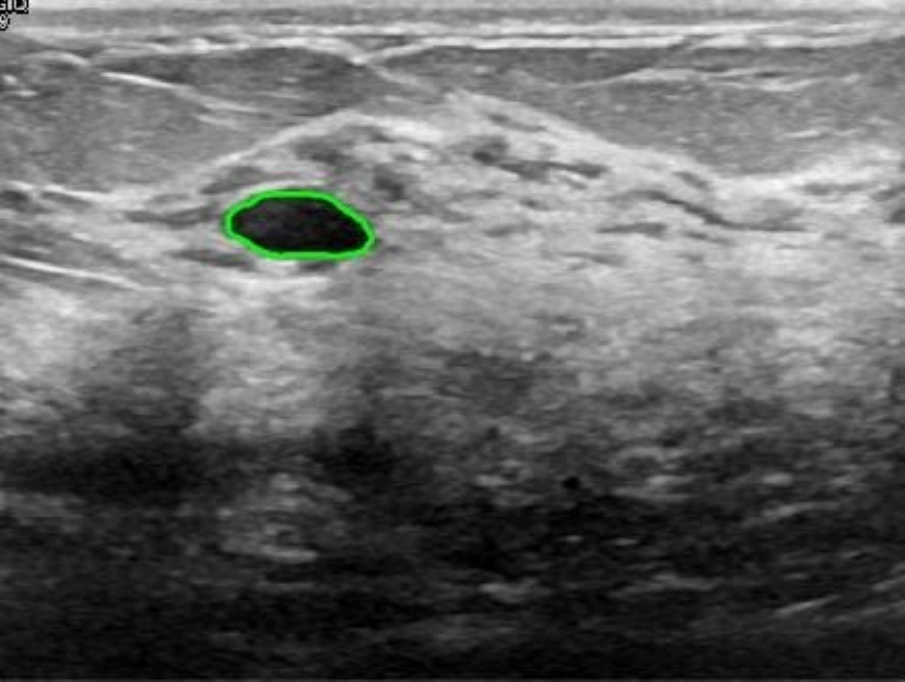}%
		\label{fig_49}}
	\subfloat{\includegraphics[width=0.8in,height=0.7in]{ 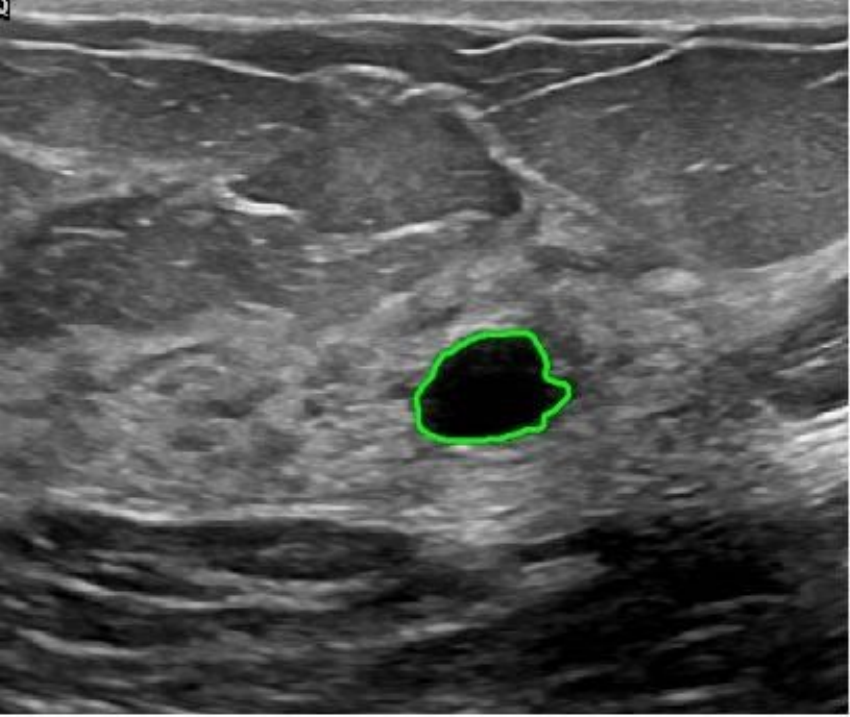}%
		\label{fig_50}}
	\hfil
	\subfloat{\includegraphics[width=0.8in,height=0.7in]{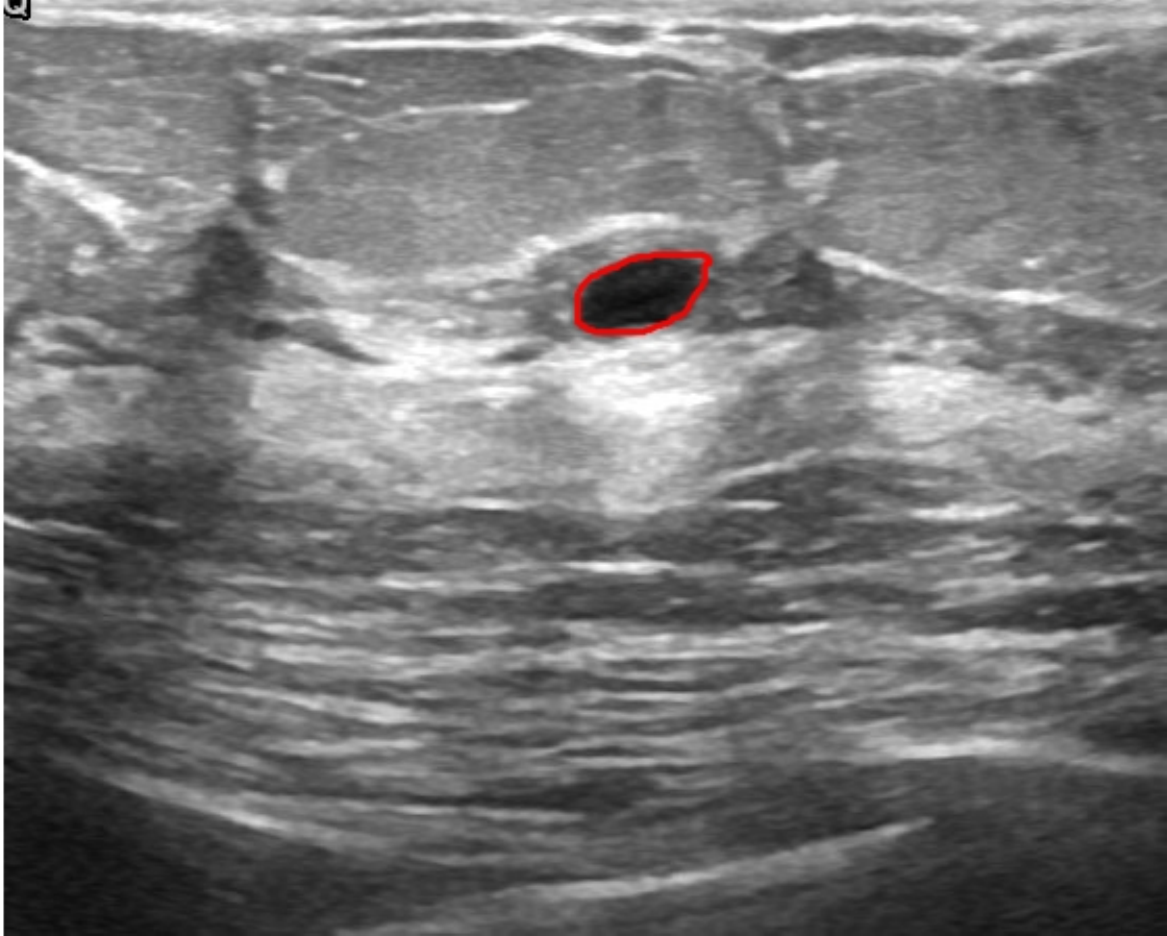}%
		\label{fig_51}}
	\subfloat{\includegraphics[width=0.8in,height=0.7in]{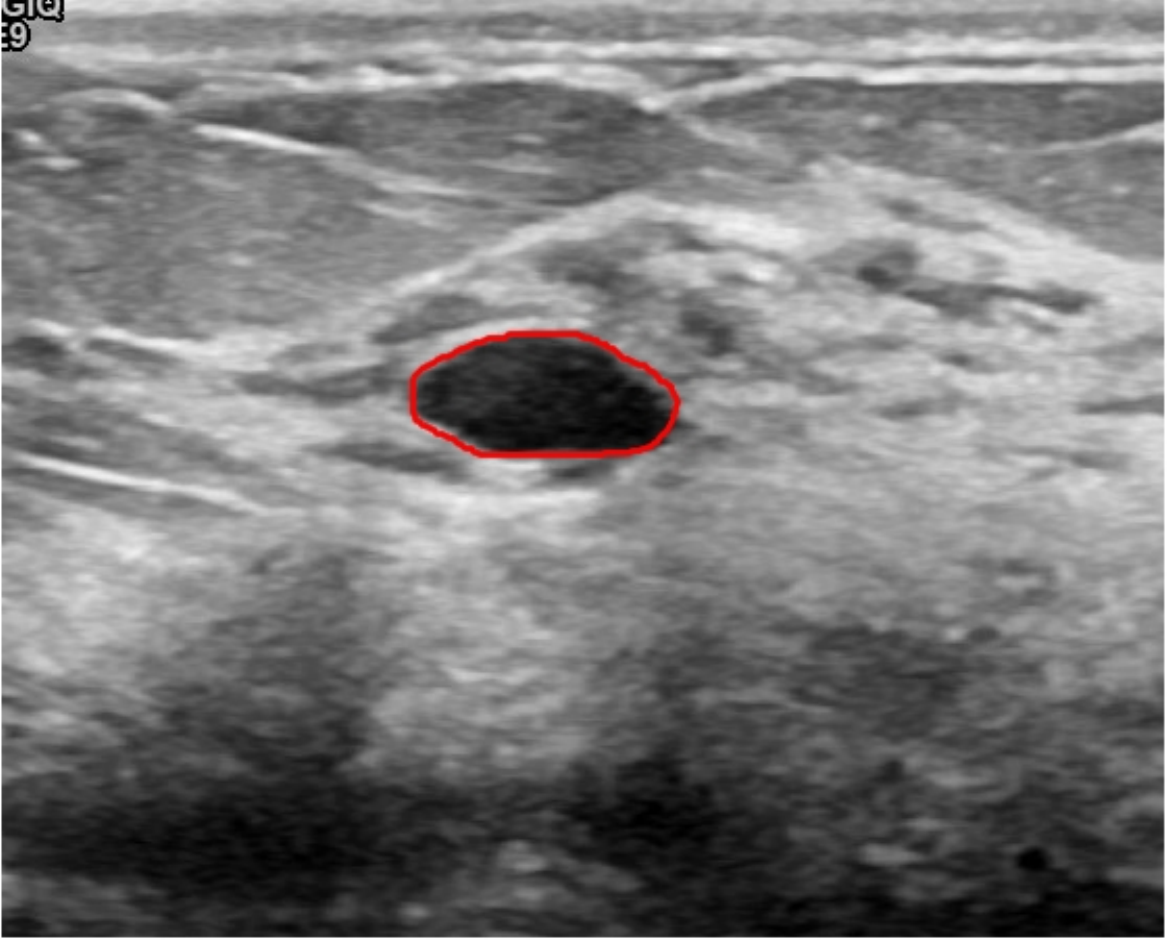}%
		\label{fig_52}}
	\subfloat{\includegraphics[width=0.8in,height=0.7in]{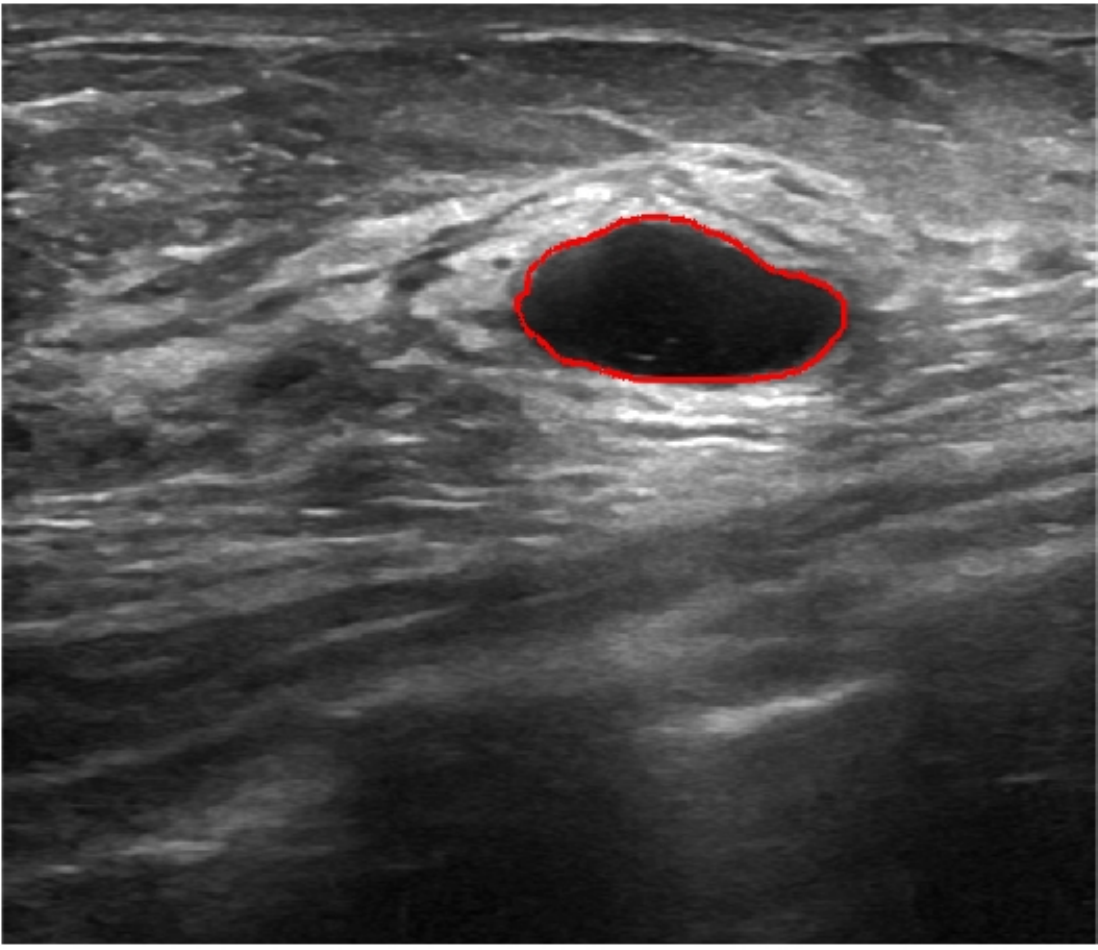}%
		\label{fig_53}}
	\subfloat{\includegraphics[width=0.8in,height=0.7in]{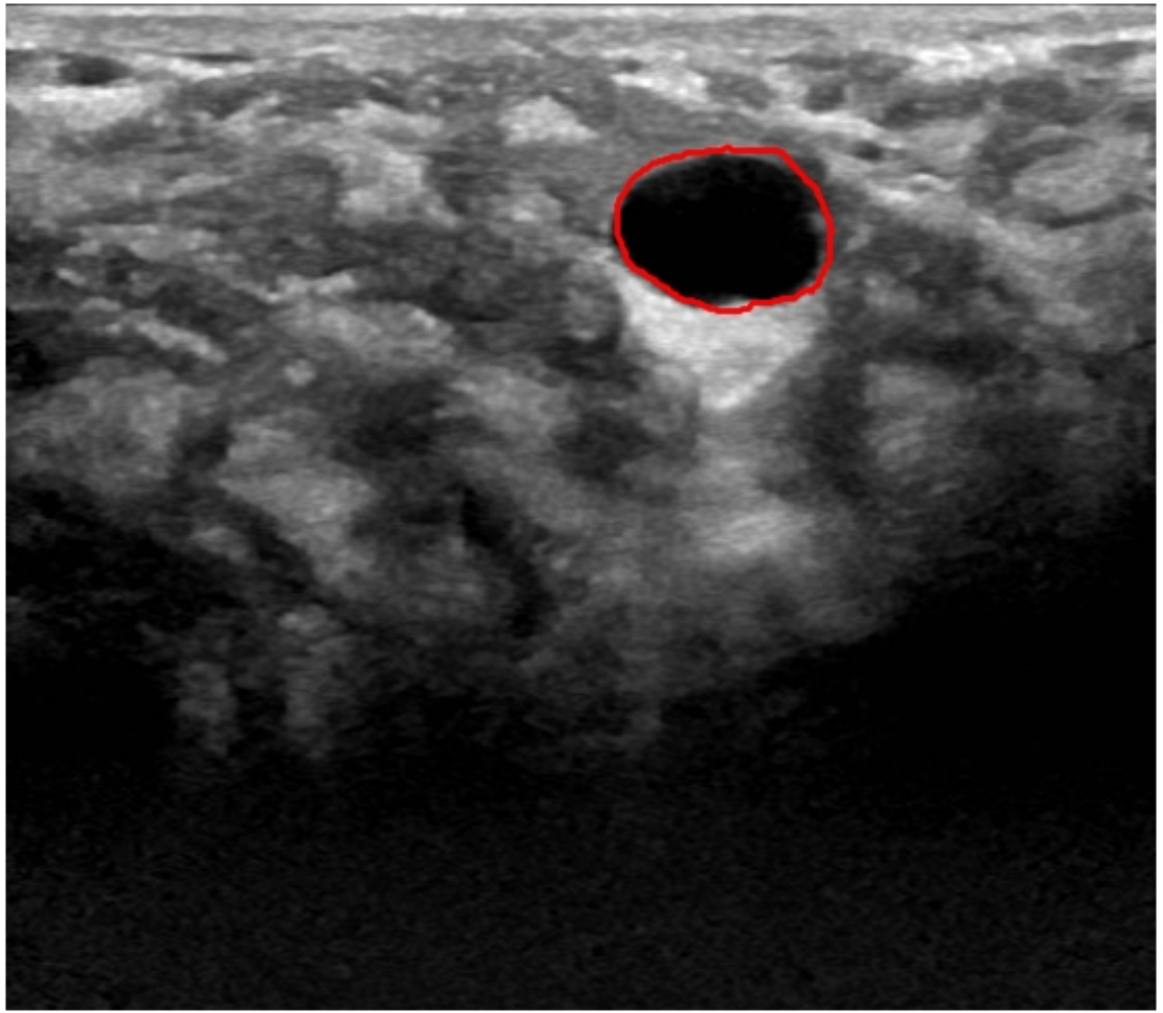}%
		\label{fig_54}}
	\subfloat{\includegraphics[width=0.8in,height=0.7in]{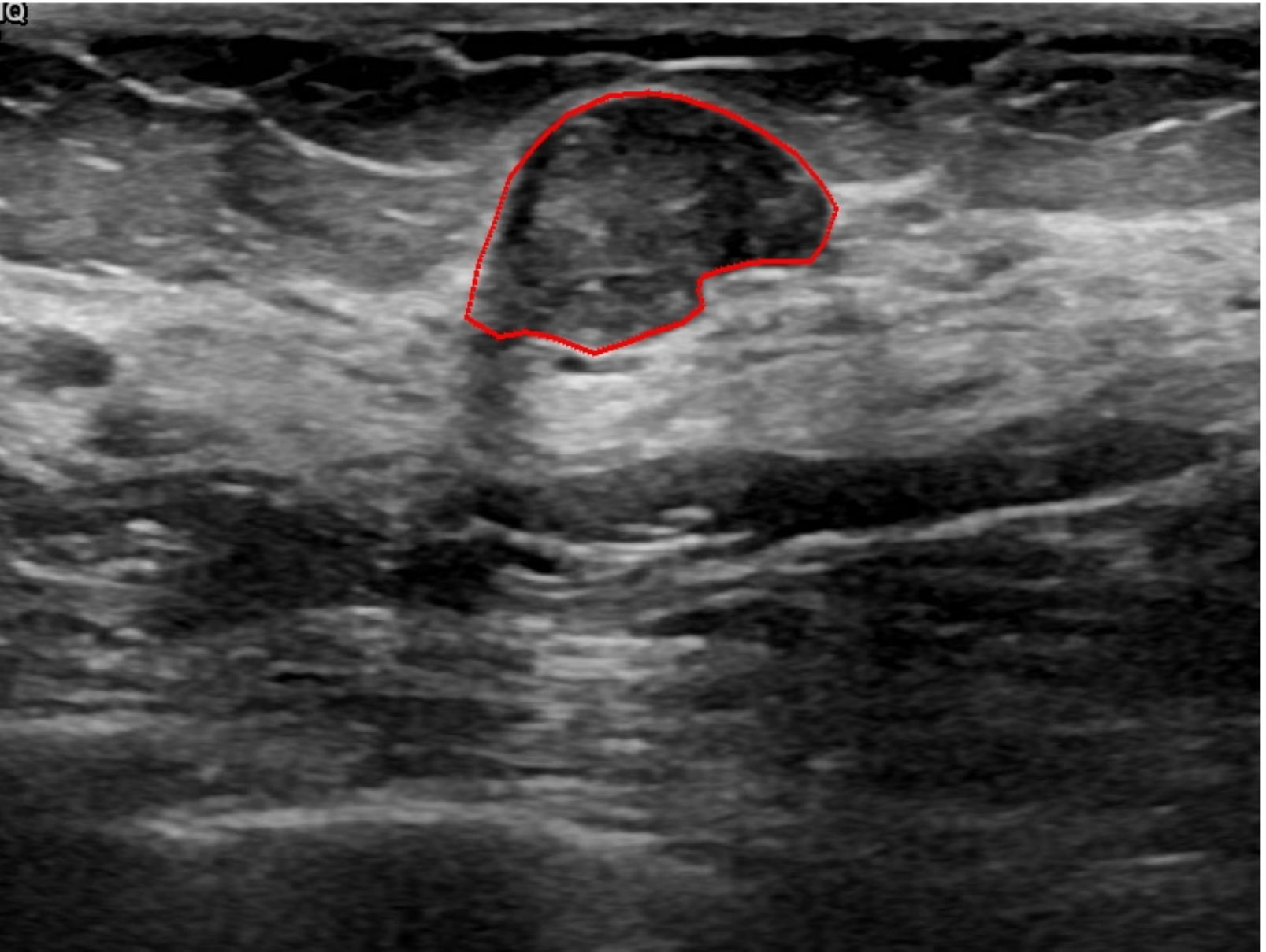}%
		\label{fig_55}}
	\subfloat{\includegraphics[width=0.8in,height=0.7in]{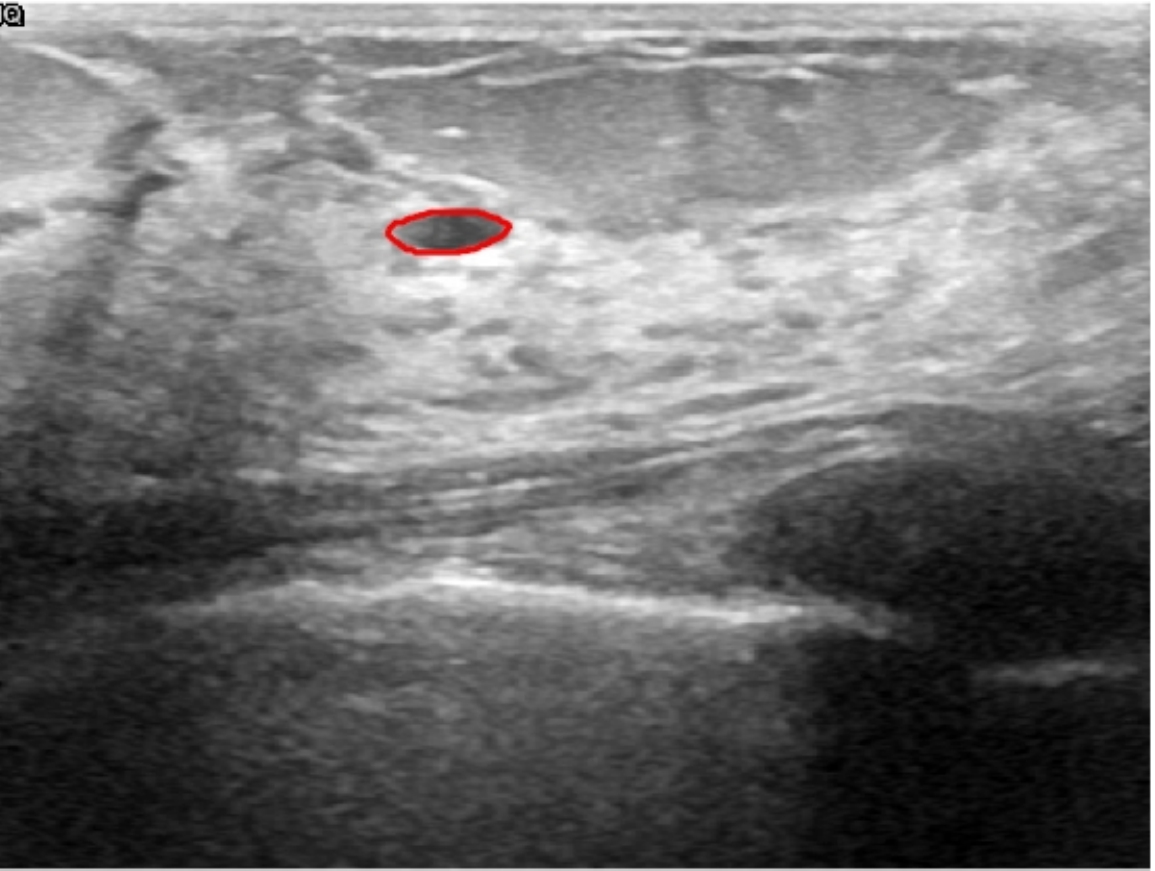}%
		\label{fig_56}}
	\subfloat{\includegraphics[width=0.8in,height=0.7in]{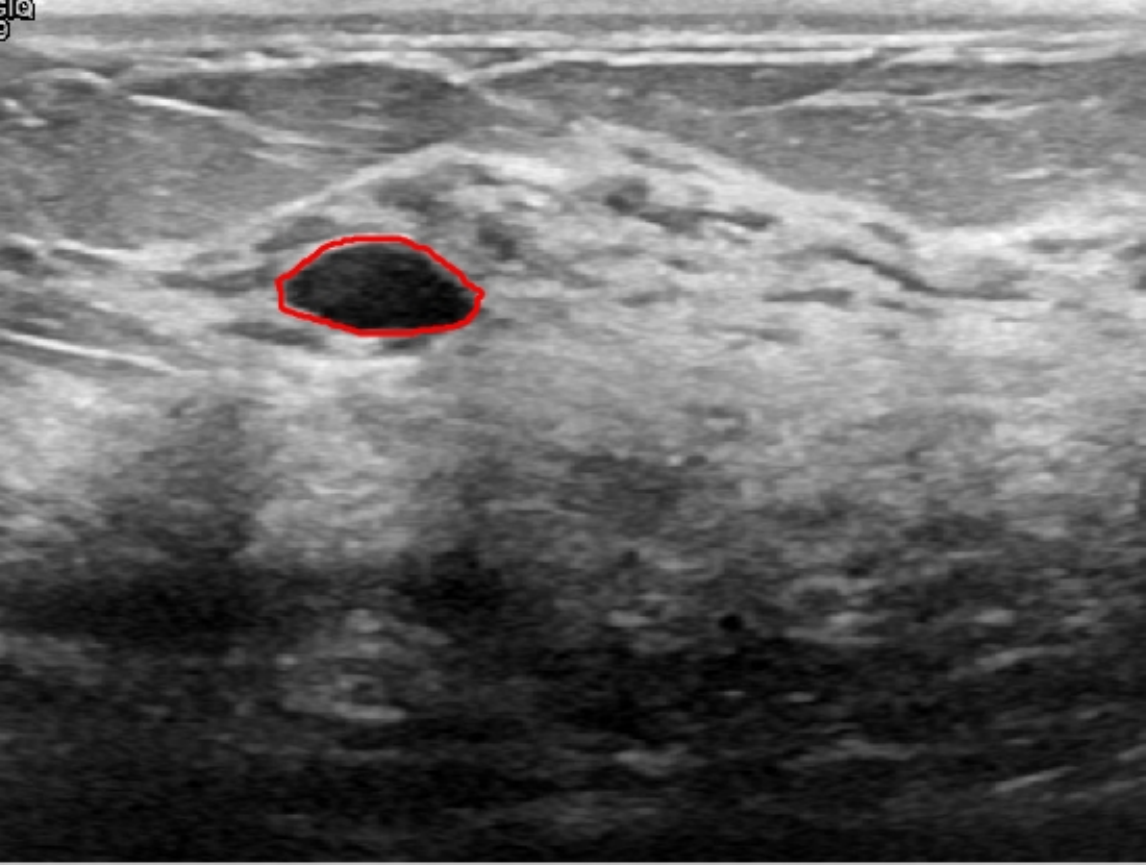}%
		\label{fig_57}}
	\subfloat{\includegraphics[width=0.8in,height=0.7in]{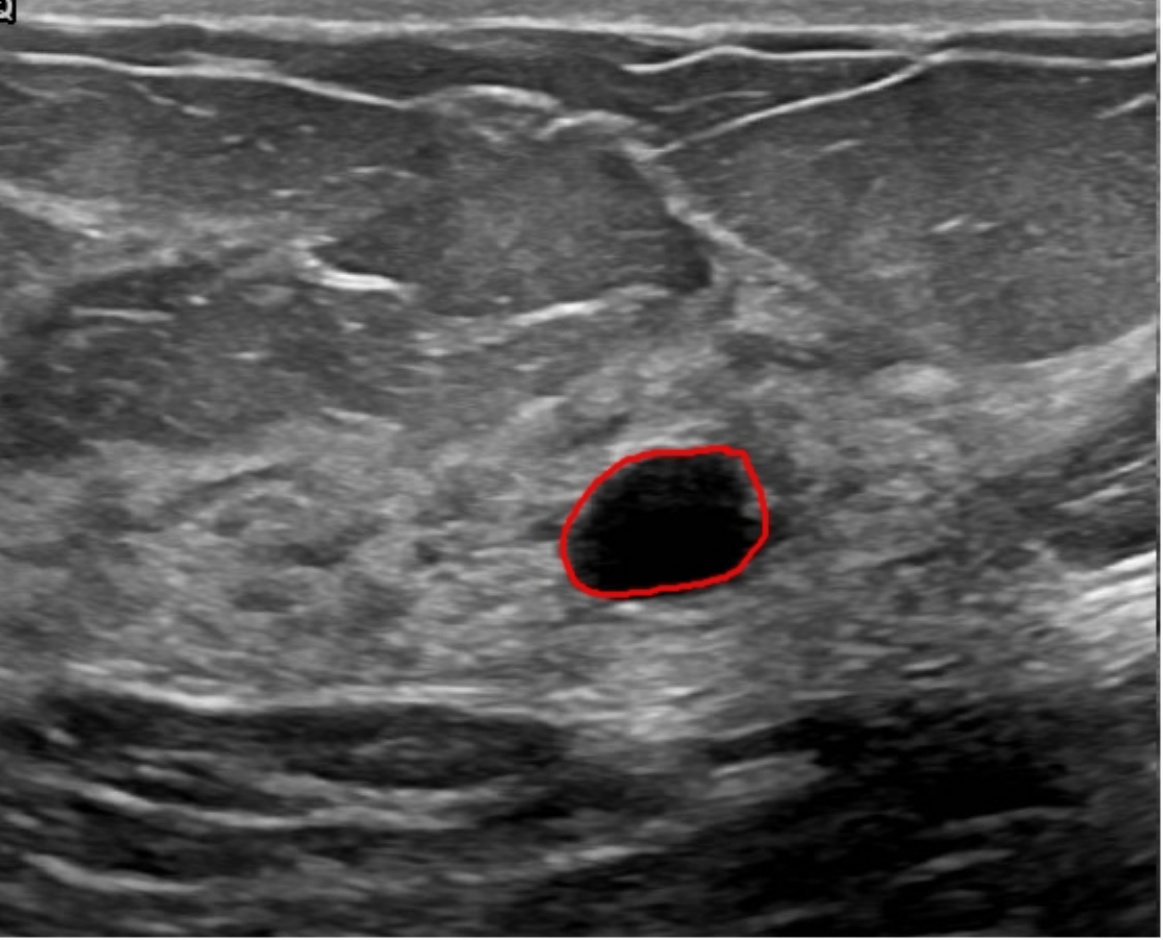}%
		\label{fig_58}}
	\hfil
	\caption{Segmentation results from evaluated models for segmenting ultrasound images. Row 1: Original images and initial contours. Row 2-5: Results from the ICTM model, RESLS model, ALF model, and RefLSM. Row 6: Ground truth.}
	\label{img8}
\end{figure*}

\subsection{Comparison with cardic MR images for right ventricle segmentation}
Cardiac MRI images often suffer from intensity inhomogeneity due to the varying magnetic susceptibility of cardiac tissues. This can lead to ambiguous boundaries, making it hard for models to consistently identify the left ventricle. Moreover, the heart is composed of several complex anatomical structures, including chambers, valves, and surrounding tissues. Accurately delineating the borders between these various components can be difficult, especially when they are closely positioned or exhibit similar intensity characteristics.
In this section, we demonstrate the effectiveness of the RefLSM in segmenting cardiac MR images. The cardiac MR images are standardized to a size of $256\times256$. To enhance the credibility of the proposed model, we segmented 5 cardiac MR images compared the results with 6 other models, including the ALF \cite{MA2019201}, LoGRSF \cite{DING2017224}, ABC \cite{WENG2021115633}, RESLS\cite{8765635}, ICTM \cite{WANG2022108794} and FeaACM \cite{XUE2024110673} models. Fig. \ref{img4} displays the partial results from all the segmentation models. It is clear that the segmentation results from the RefLSM closely align with the ground truth of the left ventricle segmentation. However, the ALF model demonstrates relatively poor ability to capture the overall structure of the left ventricle; it exhibits limitations with precise boundary detection, particularly in areas of intensity inhomogeneity. Consequently, the evolution curve of the ALF fits into an unstable wave-like result. Suffering from intensity inhomogeneity and surrounding anatomical structures, the LoGRSF model, ABC model and RESLS model all represent over-segmentation and under-segmentation to some extent. Additionally, the results in the 6th column of Fig. \ref{img4} show multiple instances where where the ICTM model inaccurately identifies adjacent myocardial tissues as part of the left ventricle due to their similar intensity, leading to false positives. While the FeaACM demonstrates some ability to segment the left ventricle, it is not robust and fails to segment the 5th row of Fig. \ref{img4}. The 2nd and 3rd rows also include small adjacent excessive regions. In contrast, our proposed model demonstrates marked improvements in accurately capturing the left ventricle's boundaries. It effectively addresses challenges associated with severe intensity inhomogeneity and noise, resulting in better segmentations. The proposed method appear to maintain better structural integrity in the segmented images, closely aligning with the ground truth.

To quantitatively measure the segmentation results of each model, we compute the Dice and Precision values for left ventricle segmentation. Moreover, we opted for a voilin plot to effectively compare the RefLSM with six other models, as shown in Fig. \ref{img17}. The results clearly demonstrate that the RefLSM yields superior Dice and Precision values than the other representative models, indicating its closer alignment with the ground truth and highest segmentation accuracy.

\subsection{Comparison with cardic MR images for left ventricle segmentation}

In this section, we continue to conduct experiments on cardiac MR images. However, we focus on the right ventricle, which is equally important as the left ventricle in cardiac MR diagnoses. In the near short-axis view, the left ventricle has a crescent shape, while the right ventricle is closer to circular. This does not imply that the right ventricle is easier to segment; in the MR images we use, the right ventricle often shows low-signal-intensity shadows, which may be caused by the imaging characteristics of certain tissues or fluids, leading to interference in segmentation. Moreover, there is a layer of myocardial tissue surrounding the right ventricle that overlaps with the boundary of the right ventricle, posing challenges for accurate identification.

Fig.~\ref{img6} presents the segmentation results of the right ventricle from the DRLSE-ADMM \cite{WALI2023109105}, ABC model \cite{WENG2021115633}, RESLS \cite{8765635}, the RefLSM, and the ground truth. 
 The results from the DRLSE-ADMM model in the 1st row reveal severe local minima due to low-signal-intensity shadows in the right ventricle. Additionally, both the ABC and RESLS models fail to correctly identify the right ventricle boundary because of the interference from surrounding myocardial tissue, as seen in the 1st and 3rd columns of Fig.\ref{img6}. In comparison, the proposed model gets more satisfactory segmentation results that are closer to the ground truth. Similarly, we evaluate the segmentation performance of all models by calculating the Dice and Precision values and drawing Fig. \ref{img18}. This overlay plot presents the best, median and worst segmentation results, along with the overall range of outcomes that it is clear that the proposed model segments the right ventricle with greater accuracy than the other models and performs well in different images.

\begin{table*}[!ht]
\centering
\caption{Precision values, CPU time (s), and iterations of the RefLSM for segmenting ultrasound images in Fig.~\ref{img7}.}
\renewcommand{\arraystretch}{1.15}
\small
\setlength{\tabcolsep}{6pt}
\begin{tabular}{lcccccccc|c}
\toprule
Metric & Img.1 & Img.2 & Img.3 & Img.4 & Img.5 & Img.6 & Img.7 & Img.8 & Mean \\
\midrule
Precision $\uparrow$ & 0.9956 & 0.9572 & 0.9891 & 0.9823 & 0.9721 & 0.9820 & 0.9956 & 0.9896 & \textbf{0.9812} \\
Time (s) $\downarrow$ & 1.593  & 1.771  & 1.799  & 1.682  & 2.105  & 2.003  & 1.532  & 1.841  & 1.790 \\
Iterations & 5 & 6 & 8 & 8 & 6 & 6 & 5 & 6 & -- \\
\bottomrule
\end{tabular}
\label{t2}
\end{table*}

\begin{table*}[!ht]
\centering
\caption{Comparison of Dice and Precision values for evaluated models on ultrasound image segmentation in Fig.~\ref{img8}.}
\renewcommand{\arraystretch}{1.15}
\small   % 设置字体比正文小
\setlength{\tabcolsep}{6pt} % 控制列间距，避免太挤
\begin{tabular}{lccccccccc|c}
\toprule
Metric & Model & Img.1 & Img.2 & Img.3 & Img.4 & Img.5 & Img.6 & Img.7 & Img.8 & Mean \\
\midrule
\multirow{4}{*}{Precision $\uparrow$} 
 & ICTM   & 0.9276 & 0.9862 & 0.9613 & 0.9783 & 0.8917 & 0.9384 & 0.9913 & 0.9477 & 0.9528 \\
 & RESLS  & 0.8702 & 0.9843 & 0.9785 & 0.9506 & 0.9341 & 0.8639 & 0.9388 & 0.8782 & 0.9248 \\
 & ALF    & 0.8976 & 0.7991 & 0.8993 & 0.8969 & 0.7687 & 0.8811 & 0.8667 & 0.7740 & 0.8479 \\
 & RefLSM & 0.9852 & 0.9931 & 0.9869 & 0.9954 & 0.9851 & 0.9317 & 0.9984 & 0.9537 & \textbf{0.9787} \\
\midrule
\multirow{4}{*}{Dice $\uparrow$} 
 & ICTM   & 0.8733 & 0.9393 & 0.9593 & 0.9202 & 0.9491 & 0.8714 & 0.9131 & 0.9104 & 0.9170 \\
 & RESLS  & 0.9023 & 0.8808 & 0.7826 & 0.9281 & 0.8743 & 0.9115 & 0.8811 & 0.8959 & 0.8821 \\
 & ALF    & 0.7957 & 0.7208 & 0.8244 & 0.7533 & 0.8252 & 0.8034 & 0.6074 & 0.7658 & 0.7620 \\
 & RefLSM & 0.9421 & 0.9169 & 0.9654 & 0.9116 & 0.9451 & 0.9392 & 0.9056 & 0.9201 & \textbf{0.9305} \\
\bottomrule
\end{tabular}
\label{t3}
\end{table*}

\begin{figure}[!t]
	\centering
	\subfloat[]{\includegraphics[width=2.5in]{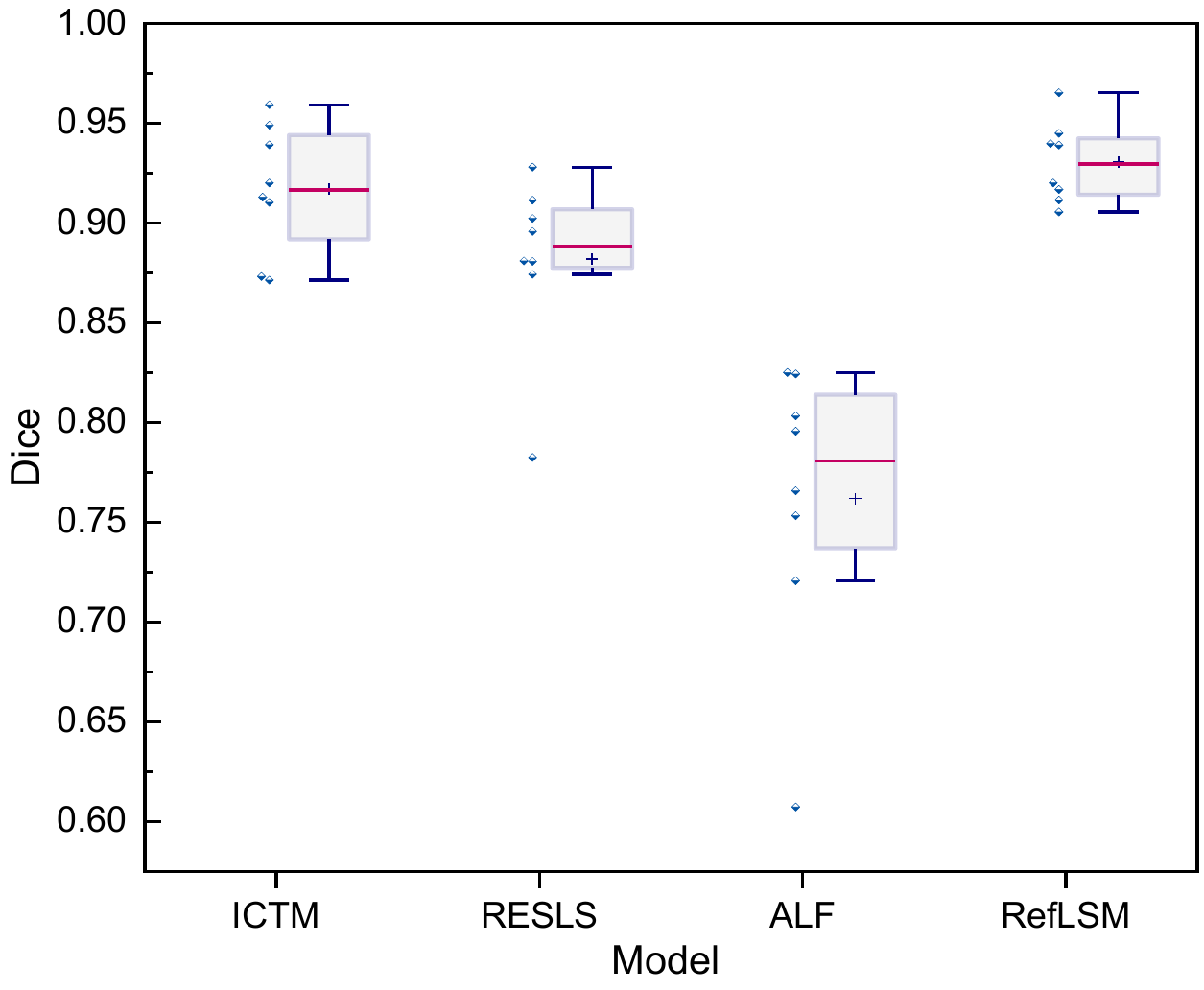}%
		\label{fig_159_case}}
	\subfloat[]{\includegraphics[width=2.5in]{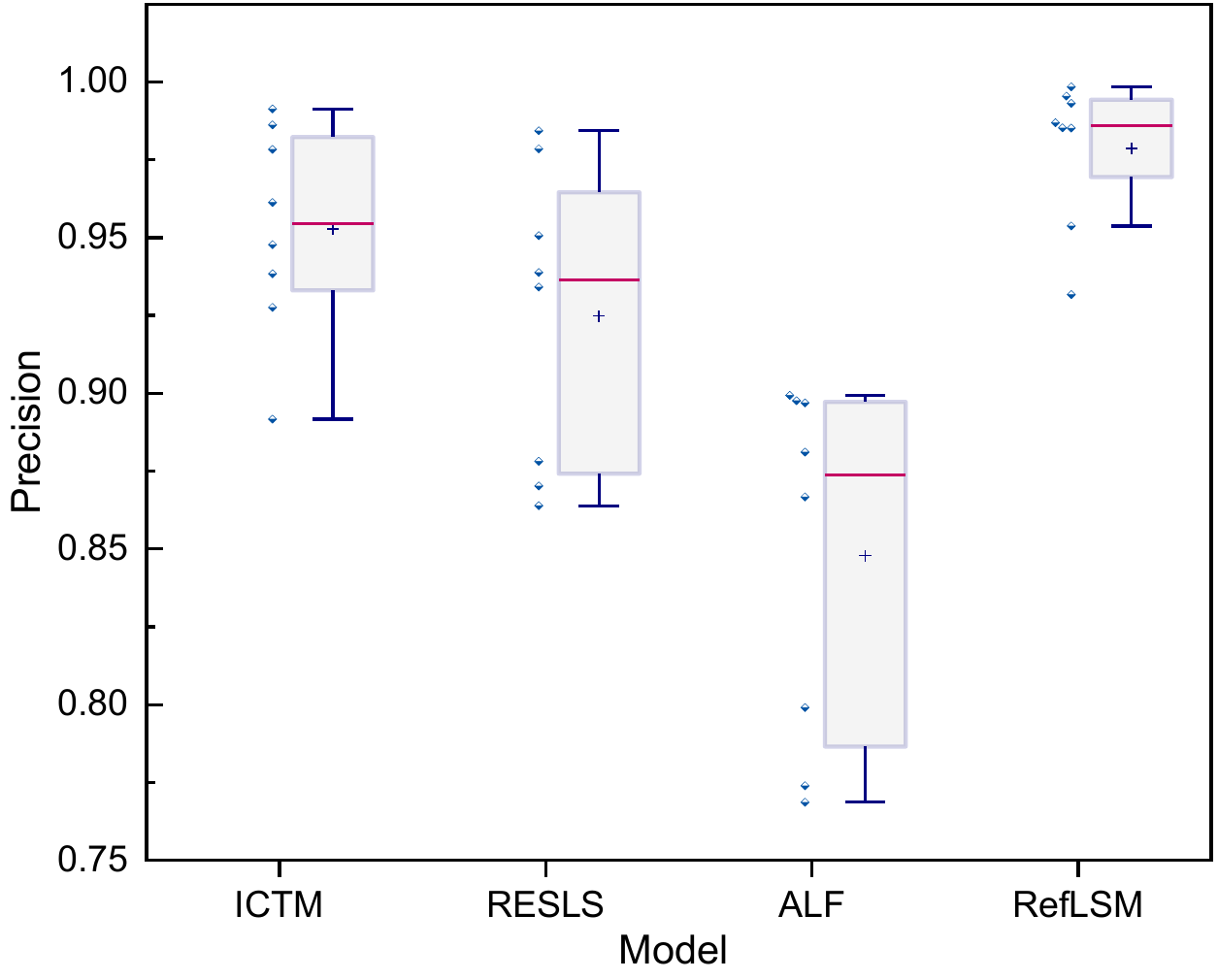}%
		\label{fig_160_case}}
	\caption{Dice values and Precision values of different models for breast ultrasound images.}
	\label{img20}
\end{figure}

\subsection{Comparison with breast ultrasound images segmentation}
Unlike MR images mentioned above, ultrasound images offer real-time imaging but usually have lower resolution and contrast, making it difficult to differentiate between adjacent tissues and often introducing artifacts and noise that complicate segmentation. However, the RefLSM is robust to high levels of noise and intensity inhomogeneity, regardless of the type of medical image. To validate this, we perform segmentation comparison experiments on breast ultrasound images and calculate the Dice coefficient and Precision values for all models under evaluation.

In Fig. \ref{img7}, the 1st row displays original images and initial contours for the RefLSM, while the second row shows the RefLSM's segmentation results. Even with high levels of noise and severe intensity inhomogeneity caused by acoustic shadowing and motion artifacts, our model successfully identifies object edges, achieving an average Precision value of 0.9812. Table \ref{t4} shows that the RefLSM requires no more than 10 iterations and has an average processing time of only 1.79 seconds to segment images in Fig. \ref{img7}. 

To further validate the effectiveness of our model, we compared it with the ICTM, RESLS, and ALF models, represented in the 2nd, 3rd, and 4th rows of Fig.~\ref{img8}, respectively. The 5th row displays our proposed model. 
  To better compare the segmentation results, we presented the initial contours in the 1st row, and the ground truth in the 6th row. We can observe that the ALF model struggles with boundary identification and has significant misclassification due to the boundary confusion in ultrasound images commonly caused by overlapping of soft tissues. While the ICTM and RESLS models perform well in some cases like column 1,4, and 7, they still face under-segmentation and over-segmentation problems when images are low contrast or have complex edges like column 1, 3 and 5. In addition to visual comparisons, we calculated the Precision and Dice coefficient for each model, as shown in Table \ref{t2}. Clearly, the RefLSM outperforms all the other models in terms of mean score. Based on these results, we draw Fig. \ref{img20} to present the data distribution and facilitates comparison of different models' performance across various breast ultrasound images. Although the ICTM model shows relatively close performance, it has nearly double the computational cost of the RefLSM. Therefore,the RefLSM demonstrates excellent segmentation capability and efficiency when processing ultrasound images.
\begin{figure*}[!t]
	\centering
	\subfloat{\includegraphics[width=1in,height=1in]{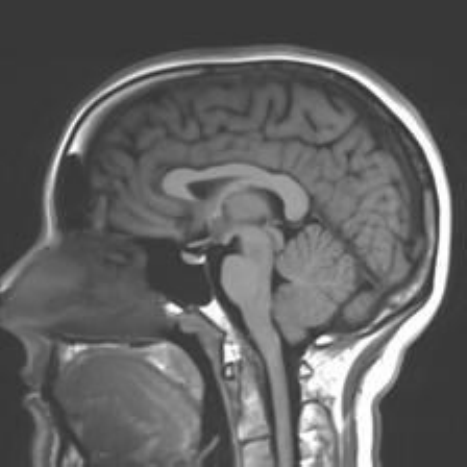}%
		\label{fig61}}\vspace{-3mm}\hspace{-1.5mm}
	\subfloat{\includegraphics[width=1in,height=1in]{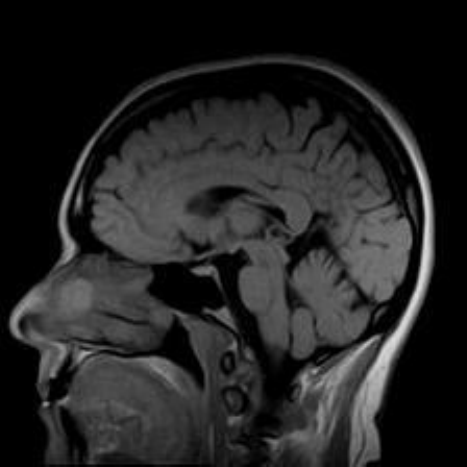}%
		\label{fig62}}
	\subfloat{\includegraphics[width=1in,height=1in]{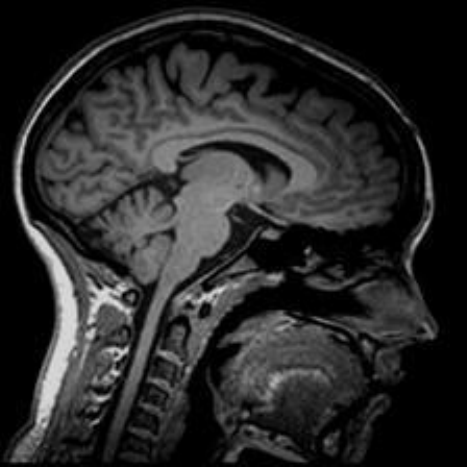}%
		\label{fig63}}
	\subfloat{\includegraphics[width=1in,height=1in]{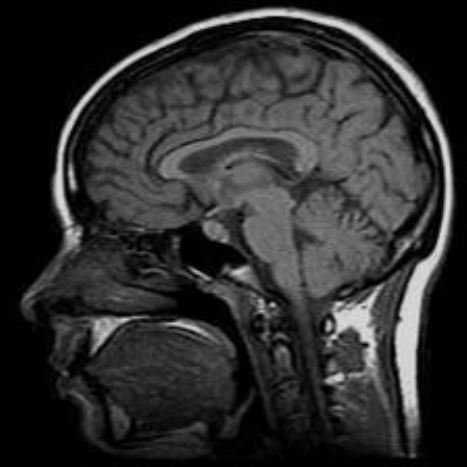}%
		\label{fig65}}
	\subfloat{\includegraphics[width=1in,height=1in]{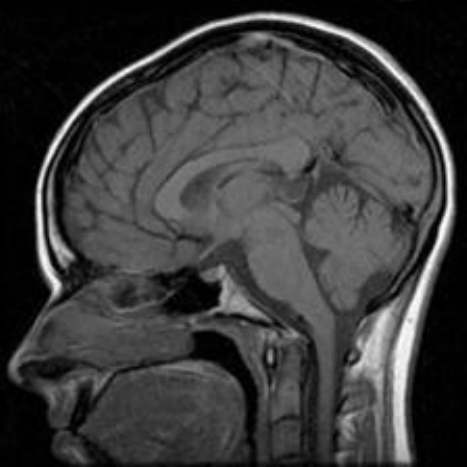}%
		\label{fig66}}
	\subfloat{\includegraphics[width=1in,height=1in]{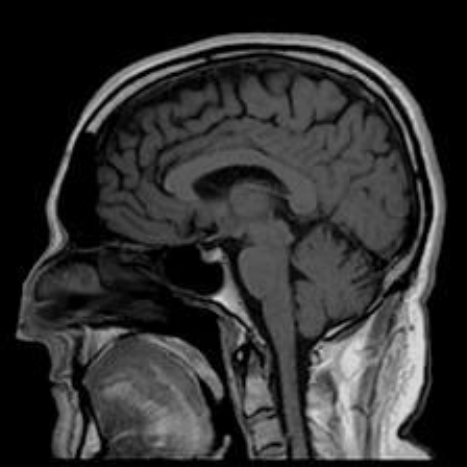}%
		\label{fig67}}
	\hfil
	\subfloat{\includegraphics[width=1in,height=1in]{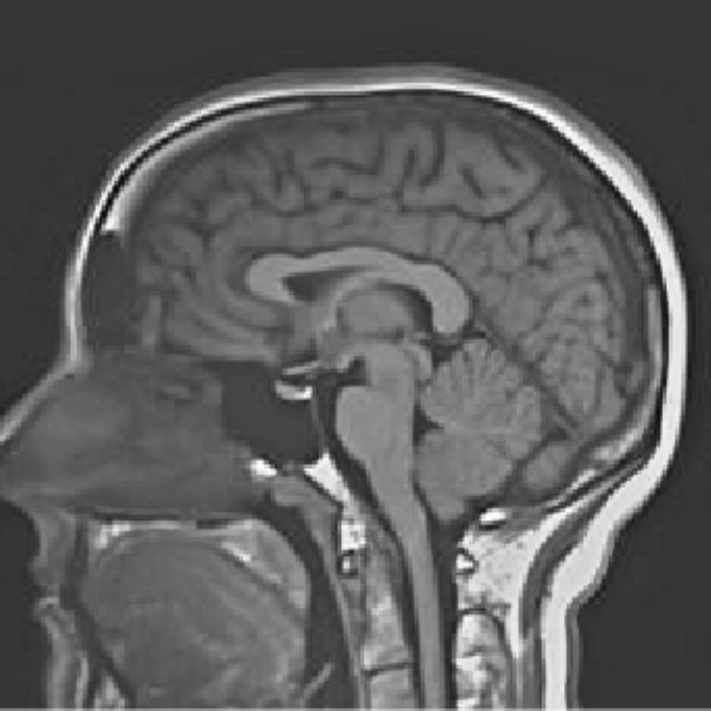}%
		\label{fig69}}
	\subfloat{\includegraphics[width=1in,height=1in]{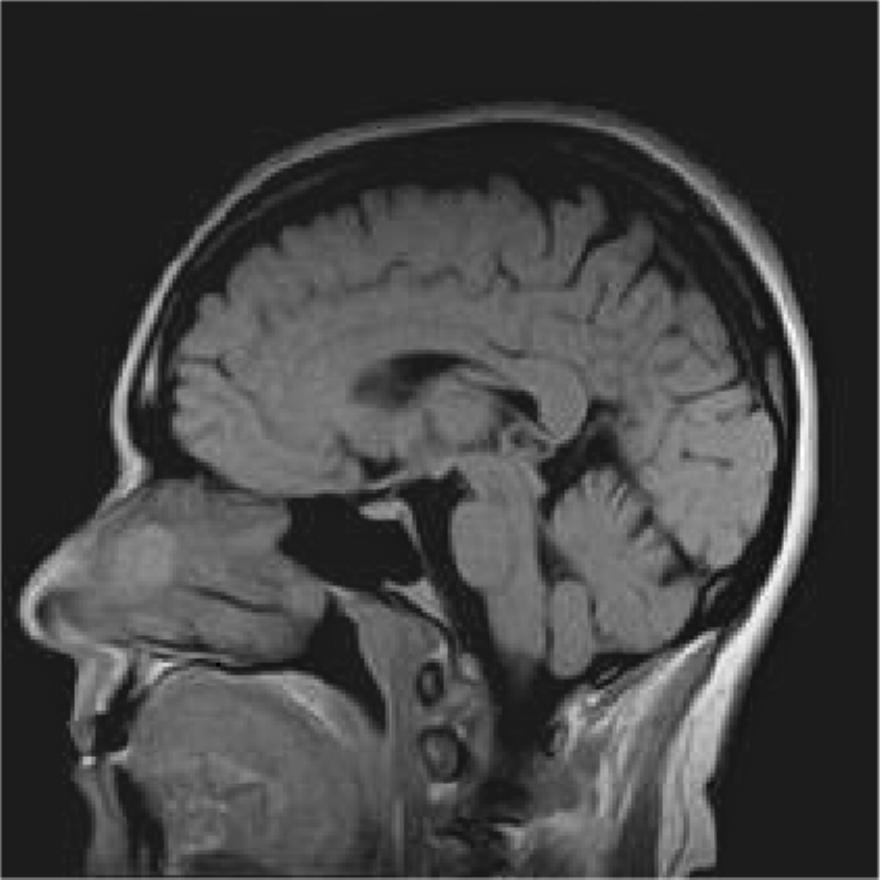}%
		\label{fig70}}
	\subfloat{\includegraphics[width=1in,height=1in]{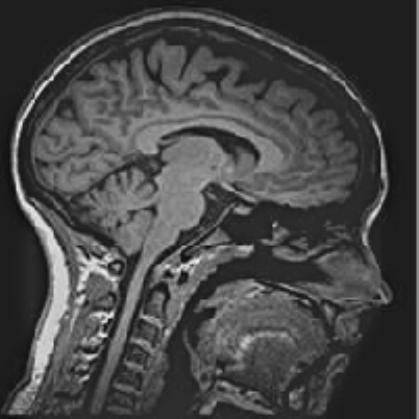}%
		\label{fig71}}
	\subfloat{\includegraphics[width=1in,height=1in]{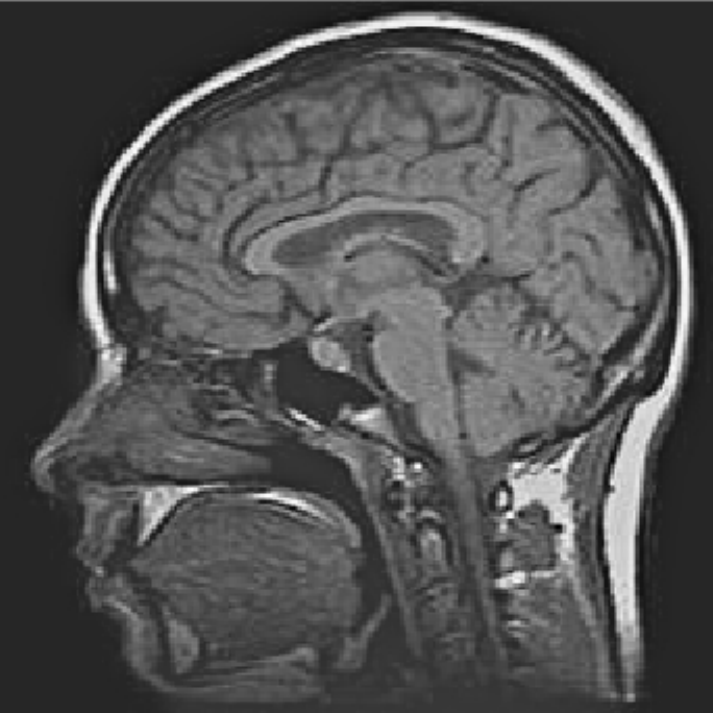}%
		\label{fig72}}
	\subfloat{\includegraphics[width=1in,height=1in]{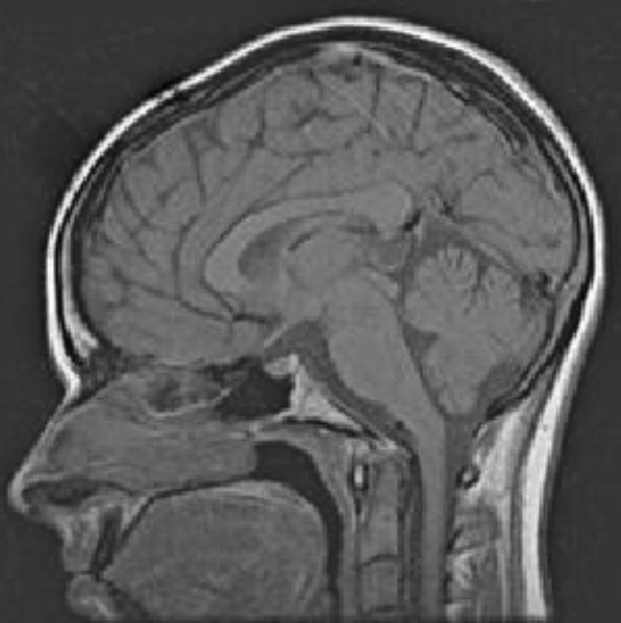}%
		\label{fig73}}
	\subfloat{\includegraphics[width=1in,height=1in]{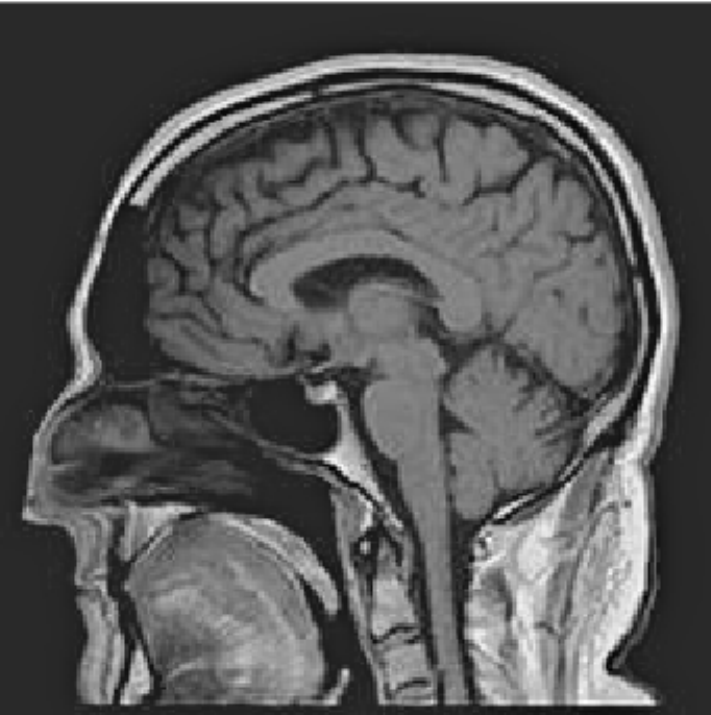}%
		\label{fig74}}
	\hfil
	\caption{Bias field correction results from the RefLSM for different Brain tumor MR images. Row 1: Original images. Row 2: Results from the RefLSM.}
	\label{img9}
\end{figure*}

\begin{figure}[!t]
	\centering
	\subfloat{\includegraphics[width=1.8in,height=1.4in]{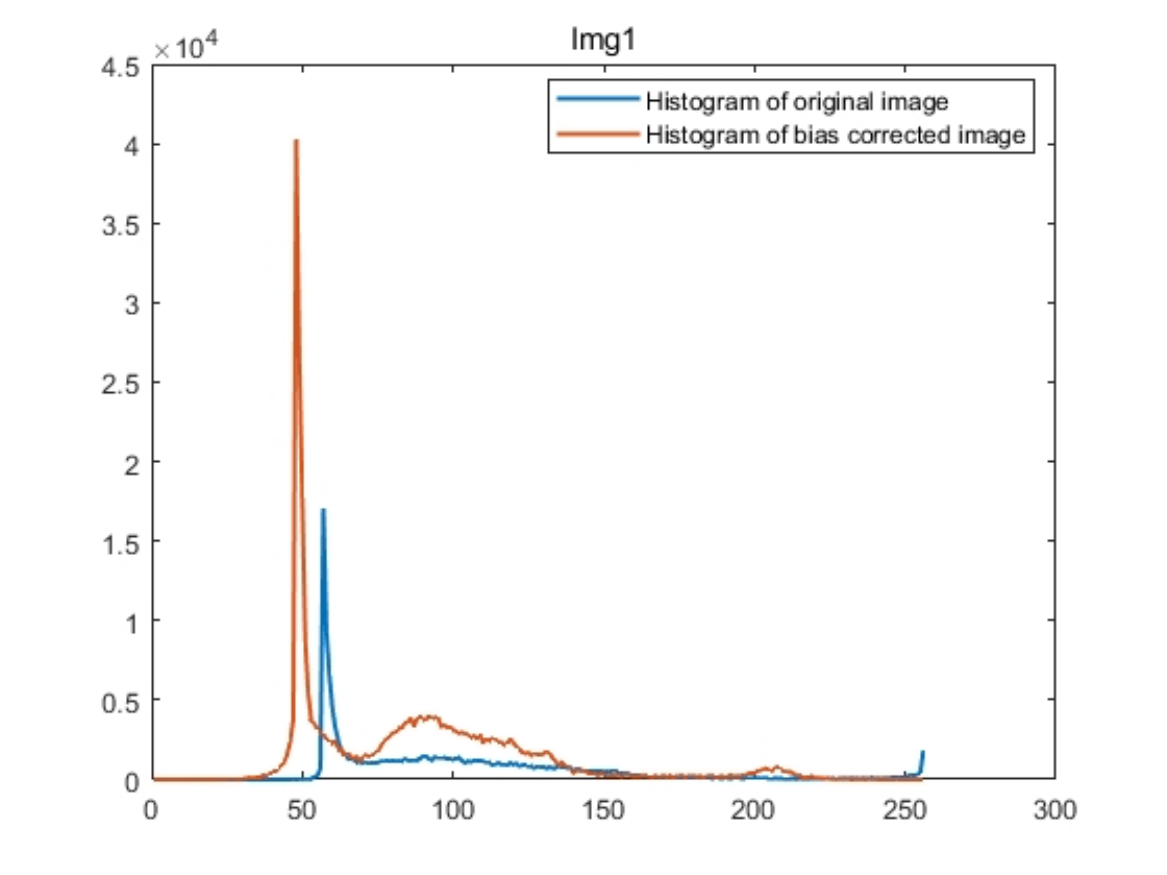}%
		\label{fig1}}\vspace{-3mm}\hspace{-1.5mm}
	\subfloat{\includegraphics[width=1.8in,height=1.4in]{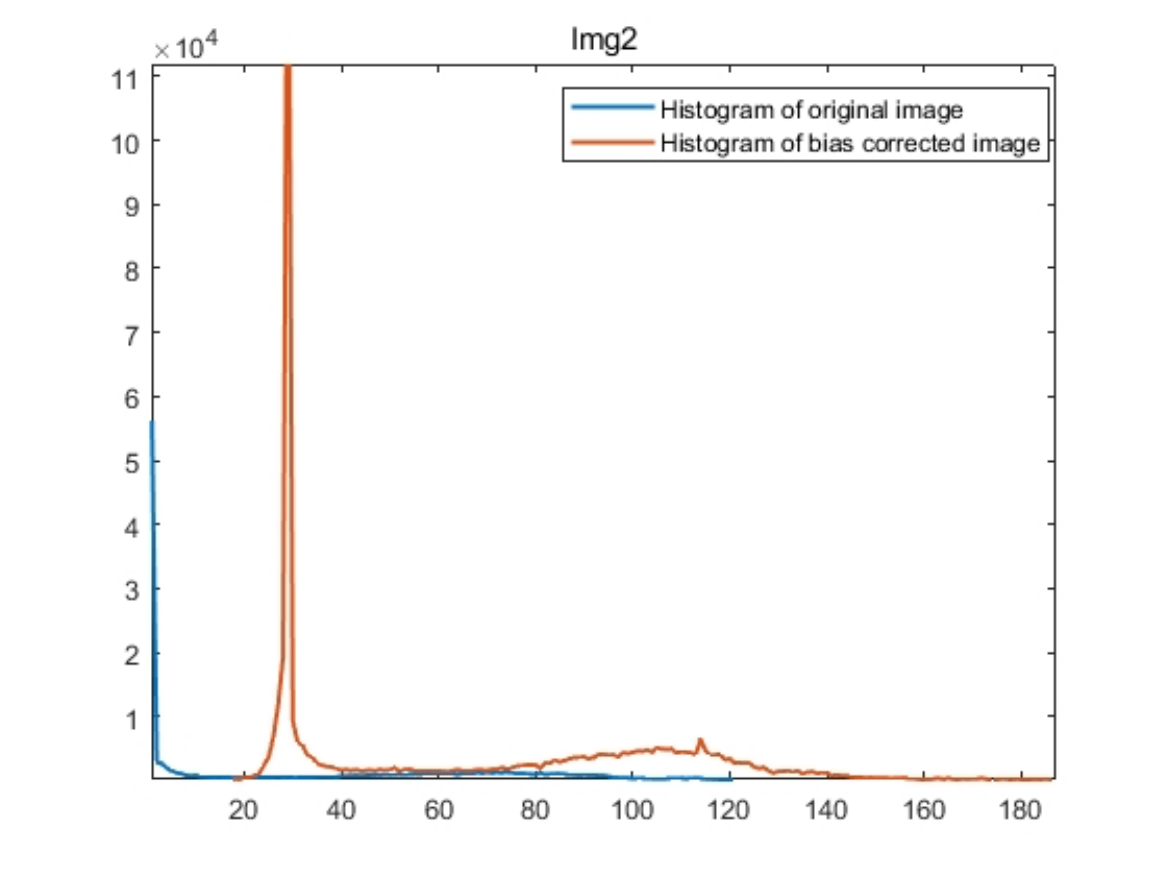}%
		\label{fig2}}
	\subfloat{\includegraphics[width=1.8in,height=1.4in]{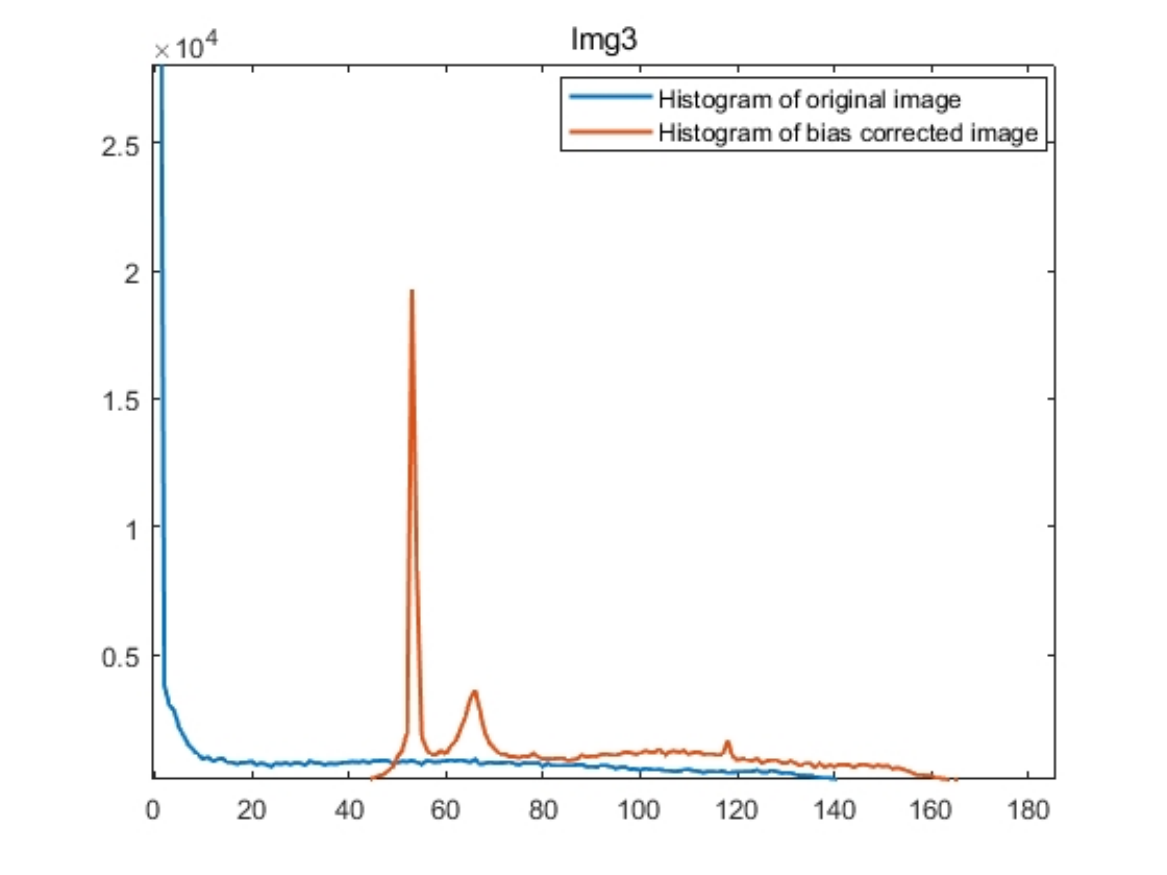}%
		\label{fig3}}
	\hfil
	\subfloat{\includegraphics[width=1.8in,height=1.4in]{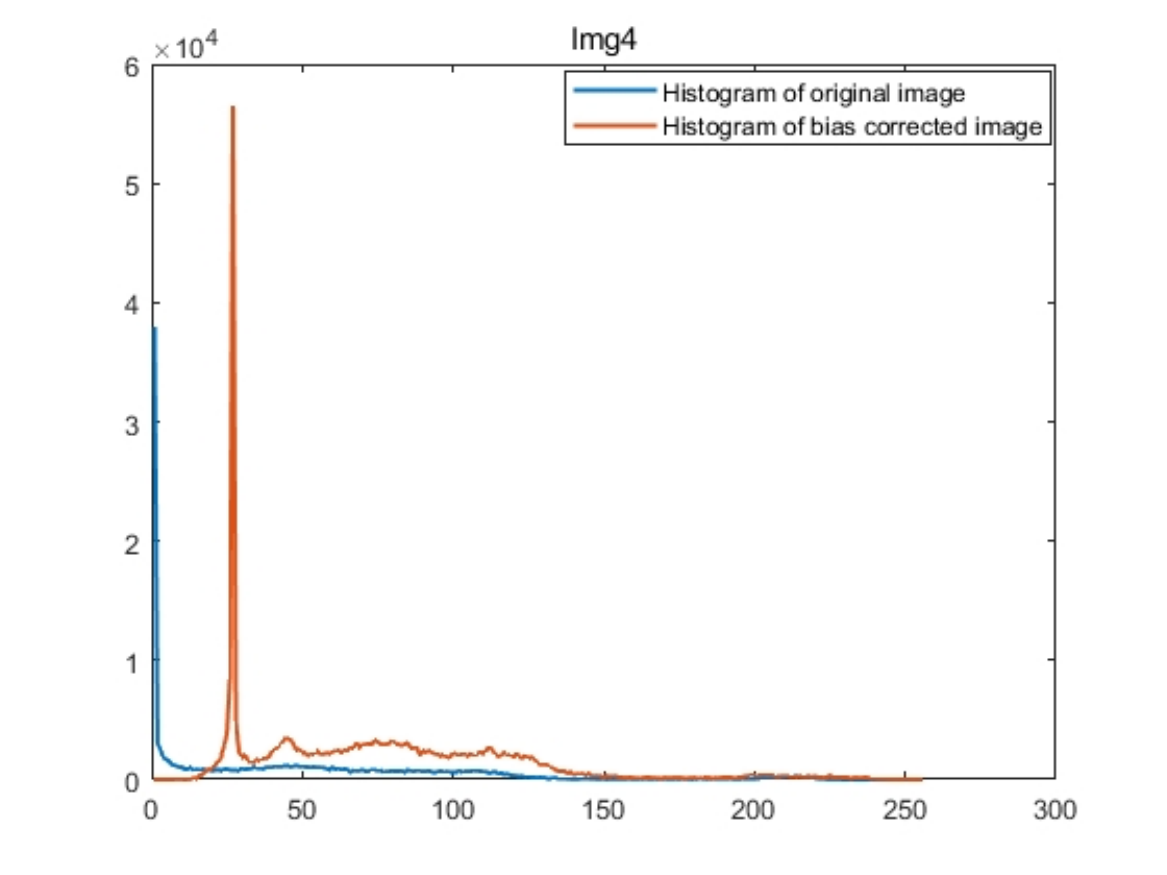}%
		\label{fig9}}
	\subfloat{\includegraphics[width=1.8in,height=1.4in]{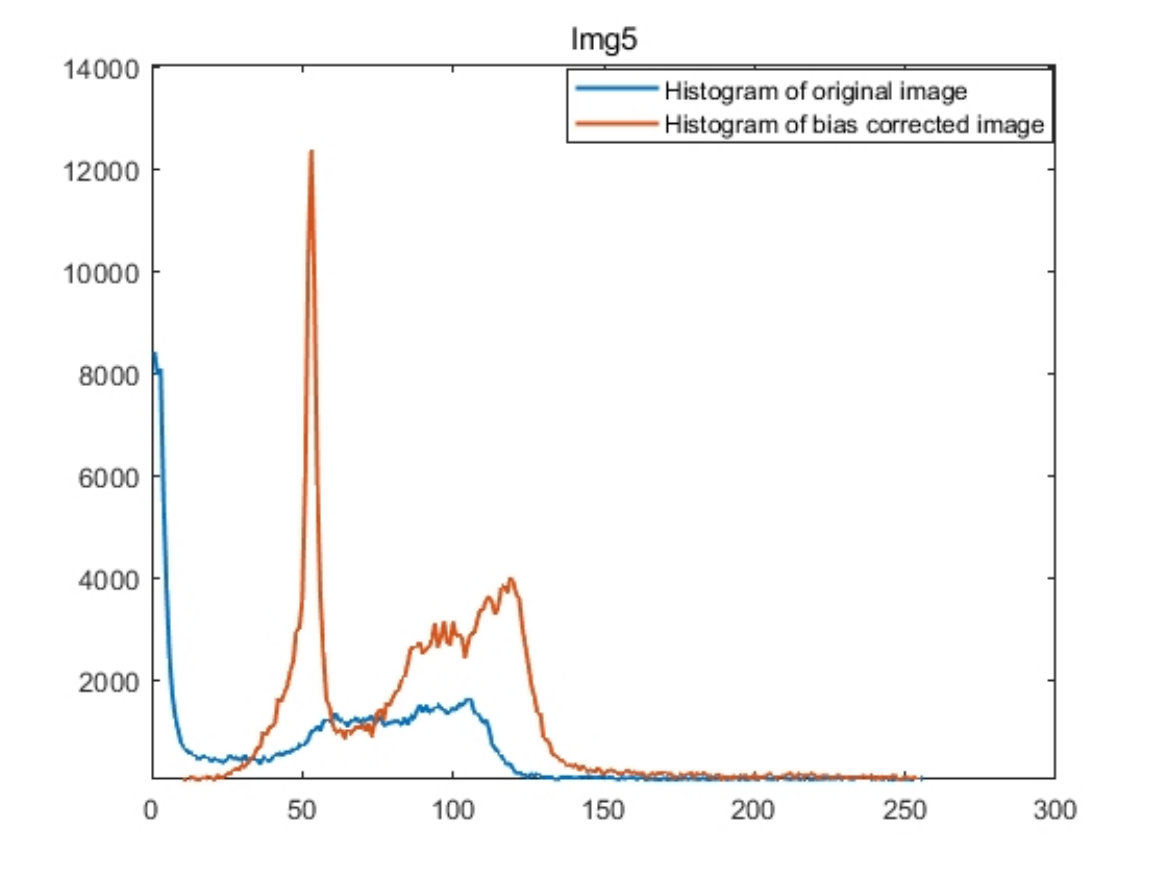}%
		\label{fig10}}
	\subfloat{\includegraphics[width=1.8in,height=1.4in]{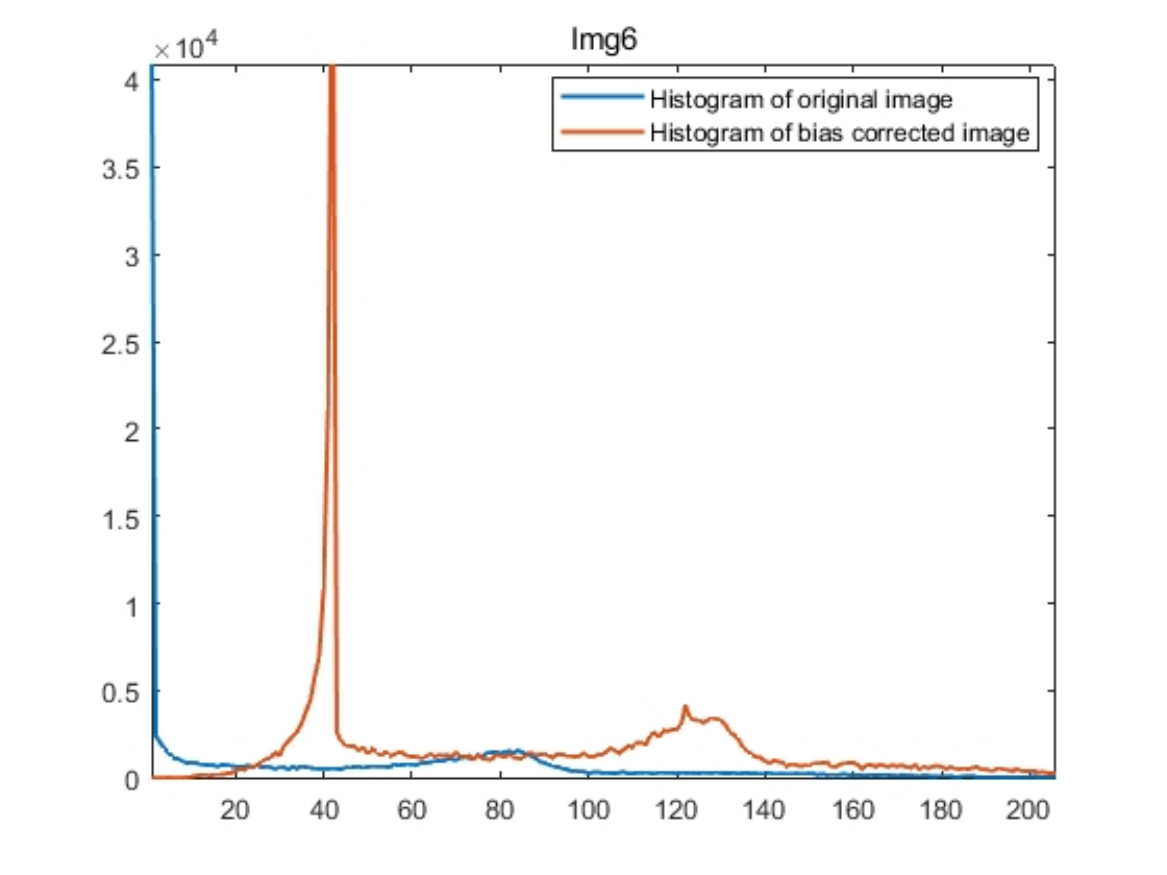}%
		\label{fig11}}
	\hfil
	\caption{Histograms comparison between original images and corrected images shown in Fig. \ref{img9}}
	\label{img10}
\end{figure}
\begin{figure}[!t]
	\centering
	\subfloat[]{\includegraphics[width=1.4in,height=1.4in]{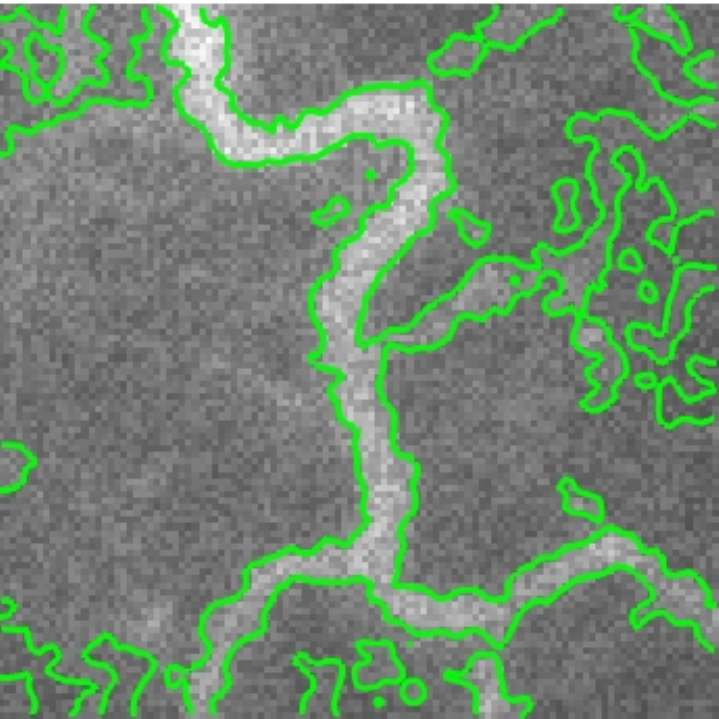}%
		\label{fig12}}
	\subfloat[]{\includegraphics[width=1.4in,height=1.4in]{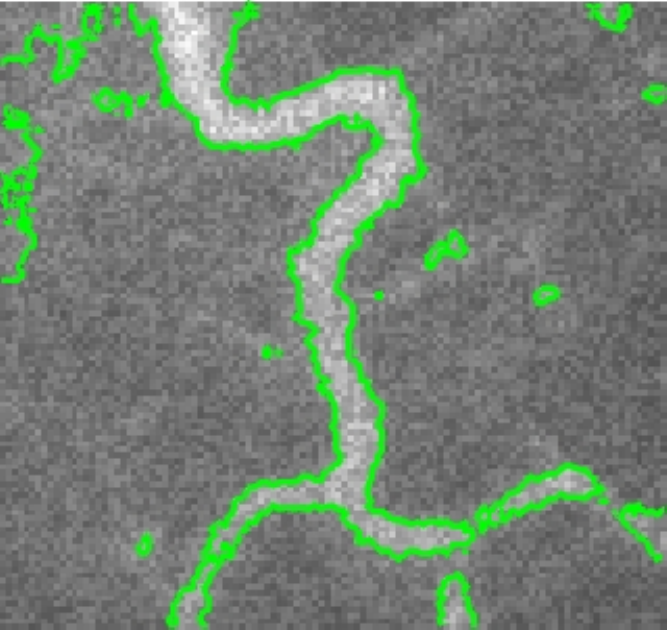}%
		\label{fig13}}
	\subfloat[]{\includegraphics[width=1.4in,height=1.4in]{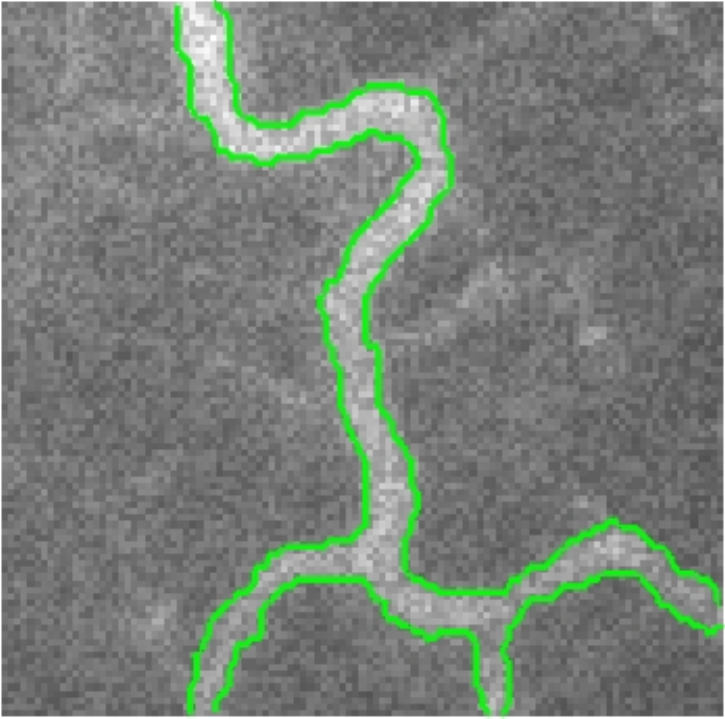}%
		\label{fig14}}
	\hfil
	\caption{Segmentation results from evaluated models for segmenting real blood vessel images. (a): The segmentation result from the LIF model, (b): The segmentation result from the LCV model, (c) The segmentation result from the RefLSM.}
	\label{img11}
\end{figure}
\subsection{Further Analysis}
\subsubsection{Bias Field Correction Analysis}
Fig.~\ref{img9} illustrates the performance of the RefLSM in correcting bias fields. 
 The 1st row presents original images characterized by significant intensity inhomogeneity. In contrast, the 2nd row displays the bias-corrected images produced by our method, highligting a more uniform appearance that facilitates segmentation. Additionally, histograms of the original images and corrected results are depicted in Fig. \ref{img10}. It is evident that the corrected results have higher peaks compared to the original images. Moreover, the corrected images exhibit greater uniformity, indicating easier segmentation and validating the effectiveness of the RefLSM's bias field estimation term.

\begin{figure*}[!t]
	\centering
	\subfloat{\includegraphics[width=1in,height=1in]{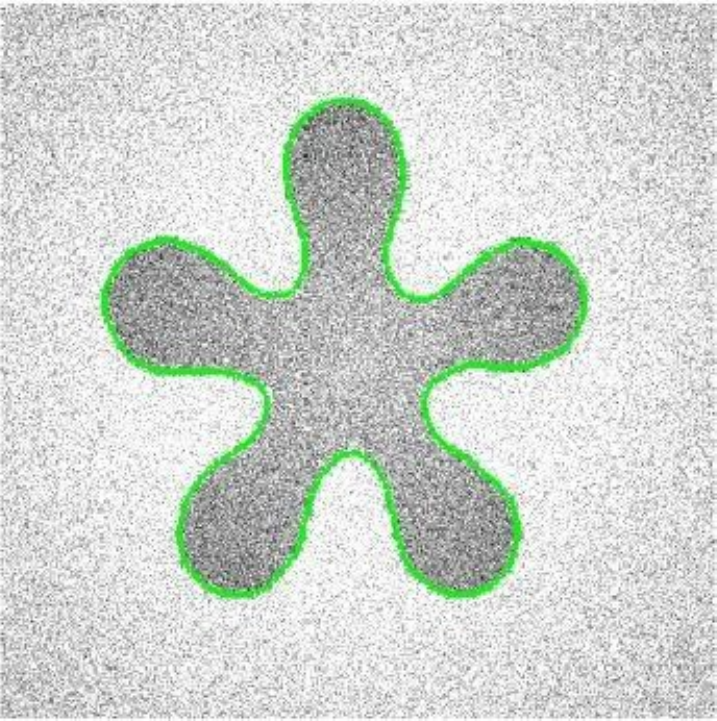}%
		\label{fig15}}\vspace{-3mm}\hspace{-1.5mm}
	\subfloat{\includegraphics[width=1in,height=1in]{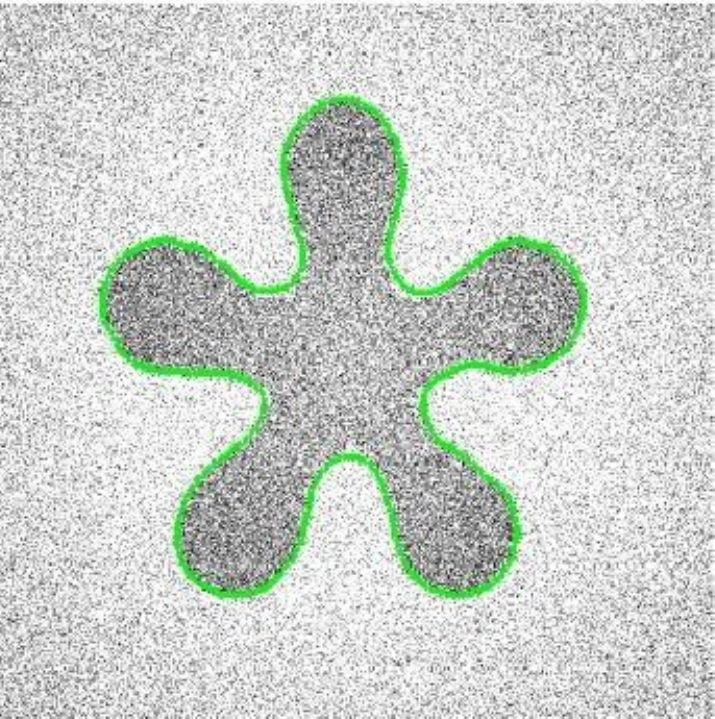}%
		\label{fig16}}
	\subfloat{\includegraphics[width=1in,height=1in]{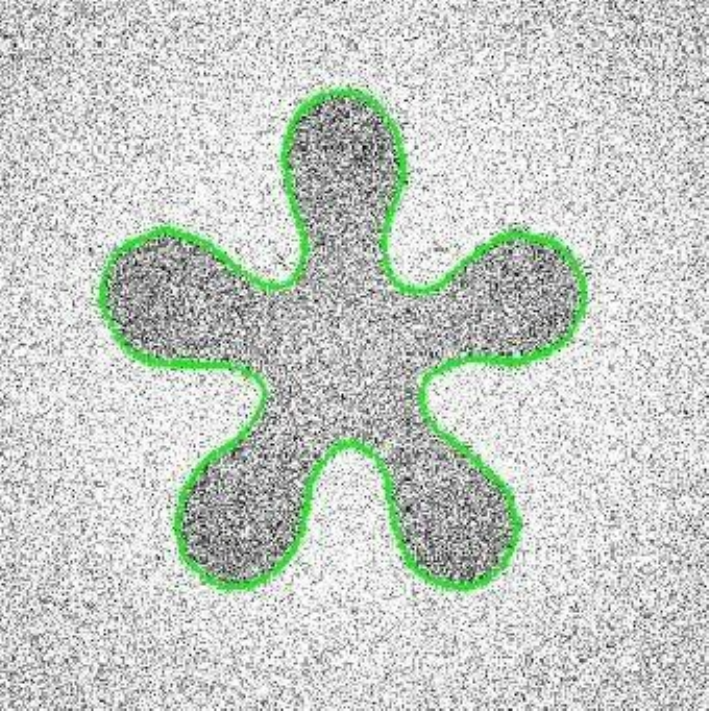}%
		\label{fig17}}
	\subfloat{\includegraphics[width=1in,height=1in]{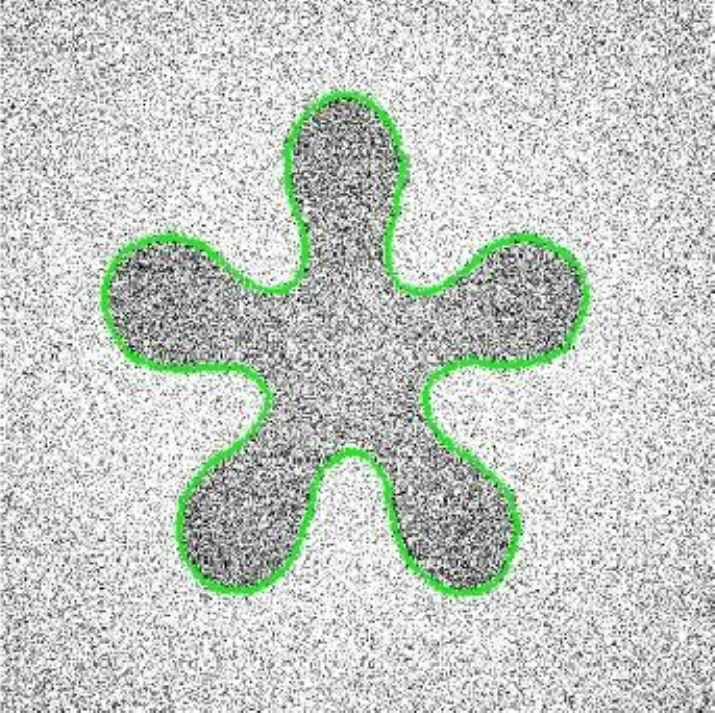}%
		\label{fig18}}
	\subfloat{\includegraphics[width=1in,height=1in]{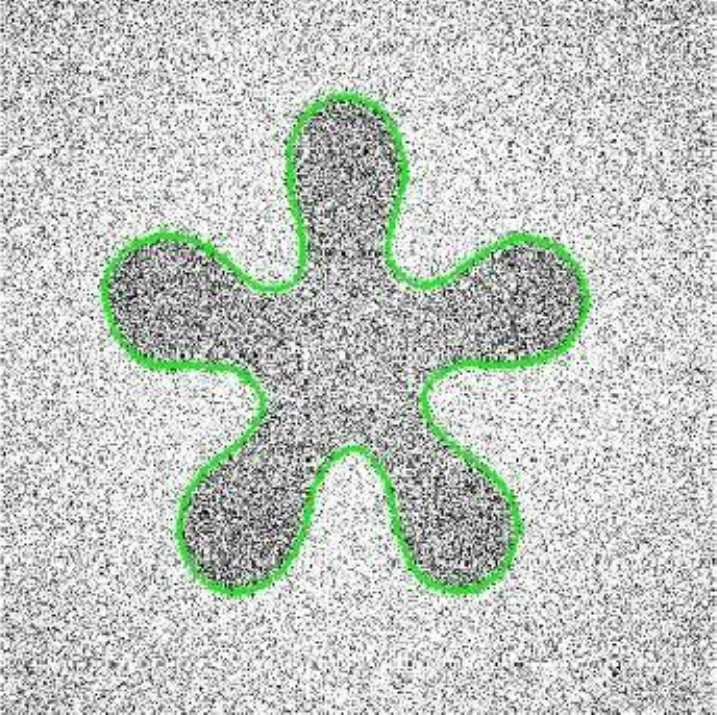}%
		\label{fig19}}
	\subfloat{\includegraphics[width=1in,height=1in]{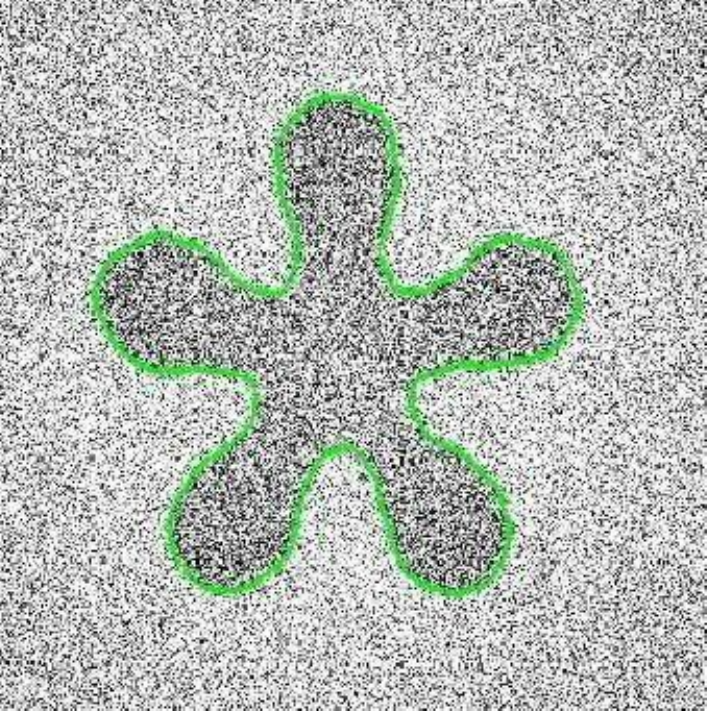}%
		\label{fig20}}
	\hfil
	
	\subfloat{\includegraphics[width=1in,height=1in]{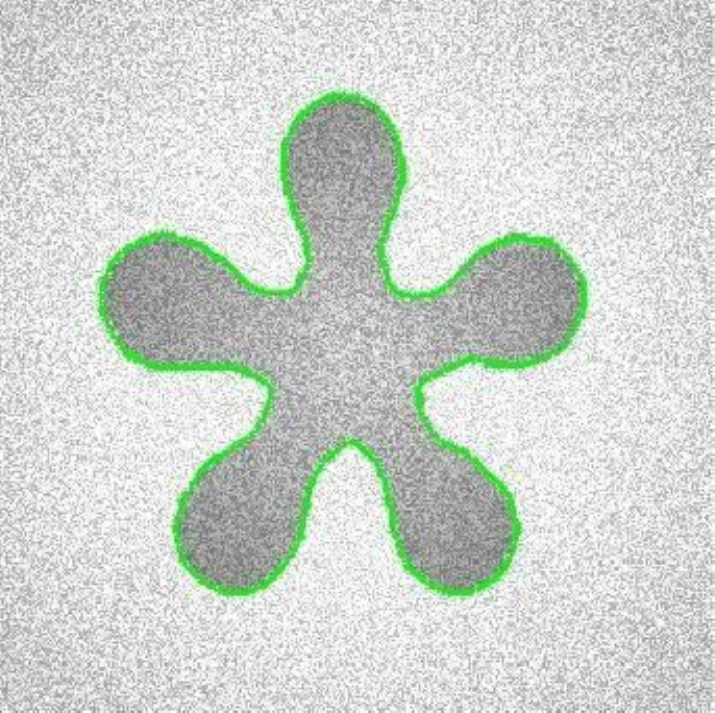}%
		\label{fig21}}\vspace{-3mm}\hspace{-1.5mm}
	\subfloat{\includegraphics[width=1in,height=1in]{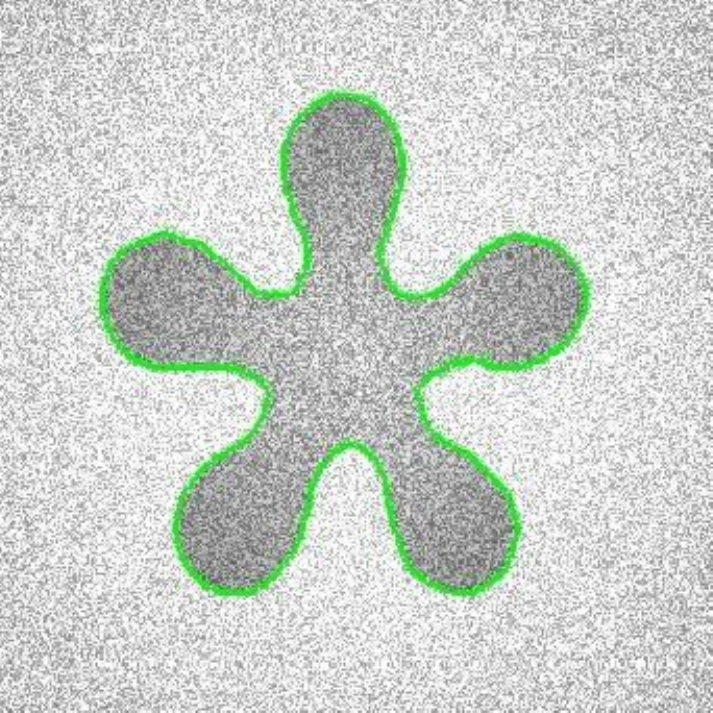}%
		\label{fig22}}
	\subfloat{\includegraphics[width=1in,height=1in]{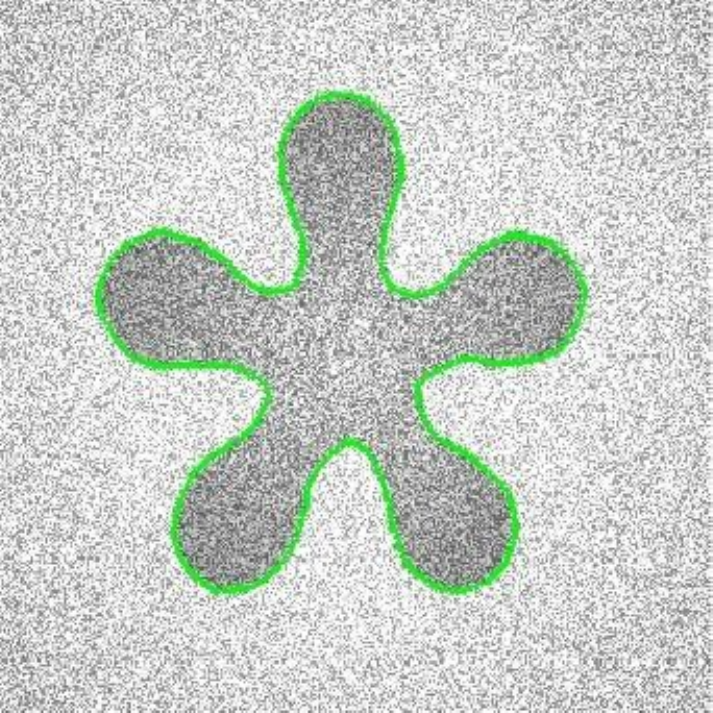}%
		\label{fig23}}
	\subfloat{\includegraphics[width=1in,height=1in]{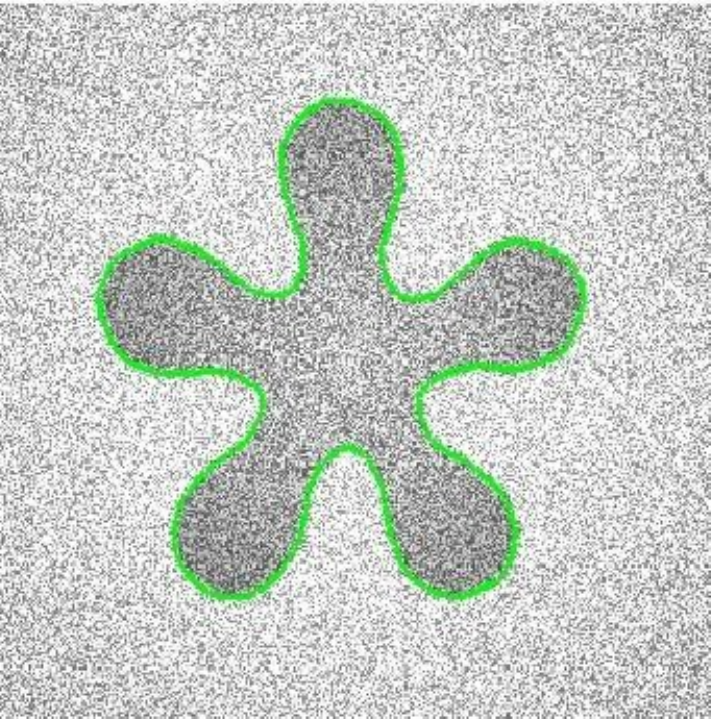}%
		\label{fig24}}
	\subfloat{\includegraphics[width=1in,height=1in]{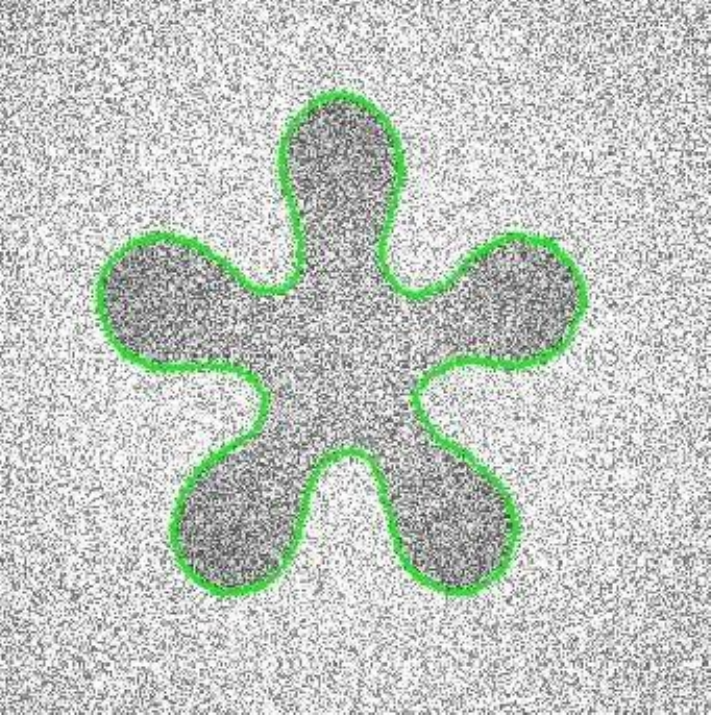}%
		\label{fig25}}
	\subfloat{\includegraphics[width=1in,height=1in]{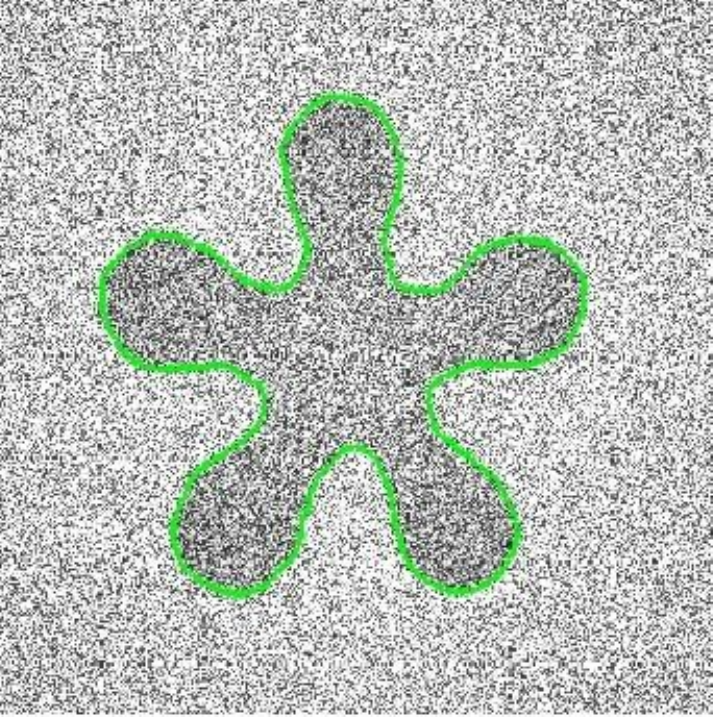}%
		\label{fig26}}
	\hfil
	
	\subfloat{\includegraphics[width=1in,height=1in]{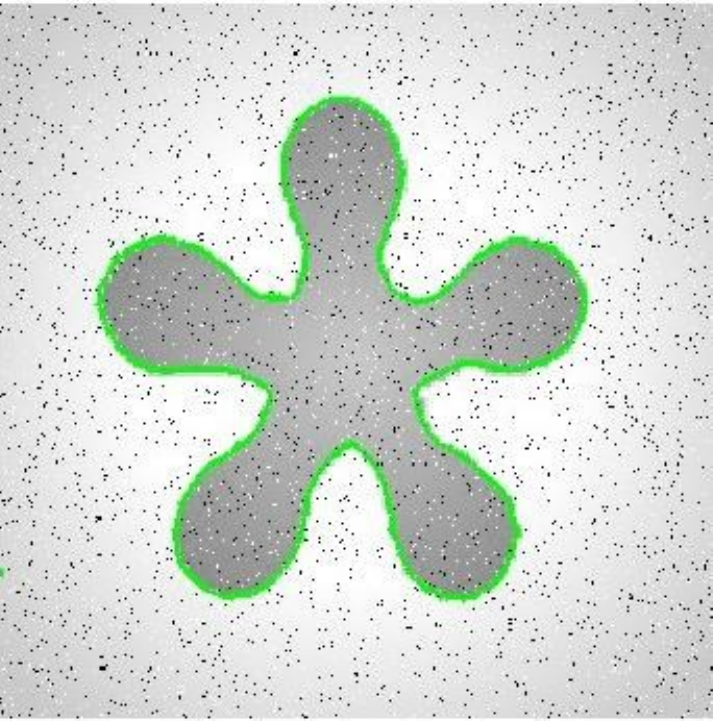}%
		\label{fig27}}
	\subfloat{\includegraphics[width=1in,height=1in]{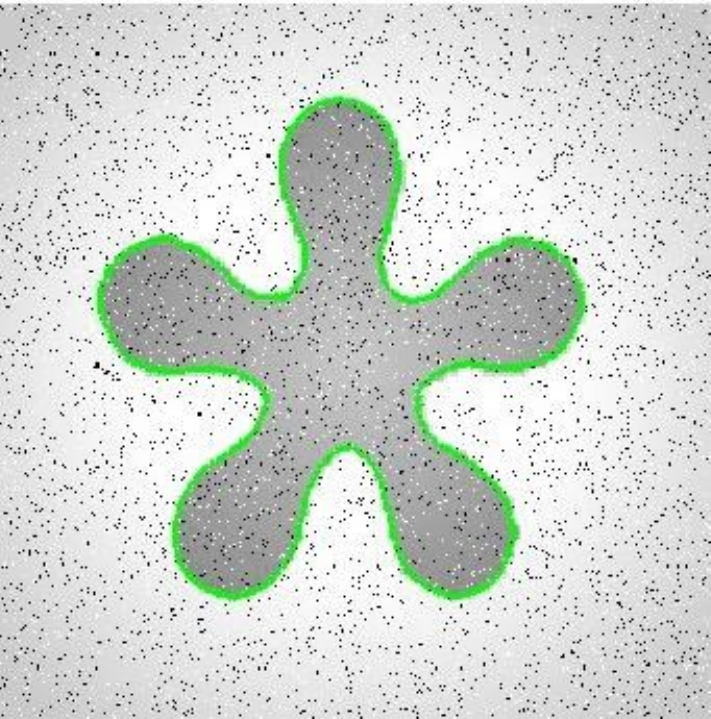}%
		\label{fig28}}
	\subfloat{\includegraphics[width=1in,height=1in]{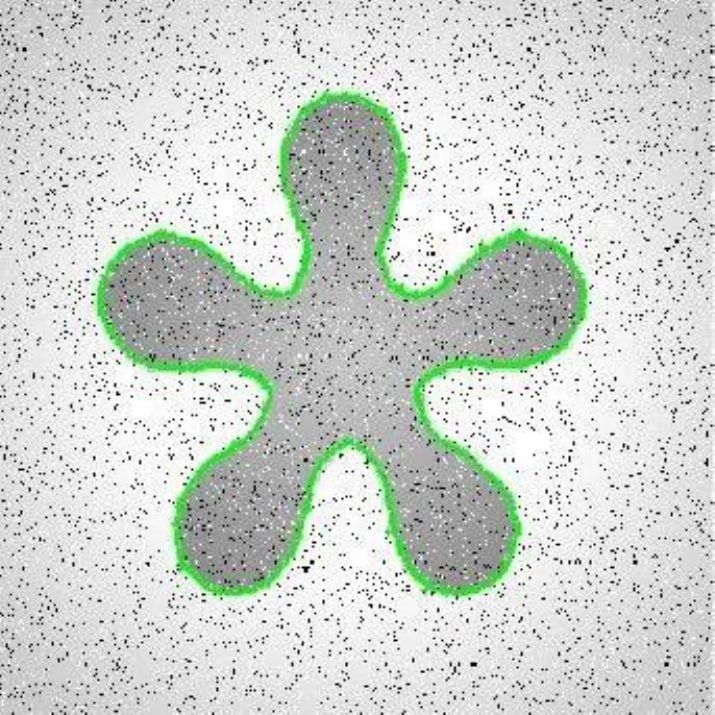}%
		\label{fig29}}
	\subfloat{\includegraphics[width=1in,height=1in]{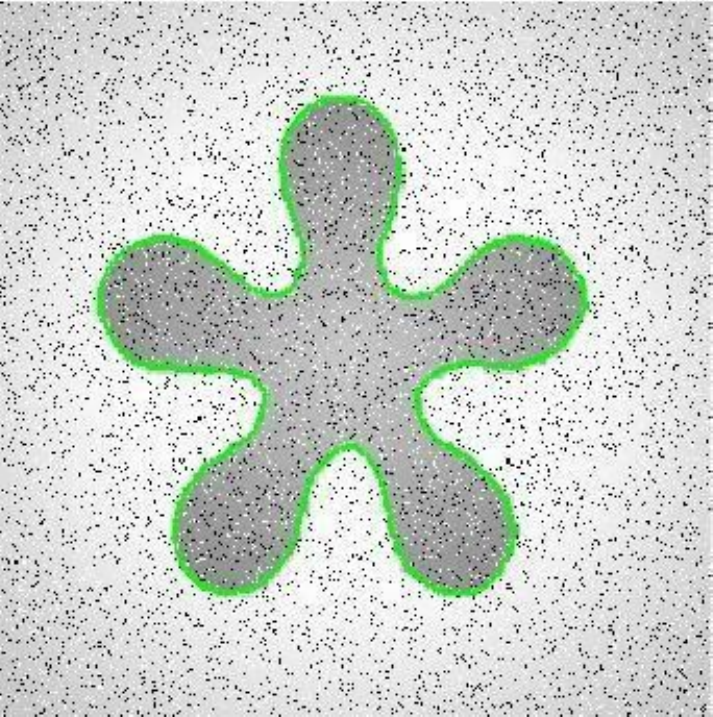}%
		\label{fig30}}
	\subfloat{\includegraphics[width=1in,height=1in]{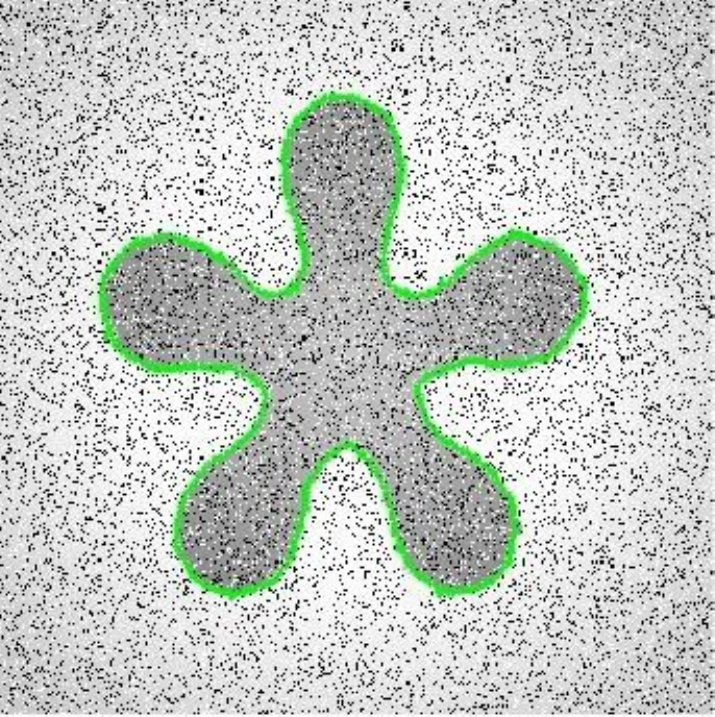}%
		\label{fig31}}
	\subfloat{\includegraphics[width=1in,height=1in]{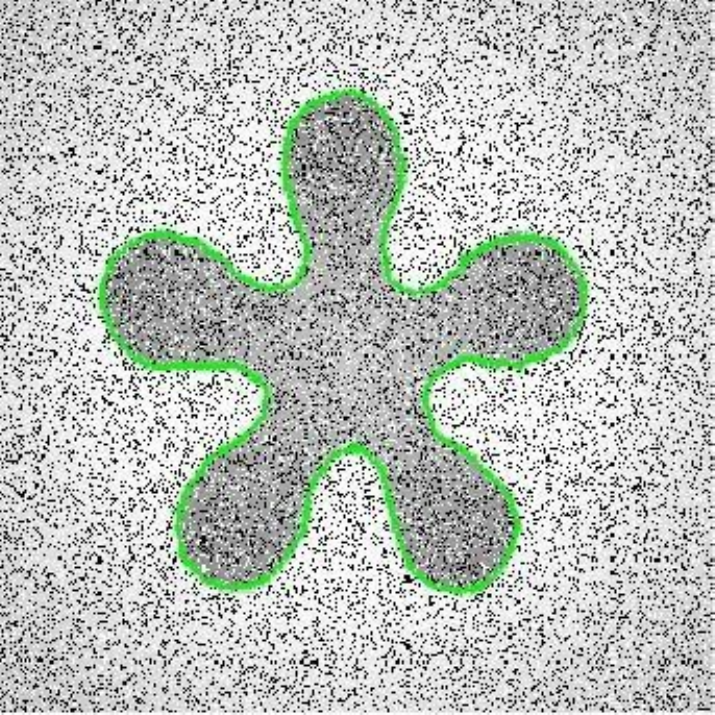}%
		\label{fig32}}
	\hfil
	\caption{Segmentation results from the RefLSM for a synthetic image added various types of noise at different levels. Row 1: Gaussian noise. Row 2: Salt and pepper noise. Row 3: Speckle noise. From left to right: Results of different noise density levels: 0.02, 0.04, 0.06, 0.08, 0.1, and 0.2.}
	\label{img12}
\end{figure*}

\begin{figure}[!t]
	\centering
	\subfloat[]{\includegraphics[width=2.3in,height=2in]{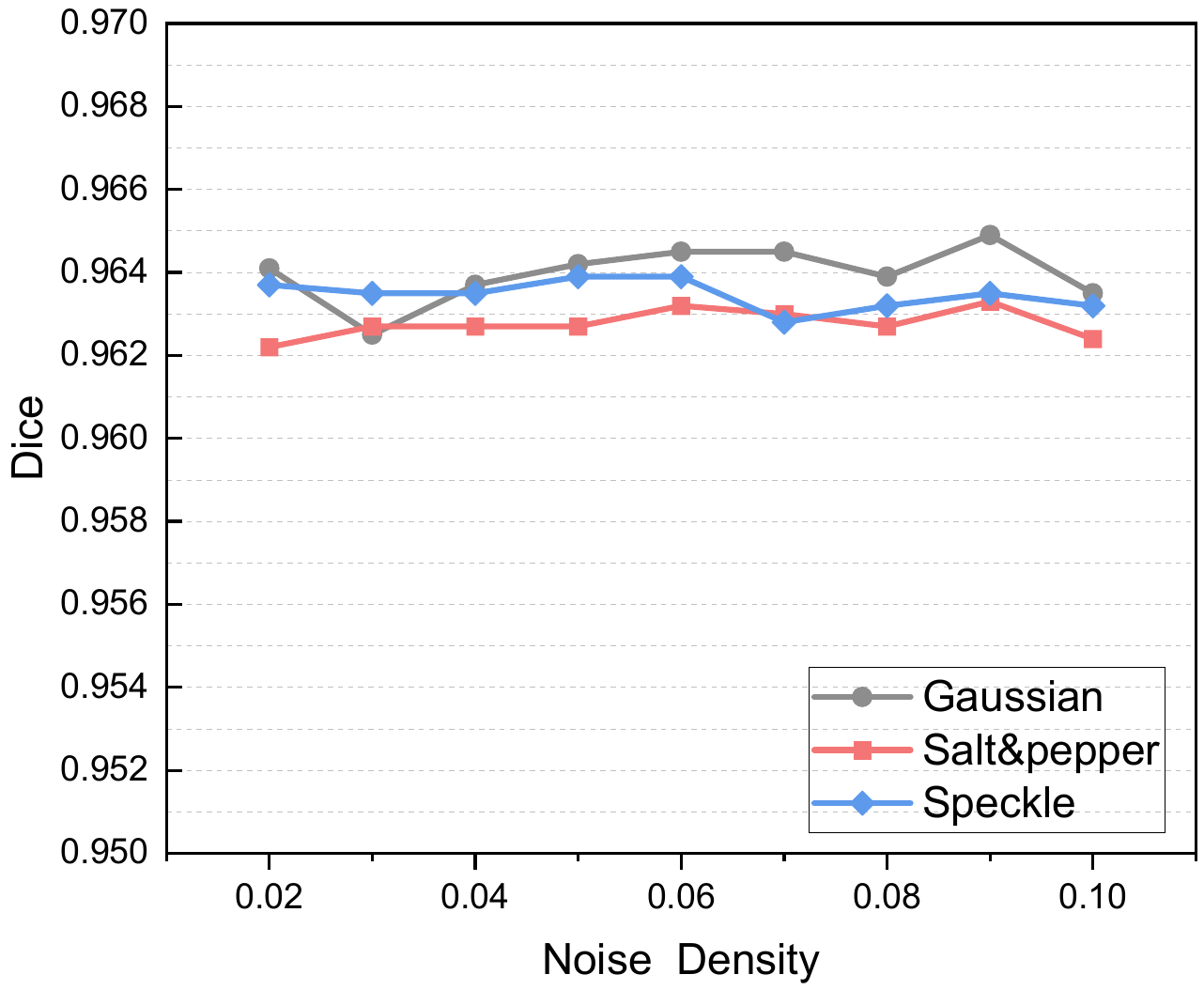}%
		\label{fig33}}
	\subfloat[]{\includegraphics[width=2.3in,height=2in]{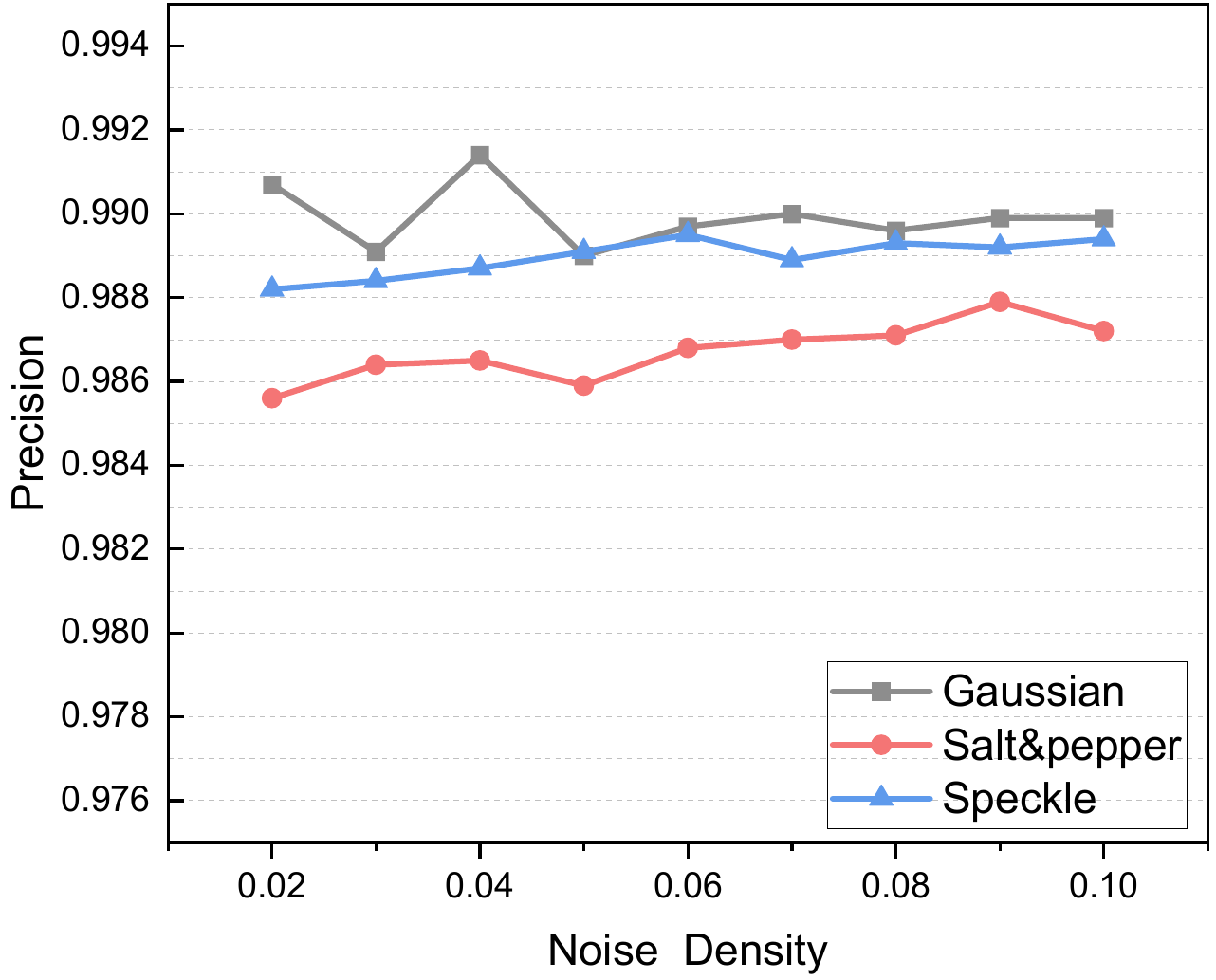}%
		\label{fig34}}
	\hfil
	\caption{Quantitative comparison of the RefLSM model for segmenting images with different kinds and levels of noise distributions shown in Fig. \ref{img12}}
	\label{img13}
\end{figure}

\subsubsection{Denoise Analysis}
General level set models often struggle with efficiency and accuracy in segmenting images with high noise levels as the interference from high noise often makes it difficult to identify image boundaries. In this section, we assess the denoising ability of the RefLSM.
 Fig. \ref{img11} (a) and Fig. \ref{img11} (b) display poor segmentation results from LIF \cite{ZHANG20101199} and LCV \cite{WANG2010603} methods under Gaussian noise interference, while our model, as shown in Fig. \ref{img11} (c), demonstrates effective segmentation and robustness against noise. To further validate this, we conducted experiments on synthetic images with various types of noise, including Gaussian noise, salt and pepper noise, and speckle noise, at different noise density levels: 0.02, 0.04, 0.06, 0.08, 0.1, and 0.2, as illustrated in Fig. \ref{img12}. The synthetic images are standardized to a size $336\times336$.
Moreover, the Precision and Dice values calculated for the RefLSM across these different noise levels remain visually consistent. As shown in Fig. \ref{img13}(a) and Fig. \ref{img13}(b), the model matains Precision values above 0.98 and Dice values above 0.96, indicating that segmentation results are marginally affected by various types of noise at different density levels. This experiment further demonstrates the proposed method’s robustness against noise.
\begin{figure}[!t]
	\centering
	\subfloat{\includegraphics[width=2.5in]{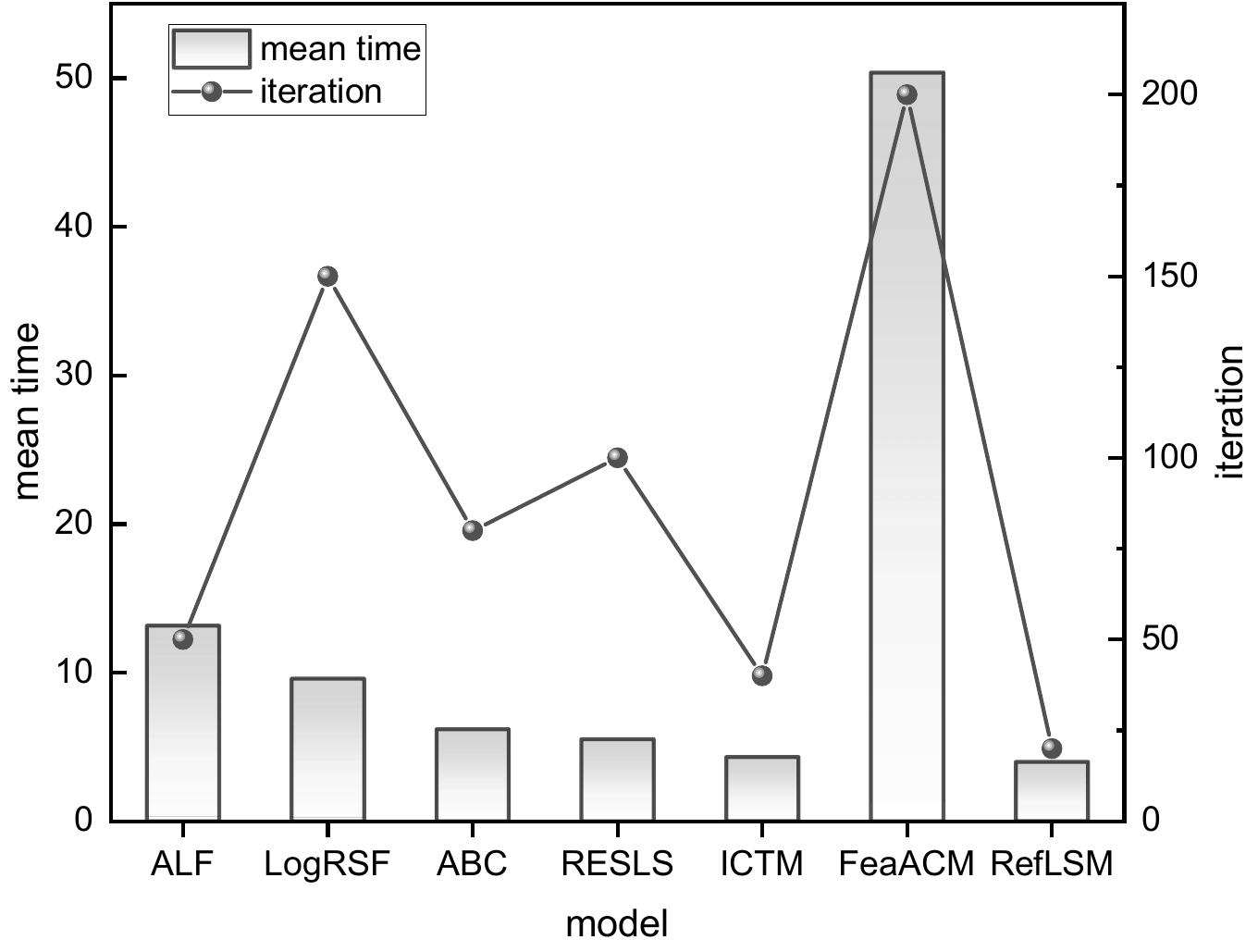}%
		\label{fig133}}
	\hfil
	\caption{ Mean CPU time(s) and iterations and of different models for segmenting Right ventricle images shown in Fig. \ref{img4}}
	\label{img19}
\end{figure}
\begin{table}[!ht]
\centering
\caption{Comparison of iterations and average CPU time (s) of different models for segmenting right ventricle images in Fig.~\ref{img6}.}
\renewcommand{\arraystretch}{1.2}
\small
\setlength{\tabcolsep}{6pt}
\begin{tabular}{lccccccc}
\toprule
Metric & RESLS & ALF & ABC & ICTM & FeaACM & LoGRSF & RefLSM \\
\midrule
Iterations & 100 & 50 & 80 & 40 & 200 & 150 & \textbf{20} \\
Time (s) $\downarrow$ & 5.515 & 13.159 & 6.186 & 6.517 & 50.373 & 9.599 & \textbf{4.003} \\
\bottomrule
\end{tabular}
\label{t4}
\end{table}

\subsubsection{Computation Efficiency Analysis}
In addition to segmentation performance, the efficiency of the model is also a crucial evaluation factor. Table \ref{t4} presents the iterations and the average CPU time of seven evaluated models for segmenting right ventricle images. Since the RefLSM directly considers the reflectance component of images and we employ the alternating direction method to implement the algorithm, the computation efficiency of the RefLSM is excellent that it converges in just 20 iterations for right ventricle images. Furthermore, we created Fig. \ref{img19} based on Table. \ref{t4} for data visualization. In comparison to other models, it is obvious that the RefLSM requires the shortest average CPU time and the least iterations for segmenting right ventricle images. Thus, when taking both accuracy and computation efficiency into account, the RefLSM is both accurate and fast for segmenting right ventricle images.
\begin{figure*}[!t]
	\centering
	\subfloat[]{\includegraphics[width=0.9in,height=0.9in]{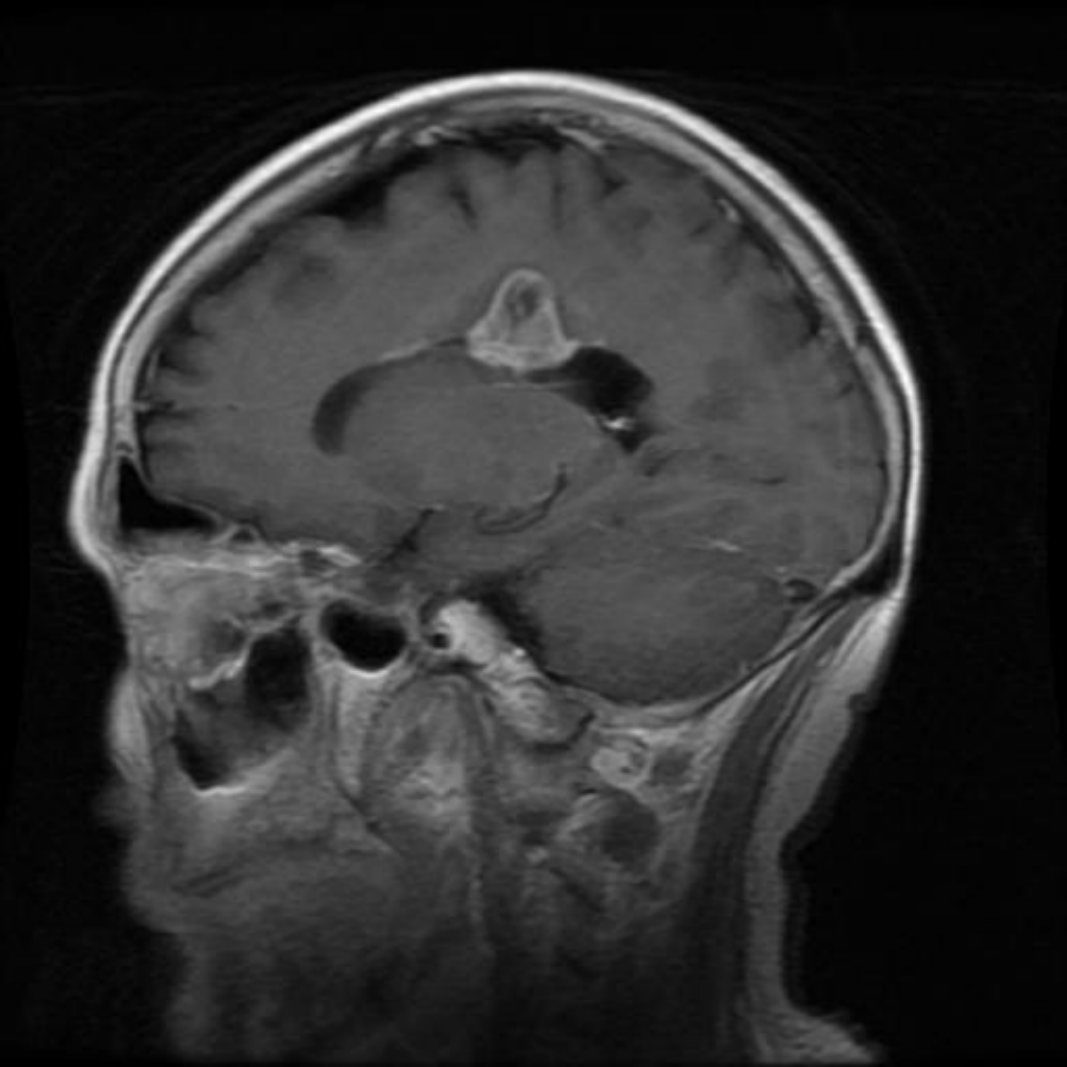}%
		\label{fig35}}
	\subfloat[]{\includegraphics[width=0.9in,height=0.9in]{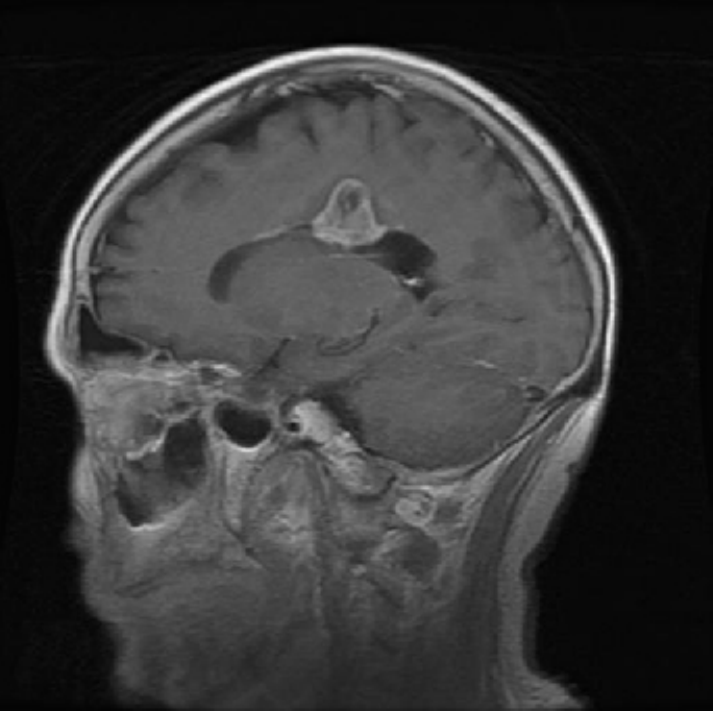}%
		\label{fig36}}
	\subfloat[]{\includegraphics[width=0.9in,height=0.9in]{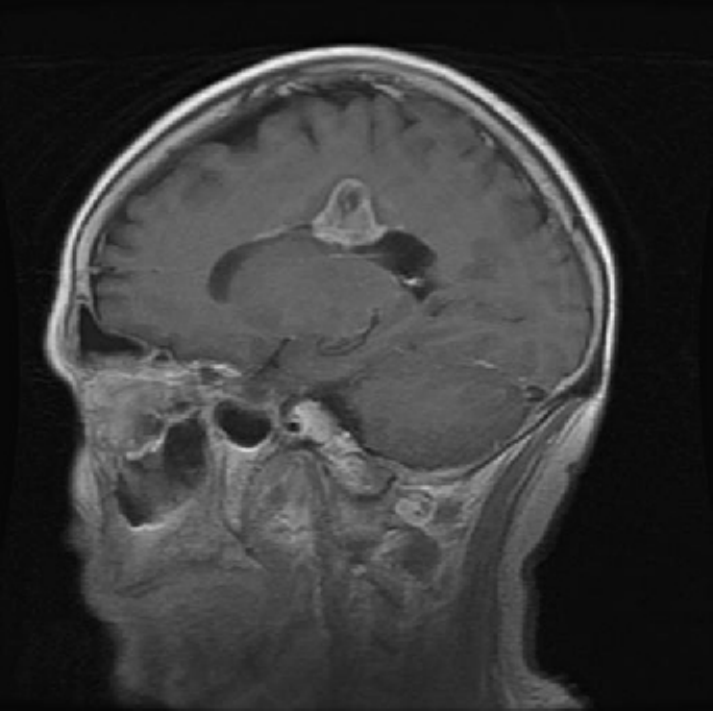}%
		\label{fig38}}
	\subfloat[]{\includegraphics[width=0.9in,height=0.9in]{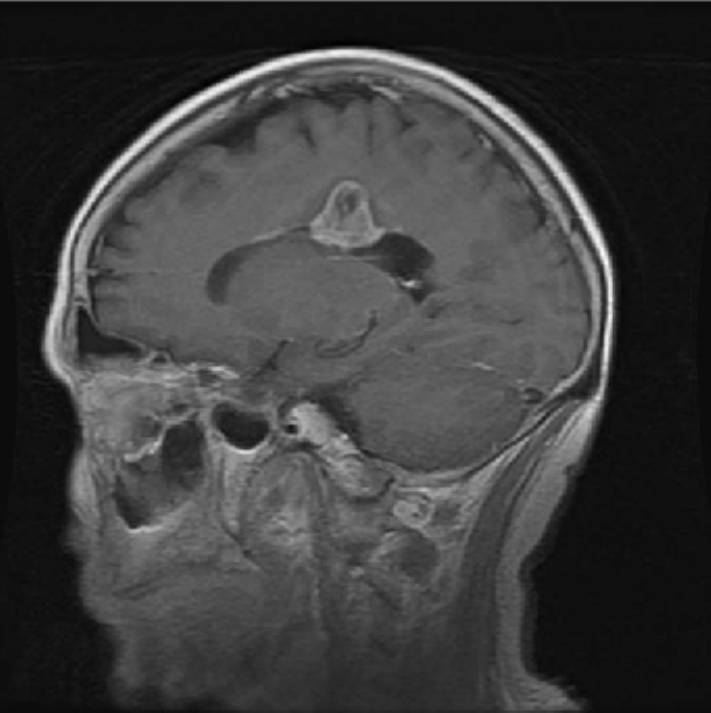}%
		\label{fig39}}
	\subfloat[]{\includegraphics[width=0.9in,height=0.9in]{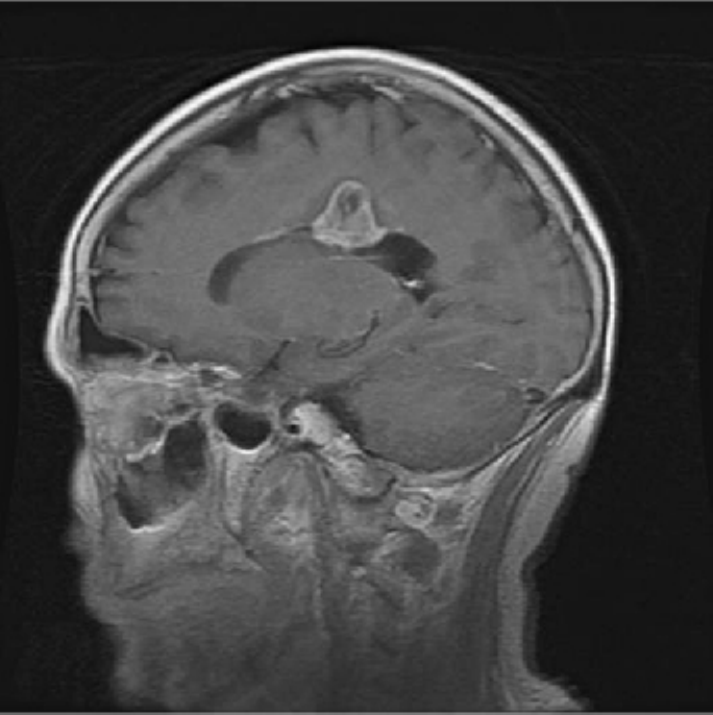}%
		\label{fig40}}
	\subfloat[]{\includegraphics[width=0.9in,height=0.9in]{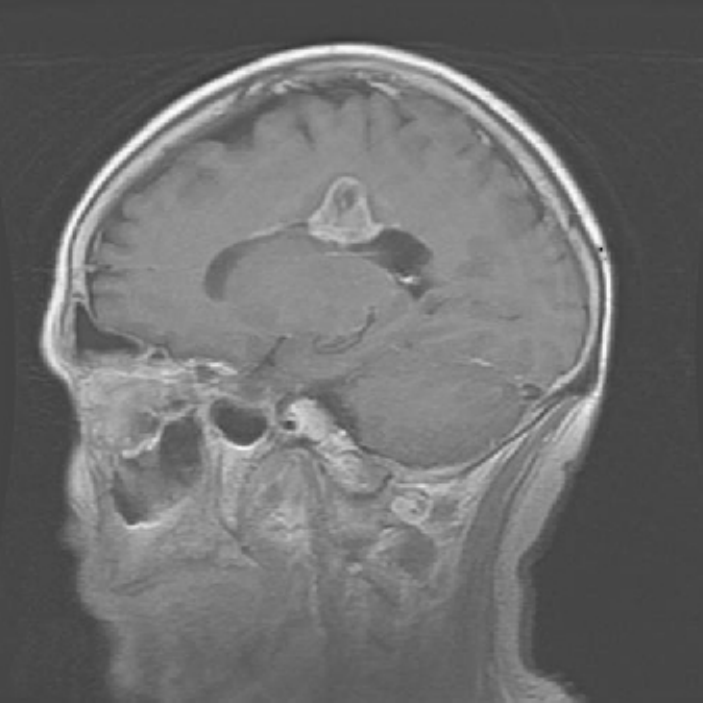}%
		\label{fig37}}
	\subfloat[]{\includegraphics[width=0.9in,height=0.9in]{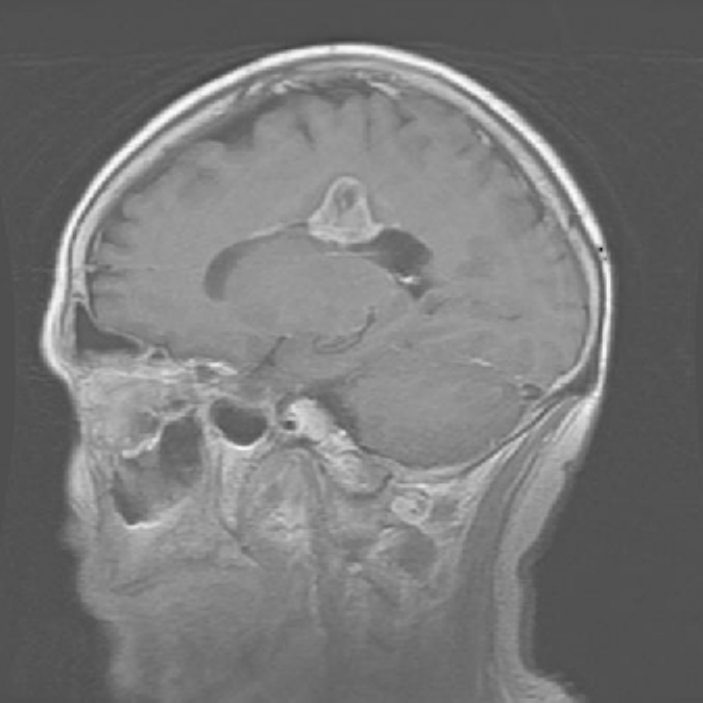}%
		\label{fig41}}
	\hfil
	\caption{Bias field correction results from the RefLSM with different $\tau$ for Brain tumor MR images. From left to right: The original image, corrected results of $\tau =0.001, \tau= 0.04, \tau = 0.12, \tau= 0.16, \tau = 0.17, \tau = 0.18$.}
	\label{img15}
\end{figure*}

\begin{table}[!ht]
\centering
\caption{Comparison of RTG values of the RefLSM with different $\lambda_3$ for bias field correction results in Fig.~\ref{img15}.}
\renewcommand{\arraystretch}{1.2}
\small
\setlength{\tabcolsep}{6pt}
\begin{tabular}{lcccccc}
\toprule
Image & $\tau=0.001$ & $\tau=0.04$ & $\tau=0.12$ & $\tau=0.16$ & $\tau=0.17$ & $\tau=0.18$ \\
\midrule
RTG $\uparrow$ & 1.0378 & 1.0536 & 1.1408 & \textbf{1.1638} & 0.9391 & 0.8930 \\
\bottomrule
\end{tabular}
\label{t5}
\end{table}

\subsubsection{Parameters Analysis}
In preceding experiments, specific parameter selections have been provided. In this section, we focus on the analysis of the weight $\tau$ of the prior term in Eq. \eqref{final}, which is crucial for re-balancing the intensity and improving low-brightness areas of images concurrently. Moreover, $\tau$ plays a crucial role in the bias field correction and the whole segmentaion process. Fig. \ref{img15} presents the bias field correction results of the RefLSM using different $\tau$ for brain tumor MR images. Moreover, we use the RTG metric, with the tenengrad (TG) of the original image as a benchmark, to quantitatively assess the image quality. Specifically, RTG values of the correction results in Fig. \ref{img15} are computed to evaluate the influence of $\tau$, as shown in Table \ref{t5}. If the RTG is greater than 1, the image quality surpasses the original, with larger values indicating better correction. If the RTG is less than 1, the quality is inferior to the original, suggesting possible under-correction or over-correction. Under our multiple tests, we found that when the value of $\tau$ is around 0.16, the highest RTG can be obtained. When $\tau$ is less than 0.16, RTG is positively correlated with $\tau$, demonstrating the validity of the Prior Constraint Term. When $\tau$ exceeds a certain value, RTG decreases as $\tau$ increases due to over-correction that enhances the entire image's brightness, as shown in Fig. \ref{img15} (f) and (g). As a result, we typically set $\tau = 0.16$ in most of our experiments to balance bias field correction and image segmentation.
\section{Conclusion}\label{section:E}
In this paper, we present an innovative variational level set model based on reflectance for segmenting medical image. The proposed model draw from Retinex theory and incorporates reflectance component, thereby reducing the interference of bias field and improving the segmentation accuracy of complex medical images characterized by severe intensity inhomogeneity and influence from surrounding tissue. By integrating the reflectance component into the level set energy functional, we can capture the structual texture information without being affected by bias field during the segmentation. Furthermore, the linear structural prior can effectively restore image intensity, resulting in improved segmentation accuracy for the model. In addition, the RefLSM model robustly segments noisy images by incorporating a relaxed binary level set into the variational model. Comparison experiments conducted on medical images (both MR and ultrasound) as well as synthetic images with different noise types demonstrate that the RefLSM method achieves superior accuracy and effectiveness compared to most leading methods. We also design experiments to validate the model’s ability in Bias Field Correction.
%	\centering
%		\includegraphics[<options>]{}
%	  \caption{}\label{fig1}
%\end{figure}

% Uncomment and use as the case may be
%\begin{theorem} 
%\end{theorem}
\section*{Acknowledgements}
The presented work was supported by the National Science and Technology Innovation 2030 of China Next-Generation Artificial Intelligence Major Project (Grant No.2018AAA0101800); Innovation Group Science Fund of Chongqing Natural Science Foundation (No.cstc2019jcyj-cxttX0003).
% Here goes the abstractt

\bibliography{ref}

\end{document}